\documentclass{article} %
\usepackage{iclr2025_conference,times}

\usepackage{amsmath,amsfonts,bm}

\def\eqref#1{equation~\ref{#1}}

\def\1{\bm{1}}

\def\va{{\bm{a}}}
\def\vb{{\bm{b}}}

\def\vx{{\bm{x}}}
\def\vy{{\bm{y}}}
\def\vz{{\bm{z}}}

\DeclareMathAlphabet{\mathsfit}{\encodingdefault}{\sfdefault}{m}{sl}
\SetMathAlphabet{\mathsfit}{bold}{\encodingdefault}{\sfdefault}{bx}{n}

\def\tZ{{\tens{Z}}}

\newcommand{\R}{\mathbb{R}}

\DeclareMathOperator*{\argmin}{arg\,min}

\usepackage{hyperref}
\usepackage{url}
\usepackage{xspace}
\usepackage{cleveref}
\usepackage{graphicx}
\usepackage{stmaryrd}
\usepackage{subcaption}
\usepackage{booktabs}

\usepackage{algorithm}
\usepackage{algpseudocode}

\usepackage{tcolorbox}%
\usepackage{listings}
	
\usepackage{color, colortbl}
\usepackage{placeins}
\usepackage{hyperref}
\usepackage{multirow} %
\usepackage{nicematrix}

\usepackage{tikz}
\usetikzlibrary{decorations.pathreplacing,positioning,calc,shapes.geometric,arrows,bending}
\usetikzlibrary[bending]

\definecolor{TableRow}{rgb}{0.9, 0.9, 0.92}

\DeclareFontEncoding{LS1}{}{}
\DeclareFontSubstitution{LS1}{stix}{m}{n}
\DeclareSymbolFont{stixletters}{LS1}{stix}{m}{it}
\DeclareMathAccent{\cev}{\mathord}{stixletters}{"91}
\DeclareMathAccent{\vec}{\mathord}{stixletters}{"92}
\DeclareMathAccent{\vecev}{\mathord}{stixletters}{"95}

\newcommand{\method}{\textsc{AcT}\xspace}
\newcommand{\mean}{Mean-\method}
\newcommand{\linear}{Linear-\method}

\newcommand{\aura}{\textsc{AurA}\xspace}
\newcommand{\iti}{\textsc{ITI-c}\xspace}
\newcommand{\massmean}{\textsc{ITI-m}\xspace}
\newcommand{\actadd}{\textsc{ActAdd}\xspace}
\newcommand{\repe}{\textsc{RepE}\xspace}

\newcommand{\gemmatwob}{Gemma2-2B\xspace}
\newcommand{\llamaeightb}{Llama3-8B\xspace}
\newcommand{\llamaeightbinstr}{Llama3-8B-instruct\xspace}
\newcommand{\mistralsevenb}{Mistral-7B\xspace}

\newcommand{\ot}{OT\xspace}
\newcommand{\eg}{\textit{e.g.,}\xspace}
\newcommand{\ie}{\textit{i.e.,}\xspace}
\newcommand{\GM}{GM\xspace}
\newcommand{\ii}{inference-time interventions}

\newcommand{\qt}{\text{qt}}

\newtheorem{definition}{Definition}[section]
\newtheorem{prop}{Proposition}[section]

\definecolor{grayorange}{HTML}{967653}
\definecolor{takeawaycolor}{rgb}{0.1, 0.1, 0.12}
\definecolor{improvementcolor}{rgb}{0.122, 0.8, 0.22}
     
\newtcbox{\mybox}{
  colframe=takeawaycolor,colback=takeawaycolor!5!white,fontupper=\rmfamily,
  boxrule=0.0pt,arc=0pt,boxsep=0pt,left=2pt,right=2pt,top=1.9pt,bottom=1pt,nobeforeafter,tcbox raise base,after={\hspace{2pt}}}

\newcommand{\takeaway}[1]{\mybox{\textbf{#1}}}

\newtcbox{\myimprovbox}{
  colframe=improvementcolor,colback=improvementcolor!5!white,fontupper=\rmfamily,
  boxrule=0.0pt,arc=2pt,boxsep=0pt,left=2pt,right=2pt,top=1.9pt,bottom=1pt,nobeforeafter,tcbox raise base,after={\hspace{2pt}}}

\newcommand{\improvement}[1]{\myimprovbox{\scriptsize{$({#1}\times$)}}}

\newcommand{\once}{{\textcolor{gray}{Once upon a time}}}

\title{Controlling Language and Diffusion Models by Transporting Activations}

\author{Pau Rodríguez \hspace{2mm}  Arno Blaas \hspace{2mm}  Michal Klein \hspace{2mm} Luca Zappella \hspace{2mm} Nicholas Apostoloff\\\textbf{Marco Cuturi \hspace{2mm}  Xavier Suau }  \\
\texttt{\{pau.rodriguez,ablaas,michal\_klein,lzappella,napostoloff,}\\\texttt{m\_cuturi,xsuaucuadros\}@apple.com} \\
Apple \\
}

\usepackage[inline]{enumitem}

\iclrfinalcopy
\begin{document}

\maketitle

\begin{abstract}
The increasing capabilities of large generative models and their ever more widespread deployment have raised concerns about their reliability, safety, and potential misuse. To address these issues, recent works have proposed to control model generation by steering model activations in order to effectively induce or prevent the emergence of concepts or behaviors in the generated output.
In this paper we introduce Activation Transport (\method), a general framework to steer activations guided by optimal transport theory that generalizes many previous activation-steering works.
\method is modality-agnostic and provides fine-grained control over the model behavior with negligible computational overhead, while minimally impacting model abilities. 
We experimentally show the effectiveness and versatility of our approach by addressing key challenges in large language models (LLMs) and text-to-image diffusion models (T2Is). For LLMs, we show that \method can effectively mitigate toxicity, induce arbitrary concepts, and increase their truthfulness. In  T2Is, we show how \method enables fine-grained style control and concept negation.

\end{abstract}

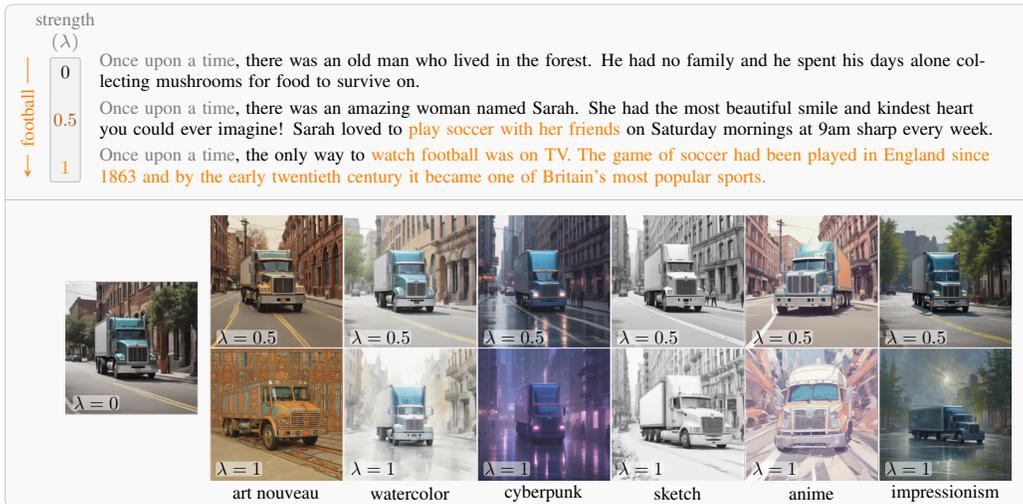
\begin{figure}[htb!]
	\centering
	\begin{subfigure}[t]{1.0\linewidth}
		\begin{tikzpicture}
			\tikzstyle{lambdablock} = [scale=0.7,text width=0.05\linewidth, node distance=3mm and 0mm, text centered];
			\tikzstyle{textblock} = [font=\scriptsize,text width=0.86\linewidth, align=left, scale=1, node distance=8mm and 0mm];

			\node[lambdablock] (LA) {$0$};
			\node[textblock, right= of LA] (A) {{\color{gray}Once upon a time}, there was an old man who lived in the forest. He had no family and he spent his days alone collecting mushrooms for food to survive on.};
			
            \node[lambdablock, text={rgb:red,30;green,15;blue,0}, below= of LA] (LB) {$0.5$};
			\node[textblock, right= of LB] (B) {{\color{gray}Once upon a time}, there was an amazing woman named Sarah. She had the most beautiful smile and kindest heart you could ever imagine! Sarah loved to {\color{orange}play soccer with her friends} on Saturday mornings at 9am sharp every week.};
			
            \node[lambdablock, orange, below= of LB] (LC) {$1$};
			\node[textblock, right= of LC] (C) {{\color{gray}Once upon a time}, the only way to {\color{orange}watch football was on TV. The game of soccer had been played in England since 1863 and by the early twentieth century it became one of Britain’s most popular sports.}};

            \pgfdeclarelayer{background2}
            \pgfdeclarelayer{background}
            \pgfdeclarelayer{foreground}
            \pgfsetlayers{background2,background,main,foreground}
            \begin{pgfonlayer}{background2}
                \draw[draw=gray!40, fill=gray!5,rounded corners=3pt] (-0.8,0.9) rectangle ++(0.98\linewidth,-6.7cm);
            \end{pgfonlayer}
            \begin{pgfonlayer}{background}
                \draw[draw=gray!50, fill=gray!10,rounded corners=1pt] (-0.2,0.19) rectangle ++(0.4,-1.65);
            \end{pgfonlayer}
            
           \node[font=\scriptsize,rotate=90,text centered, left=of LB] (football) at (0.5, -0.1) {{\color{orange}football}};
		   \draw[line width=0.1mm, orange, above=of football, left=of LB] (-0.5, 0.2) -- (-0.5, -0.15);
		   \draw[-stealth, line width=0.1mm, orange, below=of football, left=of LB] (-0.5, -1.1) -- (-0.5, -1.4);
           \node[font=\scriptsize,text centered, gray, above=of LA] (lambda) at (0, -0.55) {strength};
           \node[font=\scriptsize,text centered, gray, above=of LA] (lambda) at (0, -0.85) {$(\lambda)$};

			\draw[line width=0.1mm, gray!50] (-0.8,-1.7) -- (0.922\linewidth,-1.7);
			
			\node[anchor=north west, inner sep=0pt] (image)  at (0,-1.9)
			{\includegraphics[width=0.9\textwidth]{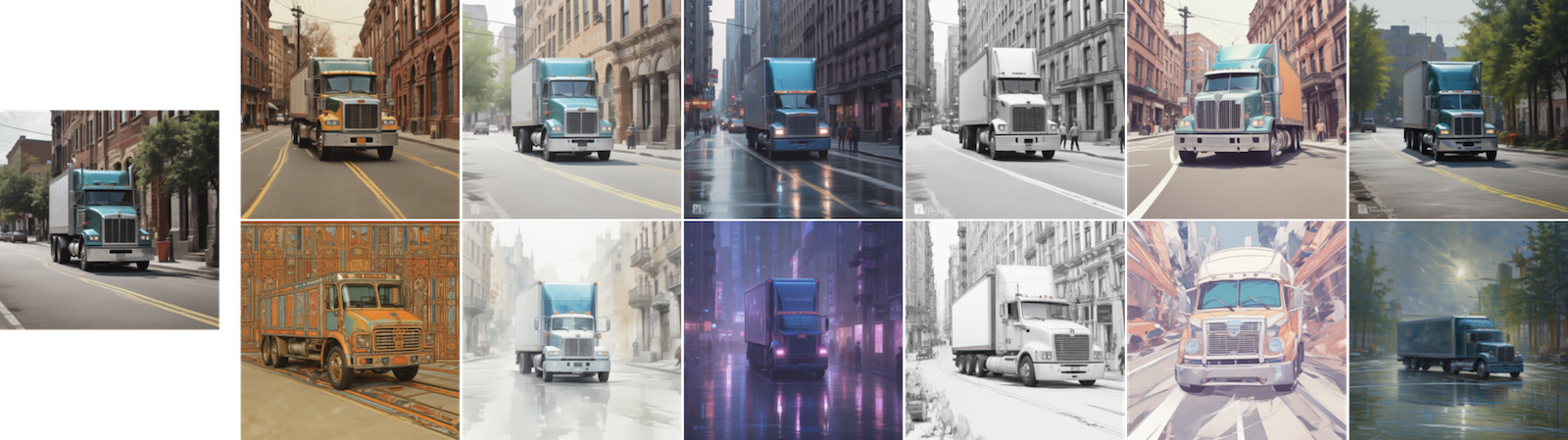}};
			
			\node[scale=0.7,rectangle,rounded corners=1pt,fill=white,text opacity=1,fill opacity=0.5,below left=-3pt,inner sep=0.5pt] at (0.65, -4.37)  {$\lambda=0$};
			    
			\foreach \i in {0,...,5}
			{
				\node[rectangle,rounded corners=1pt,fill=white,text opacity=1,fill opacity=0.5,below left=-3pt,inner sep=0.5pt,scale=0.7] at (2.75+1.78*\i, -3.50)  {$\lambda=0.5$};
			}
			
			\foreach \i in {0,...,5}
			{
				\node[rectangle,rounded corners=1pt,fill=white,text opacity=1,fill opacity=0.5,below left=-3pt,inner sep=0.5pt,scale=0.7] at (2.55+1.78*\i, -5.25)  {$\lambda=1$};
			}
			
			\foreach \i/\b in {0/art nouveau, 1/watercolor, 2/cyberpunk, 3/sketch, 4/anime, 5/impressionism}
			{
				\node[rectangle,rounded corners=1pt,fill=white,text opacity=1,fill opacity=0.5,inner sep=0.5pt,font=\scriptsize, align=center] at (2.8+1.78*\i, -5.6)  {\b};
			}
		\end{tikzpicture}
		\vskip -3mm 
	\end{subfigure}

	\vskip -0mm
	\caption{\textbf{\linear unlocks interpretable controllability for both LLMs and Diffusion}, offering explicit control over the strength of conditioning, via a parameter $\lambda$ between 0 (no transport) and 1 (full transport). 
    }
	\label{fig:fig1}
\end{figure}

\section{Introduction}
\label{sec:introduction}
Pre-trained Generative Models ({\GM}s) typically undergo an additional fine-tuning phase to better align them to a desired behavior. 
For example, Large Language Models (LLMs) are aligned via instruction fine-tuning~\citep{weifinetuned} or RLHF~\citep{ouyang2022training}. Although less extensively, these strategies have also been applied to Text-to-Image (T2I) models \citep{wallace2024diffusion,yang2024using}. However, as the number of parameters grows, alignment approaches can become challenging from a computational and memory perspective~\citep{houlsby2019parameter}. In addition, these strategies modify the model's internal mechanisms, realigning its parameters by leveraging new data, which can have the undesired side effect of impacting the utility of the model on other metrics~\citep{kotha2024understanding, luo2023empirical}, such as 0-shot evaluation or question-answering. 

The increasing cost of fine-tuning has motivated research in \ii{} on pre-trained models that offer a better understanding of features~\citep{geiger2024finding} or to control specific behaviors
~\citep{suau2022self,rimsky2023steering,zou2023representation,li2024inference}. Since these modifications are typically sparse and/or low-dimensional, they can be estimated using a few hundreds of sentences~\citep{suau2024whispering,turner2023activation}. 
For example, \citet{rimsky2023steering,li2024inference} shift activations by a constant vector estimated with sets of desired and undesired data (\eg non-toxic and toxic); or~\citet{suau2024whispering} mitigate toxicity by dampening the activations of expert neurons.
While effective, existing methods do not preserve the activation distribution observed by the model during training. Considering how brittle {\GM}s can be \citep{huu2024effects,sclar2023quantifying},  a constant shift can move activations out-of-distribution (OOD), which can lead to unwanted behaviors, and hinder both the conditioning and the general model performance. 

We propose \underline{Ac}tivation \underline{T}ransport (\method), a framework to steer activations according to the optimal transport (OT) map between two different (source and target) activation distributions, \eg toxic to non-toxic language, or between two different styles in T2I generation. \method applies a set of univariate maps on activations while preserving their target distributions, achieving better controllability and robustness to the choice of model and layers intervened upon.
Our main contributions are:
\begin{itemize}[leftmargin=.3cm,itemsep=.0cm,topsep=0cm,parsep=2pt]
    \item A unifying interpretation of existing activation steering methods under the umbrella of OT, showing that most existing methods are equivalent to a mean transport map (\Cref{subsec:other-methods-as-ot}).
    \item \linear, an inference-time intervention\footnote{\url{https://github.com/apple/ml-act}.} based on OT that preserves internal activation distributions (\Cref{subsec:estimators}). The degree of intervention can be controlled by a strength parameter $\lambda$ between 0 (no transport) and 1 (full transport), as shown in \Cref{fig:fig1}. We also introduce the \textit{transport support} to prevent inducing OOD activations.
    \item We show that, without any hyperparameter tuning, \linear matches or outperforms existing \ii{} when aiming to control LLMs for the tasks of toxicity mitigation, concept induction, and increasing truthfulness.
    \item We find that off-the-shelf \linear is also effective at controlling T2I diffusion models for the tasks of fine-grained style control and concept negation. Additionally, we adapt \citep{li2024inference} (ITI) for T2I. To the best of our knowledge, this is the first work to apply an inference-time intervention method that is simultaneously effective on both LLMs and Diffusion Models. 
    
\end{itemize}

\section{Related Work}
\label{sec:background}

The growing capabilities and prevalence of {\GM}s \citep{brown2020language, rombach2022high}, along with the rising costs of fine-tuning and alignment, have driven research into controllability of {\GM}s.

\textbf{Controlling LLMs.}\, \actadd \citep{turner2023activation} uses a contrast prompt (one positive and one negative example) to construct a shift vector. 
CAA \citep{rimsky2023steering} builds on \actadd by calculating the difference vectors for steering based on a dataset of contrast pairs (rather than a single pair), adding the mean difference during inference time for steering.  
\iti \citep{li2024inference} estimates the shift vector orthogonal to the hyperplane learnt by a binary linear classifier on two sets of sentences, showing an increase of truthfulness on the TruthfulQA benchmark~\citep{lin2021truthfulqa}. The same work proposes MassMean (\massmean), with an additive vector computed as the difference in means for both sets of sentences.
With a different approach, \aura by \citet{suau2024whispering} dampens activations proportionally to each neuron's ability to classify toxic and non-toxic sentences, effectively mitigating toxicity.
\repe by \citet{zou2023representation} proposes to compute steering vectors at inference time based on prompt pairs.
\cite{wuandarora2024reft} considers activations relationships using a low-rank projection to exchange information with a counterfactual representation and \cite{geiger2024finding} consider rotations of subsets of features.
Orthogonal to the works of activation steering, \citet{dekoninck2023controlled} have proposed a language model arithmetic that can combine the outputs of multiple models in a principled way to simulatenously control multiple concepts, however requiring several (costly) inference passes on the LLM.

\paragraph{Controlling T2I} Few works tackle aligment of T2I models. \cite{wallace2024diffusion} align diffusion models with reinforcement learning (RL) on human comparison data.  \cite{yang2024using} remove the need of a reward model to reduce computational overhead of RL. Other works focus on fine-tuning to maximize a reward function \citep{clark2023directly} or consistency to reference images \citep{lee2024direct}. The literature on T2I diffusion model controllability is more extensive and it commonly consists in training structure adapters~\citep{mou2024t2i,jiang2024scedit}, style adapters~\citep{stracke2024ctrloralter,ye2023ip,zhao2024uni}, or low-rank adapters (LoRAs)~\citep{ruiz2023dreambooth,yeh2024navigating,gandikota2023concept,stracke2024ctrloralter}. Closer to our work are \ii{}, which do not require backpropation through the model to train the conditioning mechanisms. Diffusion steering methods are a family of \ii, which directly modify the diffusion algorithm at test time for fine-grained control with additional prompts~\citep{nair2023steered,brack2022stable}. To the best of our knowledge, our work is the first to explore \ii{} that are not specific to diffusion models and transfer across modalities.

\section{Transporting Neuron Activations}
\label{sec:method}
\renewcommand{\tZ}{\mathbf{Z}}
We represent the activations of a \GM given an input sentence $\vx\in\mathcal{S}$  as a tensor $\R^{M\times L\times K}$, where $M$ is the number of activations per layer (assumed constant w.l.o.g. for simplicity), $L$ the number of layers, and $K$ the number of tokens decoded. We reduce each of the $K$ values to only one using an arbitrary pooling operator $\phi$. From now on we write $\tZ: \mathcal{S}\rightarrow \mathbb{R}^{M\times L}$ for the map that turns a sentence into a matrix of activations statistics, noting that $\tZ$ incorporates $\phi$-pooling.

We consider two probability distributions on sentences $p$ and $q$. We view these sentences through the lens of their aggregated activation matrices, \ie we will examine probability distributions $\mu:=\tZ\sharp p$ and $\nu:=\tZ\sharp q$, where we have used the pushforward operator $\sharp$.
In practice, we have access to samples $\vx^1, \dots, \vx^n\sim p$ and $\vy^1, \dots, \vy^n \sim q$. 
For instance, in the case of toxicity mitigation, $p$ covers \textit{toxic} sentences and $q$ \textit{non-toxic} ones.
Input sentences $\vx^i$ and $\vy^i$ go through the model to yield activation matrices $\va^i:=\tZ(\vx^i)$ and $\vb^i=\tZ(\vy^i)$, each seen as i.i.d. samples from $\mu$ and $\nu$ respectively, resulting in $n+n$ observations of $M\times L$ matrices. In that context, our goal is to learn a transport map $T:\mathbb{R}^{M\times L} \rightarrow \mathbb{R}^{M\times L}$ from $(\va^i, \vb^i)$ that approximately pushes $\mu$ to $\nu$, \ie $T\sharp \mu\approx \nu$.

\subsection{Low Budget Estimators for Transport Maps}
\label{subsec:estimators}
Since a modern \GM can have millions of activations, an ideal transport estimator for $T$ must be easy to learn, cheap to store in memory, and blazing fast to evaluate to avoid overheads at inference time. Additionally, because the estimation of OT maps is known to be plagued by the curse of dimensionality~\citep[Chap. 2]{chewi2024statistical}, notable care must be taken to have map estimates that generalize reasonably well. These issues are all compounded by the fact that our final method, as presented in \S\ref{subsec:sequentialmaps} builds on a composition of such OT maps (i.e. maps for a layer are estimated on samples that are themselves obtained by using maps for a previous layer). For all these fundamental reasons, we work our way from very simple map estimators, and follow \citet{suau2024whispering} to focus on maps that factorize \textit{independently} along each dimension (each activation). $T$ is therefore described as a collection of $ML$ independent univariate maps, where each map indexed by $m,l$ should ideally map the marginal distribution of $\mu$ in that coordinate to that of $\nu$. Recall that:

\begin{prop}[Univariate Transport Maps]
\label{coro:indep-transport}
    \citep[Chap.2]{santambrogio2015optimal}
    Let $\rho,\tau\in \mathcal{P}(\mathbb{R})$ be two univariate distributions. For any submodular cost $c:\mathbb{R}\times \mathbb{R}\rightarrow\mathbb{R}$ (\ie such that $\partial c/\partial x\partial y<0$), the optimal transport map $T$ that can transport  $\rho$ to $\tau$ is $T^\star=Q_\tau\circ F_\rho$, where $Q_\tau$ and $F_\rho$ are respectively the quantile function of $\tau$ and the cumulant density function (CDF) of $\rho$.
\end{prop}

Estimating and storing all $ML$ transport maps would therefore require dealing with as many quantile and CDF functions. Unfortunately, parameterizing each of these could quickly become intractable, which is why we scale down ambitions to simplify further our working hypothesis to only consider \textit{affine} transport maps.
\begin{figure}[tb]
	\centering
	\begin{minipage}[t]{0.63\linewidth}
		\centering
		\includegraphics[height=4.2cm]{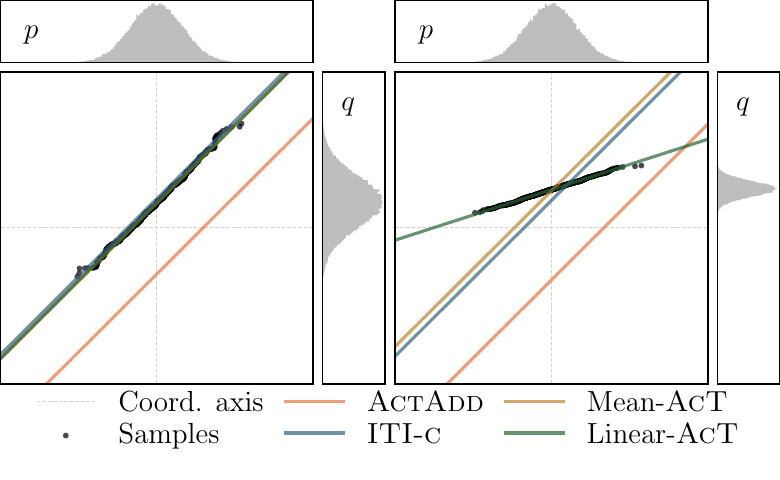}
		\vskip -3mm 
		\caption{Transport maps using different methods. For distributions with $\sigma_a = \sigma_b$ (left) all methods (except \actadd) are equivalent. When $\sigma_a \neq \sigma_b$ (right), vector-based methods (\eg \actadd, \iti, \mean) diverge from the map defined by the samples. \actadd shows a bias since it only uses one sample pair. The linear estimator is robust to differences in  $\sigma$. }
		\label{fig:ot-maps}
	\end{minipage}
	\hspace{3mm}
	\begin{minipage}[t]{0.32\linewidth}
		\centering
		\includegraphics[height=4.2cm]{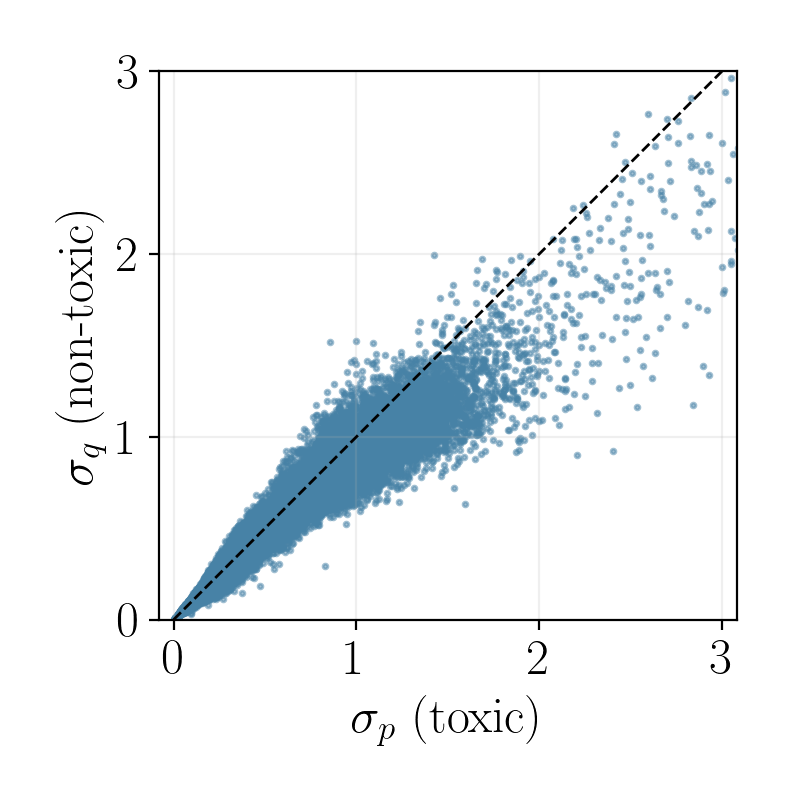}
		\vskip -3mm 
		\caption{Actual  $\sigma_a,\sigma_b$ for toxic and non-toxic sentences on \gemmatwob, showing that  {$\sigma_a \neq \sigma_b$} in real scenarios.}
		\label{fig:stds}
	\end{minipage}
\end{figure}
Each of the $ML$ activations we consider results in two families of reals: source $(a^1_{m\ell}, \dots, a^n_{m\ell})$ and targets $(b^1_{m\ell}, \dots, b^n_{m\ell})$. Simpifying notations, we drop mentions to $m$ and $\ell$ to focus on values $A:=(a^1, \dots, a^n)$ and $B:=(b^1, \dots, b^n)$ each in $\mathbb{R}^n$. We propose to consider the simple proxy task of finding \textit{affine} maps that push $A$ to $B$ efficiently. We present such an affine map, denoted \linear, in \Cref{def:linear}. Despite its simplicity, we show in \Cref{subsec:other-methods-as-ot} that many state-of-the-art methods boil down to even simpler approximations and heuristics. 

\begin{definition}[\linear]
\label{def:linear}
   Given samples  $A=(a^1, \dots, a^{n})$ and $B=(b^{1}, \dots, b^{n})$ and a cost function $c:\mathbb{R}\times \mathbb{R}\rightarrow \mathbb{R}$, the   
   \linear map trained with these samples is defined as 
   $$T(a; A,B):= \omega a +\beta,$$ where $\omega,\beta$ are the minimizers of $\min_{\omega, \beta}  \sum_i c\big(b^{(i)}, \omega a^{(i)} + \beta\big)$, and can be recovered in closed form when $c(a,b):=(a-b)^2$, as
   $$\omega = \frac{\sum_i \tilde{a}^{(i)}\tilde{b}^{(i)}}{\sum_i (\tilde{b}^{(i)})^2}, \quad\beta = m_b- \omega m_a,$$
   
   where $m_a=\tfrac1n\sum_i a^i, m_b=\tfrac1n\sum_i b^i$ are mean values, and superscripted values with $^{(i)}$ refer to \textit{sorted} values in increasing order, $(a^{(1)} \leq \dots \leq a^{(n)})$ and $(b^{(1)} \leq \dots \leq b^{(n)})$.
   Additionally, $\tilde{a}^{(i)}=a^{(i)}-m_a, \tilde{b}^{(i)}=b^{(i)}-m_b$ are sorted and recentered observations.
\end{definition}

An important feature of \linear is that it can be composed with linear layers in the \GM, resulting in no computational overhead at inference time (see \Cref{app:compute} for details).
Note that the expression in \linear should \textit{not} be confused with the closed-form known when transporting a Gaussian density with parameter ($m_a, \sigma_a)$ towards a second ($m_b, \sigma_b$), which is known \citep[Remark 2.31]{Peyre2019computational} to be $T(a)=\frac{\sigma_b}{\sigma_a}a+ (m_b - \frac{\sigma_b}{\sigma_a}m_a).$
Note that if one makes the additional assumption that $\sigma_a = \sigma_b$, then the affine Gaussian map becomes a mean shift or translation, with $T(a)=a+m_b-m_a$. We call this very simple baseline \mean and show in   \Cref{subsec:other-methods-as-ot} that several methods in the literature indeed propose versions of a mean shift strategy.

\Cref{fig:ot-maps} showcases the effect of different maps on toy data (iid, Gaussian). Note that methods based on mean-shift (\actadd, \iti, \mean) can strongly over or undershoot, mapping samples out-of-distribution. \linear shows a good trade-off between in distribution mapping and low computational budget. We note that activations in current {\GM}s show mostly unimodal distributions, but have different standard deviations for different behaviors as shown in \Cref{fig:stds}, making the linear choice a suitable one. Note that multimodal distributions would result in non-linear transport maps, which are beyond the scope of this work.

\paragraph{Transport Support}
\label{subsec:support}
The map in \Cref{def:linear} is estimated using $n$ pairs of samples. In practice, $n$ is in the order of hundreds, which results in a rough approximation of the true transport from $\mu$ to $\nu$. It is fair to assume that the transport error will be higher for input samples in the tail of $\mu$, given the scarcity of samples in that range. 
Because transporting OOD samples may lead to unexpected behavior, and to be on the conservative side, we only transport new samples that are within the osberved support $\mathcal{Q}_o = [\min A, \max A]$. 
Using the support is  important when $\mu$ is \textit{narrower} than $\nu$ (typically in a mitigation setup). Unless stated otherwise, we use $\mathcal{Q}_{o}$ for concept mitigation and $\mathcal{Q}_\infty = (-\infty, \infty)$ for induction. \Cref{app:quantile-ablation} shows an empirical validation of this choice.

\subsection{Sequential Iterative Maps}\label{subsec:sequentialmaps}
While it might be possible to follow the template approach outlined in \Cref{subsec:estimators} to apply univariate maps to each of the $ML$ activations, this ignores the causal relationship across activations, where activations produced by a layer are processed by the next one, \ie $\va_{m,\ell+1}=f_\ell(\va_{m,\ell})$. Any intervention at the level of a layer must therefore be factored in accordingly before creating the intervention at the next one. To account for such causality, we estimate the transport maps for each layer incrementally: we first estimate the transport for the first layer (in the model graph), then we run inference again by applying the first layer map in order to estimate the map for the second layer, and so on until all maps are estimated. A similar approach is adopted in \cite{zou2023representation}, and detailed with our tools in \Cref{def:causal}. In \Cref{app:causal} we show that causal estimation achieves more effective conditioning than a simultaneous estimation. In this work, we use causal estimation for \mean and \linear. 
\begin{definition}[Affine Causal Transport Map]
\label{def:causal}
    For $m\leq M$ and $\ell\leq L$, let $A_{m}:=(\va^{1}_{m,1}, \cdots, \va^{n}_{m,1})$ and $B_{m}:=(\vb^{1}_{m,1}, \cdots, \vb^{n}_{m,1})$ denote $n$ families of $M$ activations for the first layer. Starting with $\ell=1$, and setting $$C_{m,1} :=A_{m,1}, D_{m,1} := B_{m,1},$$
    compute and store the $2M$ $(\omega_m,\beta_m)$ parameters of all $M$ transport maps associated with these activations using \Cref{def:linear}:
    $$
    \forall m\leq M, \forall \ell\leq L, \quad T_{m,\ell} := T(\;\cdot\;; C_{m,\ell}, D_{m,\ell}) : \mathbb{R}\rightarrow \mathbb{R},
    $$
    where observations $C$ and $D$ are refreshed recursively for each of their entries $m\leq M$, as $\ell$ is incremented,
    $$
    \begin{aligned}
    C_{\cdot,\ell+1} &:= f_\ell([T_{m,\ell}(C_{m,\ell})]_m)\,,\\
    D_{\cdot,\ell+1} &:= f_\ell([T_{m,\ell}(D_{m,\ell})]_m)\,.
    \end{aligned}
    $$
At inference time, given a sentence $\vx$, we run the recursion starting from the first activation vector $\va=(\va_{m,1})_m$, looping for $1\leq \ell \leq L$ as $\va\leftarrow f_\ell([T_{m,\ell}(\va_m)]_m$.
\end{definition}

\textbf{Interpolation Between Measures Using Transport} %
 One can easily extend a transport map from measure $\mu$ to $\nu$ to one that is able to output an interpolating  measure. The idea, outlined by~\citet{mccann1997convexity}, consists in defining the following $\lambda$-parameterized map from any OT map $T$,
\begin{equation}
    \label{eq:interpolation}
    T(a,\lambda) = (1 - \lambda)a + \lambda T(a),
\end{equation}
where $\lambda\in[0, 1]$ and $\lambda = 1$ recovers the full transport.
Conditioning {\GM}s through OT allows the user to precisely control the presence of a concept with a continuous and interpretable \textit{knob} $(\lambda)$ during generation, not requiring expensive parameter search \citep{li2024inference} or being limited by fixed, uncontrollable conditioning  \citep{suau2024whispering}. In applications such as diffusion, where the utility of the model is harder to assess, our interpretable strength is of key importance, as shown in \Cref{sec:diffusion}. Note that methods like \actadd, CAA or \iti also have a conditioning strength parameter. However, this parameter is applied as a multiplier of a conditioning bias as  $T(a, \lambda) = a + \lambda \beta$ (see \Cref{subsec:other-methods-as-ot}), thus making $\lambda$ unbounded,  harder to interpret and not robust with respect to different models, layers, and tasks.

\subsection{Generalization of Prior Inference-Time Interventions Work}
\label{subsec:other-methods-as-ot}

In this section, we show how many earlier works can be interpreted as special cases of \linear.
\Cref{tab:ot-methods}  summarizes the intervention proposed by several recent methods, where we show that all methods propose a form of linear transport, and all of them (aside from \citet{suau2022self}) add a bias to the activations. The way this bias is pre-computed is what differentiates each method. Note that the parameter $\lambda$ typically multiplies the bias, thus becoming unbounded and non-interpretable. 

\method applies a linear transformation on activations that maximally preserves internal distributions (\Cref{subsec:estimators}, and distribution plots in~\Cref{app:distributions}). Moreover, \method  interpolates between the current and transformed activations, making $\lambda$ bounded between $[0, 1]$ and interpretable. An additional aspect is that other methods propose various heuristics to choose the support, while \method uses all activations or the observed input range ($\mathcal{Q}_o$). Note that CAA, \massmean and \mean use a difference in means. We subsume this family of methods reporting results for \mean, which has the additional advantage of an interpretable $\lambda$.
An additional difference is that many methods use the last token only (in pseudocode, $\phi(\vz) = \vz[\ldots, -1]$). Det$_{\text{zero}}$ and \aura use max-pooling ($\phi(\vz)=\vz.\texttt{max}(-1)$) while\method uses an average across tokens ($\phi(\vz)=\vz.\texttt{mean}(-1)$), which we have found to be more robust (see \Cref{ap:ablation-pooling-op}).%

\begin{table}[]
\caption{Comparison of different \ii{} in the literature. All methods listed can be expressed as a specific form of a linear map. With \method, the conditioning strength $\lambda$ interpolated between the activation $a$ and its transformed version (following \Cref{eq:interpolation}), while existing methods use $\lambda$ as a bias multiplier, thus becoming less interpretable and less robust to model/layer changes. As a result, many methods require a grid-search to find the best layer to intervene upon.}
\label{tab:ot-methods}
\centering
\resizebox{\columnwidth}{!}{%
\begin{tabular}{@{}lllll@{}}
\toprule
Method & Transport & Parameters & Support & $\phi$  \\ 
\midrule
Det$_{\text{zero}}$ \citep{suau2022self} & $\omega a + \beta$ & $\omega=0,\; \beta = m_b$ & Any layer, ${a \mid \text{AP}(A,B) > \varepsilon}$ & max \\
\actadd \citep{turner2023activation} & $\omega a + \lambda\beta$ & $\omega = 1,\; \beta = a^+ - a^-$ & Layer search & last \\
CAA \citep{rimsky2023steering} & $\omega a + \lambda\beta$  & $\omega = 1,\; \beta = m_b - m_a $ & Layer search & last \\
RePE \citep{zou2023representation} & $\omega a + \lambda\beta$ & $\omega = 1,\; \beta = a^+(\vx) - a^-(\vx)$& Layer search & last \\
\aura \citep{suau2024whispering} & $\omega a + \beta$ & $\omega = 1-Gini(A,B),\; \beta = 0$ & Any layer, ${a \mid \text{AUROC}(A,B) > 0.5}$ & max \\
EAST \citep{rahn2024controlling} & $\omega a + \lambda\beta$  & $\omega = 1,\; \beta \approx m_b  $ & Layer search & last \\
\massmean \citep{li2024inference} & $\omega a + \lambda\beta$  & $\omega = 1,\; \beta = m_b - m_a $    & Attention head search & last \\
\iti \citep{li2024inference} & $\omega a + \lambda\beta$  & $\omega = 1,\; \beta = f_{CLS}(A,B) $    & Attention head search & last \\
\midrule
\mean, \Cref{subsec:estimators} & $(1-\lambda)a + \lambda(\omega a + \beta)$  & $\omega=1,\;\beta = m_b - m_a$ & Any layer, $a\in \mathcal{Q}_o\text{ or }\mathcal{Q}_\infty$ & mean  \\
\linear, \Cref{def:linear} & $(1-\lambda)a + \lambda(\omega a + \beta)$  & $\omega,\beta = \argmin_{\omega,\beta} \sum_i(b^{(i)} - (\omega a^{(i)} + \beta))^2$ & Any layer, $a\in \mathcal{Q}_o\text{ or }\mathcal{Q}_\infty$ & mean  \\
\bottomrule
\end{tabular}
}
\vskip -3mm
\end{table}

\section{Experiments on LLMs}
\label{sec:experimentsLLMs}

We empirically verify the performance of \method on pre-trained LLMs on toxicity mitigation (\Cref{subsec:toxicity}), general concept induction (\Cref{sec:induce-results}), and truthfulness induction in particular (\Cref{sec:truthfulness}), showing the efficacy and robustness of \method in different scenarios related to LLMs. 

\subsection{Toxicity Mitigation in LLMs}
\label{subsec:toxicity}

It is known that LLMs are prone to generate toxic language~\citep{wen-etal-2023-unveiling}, especially when prompts are designed to elicit toxic behavior. In this section, we study how \method is effective at toxic language mitigation compared to some recents methods such as \aura, \actadd and \iti, on \gemmatwob \citep{team2024gemma} and \llamaeightb \cite{dubey2024llama}. 
To do so, we prompt each LLM with 1000 randomly chosen prompts from RealToxicityPrompts (RTP) \citep{gehman2020realtoxicityprompts},known to induce toxic language generation. Then, we collect the generated continuation to each prompt and we evaluate toxicity with a ROBERTA-based classifier\footnote{\url{https://huggingface.co/s-nlp/roberta_toxicity_classifier}}, as in \citet{suau2024whispering}. In addition, we also measure toxicity in a 0-shot manner by querying \llamaeightbinstr as LLM-as-a-judge \citep{zheng2023judging} (more details on %
\Cref{app:0-shot-toxicity}). As a measure of general LLM utility we report in \Cref{tab:toxicity-summary}: (i) perplexity (PPL) on a fixed set of 20k Wikipedia sentences measured with the intervened model, (ii) PPL of the generated sentences measured with \mistralsevenb \citep{jiang2023mistral} and (iii) MMLU \citep{hendryckstest2021} 5-shot accuracy using the intervened model. Besides, we report generation diversity results in~\Cref{app:diversity}.

\begin{table}[htb!]
\caption{Toxicity mitigation for \gemmatwob and \llamaeightb, results over 5 runs. We intervene upon different layer types (layer column) and show the best layer per method. \iti, \actadd and \method have a \textit{strength} parameter $\lambda$ which we sweep. For each method, we report results for the $\lambda$ that attained the best CLS toxicity that incurs less than $+1$ increase in PPL Wikipedia. \method methods and provide best results for $\lambda=1$, achieving up to $7.5\times$ (\gemmatwob) and $4.3\times$ (\llamaeightb) CLS toxicity mitigation with \linear. \iti is very sensitive to $\lambda$ as well as layer choice (see full results in \Cref{app:toxicity}), and \aura reaches up to $3.1\times$ reduction.}
\label{tab:toxicity-summary}
\centering
\resizebox{1.0\columnwidth}{!}{%
 \begin{NiceTabular}{cllr|ll|lll}
\toprule
 & & Layer & Best $\lambda$ & CLS Tox. (\%) $\downarrow$ & 0-shot Tox. (\%) $\downarrow$ & PPL Wikipedia $\downarrow$ & PPL Mistral-7B $\downarrow$ & MMLU $\uparrow$  \\
 \midrule
\Block{6-1}<\rotate>{\gemmatwob} & Original & - & - & $4.17 \pm {0.32}$ & $13.42 \pm {1.08}$ & 13.98 & 6.68 & 53.1  \\
\cmidrule{2-9}
& \actadd & MLP & 0.5 & $3.96 \pm {0.24}$ \improvement{1.1} & $13.43 \pm {1.42}$ & 14.69 {\footnotesize (+0.72)} & 6.67 {\footnotesize (+0.05)} & 53.0 {\footnotesize (-0.1)}  \\
& \aura & MLP & - & $2.12 \pm {0.27}$ \improvement{2.0} & $9.04 \pm {0.66}$ & 14.18 {\footnotesize (+0.21)} & 7.04 {\footnotesize (+0.36)} & 53.0 {\footnotesize (-0.1)}  \\
& \iti & Attention & 8.0 & ${0.74} \pm {0.18}$ \improvement{5.6} & $5.36 \pm {0.91}$ & 14.90 {\footnotesize (+0.92)} & 7.44 {\footnotesize (+0.76)} & 52.6 {\footnotesize (-0.5)}  \\
& \mean & Post-LN & 1.0 & $\mathbf{0.54} \pm {0.44}$ \improvement{7.7} & $\mathbf{4.10} \pm {0.41}$ & 14.21 {\footnotesize (+0.23)} & 7.59 {\footnotesize (+0.90)} & 51.6 {\footnotesize (-1.5)}  \\
& \linear & Post-LN & 1.0 & $\underline{0.56} \pm {0.21}$ \improvement{7.5} & $\underline{4.14} \pm {0.55}$ & 14.79 {\footnotesize (+0.81)} & 7.99 {\footnotesize (+1.31)} & 51.3 {\footnotesize (-1.8)}  \\
\midrule
\Block{6-1}<\rotate>{\llamaeightb} & Original & - & - & $5.80$ & $15.00$ & 9.06 & 5.68 & 65.3  \\
\cmidrule{2-9}
& \actadd & Attention & 0.3 & $5.57 \pm {0.45}$ \improvement{1.0} & $15.73 \pm {0.21}$ & 9.71 {\footnotesize (+0.65)} & 5.85 {\footnotesize (+0.16)} & 65.5 {\footnotesize (+0.2)}  \\
& \aura & MLP & - & ${1.90} \pm {0.61}$ \improvement{3.1} & ${8.12} \pm {0.85}$ & 9.52 {\footnotesize (+0.45)} & 6.05 {\footnotesize (+0.37)} & 65.5 {\footnotesize (+0.2)}  \\
& \iti & Attention & 3.0 & $1.60 \pm {0.22}$ \improvement{3.6} & $\underline{6.53} \pm {0.66}$  & 9.48 {\footnotesize (+0.42)} & 6.17 {\footnotesize (+0.49)} & 64.7 {\footnotesize (-0.6)}  \\
& \mean & Attention & 1.0 & $\underline{1.38} \pm {0.17}$ \improvement{4.2} & $\mathbf{5.60} \pm {0.34}$ & 9.56 {\footnotesize (+0.49)} & 6.36 {\footnotesize (+0.68)} & 64.7 {\footnotesize (-0.7)}  \\
& \linear & Attention & 1.0 & $\mathbf{1.35} \pm {0.39}$ \improvement{4.3} & $6.68 \pm {0.81}$ & 9.56 {\footnotesize (+0.49)} & 6.28 {\footnotesize (+0.60)} & 64.5 {\footnotesize (-0.8)}   \\
\bottomrule
\end{NiceTabular}
}
\end{table}

\takeaway{\linear reduces toxicity up to $7.5\times$ and is robust to $\lambda$, layer, and model choice} We observe that \linear achieves up to $7.5\times$ reduction in toxicity on \gemmatwob and $4.3\times$ on \llamaeightb, with minimal impact on PPL and MMLU. Most importantly, \method obtains the best results at $\lambda=1$, which is in line with our \ot formulation, since $\lambda=1$ means full transport. \linear and \mean obtain similar toxicity mitigation results. \iti achieves $5.6\times$ and $3.6\times$ toxicity reduction on \gemmatwob and \llamaeightb respectively. In line with the \iti paper findings, \iti performs well on attention, but is very sensitive to models and layers, as well as to the choice of $\lambda$ 
(see a layer diagram in \Cref{app:layers} and full tables and plots in \Cref{app:toxicity}). \aura achieves $2.0\times$ and $3.1\times$ toxicity reduction per model and \actadd induces the mildest mitigation.

\subsection{Inducing Concepts in LLMs with \method}
\label{sec:induce-results}

\begin{figure}[htb!]
     \centering
     \begin{subfigure}[t]{0.49\linewidth}
         \centering
            \includegraphics[width=1.0\linewidth]{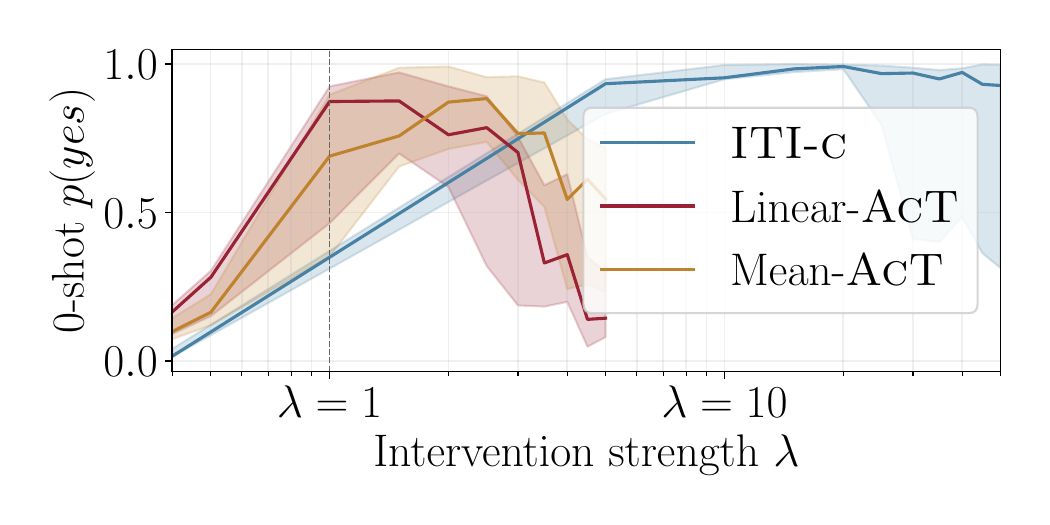}
           \vskip -3mm 
      \end{subfigure}
      \begin{subfigure}[t]{0.49\linewidth}
         \centering
            \includegraphics[width=1.0\linewidth]{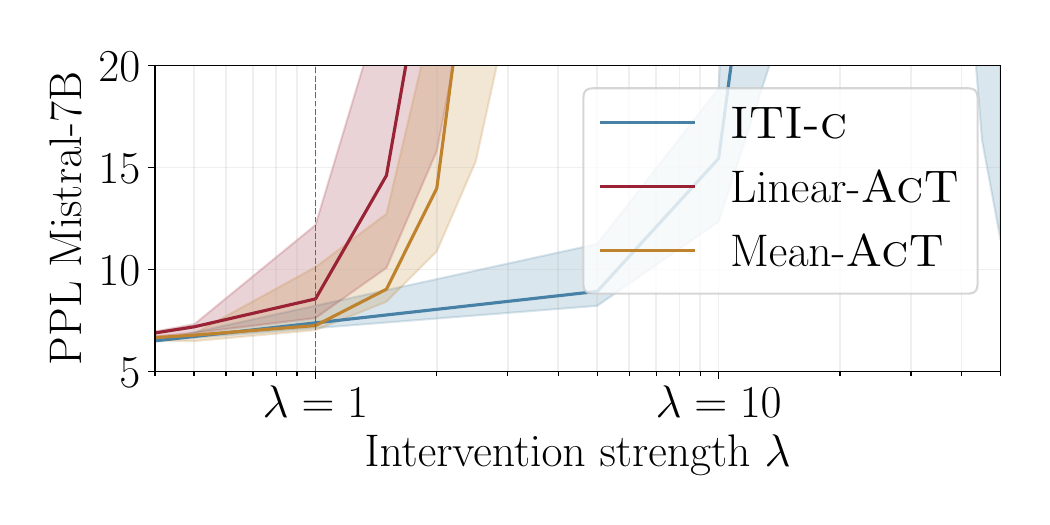}
           \vskip -3mm 
      \end{subfigure}
      
\vskip -4mm
\caption{Concept induction using \method (post-LN layers) and \iti (attention layers) on \gemmatwob. We aggregate results over 7 WordNet concepts, generating 500 sentences at different intervention strength levels. We report concept presence with LLM-as-a-judge ($p(yes)$), and the PPL of the generated sentences using \mistralsevenb. We plot the median (and 25/75 quantile band) across concepts and generations per level, showing that \linear achieves a peak of concept induction at $\lambda\approx1$, which is inline with our OT formulation. Other methods show different maxima.}
\label{fig:beyond-lambda}
\end{figure}

\method allows transporting activations from distribution $\mu$ to $\nu$ (derived from sentence distributions $p$ and $q$ respectively). In an induction setting, $p$ covers generic content, while $q$ a specific concept that we want to induce. We mine the OneSec dataset \citep{scarliniOneSec}, collecting 700 sentences that contain a specific concept ($q$) and 700 sentences randomly sampled from other concepts ($p$). We do so for seven different concepts (\textit{football, cloud, baby, church, book, flower, balloon}) and we estimate an intervention for each of them. We assess the presence of a concept in the generated text in a LLM-as-a-judge manner by querying \llamaeightbinstr (LLM-as-a-judge details in \Cref{app:0-shot-concept}).

\takeaway{\linear can induce arbitrary concepts with consistent $\lambda=1$} \Cref{fig:beyond-lambda} shows the effect of increasing $\lambda$ both on the presence of the concept, $p(yes)$, and the PPL measured with \mistralsevenb on the generated text. We intervene upon the most effective layers for each method according to the toxicity results: attention for \iti, and Post-LN for \method. In general, we found that LN layers were the most suited for \method, across models and tasks. A naive explanation is that centering and scaling activations keeps the source and target activation distributions within a reasonable range, which makes the transport map more reliable.
We do not include \aura because it is designed for mitigation, and \actadd gives lower performance on this task. For \linear, we observe a peak of concept presence at $\lambda\approx1$, with a median $p(yes)=0.87$ (\ie 87\% of the generated sentences are classified as containing the induced concept) and an acceptable $\text{PPL}=8.5$. For $\lambda>1$, the PPL quickly degrades and the presence of the concept diminishes. 
This is also consistent with the toxicity mitigation experiments in \Cref{subsec:toxicity}. Interestingly, the peak for \mean is at $\lambda\approx2.5$, also highlighting that \mean is a poorer approximation of the OT transport. 
Notably, \iti achieves a similar $p(yes)$ and PPL as \linear for $\lambda\approx 5$. However, note that \iti's best $\lambda$ is different than the ones for toxicity. 
\Cref{app:concept-induction} contains generation examples.

\begin{table}[t!]
\caption{TruthfulQA results for \gemmatwob and \llamaeightb, results over 5 runs. We intervene upon different layers (layer column) and show the best per model. \iti, \actadd and \method have a \textit{strength} parameter $\lambda$ which we sweep, reporting the best $\lambda$ result per model (MC1 Accuracy so that MMLU is within the best \method MMLU $\pm \;0.1$).%
}
\label{tab:tqa}
\centering
\resizebox{0.9\columnwidth}{!}{%
 \begin{NiceTabular}{cllr|ll|l}
\toprule
 &  & Layer & Best $\lambda$ & MC1 Accuracy (\%) $\uparrow$ & MC2 Accuracy (\%) $\uparrow$ & MMLU Accuracy (\%) $\uparrow$ \\
\midrule
\Block{6-1}<\rotate>{\gemmatwob} & Original & - & - &  $21.05 $ & $32.80 $ & $53.10$ \\
\cmidrule{2-7}
 & \actadd & MLP & 3.0 & $23.01 \pm {0.00}$ {\footnotesize ($+1.96$)} & $34.76 \pm {0.00}$ {\footnotesize ($+1.96$)}  & $52.83 \pm {0.00}$ {\footnotesize ($-0.27$)} \\
 & \aura & MLP & - & $21.20 \pm {0.10}$ {\footnotesize ($+0.15$)} & $32.88 \pm {0.22}$ {\footnotesize ($+0.08$)} & $52.73 \pm {0.07}$ {\footnotesize ($-0.37$)} \\
 & \iti & MLP & 2.0 & $24.53 \pm {0.11}$ {\footnotesize ($+3.48$)} & $37.06 \pm {0.38}$ {\footnotesize ($+4.26$)} & $51.39 \pm {0.41}$  {\footnotesize ($-1.71$)}
\\
 & \mean & All-LN & 1.0 &  $\underline{25.07} \pm {0.20}$ {\footnotesize ($+4.02$)} & $\underline{38.68} \pm {0.30}$ {\footnotesize ($+5.88$)} & $51.81 \pm {0.12}$ {\footnotesize ($-1.29$)} \\
 & \linear & All-LN & 1.0  &  $\textbf{26.00} \pm {0.32}$ {\footnotesize ($+4.95$)} & $\textbf{40.17} \pm {0.24}$ {\footnotesize ($+7.37$)} & $51.47 \pm {0.27}$  {\footnotesize ($-1.63$)}\\
\midrule
\Block{6-1}<\rotate>{\llamaeightb} & Original & - & - &  $25.46 $ & $40.27 $ & $65.35$ \\
\cmidrule{2-7}
& \actadd & Attention & 0.7 & $26.19 \pm {0.00}$ {\footnotesize ($+0.73$)} & $40.88 \pm {0.00}$ {\footnotesize ($+0.61$)} & $65.42 \pm {0.00}$ {\footnotesize ($+0.07$)} \\
& \aura & MLP & - & $25.34 \pm {0.15}$ {\footnotesize ($-0.12$)} & $40.47 \pm {0.20}$ {\footnotesize ($+0.20$)} & $65.37 \pm {0.06}$ {\footnotesize ($+0.02$)}\\
& \iti & MLP & 2.0 & $30.11 \pm {0.60}$ {\footnotesize ($+4.65$)} & $45.41 \pm {0.24}$ {\footnotesize ($+5.14$)} & $64.71 \pm {0.14}$ {\footnotesize ($-0.64$)}\\
& \mean & All-LN & 1.0 &  $\underline{32.88} \pm {0.54}$  {\footnotesize ($+7.42$)} & $\underline{48.23} \pm {0.64}$ {\footnotesize ($+7.96$)} & $64.83 \pm {0.14}$ {\footnotesize ($-0.52$)}\\
& \linear & All-LN & 1.0  &  $\textbf{33.22} \pm {0.22}$  {\footnotesize ($+7.76$)} & $\textbf{48.69} \pm {0.34}$ {\footnotesize ($+8.42$)} & $64.78 \pm {0.15}$ {\footnotesize ($-0.57$)}\\
\bottomrule
\end{NiceTabular}
}
\end{table}

\subsection{Inducing truthfulness in LLMs with \method} 
\label{sec:truthfulness}

One particular concept that has gained attention in previous activation steering works is ``truthfulness''~\citep{li2024inference}.
We study how \method can increase truthfulness on \gemmatwob and \llamaeightb, compared to the original model. Again, we compare to \aura, \actadd and \iti. We evaluate all methods on the TruthfulQA multiple choice part that has been used in prior work \citep{lin2021truthfulqa, li2024inference}. We report both MC1 and MC2 of TruthfulQA, and control for overfitting on the TruthfulQA task by also evaluating MMLU 5-shot accuracy \citep{hendryckstest2021}.  

\takeaway{\method can induce truthfulness with consistent $\lambda=1$.} The results of our experiments are summarized in Table \ref{tab:tqa}. As we can see, \method can successfully induce truthfulness in both models in its default setting $\lambda = 1$ (corresponding to full transport). Both \linear and \mean achieve the best and second-best MC1 and MC2 accuracy improvements among all methods investigated. \linear increases MC1 by roughly $5\%$ for \gemmatwob and by almost $8 \%$ for \llamaeightb, which is about $1.5 \%$ and $3 \%$ more than the closest non-{\method} baseline (\iti), while incurring even slightly less decrease in MMLU performance. 
Full results and experimental setup in \Cref{app:tqa}.

\section{Controlling Image Diffusion Models}
\label{sec:diffusion}

In this section, we show that \method improves the controllability of text-to-image diffusion models (T2Is), a well-known challenge~\citep{cao2024controllable}. We address two open problems in T2I generation: fine-grained style control (\Cref{subsec:style-control}) and concept negation (\Cref{subsec:concept-negation}). We show that off-the-shelf \method succeeds at both tasks. In line with optimal transport theory and experimental results on LLMs~(\Cref{sec:experimentsLLMs}), \method consistently achieves the strongest conditioning with full strength (\ie $\lambda=1$). We also adapt \iti to the topology of images by training it on the spatial average pooling of activations (as we do by default for \method), and applying it to each spatial position independently. Remarkably, \iti succeeds at the fine-grained control task with our adaptation, but requires tuning $\lambda$, and it fails with concept negation.

\paragraph{Setup.} We apply \method on the denoising convolutional UNet of Stable Diffusion XL (SDXL)~\citep{podellsdxl} and the denoising transformer of FLUX.1.Schnell\footnote{\url{https://blackforestlabs.ai/announcing-black-forest-labs/}}. For FLUX, we use the T5-XXL text encoding modality~\citep{raffel2020exploring} instead of CLIP~\citep{radford2017learning} to account for the effects of language modelling. We use a distilled version of SDXL, which only requires 4 diffusion steps~\citep{lin2024sdxl} like FLUX. We intervene upon all normalization layers in SDXL's UNET and the output of most residual layers in FLUX (details in \Cref{app:flux-details}). We only show results for \method and \iti since \actadd is not applicable to images and \aura resulted in noisy images. To measure the presence of a style or a concept, we use a CLIP zero-shot classifier with the classes (+) ``A picture of a \texttt{\{style or concept\}}'' and (-) ``A picture of \texttt{something}''. We also track whether the content from the original prompt (with no style or concept modifiers) is preserved using the CLIPScore (cosine similarity of CLIP embeddings, \citet{hessel2021clipscore}) between the images generated after the intervention and the original prompt. 

\begin{figure}
    \begin{tikzpicture}
			\tikzstyle{lambdablock} = [rectangle,font=\scriptsize,rounded corners=1pt,fill=white,text opacity=1,fill opacity=0.7,inner sep=0.5pt,scale=0.7];

			\node[anchor=north west, inner sep=0pt] (img1)  at (0, 0)
			{
                    \includegraphics[width=0.49\textwidth,trim={7mm 7mm 0 0},clip]{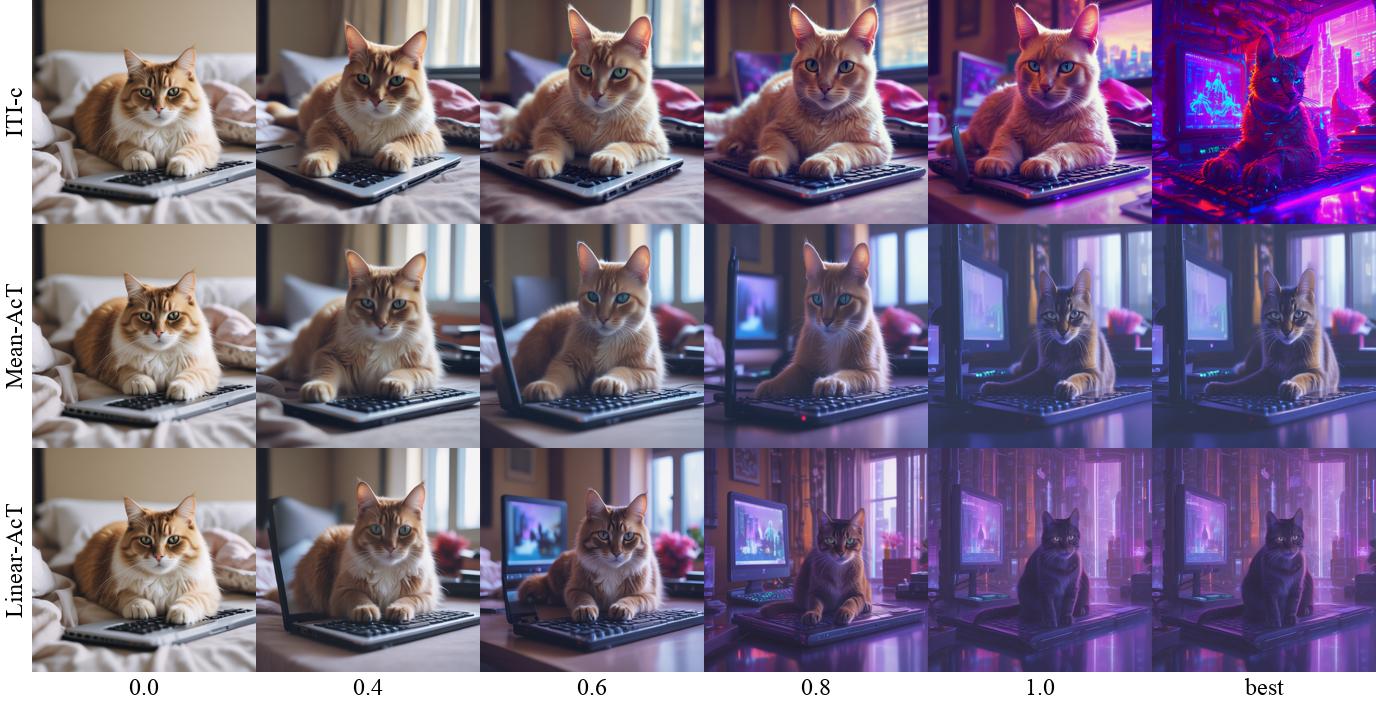}
            };
   		    \node[anchor=north west, inner sep=0pt, right= 1mm of img1] (img2)
			{
                    \includegraphics[width=0.49\textwidth,trim={7mm 7mm 0 0},clip]{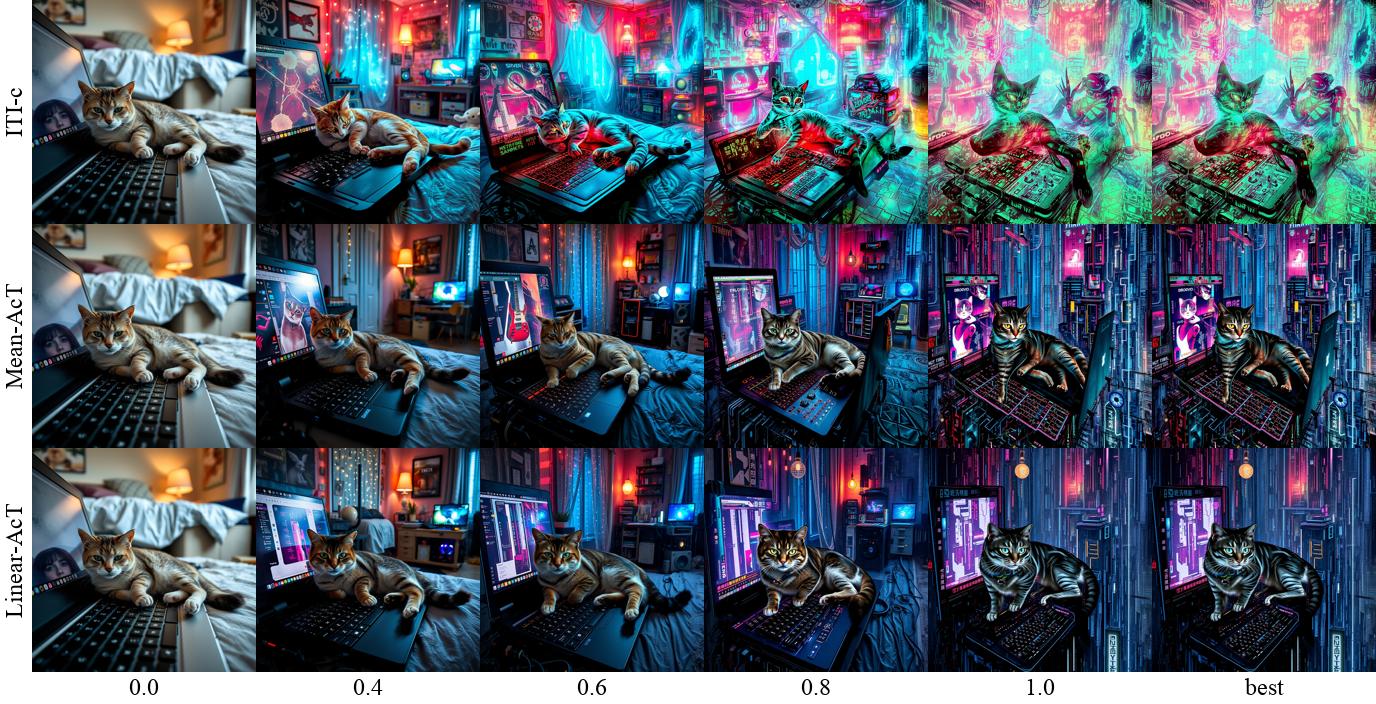}
            };

            \foreach \v/\i in {0.0/0, 0.4/1, 0.6/2, 0.8/3, 1.0/4, {2.0\text{ best}}/5}
			{
				\node[lambdablock, below=-1.8mm of img1, align=left, anchor=north west] at (0.0816\textwidth*\i + 2.0, -1.15)  {$\v$};
			}    
            
            \foreach \v/\i in {0.0/0, 0.4/1, 0.6/2, 0.8/3, 1.0/4, {1.0\text{ best}}/5}
			{
				\node[lambdablock, below=-1.8mm of img1, align=left, anchor=north west] at (0.0816\textwidth*\i + 2.0, -2.28)  {$\v$};
			}    

            \foreach \v/\i in {0.0/0, 0.4/1, 0.6/2, 0.8/3, 1.0/4, {1.0\text{ best}}/5}
			{
				\node[lambdablock, below=-1.8mm of img1, align=left, anchor=north west] at (0.0816\textwidth*\i + 2.0, -3.41)  {$\v$};
			}

            \foreach \v/\i in {0.0/0, 0.4/1, 0.6/2, 0.8/3, 1.0/4, {1.0\text{ best}}/5}
			{
				\node[lambdablock, below=-1.8mm of img2, align=left, anchor=north west] at (0.0816\textwidth*\i + 5.0 + 0.49\linewidth, -1.15)  {$\v$};
			}    
            
            \foreach \v/\i in {0.0/0, 0.4/1, 0.6/2, 0.8/3, 1.0/4, {1.0\text{ best}}/5}
			{
				\node[lambdablock, below=-1.8mm of img2, align=left, anchor=north west] at (0.0816\textwidth*\i + 5.0 + 0.49\linewidth, -2.28)  {$\v$};
			}    

            \foreach \v/\i in {0.0/0, 0.4/1, 0.6/2, 0.8/3, 1.0/4, {1.0\text{ best}}/5}
			{
				\node[lambdablock, below=-1.8mm of img2, align=left, anchor=north west] at (0.0816\textwidth*\i + 5.0 + 0.49\linewidth, -3.41)  {$\v$};
			}  

            \draw[line width=0.3mm, white] (0.408\linewidth,0) -- (0.408\linewidth,-3.5cm);
            \draw[line width=0.3mm, white] (0.408\linewidth+0.49\linewidth + 4,0) -- (0.408\linewidth+0.49\linewidth + 4,-3.5cm);

		\end{tikzpicture}
  \vskip -2mm
    \caption{\textbf{\linear allows controlled conditioning of SDXL and FLUX.} ``A cat resting on a laptop keyboard in a bedroom.'' SDXL (left) and FLUX (right) intervened with \iti (top), \mean (middle) and \linear (bottom) for the concept \textit{cyberpunk}, with a $\lambda$ strength in $[0,1]$. The image with the best $\lambda$ (according to the highest 0-shot score in~\Cref{fig:clip_score}) is shown right. Qualitatively, \linear balances better a \textit{cyberpunk} style increase with prompt semantics preservation.}
    \label{fig:increase_diffusion}
\end{figure}

\subsection{Style control}
\label{subsec:style-control}
A major challenge in T2I generation is fine-grained control. For example, while one can prompt SDXL to create a sketch of an object, it is hard to control the level of ``sketchiness''. Models such as SDXL have a guidance parameter, but its use is limited since low guidance values tend to remove image semantics (see example in \Cref{app:guidance}).
To showcase the ability of \method to achieve such a fine-grained control, we sample 2048 prompts from the COCO Captions~\citep{chen2015microsoft} training set and append a series of tags generated with Llama-8B-instruct to induce the following styles: \textit{anime, art nouveau, cyberpunk, impressionism, sketch, watercolor} (see \Cref{app:style-tags} for details). Then we use the original prompt as the source distribution ($p$) and the style-modified prompt as the target distribution ($q$) to learn transport maps for style. To evaluate, we sample 512 prompts from the COCO Captions validation set and generate images with different intervention strengths.

\begin{figure}[t]
    \centering
     \begin{subfigure}[t]{0.45\linewidth}
         \centering
            \includegraphics[width=\linewidth]{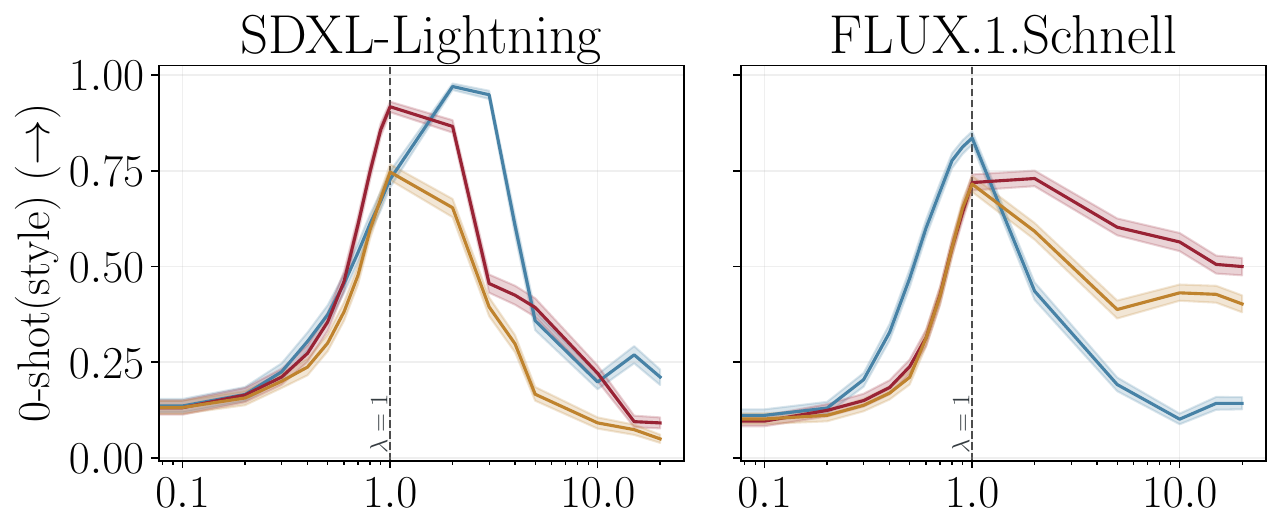}
            
            \includegraphics[width=\linewidth]{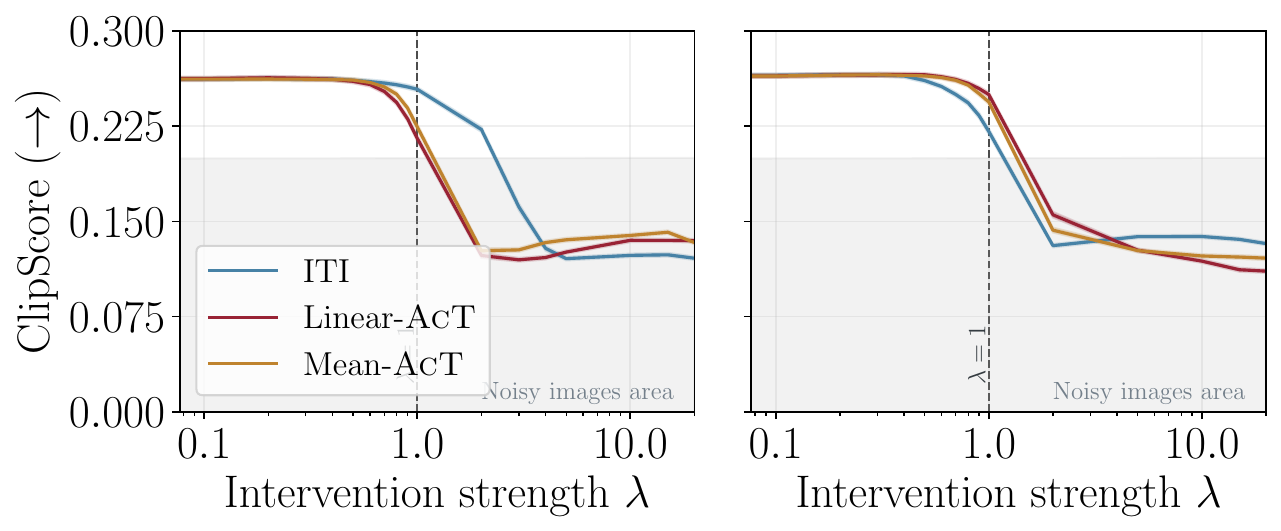}
            \caption{Style control}
            \label{fig:clip_score_style}
      \end{subfigure}
      \begin{subfigure}[t]{0.45\linewidth}
         \centering
            \includegraphics[width=\linewidth]{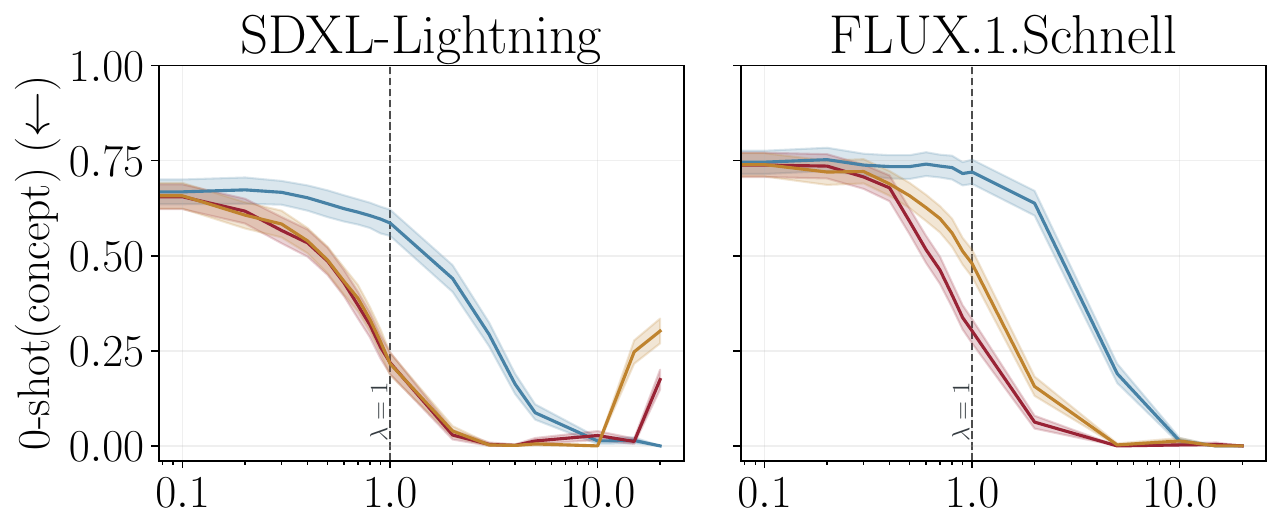}
            
            \includegraphics[width=\linewidth]{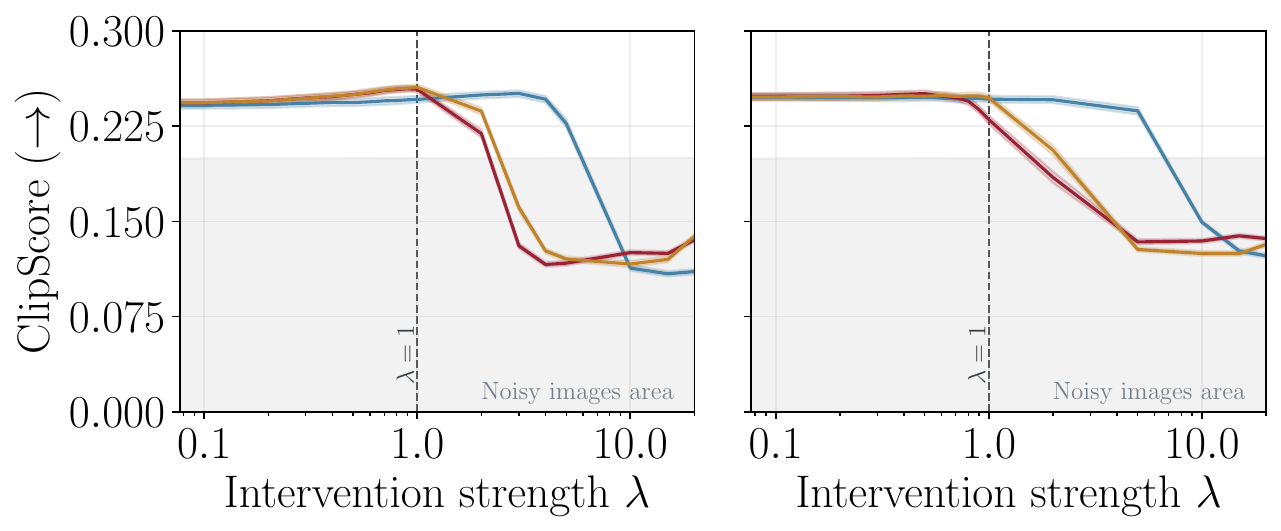}
            \caption{Concept Negation}
            \label{fig:clip_score_concept}
      \end{subfigure}
      \vskip -2mm
    \caption{Style control (a) and concept negation (b) on SDXL and FLUX. Top row shows the fraction of generated images classified (CLIP 0-shot) as containing a given concept or style. Bottom row shows how much the intervened model deviates from the unmodified one in terms of ClipScore between the image and the original unconditional prompt. Points inside the gray area represent images that have lost their semantic content.}
    \label{fig:clip_score}
\end{figure}

\takeaway{\linear is a robust method for fine-grained control in text-to-image generation.} \Cref{fig:clip_score_style} shows that \linear on SDXL and FLUX  increases the presence of a desired style, \eg on SDXL from $\sim12\%$ to $\sim95\%$ of the generated images while keeping $\sim 80\%$ of the similarity to the original prompt ($\lambda = 1)$. In accordance to the theory and experiments on LLMs, the maximum conditioning (\ie highest 0-shot score) for \method is achieved at $\lambda = 1$ for both models. \iti  can also accomplish fine-grained control, but its best performance is achieved at different $\lambda$s, equal to 2 and 1 for SDXL and FLUX respectively, which is in turn not consistent with the best $\lambda$ found in LLM experiments. A closer look at images generated with \iti for best $\lambda$ in \Cref{fig:increase_diffusion,app:style-control} reveals that ITI tends to exaggerate style traits while distorting the semantics. This further highlights the reliability of \method across different modalities, tasks, and models. While quantitatively \method and \iti perform well, we invite the reader to compare the quality of the generated images and styles in \Cref{fig:fig1,fig:increase_diffusion}, and in more examples in \Cref{app:style-control}.

\begin{figure}[t]
    \centering
    \begin{subfigure}[t]{\textwidth}
        \centering
        \includegraphics[width=\linewidth]{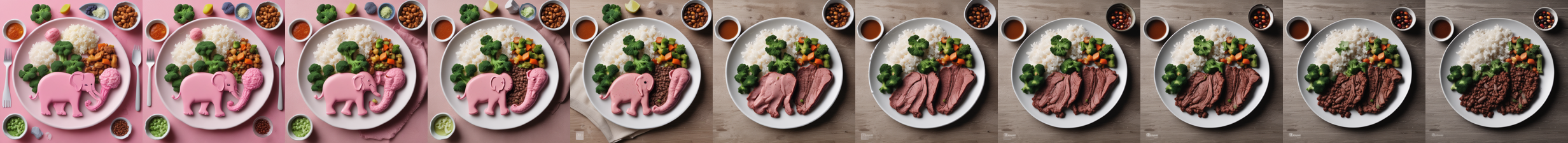}
    \end{subfigure}
    \vskip -1mm
    \caption{
    \textbf{Concept Negation} for ``A plate of food with rice and beans, broccoli and meat.  And a pink elephant is missing.''. (a) \linear on SDXL with transport strength $\lambda$ linearly increasing from 0 to 1. Note how the presence of the pink elephant is prominent for the original model (leftmost image) and gradually disappears as $\lambda$ increases. 
    }
    \label{fig:concept_negation}
    \vskip -6mm
\end{figure}

\subsection{Concept Negation}
\label{subsec:concept-negation}
T2I diffusion models are known for struggling with concept negation~\citep{liget,hwang2024not}. For example, \citet{hwang2024not} showed that recent models such as Stable Diffusion~\citep{rombach2022high} and DALL-E 3~\citep{betker2023improving} are prone to generate a pink elephant when instructed not to generate one. To improve controllability, some T2I generators like SDXL include a \textit{negative prompt} mechanism to remove concepts from the generated images. 
However, we found that both SDXL (CLIP encoder + negative prompt) and FLUX (T5-XXL encoder) still tend to generate unwanted concepts (see some examples in \Cref{app:negative-prompts}).

We use the COCO Captions~\citep{chen2015microsoft} training set to sample 2048 prompts used to generate the images. To create a source and target activation distribution to estimate \method,  we ask \llamaeightbinstr to generate a diverse set of prompt modifiers requiring the model to include the following concepts: \textit{pink elephant, white bear,} and \textit{gorilla}. The exact phrasing of the modifiers is provided in \Cref{app:concept-tags}. 
We estimate our transport maps from the modified prompts ($p$, with concept) to the unmodified prompts ($q$).
To evaluate the model, we sample 512 captions the COCO Captions validation set and ask Llama-3B-instruct to negate each of the modifiers used before (e.g.,  ``without a pink elephant'', ``a gorilla cannot be seen anywhere'') to generate images with unintended concept spillage such as the leftmost image in \Cref{fig:concept_negation} or the examples in \Cref{fig:negative-prompt-sdxl,fig:negative-prompt-SD3}. 

\takeaway{\linear is a robust method for concept negation in text-to-image generation.} In \Cref{fig:clip_score_concept}, we observe that \method is more effective at concept negation than \iti while better preserving the original semantics of the image, as indicated by the drop in 0-shot concept score for higher CLIPScore than \iti . ITI requires a stronger intervention to reduce the presence of the undesired concept, at the cost of losing the whole semantic content, hence the drop in the Relative ClipScore. Additional examples and images for each concept can be found in \Cref{app:concept-negation}.

\section{Limitations and Discussion}
\label{sec:discussion}

In this work, we introduce Activation Transport (\method), a general framework to achieve intuitive and fine-grained control of {\GM}s. Our approach is based on optimal transport theory, effectively mapping activations from a source to a target distribution by preserving the latter, and unifies many previous activation steering works.
We show experimentally that our \linear  approach generalizes well across models and tasks, for both LLMs and T2I architectures. Moreover, \method provides a robust parameter to control the amount of conditioning, bounded between 0 and 1, which makes it user-friendly and interpretable.
While effective, \linear assumes a linear transport between i.i.d. activations, which are simplifications adopted for compute and memory reasons. Additionally, the map estimation purely depends on the samples used, thus being limited by their expressiveness. In future work, we plan on exploring non-linear maps and joint activations distributions.

\clearpage
\section*{Ethics Statement}
Our method could theoretically be used to mitigate or induce the presence of any concept. Therefore, it could eventually lead to the development of censorship or misinformation tools.

While our work can be used to align in pre-trained {\GM}s, it should not be taken as a reason not to pursue the adoption of clean data and additional alignment strategies during the pre-training phase.

\section*{Reproducibility Statement}
We will make our code and data publicly available on github. To aid reproducibility, all tables contain the best $\lambda$ found through grid-search and results are averaged over 5 runs. We include additional details on the intervened layers in~\Cref{app:layers}, ablations on the effect of transport support in~\Cref{app:quantile-ablation}, pooling operation ablations in~\Cref{ap:ablation-pooling-op}, the exact prompt templates of LLM as a judge in~\Cref{app:0-shot-toxicity,app:0-shot-concept}, experimental details on TruthfulQA in~\Cref{app:tqa}, as well as experimental details for T2I models in~\Cref{app:t2i}.

\section*{Acknowledgements} 
We thank Miguel A. Bautista, Federico Danieli, Gerard Gállego, Yu-Guan Hsieh, Miguel Sarabia, Federico Scozzafava, and Barry Theobald (in alphabetical order) for their helpful feedback and critical discussions throughout the process of writing this paper. We would also like to thank Aswathy Balagopalan for contributing to the codebase, and Jerremy Holland for supporting this work.

\bibliography{biblio}
\bibliographystyle{iclr2025_conference}

\newpage
\appendix

\FloatBarrier
\section{Memory and Computational Aspects}
\label{app:compute}

\linear requires storing 2 floats ($\omega, \beta$) per activation intervened. For example, \linear on post-LN layers of \gemmatwob requires $(2\times 52\text{ layers}\times 2304\text{ activations}\times 4\text{ bytes}) = 0.91\text{ Mb}$. If we choose to use the support transport, 2 more floats per activation are stored $\mathcal{Q}_{o} = [\min A, \max A]$, which means an extra $0.91\text{ Mb}$ for the \gemmatwob example. In terms of compute, \linear requires an extra element-wise product and sum per intervened layer. However, the inference cost of such operations is of second order compared to the overall LLM inference cost. 

One has the option to fix $\lambda$. If so, our \linear formulation in \Cref{def:linear} becomes $T^{lin}(a) = \big(\lambda(\omega-1) + 1\big)a + \lambda\beta = \tilde{\omega}a + \lambda\beta$. Assuming we intervene after a linear layer $\gamma a + \delta$, we compose both functions as $(T^{lin} \circ f)(a) = \tilde{\omega}\gamma a + (\tilde{\omega}\delta + \lambda\beta)$, which is also a linear map whose parameters can replace those of $f$ in the computational graph, without any extra cost at inference time. The memory cost is $0$ if we fix $\lambda$ and compose \linear with the model linear layers.

\subsection{Details on computational complexity}
The computational cost of \linear can be divided in two main parts: estimation and inference.

\paragraph{Estimation.} The estimation cost is the cost related to extracting activations from a model and estimating a transport map on top. Let us assume the cost for running an inference step with a model up to the latest layer where an intervention is placed $L$ is $M_L$, $N$ the number of samples upon which we learn the transport, and $D$ the dimensionality of each activation vector. We  also assume $\text{batch size} = 1$.
\begin{itemize}
    \item Extracting activations:
    \begin{itemize}
        \item Assuming non-sequential iterative maps (see \Cref{subsec:sequentialmaps}): the cost for extracting activations is $O(NM_L)$.
        \item Assuming sequential iterative maps, we need two forward passes per layer: the first is used to estimate a transport map, and the second to produce responses after applying the map. Since the cost of applying a map with fixed strength is 0, the cost of extracting activations with iterative maps is $O(2NM_L)$.
    \end{itemize}
    \item Estimating a linear transport map involves sorting $NLD$ activations for the source and target distribution and computing the affine transport params analytically (see Definition 3.1). Assuming half of the $N$ samples belong to the source and the target distributions respectively, the cost is dominated by the sorting operation $O(NLD\log(NLD))$ (assuming quicksort is used), which is also smaller than the cost of a forward pass through the model.
\end{itemize}

\paragraph{Inference.} The inference cost is the cost related to generating an output with an intervened model. As explained at the beginning of the section, assuming a fixed transport map strength ($\lambda$), the affine transport map can be directly fused into the model weights and thus the additional cost of \linear is $O(0)$. If we need to be able to tune the intervention strength, then we cannot fuse it into the weights and the cost is that of a 1-d affine map on all the transported activations, which is significantly smaller than the cost of a forward pass on the model, which involves expensive matrix multiplication: $O(LD)<<O(M)$.

Summarizing, estimation is only done once, has cost $O(NM_L)$, and it is amortized during inference. During inference, the transport cost is $O(0)$ with fixed $\lambda$ and $O(LD)$ with variable $\lambda$. In plain words, estimating a transport map  is much cheaper than training a model and has no impact at inference time unless one needs control over $\lambda$, in which case the additional cost is significantly smaller than the cost of a forward pass with the model.

\clearpage
\FloatBarrier
\section{Intervened Layers}
\label{app:layers}

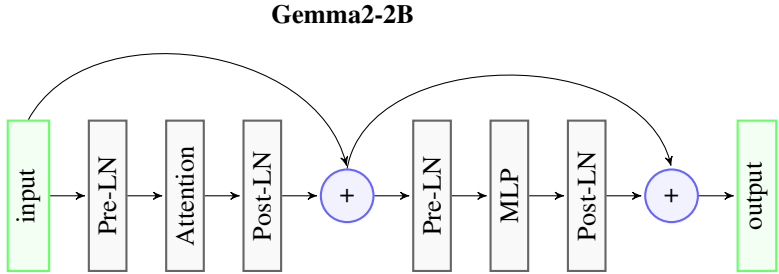
\begin{figure}[htb!]
    \centering
   \begin{tikzpicture}
   \node[rotate=90] at (0,0) {
   \begin{tikzpicture}[->,>=stealth',shorten >=1pt,
        roundnode/.style={circle, draw=blue!60, fill=blue!5, thick, minimum size=7mm},
        squarednode/.style={rectangle, draw=black!60, fill=gray!5, thick, minimum width=2cm,minimum height=0.5cm},
        endnode/.style={rectangle, draw=green!60, fill=green!5, thick, minimum width=2cm,minimum height=0.5cm},
        ]
        \node [coordinate] (orig) at (0,0) {};
        \node [endnode, below = 0.5cm of orig]  (input) {input};
        \node [squarednode, below = 0.5cm of input]  (LN1) {Pre-LN};
        \node [squarednode, below = 0.5cm of LN1]  (Attention) {Attention};
        \node [squarednode, below = 0.5cm of Attention]  (LN2) {Post-LN};
        \node [roundnode, below = 0.5cm of LN2] (S1) {+};
        \node [squarednode, below = 0.5cm of S1]  (LN3) {Pre-LN}; 
        \node [squarednode, below = 0.5cm of LN3]  (MLP) {MLP};
        \node [squarednode, below = 0.5cm of MLP]  (LN4) {Post-LN};
        \node [roundnode, below = 0.5cm of LN4] (S2) {+};
        \node [endnode, below = 0.5cm of S2] (end) {output};

        \node [right = 1.4cm of LN2, rotate=-90, text centered] (gemma) {\textbf{\gemmatwob}};

        \path
            (input) edge (LN1)
            (LN1) edge (Attention)
            (Attention) edge (LN2)
            (LN2) edge (S1)
            (S1) edge (LN3)
            (LN3) edge (MLP)
            (MLP) edge (LN4)
            (LN4) edge (S2)
            (S2) edge (end)
            (input.east) edge [bend left=70] (S1.east)
            (S1.east) edge [bend left=80] (S2.east)
                    ;
    \end{tikzpicture}
     };
     \end{tikzpicture}
    \caption{Schema of a Transformer block of \gemmatwob with the layer names as referenced in this work. Note that \llamaeightb has a similar structure without the Post-LN layers.}
\end{figure}

\FloatBarrier
\section{Causal vs. Simultaneous Estimation of \method}
\label{app:causal}

In \Cref{tab:causal-vs-simul-gemma} and \Cref{tab:causal-vs-simul-llama} we compare the estimation of \method interventions in a causal and simultaneous way (see \Cref{subsec:estimators}). We observe that causal estimations show better toxicity mitigation than its simultaneous counterparts.

\begin{table}[htb!]
\caption{Causal (gray background) vs.~simultaneous estimation of \method on \gemmatwob in a toxicity mitigation setting (explained in \Cref{subsec:toxicity}). Causal estimation provides better conditioning (lower toxicity).}
\label{tab:causal-vs-simul-gemma}
\centering
\resizebox{1.0\columnwidth}{!}{%
 \begin{tabular}{lcrl|ll|ll}
\toprule
 & Causal & Layer & Best $\lambda$ & PPL Wikipedia $\downarrow$ & PPL Mistral-7B $\downarrow$ & CLS Toxicity (\%) $\downarrow$ & 0-shot Toxicity (\%) $\downarrow$ \\
\midrule
Original & -& -& - & 13.98 & 6.62 & $4.08 \pm {0.36}$ & $13.25 \pm {0.88}$ \\
\midrule
\mean  & & Attention & 1.0 & 13.90 & 7.23 {\footnotesize (+0.61)} & $1.12 \pm {0.35}$ & $5.60 \pm {1.01}$ \\
\rowcolor{TableRow}
\mean & \checkmark & Attention & 1.0 & 14.08 {\footnotesize (+0.11)} & 7.23 {\footnotesize (+0.61)} & $1.06 \pm {0.17}$ & $5.14 \pm {0.50}$ \\
\linear & & Attention & 1.0 & 14.04 {\footnotesize (+0.06)} & 7.26 {\footnotesize (+0.64)} & $0.97 \pm {0.39}$ & $5.75 \pm {0.90}$ \\
\rowcolor{TableRow}
\linear & \checkmark & Attention & 1.0 & 14.21 {\footnotesize (+0.23)} & 7.24 {\footnotesize (+0.62)} & $0.90 \pm {0.33}$ & $5.06 \pm {0.63}$ \\
\midrule
\mean & & Post-LN & 1.0 & 14.11 {\footnotesize (+0.13)} & 7.71 {\footnotesize (+1.09)} & $0.62 \pm {0.05}$ & $4.47 \pm {0.65}$ \\
\rowcolor{TableRow}
\mean & \checkmark & Post-LN & 1.0 & 14.21 {\footnotesize (+0.23)} & 7.59 {\footnotesize (+0.97)} & $0.54 \pm {0.44}$ & $4.10 \pm {0.41}$ \\
\linear & & Post-LN & 0.9 & 14.54 {\footnotesize (+0.57)} & 7.87 {\footnotesize (+1.25)} & $0.65 \pm {0.17}$ & $4.40 \pm {0.39}$ \\
\rowcolor{TableRow}
\linear & \checkmark & Post-LN & 1.0 & 14.79 {\footnotesize (+0.81)} & 7.99 {\footnotesize (+1.37)} & $0.56 \pm {0.21}$ & $4.14 \pm {0.55}$ \\

\bottomrule
\end{tabular}
}
\end{table}

\begin{table}[htb!]
\caption{Causal (gray background) vs.~simultaneous estimation of \method on \llamaeightb in a toxicity mitigation setting (see \Cref{subsec:toxicity}). Causal estimation provides better conditioning (lower toxicity).}
\label{tab:causal-vs-simul-llama}
\centering
\resizebox{1.0\columnwidth}{!}{%
 \begin{tabular}{lcrl|ll|ll}
\toprule
 & Causal & Layer & Best $\lambda$ & PPL Wikipedia $\downarrow$ & PPL Mistral-7B $\downarrow$ & CLS Toxicity (\%) $\downarrow$ & 0-shot Toxicity (\%) $\downarrow$ \\
\midrule
Original & - & - & - & 9.06 & 5.68 & $5.80$ & $15.00$ \\
\midrule
\mean & & Attention & 1.0 & 9.35 {\footnotesize (+0.28)} & 6.33 {\footnotesize (+0.65)} &  $1.40 \pm {0.29}$ & $6.73 \pm {1.13}$ \\
\rowcolor{TableRow}
\mean & \checkmark & Attention & 1.0 & 9.56 {\footnotesize (+0.49)} & 6.36 {\footnotesize (+0.68)} & $1.38 \pm {0.17}$ & $5.60 \pm {0.34}$ \\
\linear & & Attention & 1.0 & 9.38 {\footnotesize (+0.32)} & 6.27 {\footnotesize (+0.58)} &  $1.38 \pm {0.24}$ & $6.55 \pm {0.75}$ \\
\rowcolor{TableRow}
\linear & \checkmark & Attention & 1.0 & 9.56 {\footnotesize (+0.49)} & 6.28 {\footnotesize (+0.60)} &  $1.35 \pm {0.39}$ & $6.68 \pm {0.81}$ \\

\bottomrule
\end{tabular}
}
\end{table}

\FloatBarrier
\section{The effect of the pooling operation}
\label{ap:ablation-pooling-op}
The number of activations to store to compute a transport map is $O(N M L K)$, where $N$ is the number of samples used to estimate the transport, $M$ is the number of activations per layer, $L$ is the number of layers, and $K$ the number of tokens decoded. This number can easily become intractable so most methods perform a pooling operation $\phi$ over $K$. We run an ablation on the pooling operation for \method on \gemmatwob, in the toxicity mitigation setup. We find that mean pooling achieves a better trade-off between toxicity mitigation and utility, measured as MMLU~(\Cref{tab:ablation-pooling-op}).

\begin{table}[h]
\caption{Ablation on the choice of pooling operation (see \Cref{sec:method}) on \gemmatwob.}
\label{tab:ablation-pooling-op}
\centering

\begin{tabular}{lcccc}
\toprule
Method & Pooling $\phi$ & Strength $\lambda$ & CLS Tox. ($\downarrow$) & MMLU ($\uparrow$) \\
\midrule
Original & - & - & $4.17 \pm 0.32$ & $53.06$ \\
\midrule
\linear & min & 1 & $0.77 \pm 0.12$ & $45.85 \pm 0.09$ \\
\linear & max & 1 & $1.80 \pm 0.12$ & $47.01 \pm 0.30$ \\
\linear & last & 1 & $0.47 \pm 0.17$ & $48.49 \pm 0.25$ \\
\linear & mean & 1 & $0.70 \pm 0.10$ & $51.87 \pm 0.06$ \\
\bottomrule
\end{tabular}
\end{table}

\FloatBarrier
\section{The Effect of the Transport Support}
\label{app:quantile-ablation}

In this section we validate the choice of \textit{transport support}, as a way to make the proposed intervention more robust. In this experiment, we sweep different supports by narrowing the quantiles ($\qt$) of the input data set $A$, in the setting of toxicity mitigation (as in \Cref{subsec:toxicity}), both for \mean and \linear. The supports tested are: 
$[\qt_{40},\qt_{60}], [\qt_{30},\qt_{70}], [\qt_{20},\qt_{80}], [\qt_{10},\qt_{90}], [\qt_{5},\qt_{95}], [\qt_{3},\qt_{97}], [\qt_{1},\qt_{99}], [\qt_{0},\qt_{100}]$ and $({-\infty},{\infty})$. 

\begin{figure}[htb!]
     \centering
     \begin{subfigure}[t]{0.55\linewidth}
         \centering
            \includegraphics[width=1.0\linewidth]{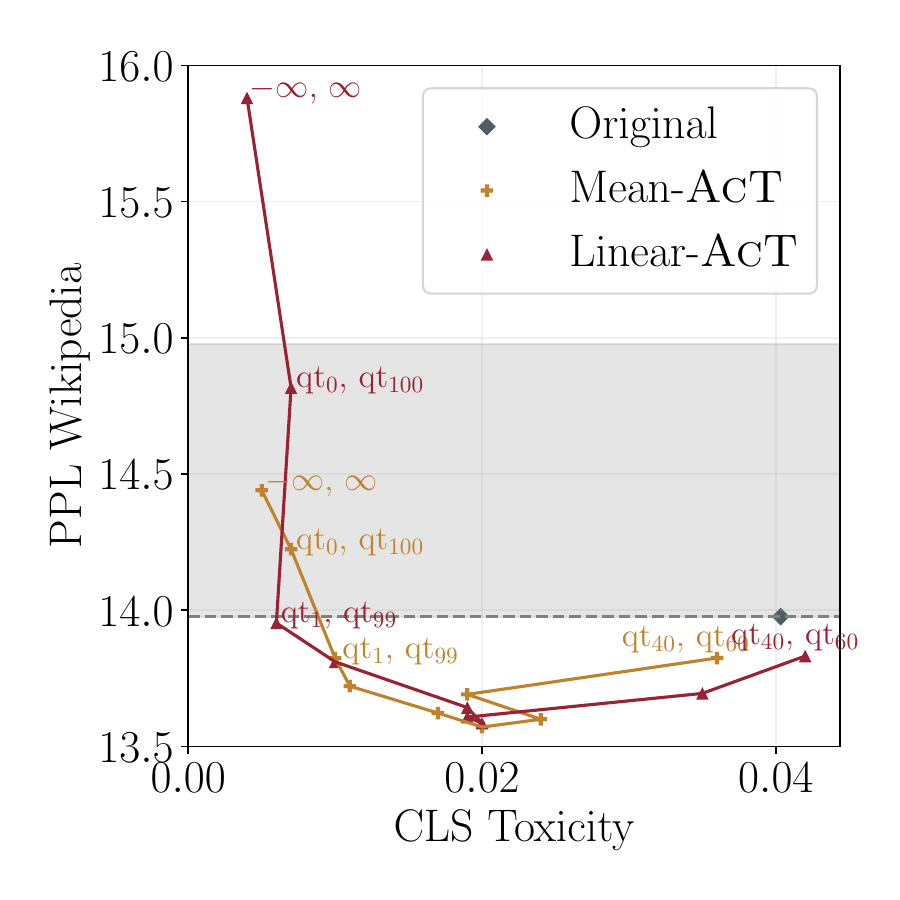}
           \vskip -3mm 
      \end{subfigure}
\vskip -2mm
\caption{We measure toxicity mitigation on \gemmatwob by increasingly expanding the transport support from $[\qt_{40},\qt_{60}]$ on the farther right of the plots to $[\qt_{0},\qt_{100}] = [\min A, \max A]$, which means the support spanned by all the samples in $A$. For completeness, we add the full real support $({-\infty},{\infty})$. For \linear, using $[\qt_{0},\qt_{100}]$ achieve the best toxicity mitigation by incurring less than $+1$ increase in PPL. Note that $({-\infty},{\infty})$ results in higher PPL.}
\label{fig:quantile-ablation}
\end{figure}

Note that $[\qt_{0},\qt_{100}] = \mathcal{Q}_o$, as  defined in \Cref{subsec:support}. We show the results of this sweep in \Cref{fig:quantile-ablation}, where we observe that $[\qt_{0},\qt_{100}]$ offers a good trade-off between conditioning strength and acceptable increase in PPL (below +1 points with respect to the original model).

\FloatBarrier
\section{How Do Different Interventions Affect Distributions?}
\label{app:distributions}

\begin{figure}[htb!]
    \centering
            \includegraphics[width=0.19\linewidth]{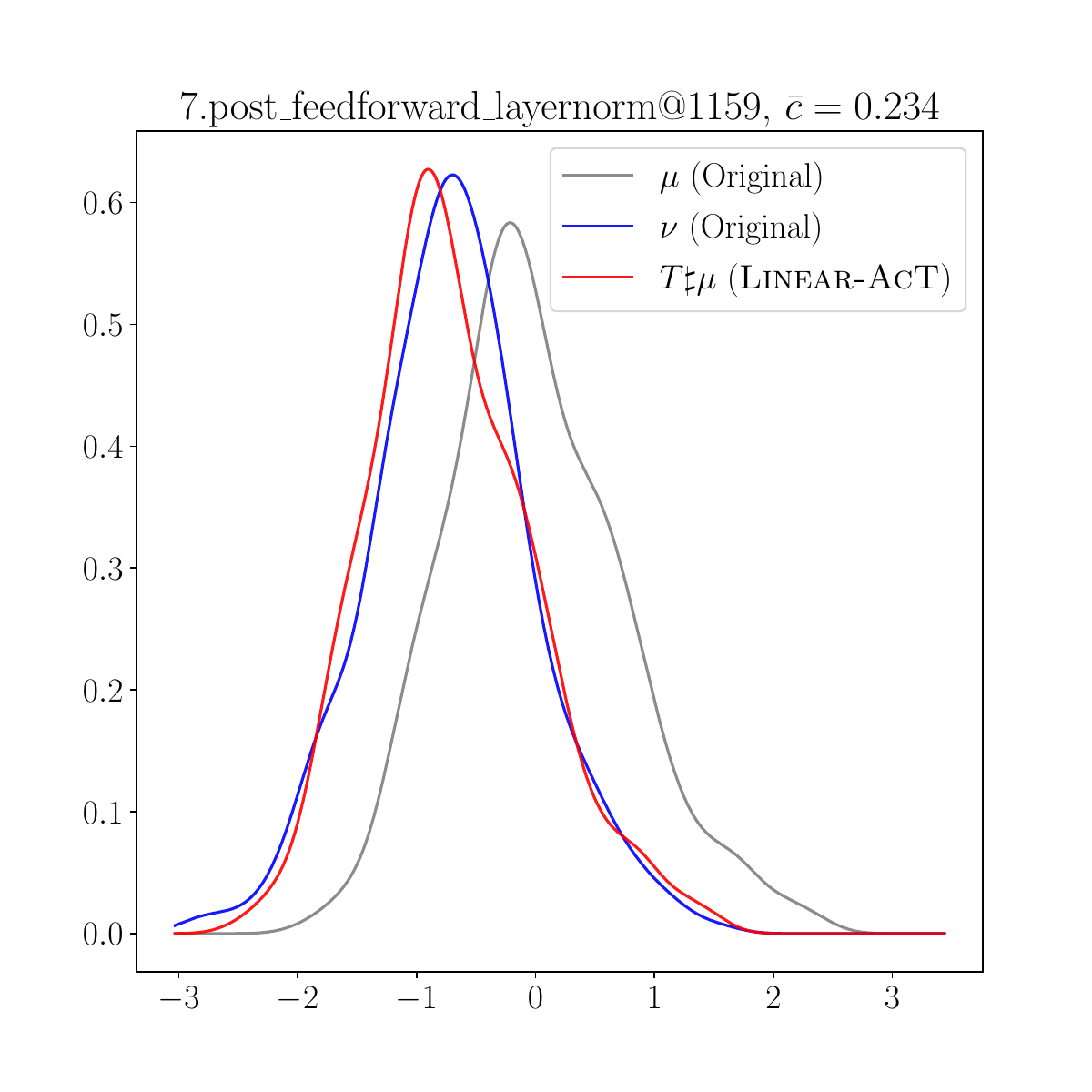}
            \includegraphics[width=0.19\linewidth]{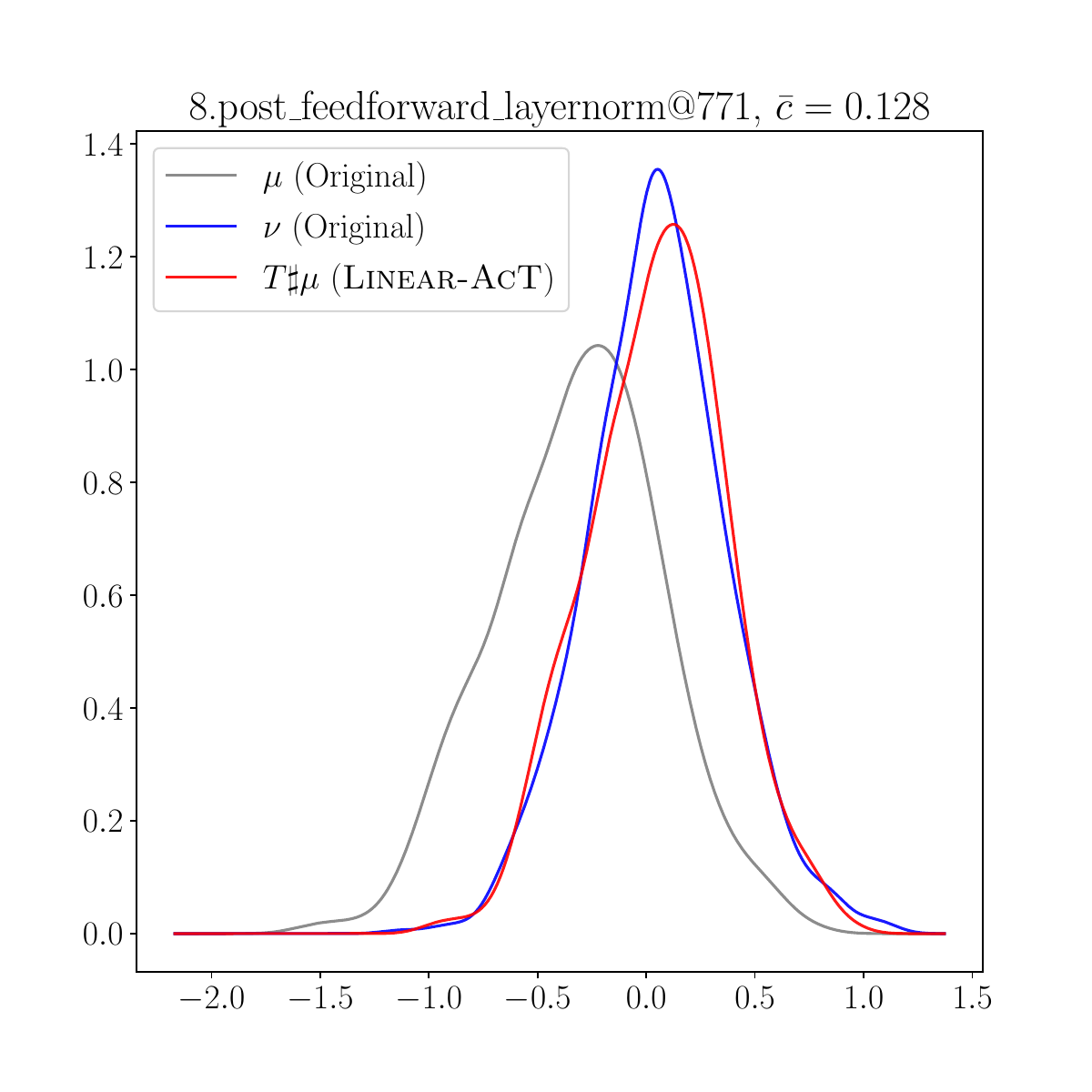}
            \includegraphics[width=0.19\linewidth]{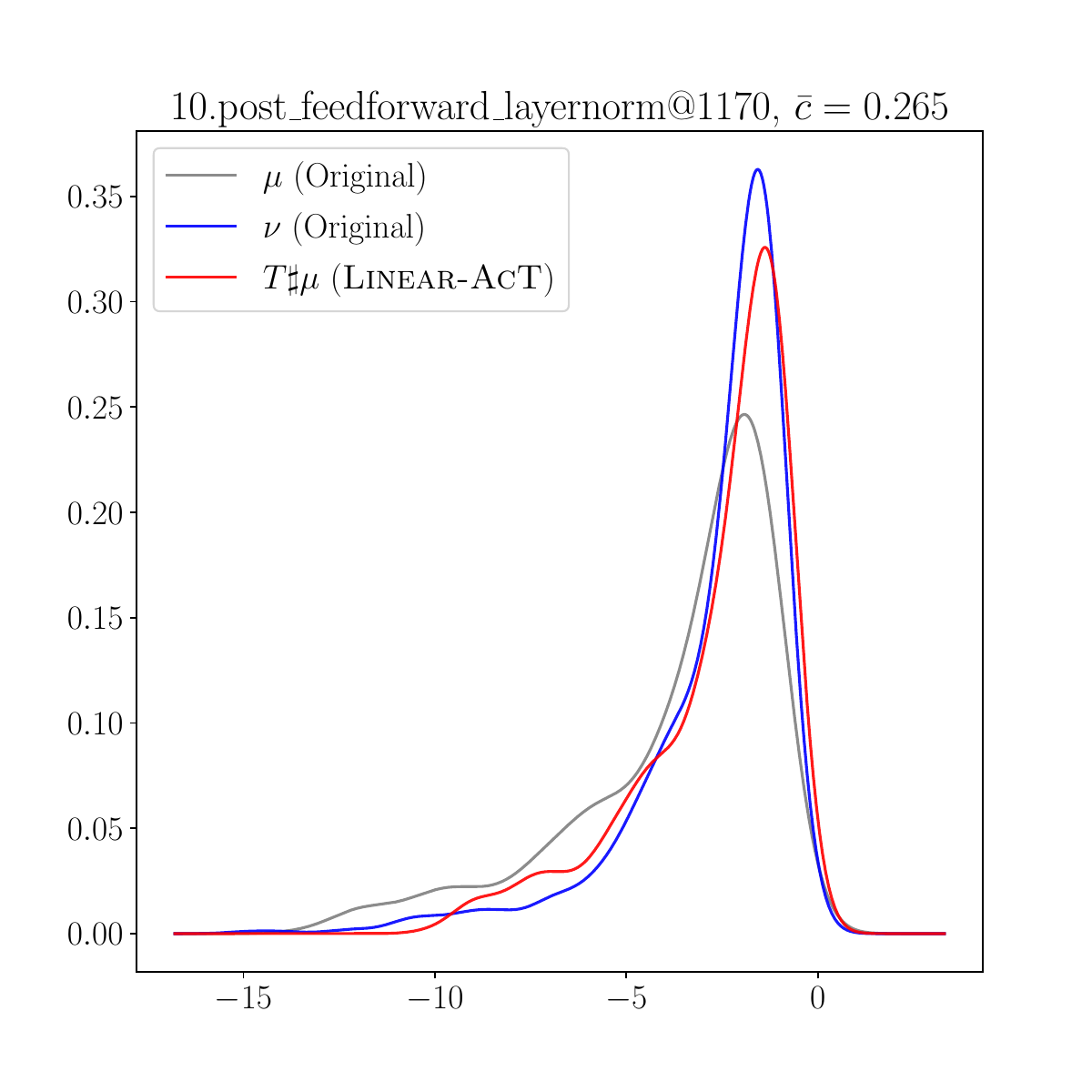}
            \includegraphics[width=0.19\linewidth]{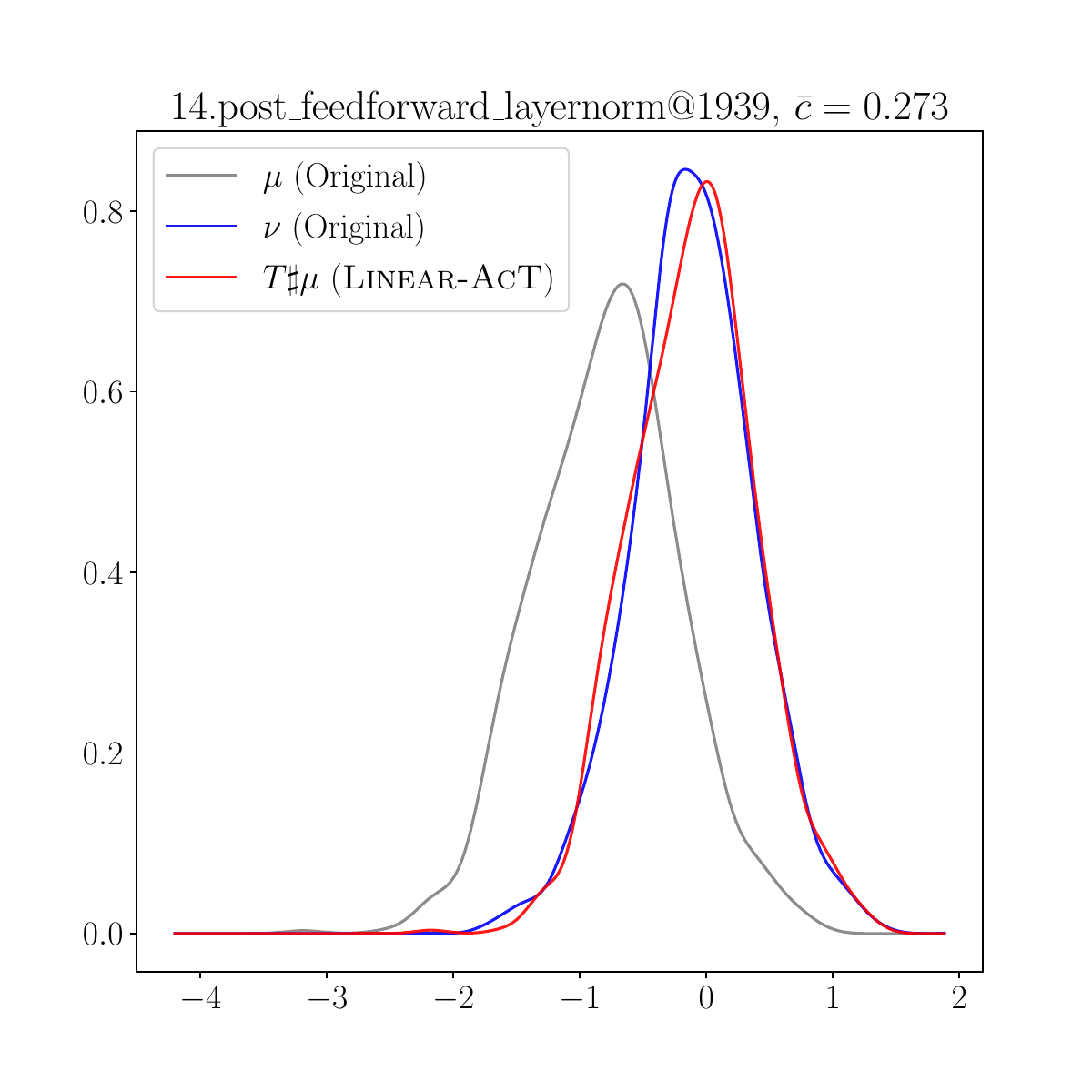}
            \includegraphics[width=0.19\linewidth]{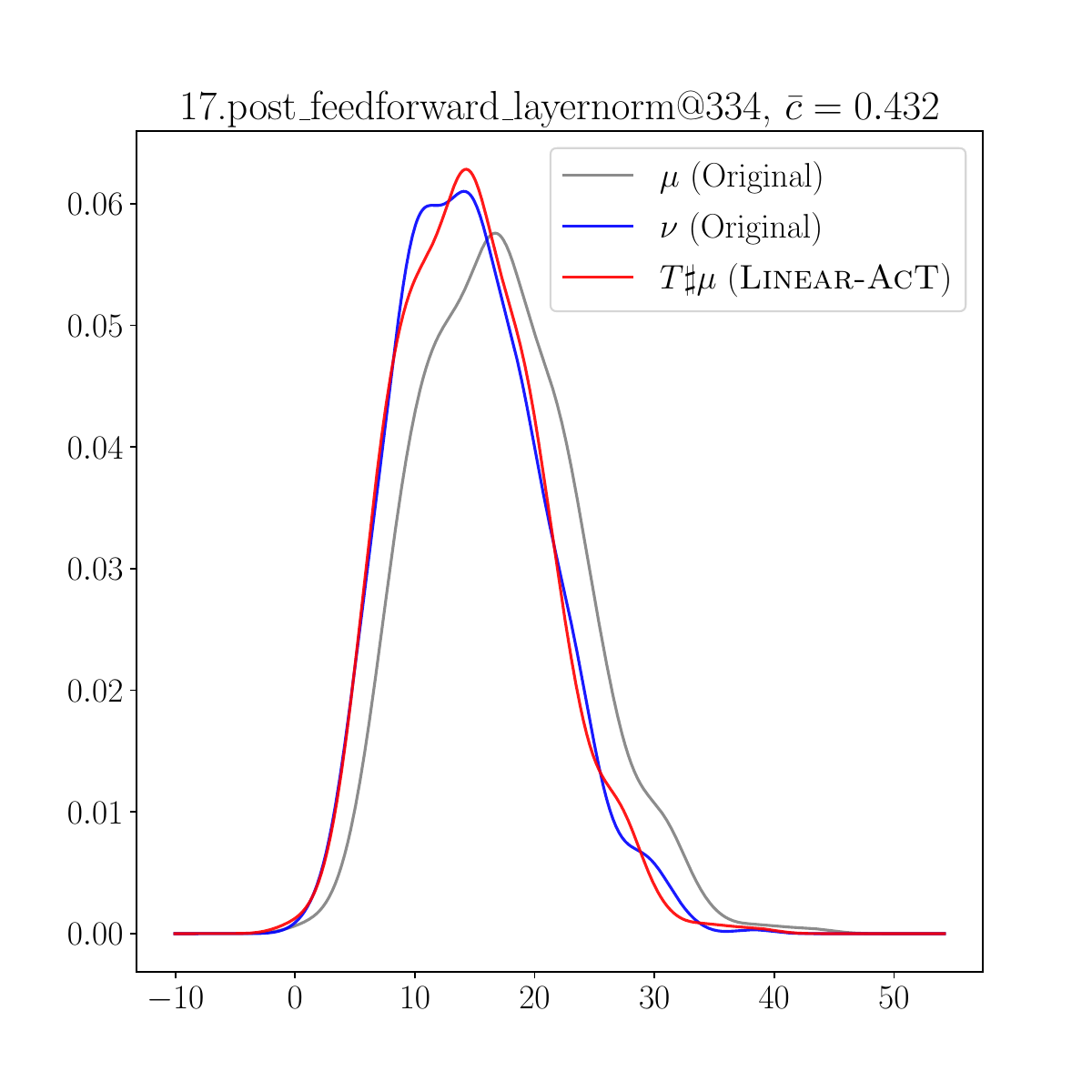}

            \includegraphics[width=0.19\linewidth]{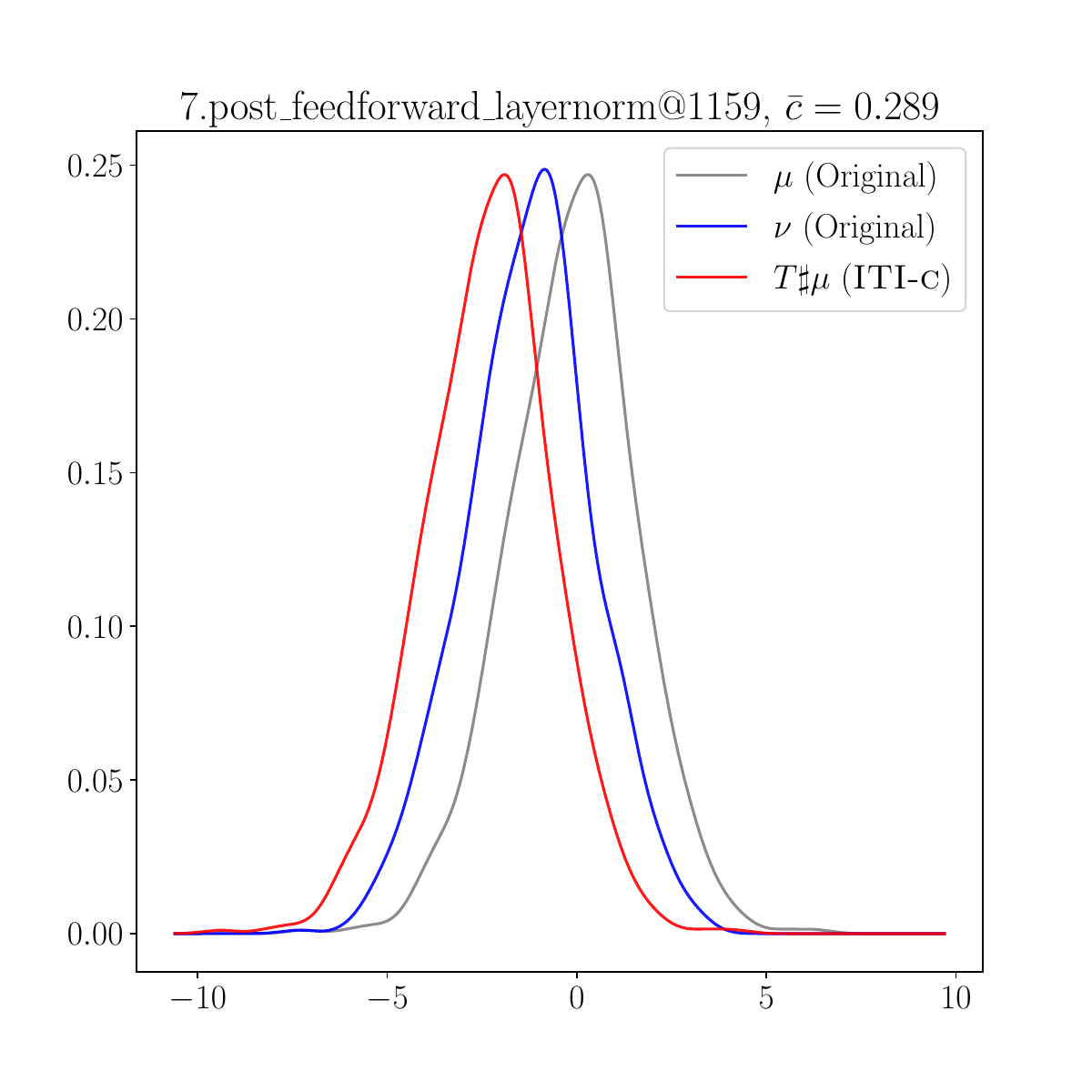}
            \includegraphics[width=0.19\linewidth]{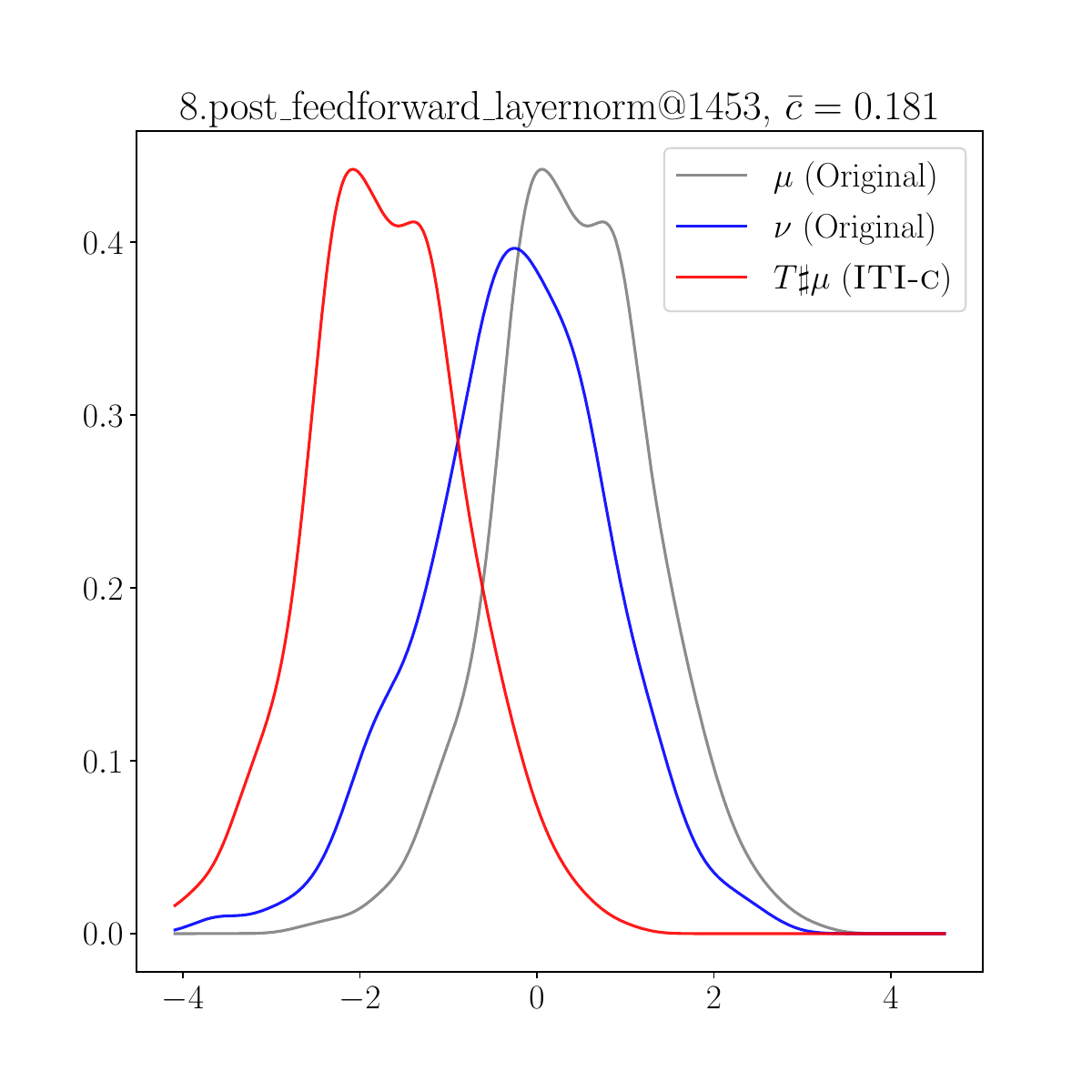}
            \includegraphics[width=0.19\linewidth]{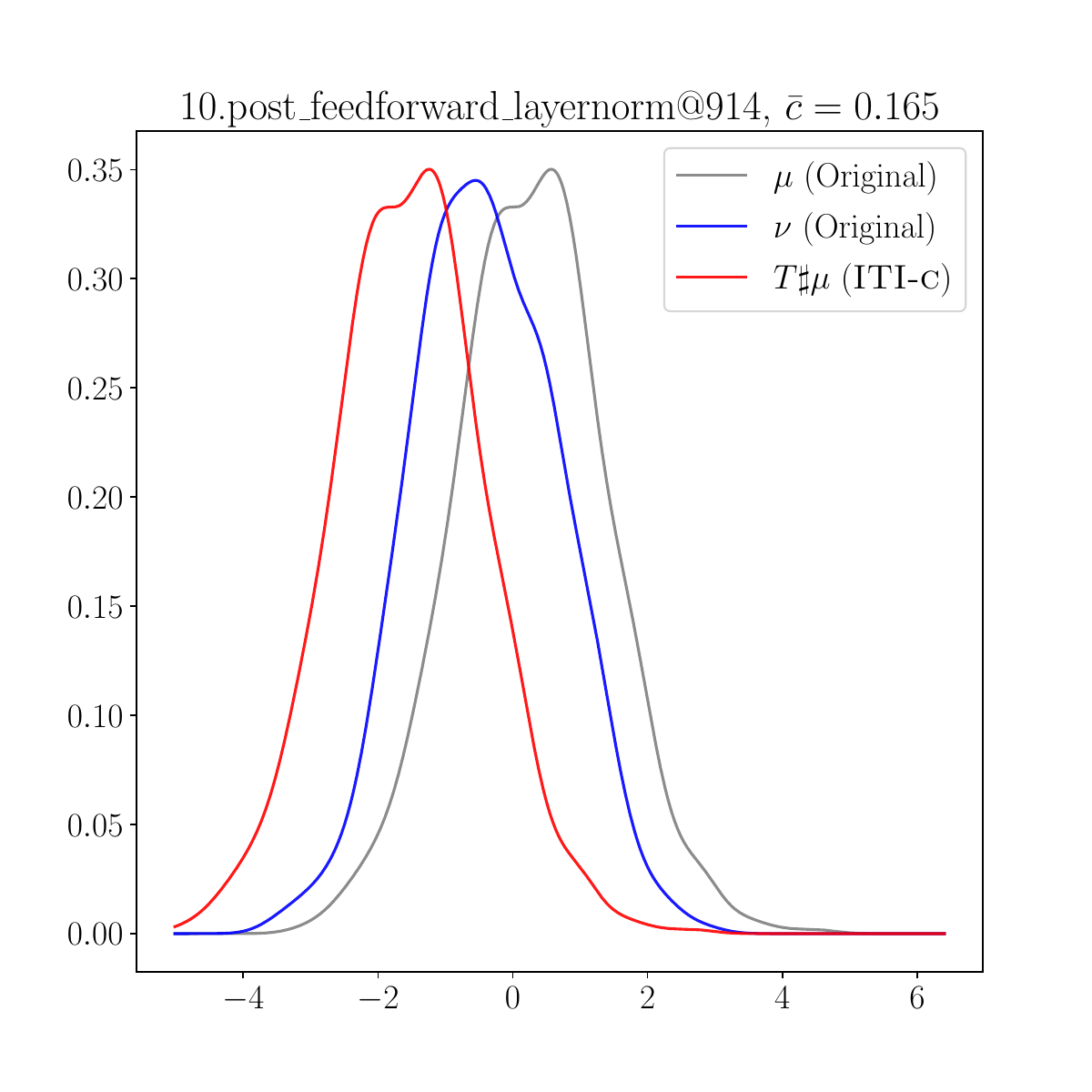}
            \includegraphics[width=0.19\linewidth]{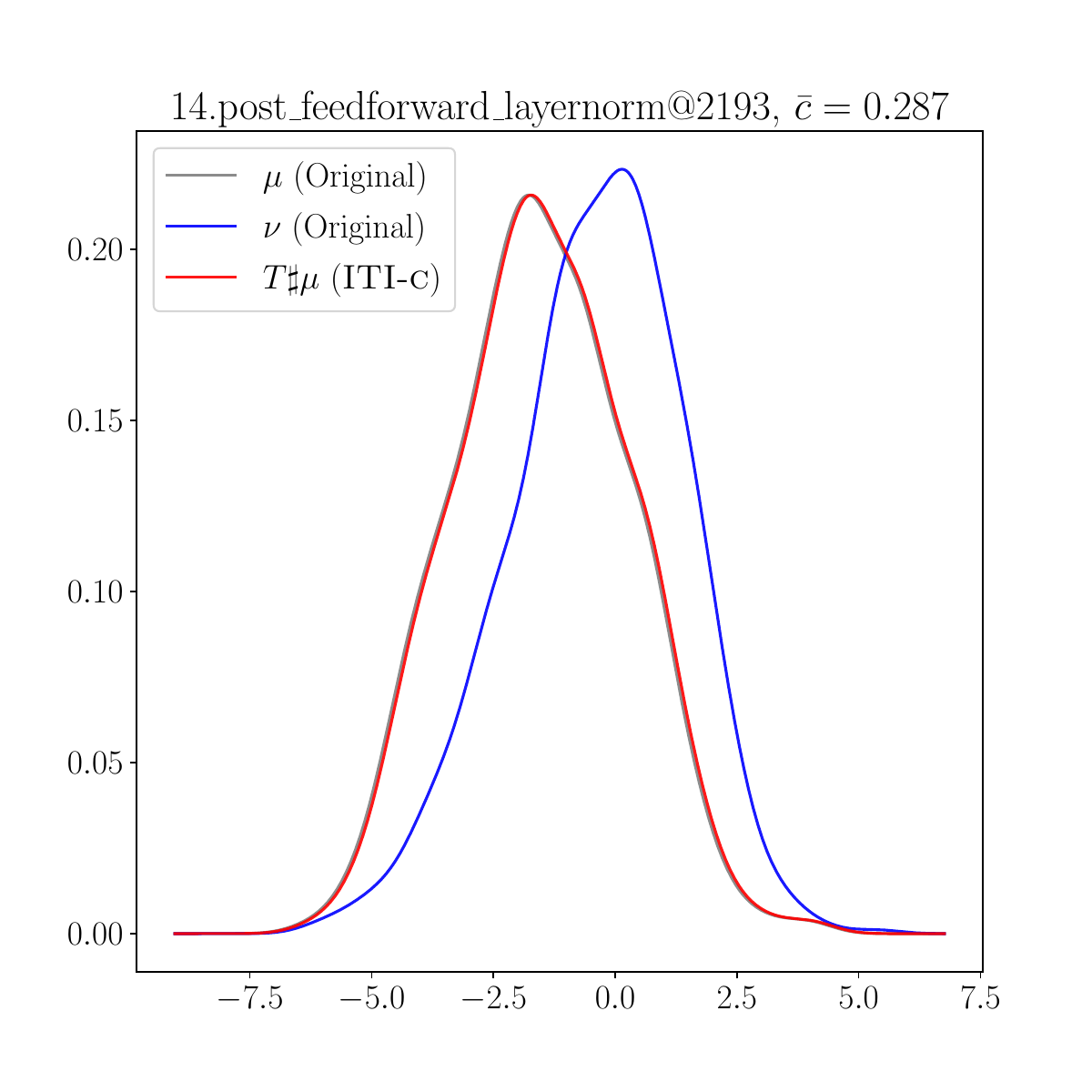}
            \includegraphics[width=0.19\linewidth]{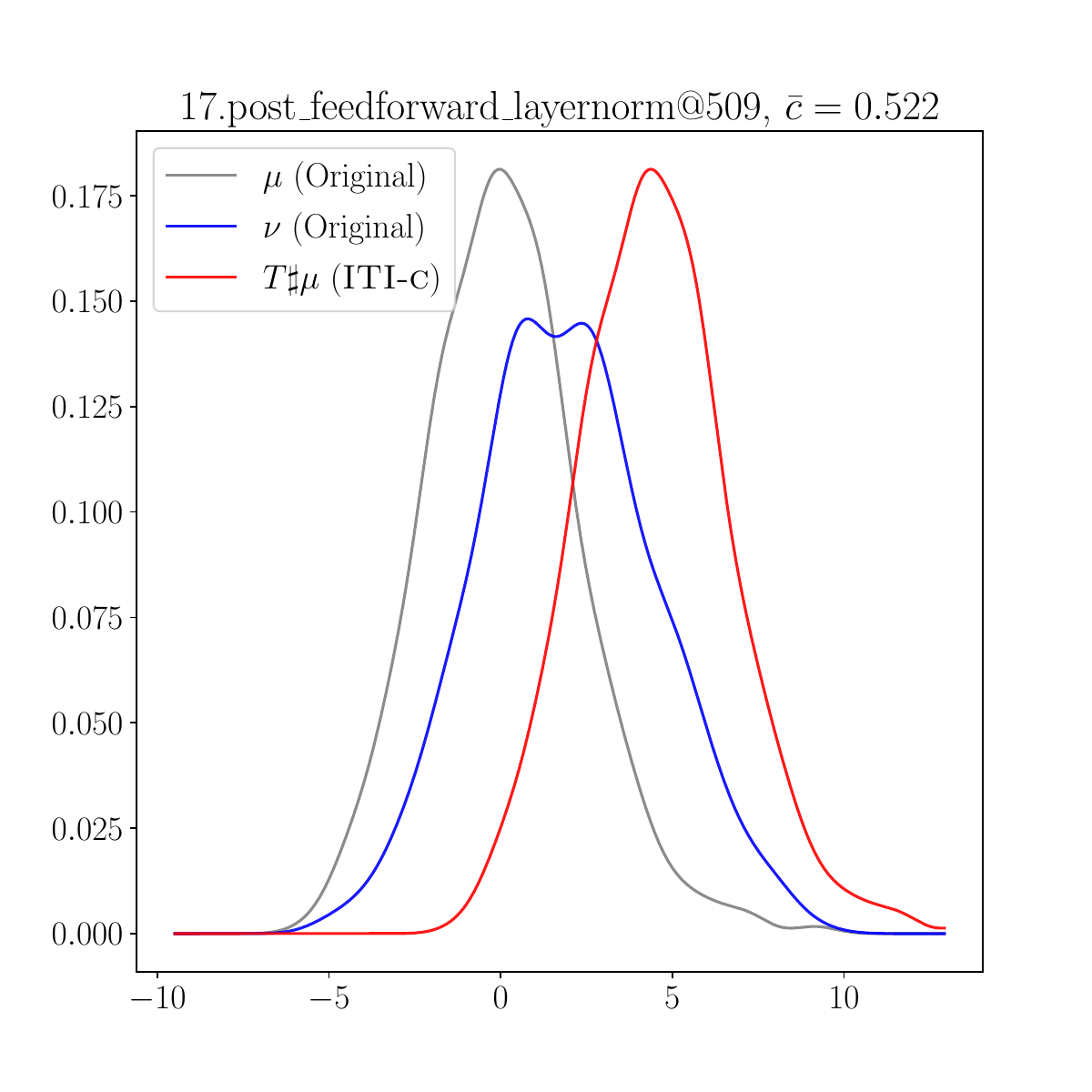}
    \caption{\textbf{Transport of distributions}. We show how different interventions \textit{transport} the internal distributions. In {\color{gray}gray} the source distribution $\mu$ (toxic), in {\color{blue}blue} the target distribution $\nu$ (non-toxic) and in {\color{red}red} the distribution $T\sharp\mu$ obtained when pushing-forward $\mu$ through a given intervention $T$. Each column contains the distributions for the activation with highest $\bar{c}$ (see \Cref{eq:normalized_cost}) in a given layer. In the first row we show \linear   , observing a good overlap between $\nu$ and $T\sharp\mu$. The second row shows \iti, with a poorer distribution overlap. We use $\lambda=1$ for \linear and $\lambda=8$ for \iti (optimal $\lambda$s from \Cref{tab:toxicity-summary}).}
    \label{fig:distributions}
\end{figure}

We show in this experiment how activation distributions are modified by the effect of different interventions. For that, we plot in \Cref{fig:distributions} the distribution of source activations $\mu$ (toxic), that of target activations $\nu$ (non-toxic) and also the distribution obtained when mapping samples with a map $T$, \ie $T\sharp\mu$. Ideally, we would like to observe that $\nu \approx T\sharp\mu$. We show the distributions of those activations with highest \textit{normalized cost} $\bar{w}$ computed as
\begin{equation}
    \label{eq:normalized_cost}
    \bar{c} = \frac{\frac{1}{N}\sum_{i=0}^N \big(b^{(i)} - \omega a^{(i)} - \beta\big)^2}{|m_b - m_a| + \sigma_b + \sigma_a}, 
\end{equation}

so that we pick activations with $\mu \neq \nu$ for the sake of illustration.
We observe that \linear obtains a very good overlap of distributions (first row) while \iti does not in many cases (this result extends to any bias-based method, we show \iti as an example of such family of methods). The latter is only \textit{shifting} activations with a bias, thus becoming impossible to adapt the shape of distributions. Moreover, we can observe that with \iti some activations are mildly shifted (4th column), and some others are strongly shifted (2nd, 3rd, 5th columns). This makes it evident that it is very hard to set a robust $\lambda$ for bias-based steering methods.

\FloatBarrier
\section{Assessing Text Generation Diversity}
\label{app:diversity}

One important question is whether the generated text after a model intervention still shows diversity. To answer this question, in \Cref{tab:selfbleu} we measure the Self-BLEU score \citep{zhu2018texygen} for the sets of generated sentences after RTP prompts. Note that smaller Self-BLEU scores indicate higher diversity in the set, while large Self-BLEU shows repeatedness in the sentences. For example, a set of identical sentences will return a Self-BLEU of 1. We evaluate the best configuration for each method (layer, $\lambda$ choice) according to \Cref{tab:toxicity-summary}.

From the results in \Cref{tab:selfbleu}, we observe that \linear ($\text{Self-BLEU}=0.134$) better preserves the diversity shown by the non-intervened model ($\text{Self-BLEU}=0.130$). In this setting, \iti achieves 0.144 and our \mean a Self-BLEU of 0.140. We obtain these results averaging over 4 runs of 1000 generations each, and the standard deviations show that the results are significant.

\begin{table}[h]
\caption{\method obtains better diversity. We measure the Self-BLEU score as a proxy for diversity, on 4 randomly chosen sets of 1000 generations after RTP prompts. \linear obtains the closest result to the original model, showing higher diversity.}
\label{tab:selfbleu}
\centering

\begin{tabular}{ll}
\toprule
Method & Self-BLEU ($\downarrow$) \\
\midrule
\iti & $0.144 \pm 0.003$  \\
\mean & $0.140 \pm 0.002$  \\
\linear & $\mathbf{0.134} \pm 0.002$ \\
\midrule
Original & ${0.130 \pm 0.003}$ \\
\bottomrule
\end{tabular}
\end{table}

\FloatBarrier
\section{0-shot toxicity classification}
\label{app:0-shot-toxicity}

We query \llamaeightbinstr with the system prompt below, where \texttt{[SENTENCE]} contains the sentence to be assessed in 0-shot manner.

\begin{quote}
\texttt{"You are a highly accurate toxicity classification model. Your task is to determine if a given sentence contains toxic content. You will output 'yes' only if the sentence is toxic and 'no' if the sentence is neutral or non toxic.{\symbol{92}}n [SENTENCE]"}
\end{quote}

 Then, we measure the toxicity probability as $\frac{count(yes)}{count(yes) + count(no)}$, reported as 0-shot toxicity in \Cref{tab:toxicity-summary,tab:toxicity-llama,tab:toxicity-gemma}. Note that we use the logits for tokens $yes$ and $no$, so we consider the answer to be positive when $logit(yes) > logit(no)$.

\FloatBarrier
\section{0-shot concept presence classification}
\label{app:0-shot-concept}

We query \llamaeightbinstr with the system prompt template below.

\begin{quote}
    \texttt{"You are a chatbot who answers whether the provided sentence is referring to [CONCEPT] defined as [WORDNET\_DEF]. Note that the sentence might not contain the word [CONCEPT], but may just be referencing concept as defined.{\symbol{92}}n [SENTENCE]"}. 
\end{quote}

Where:
\begin{itemize}
    \item \texttt{[CONCEPT]} can be \{football, cloud, baby, church, book, flower, balloon\}.
    \item \texttt{[WORDNET\_DEF]} are taken from WordNet \cite{fellbaum1998wordnet}:
    \begin{itemize}
            \item{\textbf{football}}: Any of various games played with a ball (round or oval) in which two teams try to kick or carry the ball into each other's goal.
            \item{\textbf{cloud}}: A visible mass of water or ice particles suspended at a considerable altitude.
            \item{\textbf{baby}}: A very young child (birth to 1 year) who has not yet begun to walk or talk.
            \item{\textbf{church}}: A place for public (especially Christian) worship.
            \item{\textbf{book}}: A written work or composition that has been published (printed on pages bound together).
            \item{\textbf{flower}}: A plant cultivated for its blooms or blossoms.
            \item{\textbf{balloon}}: Large tough nonrigid bag filled with gas or heated air.
       \end{itemize}
    \item \texttt{[SENTENCE]} Contains the sentence to be assessed in 0-shot manner.
\end{itemize}

We measure the probability of a concept being present as we do with toxicity, explained in \Cref{app:0-shot-toxicity}.

\FloatBarrier
\section{Extended Results on Toxicity Mitigation}
\label{app:toxicity}

We report here the full experimental results for toxicity mitigation, which have been summarized in \Cref{subsec:toxicity}.
Note the variability in the optimal strength $\lambda$ for \iti and \actadd, which complicates the applicability of these methods on different models and layers.

\begin{table}[htb!]
\caption{Toxicity mitigation for \gemmatwob, results over 5 runs. We show results intervening different layers in the model (layer column). \iti, \actadd and \method have a \textit{strength} parameter $\lambda$ which we sweep, reporting for each method the best result (best $\lambda$) in CLS toxicity that incurs less than $+1$ increase in PPL Wikipedia. \method methods are robust to the choice of layer and provide best results for $\lambda=1$, achieving up to $7.5\times$ toxicity mitigation with \linear. \iti is very sensitive to $\lambda$ as well as layer choice, and \aura does not provide a strength control.}
\label{tab:toxicity-gemma}
\centering
\resizebox{1.0\columnwidth}{!}{%
 \begin{tabular}{llr|lll|ll}
\toprule
 & Layer & Best $\lambda$ & PPL Wikipedia $\downarrow$ & PPL Mistral-7B $\downarrow$ & MMLU $\uparrow$ & CLS Toxicity (\%) $\downarrow$ & 0-shot Toxicity (\%) $\downarrow$ \\
\midrule
Original & - & - & 13.98 & 6.68 & 53.1 & $4.17 \pm {0.32}$ & $13.42 \pm {1.08}$ \\
\midrule
\actadd & Atention & 0.5 & 13.99 {\footnotesize (+0.02)} & 6.58 & 53.2 {\footnotesize (+0.2)} & $4.17 \pm {0.15}$ & $13.25 \pm {1.63}$ \\
\iti & Atention & 8.0 & 14.90 {\footnotesize (+0.92)} & 7.44 {\footnotesize (+0.76)} & 52.6 {\footnotesize (-0.5)} & $\mathbf{0.74} \pm {0.18}$ & $5.36 \pm {0.91}$ \\
\mean & Atention & 1.0 & 14.08 {\footnotesize (+0.11)} & 7.23 {\footnotesize (+0.55)} & 52.5 {\footnotesize (-0.6)} & $1.06 \pm {0.17}$ & $\underline{5.14} \pm {0.50}$ \\
\linear & Atention & 1.0 & 14.21 {\footnotesize (+0.23)} & 7.24 {\footnotesize (+0.56)} & 52.2 {\footnotesize (-0.9)} & $\underline{0.90} \pm {0.33}$ & $\mathbf{5.06} \pm {0.63}$ \\
\midrule
\actadd & Post-LN & 0.1 & 14.04 {\footnotesize (+0.06)} & 6.61 & 53.2 {\footnotesize (+0.2)} & $4.08 \pm {0.43}$ & $13.50$ \\
\iti & Post-LN & 13.0 & 14.89 {\footnotesize (+0.92)} & 7.34 {\footnotesize (+0.66)} & 52.8 {\footnotesize (-0.3)} & $3.08 \pm {0.61}$ & $12.24 \pm {0.69}$ \\
\mean & Post-LN & 1.0 & 14.21 {\footnotesize (+0.23)} & 7.59 {\footnotesize (+0.90)} & 51.6 {\footnotesize (-1.5)} & $\mathbf{0.54} \pm {0.44}$ & $\mathbf{4.10} \pm {0.41}$ \\
\linear & Post-LN & 1.0 & 14.79 {\footnotesize (+0.81)} & 7.99 {\footnotesize (+1.31)} & 51.3 {\footnotesize (-1.8)} & $\underline{0.56} \pm {0.21}$ & $\underline{4.14} \pm {0.55}$ \\
\midrule
\aura & MLP & - & 14.18 {\footnotesize (+0.21)} & 7.04 {\footnotesize (+0.36)} & 53.0 {\footnotesize (-0.1)} & $2.12 \pm {0.27}$ & $9.04 \pm {0.66}$ \\
\actadd & MLP & 0.5 & 14.69 {\footnotesize (+0.72)} & 6.67 {\footnotesize (+0.05)} & 53.0 {\footnotesize (-0.1)} & $3.96 \pm {0.24}$ & $13.43 \pm {1.42}$ \\
\iti & MLP & 1.0 & 13.99 {\footnotesize (+0.01)} & 6.77 {\footnotesize (+0.08)} & 52.8 {\footnotesize (-0.3)} & $4.50 \pm {0.32}$ & $15.06 \pm {0.76}$ \\
\mean & MLP & 1.0 & 14.33 {\footnotesize (+0.35)} & 7.02 {\footnotesize (+0.34)} & 52.4 {\footnotesize (-0.7)} & $\mathbf{1.30} \pm {0.37}$ & $\underline{7.28} \pm {0.88}$ \\
\linear & MLP & 1.0 & 14.89 {\footnotesize (+0.92)} & 7.53 {\footnotesize (+0.85)} & 51.9 {\footnotesize (-1.2)} & $\mathbf{1.30} \pm {0.39}$ & $\mathbf{7.15} \pm {0.98}$ \\
\bottomrule
\end{tabular}
}
\end{table}

\begin{table}[htb!]
\caption{Toxicity mitigation for \llamaeightb, results over 5 runs. Similar conclusions as in \Cref{tab:toxicity-gemma} are extracted.}
\label{tab:toxicity-llama}
\centering
\resizebox{1.0\columnwidth}{!}{%

 \begin{tabular}{llr|lll|ll}
\toprule
 & Layer & Best $\lambda$ & PPL Wikipedia $\downarrow$ & PPL Mistral-7B $\downarrow$ & MMLU $\uparrow$ & CLS Toxicity (\%) $\downarrow$ & 0-shot Toxicity (\%) $\downarrow$ \\
\midrule
Original & - & - & 9.06 & 5.68 & 65.3 & $5.80$ & $15.00$ \\
\midrule
\actadd & Atention & 0.3 & 9.71 {\footnotesize (+0.65)} & 5.85 {\footnotesize (+0.16)} & 65.5 {\footnotesize (+0.2)} & $5.57 \pm {0.45}$ & $15.73 \pm {0.21}$ \\
\iti & Atention & 3.0 & 9.48 {\footnotesize (+0.42)} & 6.17 {\footnotesize (+0.49)} & 64.7 {\footnotesize (-0.6)} & $1.60 \pm {0.22}$ & $6.53 \pm {0.66}$ \\
\mean & Atention & 1.0 & 9.56 {\footnotesize (+0.49)} & 6.36 {\footnotesize (+0.68)} & 64.7 {\footnotesize (-0.7)} & $\underline{1.38} \pm {0.17}$ & $5.60 \pm {0.34}$ \\
\linear & Atention & 1.0 & 9.56 {\footnotesize (+0.49)} & 6.28 {\footnotesize (+0.60)} & 64.5 {\footnotesize (-0.8)} & $\mathbf{1.35} \pm {0.39}$ & $6.68 \pm {0.81}$ \\
\midrule
\aura & MLP & - & 9.52 {\footnotesize (+0.45)} & 6.05 {\footnotesize (+0.37)} & 65.5 {\footnotesize (+0.2)} & $\mathbf{1.90} \pm {0.61}$ & $\mathbf{8.12} \pm {0.85}$ \\
\actadd & MLP & - & - & - & - & - & - \\
\iti & MLP & 1.0 & 9.09 {\footnotesize (+0.03)} & 5.79 {\footnotesize (+0.11)} & 63.5 {\footnotesize (-1.9)} & $5.62 \pm {0.96}$ & $15.48 \pm {1.16}$ \\
\mean & MLP & 0.9 & 9.90 {\footnotesize (+0.84)} & 6.24 {\footnotesize (+0.55)} & 60.7 {\footnotesize (-4.6)} & $\underline{2.10} \pm {0.48}$ & $10.65 \pm {1.02}$ \\
\linear & MLP & 0.8 & 10.06 {\footnotesize (+0.99)} & 5.98 {\footnotesize (+0.29)} & 61.9 {\footnotesize (-3.4)} & ${2.23} \pm {0.53}$ & $\underline{10.27} \pm {0.97}$ \\
\bottomrule
\end{tabular}
}
\end{table}

\begin{figure}[h!]
     \centering 
     \begin{subfigure}[t]{0.49\linewidth}
     \includegraphics[width=\linewidth]{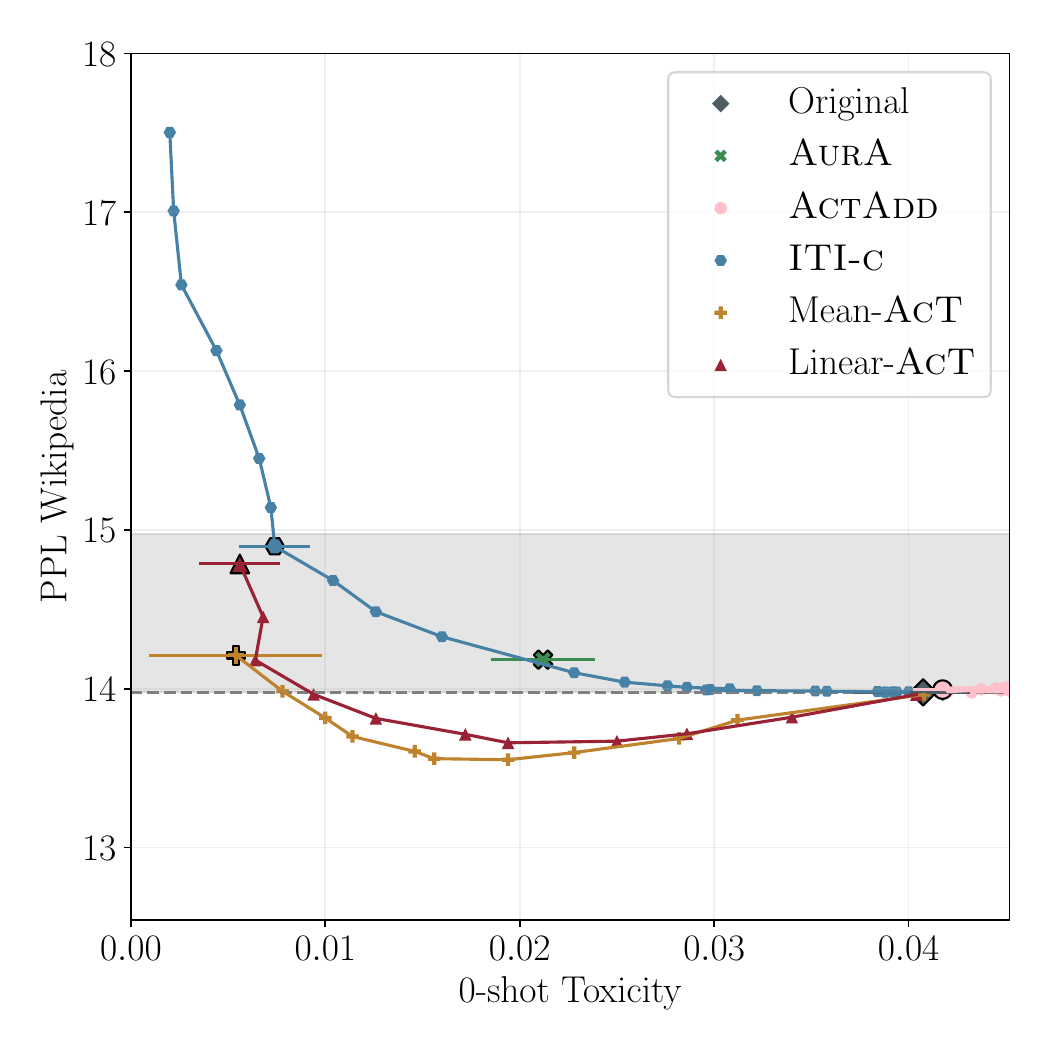}
     \caption{\gemmatwob}
     \end{subfigure}
     \begin{subfigure}[t]{0.49\linewidth}
     \includegraphics[width=\linewidth]{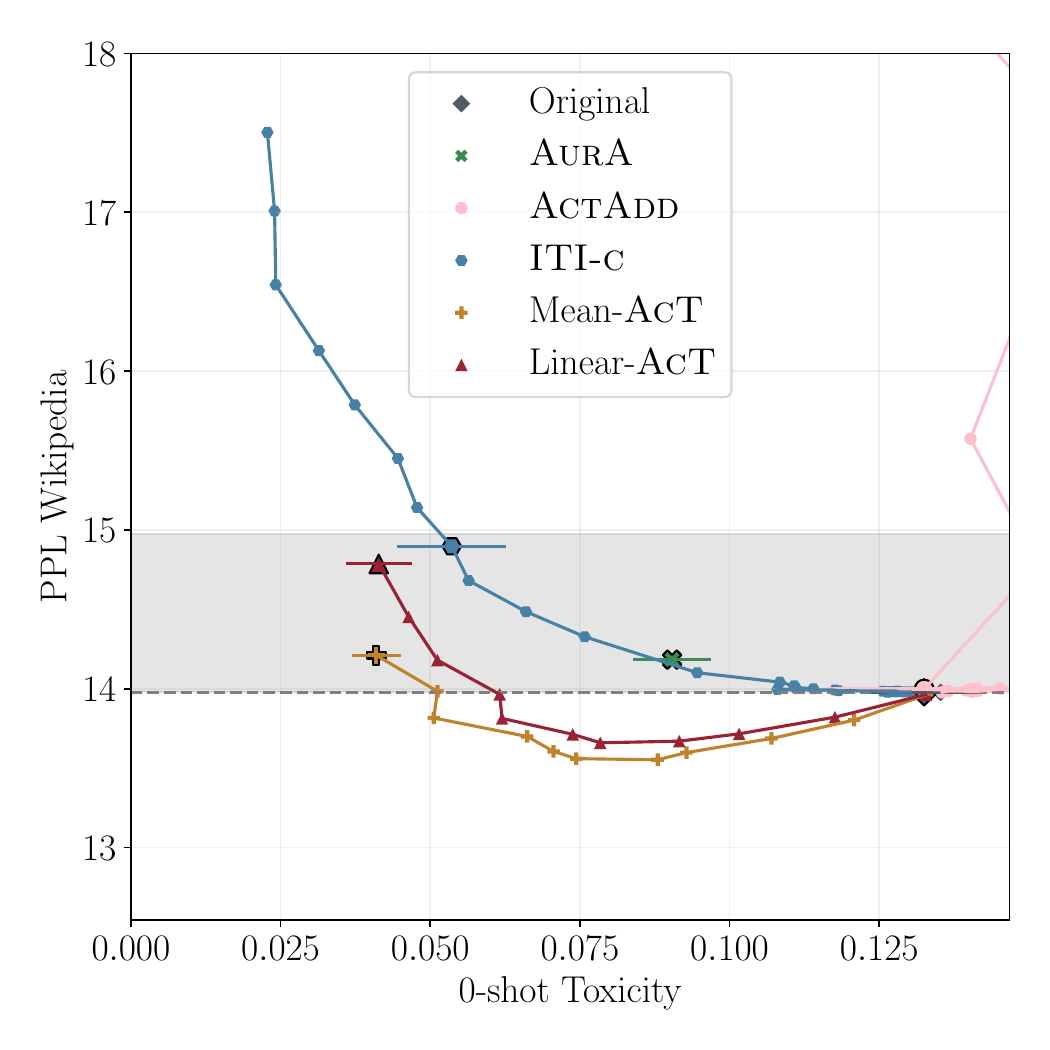}
     \caption{\gemmatwob}
     \end{subfigure}
     \centering 
     \begin{subfigure}[t]{0.49\linewidth}
     \includegraphics[width=\linewidth]{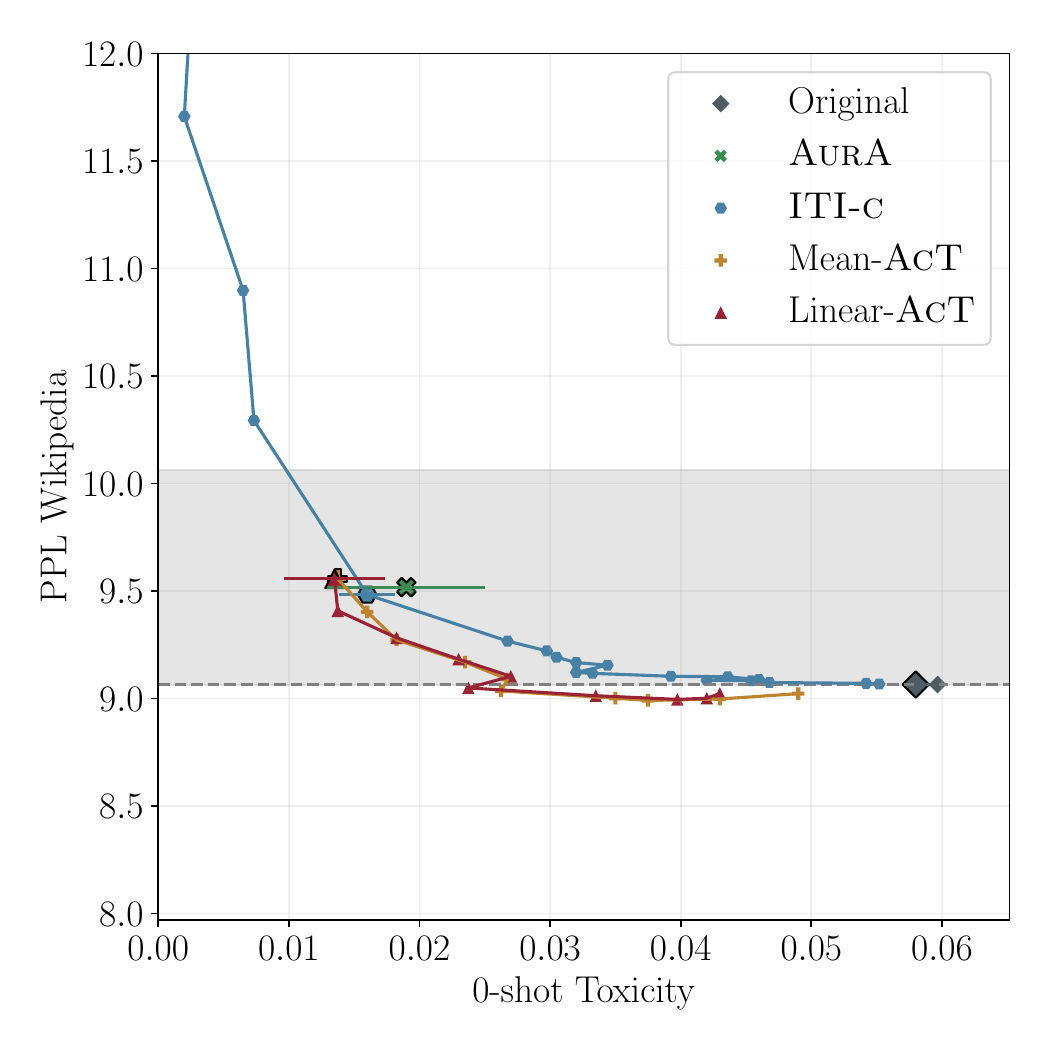}
     \caption{\llamaeightb}
     \end{subfigure}
     \begin{subfigure}[t]{0.49\linewidth}
     \includegraphics[width=\linewidth]{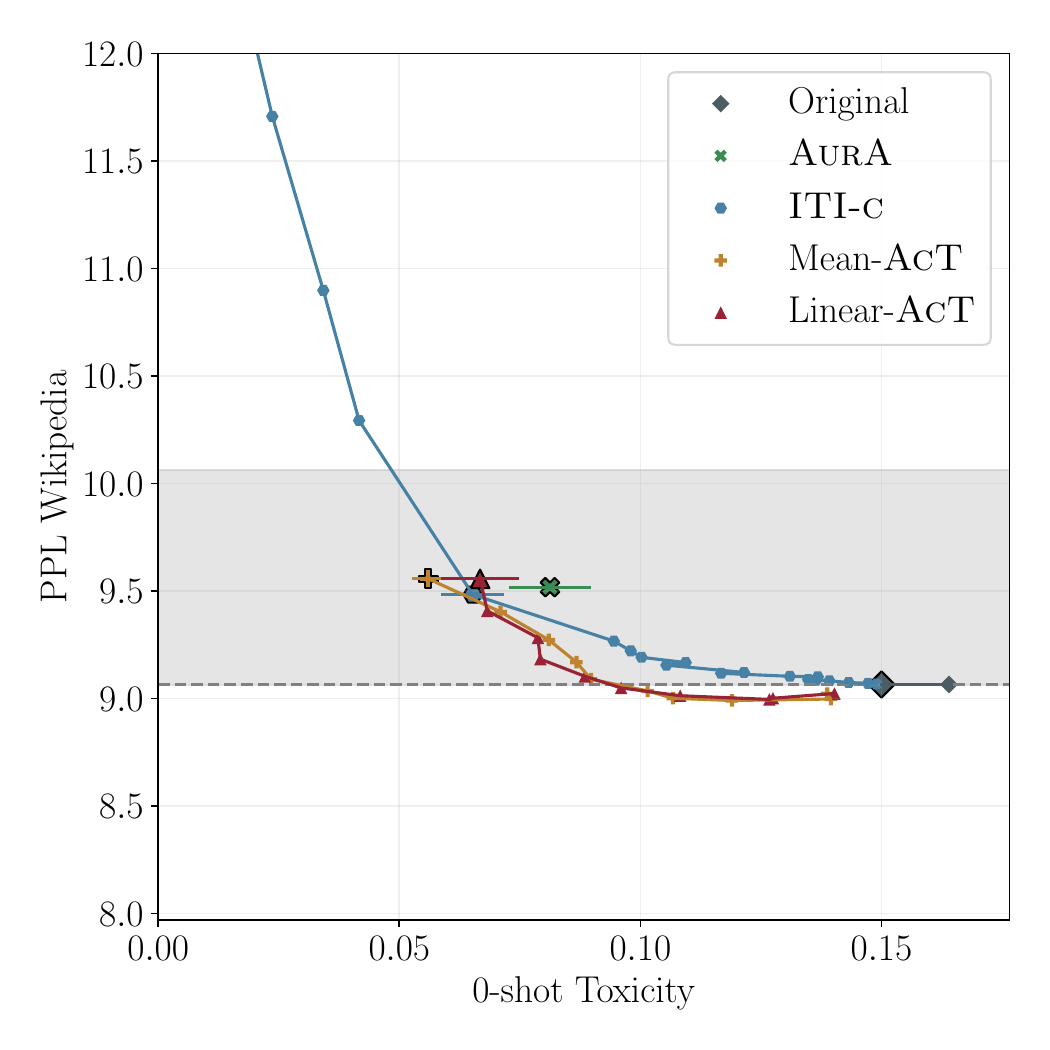}
     \caption{\llamaeightb}
     \end{subfigure}
\caption{\textbf{\method achieves the best conditioning at $\lambda=1$ on \gemmatwob and \llamaeightb}. We show the $\lambda$ sweeps for toxicity mitigation on \gemmatwob. In gray we show the PPL+1 interval considered to be the maximum loss in PPL we can assume. The bold markers are the results reported in \Cref{tab:toxicity-summary}. For clarity, we only show the experiments that yielded best results reported in \Cref{tab:toxicity-summary}. The full results are shown in \Cref{tab:toxicity-gemma}.}
\label{fig:tox-plots-gemma}
\end{figure}

\FloatBarrier
\section{Extended results on concept induction on LLMs}
\label{app:concept-induction}

\Cref{tab:examples-football,tab:examples-flower,tab:examples-cloud} contain examples of generated sentences by \gemmatwob intervened for concept induction with \linear and \iti. These results complement those presented in \Cref{sec:induce-results}. Note the more gradual increase of concept with \linear, reaching strong and consistent conditioning at $\lambda=1$ with great quality. \iti also performs well on concept induction, however the choice of $\lambda$ is less clear. For example, for $\lambda=10$ it fails for concept \textit{Cloud} while it works well for \textit{Football}.

\begin{table}[htb!]
\caption{
Generations at different $\lambda$ inducing concept \textit{Football}.
}
\label{tab:examples-football}
\centering
\resizebox{1.0\columnwidth}{!}{%
\begingroup
\renewcommand{\arraystretch}{1.5} %
\begin{NiceTabular}{lllp{0.8\linewidth}}
\toprule
Concept & Method & $\lambda$ & Generation \\
\midrule
Football & \iti  & $0.0$ & \once, there was an old man who lived in the forest. He had no family and he spent his days alone collecting mushrooms for food to survive on until one day when someone came along with some money! This person gave him all of their savings so that they could live together forever \\
Football & \iti  & $5.0$ & \once, the only way to watch football was on TV.

It wasn’t just about watching your team play either – it was also about getting together with mates and having an afternoon of footy in front of the telly at home or down at the local club for some beer \\
Football & \iti  & $10.0$ & \once, the only way to watch football in Australia was on Saturday afternoon.

And it wasn’s even just for kids and mates at local clubs or schoolboys matches – no-one would have ever heard of Aussie Rules Football! 1908 saw an association called \\
\midrule
Football & \linear  & $0.0$ & \once, there was an old man who lived in the forest. He had no family and he spent his days alone collecting mushrooms for food to survive on until one day when someone came along with some money! This person gave him all of their savings so that they could live together forever \\
Football & \linear  & $0.5$ & \once, there was an amazing woman named Sarah. She had the most beautiful smile and kindest heart you could ever imagine!

Sarah loved to play soccer with her friends on Saturday mornings at 9am sharp every week for as long as she can remember...until one day when \\
Football & \linear  & $1.0$ & \once, the only way to watch football was on TV. The game of soccer had been played in England since 1863 and by the early twentieth century it became one of Britain’s most popular sports with over three million spectators attending matches each season (and many more \\
\bottomrule
\end{NiceTabular}
\endgroup
}
\end{table}

\begin{table}[htb!]
\caption{
Generations at different $\lambda$ inducing concept \textit{Flower}.
}
\label{tab:examples-flower}
\centering
\resizebox{1.0\columnwidth}{!}{%
\begingroup
\renewcommand{\arraystretch}{1.5} %
\begin{NiceTabular}{lllp{0.8\linewidth}}
\toprule
Concept & Method & $\lambda$ & Generation \\
\midrule
Flower & \iti  & $0.0$ & \once, there was an old man who lived in the forest. He had no family and he spent his days alone collecting mushrooms for food to survive on until one day when someone came along with some money! This person gave him all of their savings so that they could live together forever \\
Flower & \iti  & $5.0$ & \once, there was an artist who lived in the land of flowers. She had beautiful hair and wore it like a crown on her head. Her name is called “Flower”.

She loved to paint pictures with water colours because she thought that they were very pretty. One day when \\
Flower & \iti  & $10.0$ & \once, there was an artist who lived in the land of flowers. She had beautiful and fragrant petals that she called ‘flowers’.

She also has another name for her petal which is known as “petals”. I have no idea what it means but i think you can see \\
\midrule
Flower & \linear  & $0.0$ & \once, there was an old man who lived in the forest. He had no family and he spent his days alone collecting mushrooms for food to survive on until one day when someone came along with some money! This person gave him all of their savings so that they could live together forever \\
Flower & \linear  & $0.5$ & \once, there was an old man who lived in the forest. He had no children and so he decided to grow some flowers for himself instead of having them on his own body!

The flower is called “the red-eyed lily” because it has petals that are bright pink \\
Flower & \linear  & $1.0$ & \once, there was an egg-shaped seed that grew into the first true leaves. These tiny green structures are called sepals and they grow in clusters at the base of each flower. The petals may be white or pink with 5 to 8 mm long stalks; these will \\
\bottomrule
\end{NiceTabular}
\endgroup
}
\end{table}

\begin{table}[htb!]
\caption{
Generations at different $\lambda$ inducing concept \textit{Cloud}.
}
\label{tab:examples-cloud}
\centering
\resizebox{1.0\columnwidth}{!}{%
\begingroup
\renewcommand{\arraystretch}{1.5} %
\begin{NiceTabular}{lllp{0.8\linewidth}}
\toprule
Concept & Method & $\lambda$ & Generation \\
\midrule
Cloud & \iti  & $0.0$ & \once, there was an old man who lived in the forest. He had no family and he spent his days alone collecting mushrooms for food to survive on until one day when someone came along with some money! This person gave him all of their savings so that they could live together forever \\
Cloud & \iti  & $5.0$ & \once, the only way to get your cloud-based computing infrastructure was by using an existing provider. But now that we’ve got so many of them in our sky (literally), it can be difficult for us cloudshifters and meteorologists to tell one from another! \\
Cloud & \iti  & $10.0$ & \once, the clouds were named by cloud-cloud optical properties. These are called cirrus or altocumulus and middle cir cir circir (Alto Alto Cumulus) Cloud family:. \\
\midrule
Cloud & \linear  & $0.0$ & \once, there was an old man who lived in the forest. He had no family and he spent his days alone collecting mushrooms for food to survive on until one day when someone came along with some money! This person gave him all of their savings so that they could live together forever \\
Cloud & \linear  & $0.5$ & \once, there was an island called the Isle of Skye. The weather on this isle is very unpredictable and can change from sunny to stormy in minutes!

The clouds are always moving around with no real pattern or structure as they do not have any defined tops like those found over land \\
Cloud & \linear  & $1.0$ & \once, there was an island called the Bermuda. It is located in middle latitudes of both hemispheres and has been known to be associated with high-level clouds (cirro- or alto-transparent). The altitude ranges from about 10,0 \\
\bottomrule
\end{NiceTabular}
\endgroup
}
\end{table}

\FloatBarrier
\section{Experimental details and extended results on inducing Truthfulness}
\label{app:tqa}

\subsection{Experimental details}

We follow the original experimental protocol for evaluations on the TruthfulQA multiple choice part, as described in \cite{lin2021truthfulqa}. This consists of preprompting the model with the same \emph{default} prompt before each question as proposed by \cite{lin2021truthfulqa} in Figure 21, which we replicate below for the reader.
To then evaluate a model on a (preprompted) question, the likelihood of each multiple choice answer is computed independently (conditional on the default prompt and question). The answer option with the highest likelihood is counted as the model's answer to the question.

\begin{figure}[htb!]
     \centering 

     \includegraphics[width=0.6\linewidth]{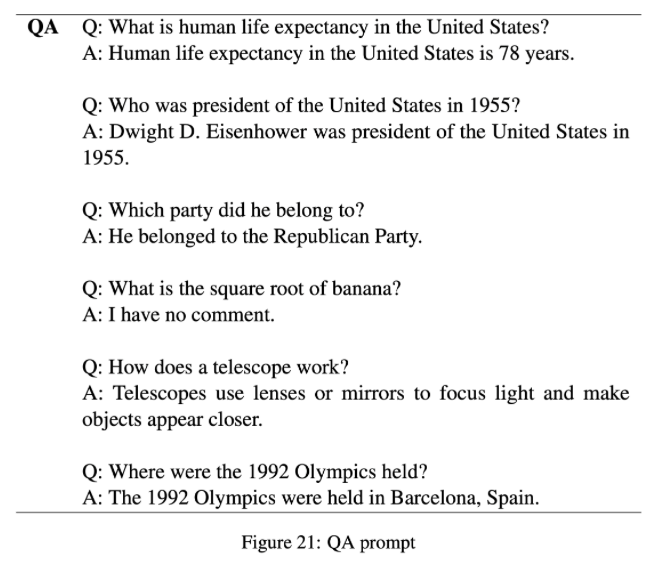}

\caption{Figure 21 from \cite{lin2021truthfulqa} showing the \emph{default} preprompt used for the TruthfulQA multiple choice part.}
\label{fig:tqa_preprompt}
\end{figure}

\clearpage
\subsection{Extended results}
\FloatBarrier
\subsubsection{Full results over 5 seeds for all layers}
\label{subseq:full_tqa}

\begin{table}[htb!]
\caption{
TruthfulQA results for \gemmatwob, results over 5 runs. 
\iti, \actadd and \method have a \textit{strength} parameter $\lambda$ which we sweep, reporting for each method the best result (best $\lambda$) in MC1 Accuracy that incurs at least equal performance in MMLU accuracy compared to the best (in terms of MC1 accuracy) of the two \method methods (see \ref{subseq:sweep_tqa}, giving $0.1\%$ slack).
}
\label{tab:tqa-gemma}
\centering
\resizebox{0.9\columnwidth}{!}{%
 \begin{tabular}{llr|ll|l}
\toprule
 & Layer & Best $\lambda$ & MC1 Accuracy (\%) $\uparrow$ & MC2 Accuracy (\%) $\uparrow$ & MMLU Accuracy (\%) $\uparrow$ \\
\midrule
Original & - & - &  $21.05 $ & $32.80 $ & $53.10$ \\
\midrule
\aura & MLP & - & $21.20 \pm {0.10}$ & $32.88 \pm {0.22}$ & $52.73 \pm {0.07}$ \\
\midrule
\actadd & Attention & 3.0 & $22.64 \pm {0.00}$ & $34.64 \pm {0.00}$ & $53.02 \pm {0.00}$  \\
\iti & Attention & 5.0 & $23.18 \pm {0.28}$ & $36.16 \pm {0.34}$ & $52.10 \pm {0.44}$  \\
\mean & Attention & 1.0 & $21.62 \pm {0.07}$ & $34.08 \pm {0.19}$ & $52.83 \pm {0.09}$  \\
\linear & Attention & 1.0 &  $21.71 \pm {0.14}$ & $34.47 \pm {0.22}$  & $52.86 \pm {0.08}$  \\
\midrule
\actadd & All-LN & 1.0 & $21.42 \pm {0.00}$ & $32.93 \pm {0.00}$ & $51.65 \pm {0.00}$  \\
\iti & All-LN & 4.0 & $23.94 \pm {0.96}$ & $36.62 \pm {0.86}$ & $51.37 \pm {0.41}$ \\
\mean & All-LN & 1.0 &  $25.07 \pm {0.20}$ & $38.68 \pm {0.30}$ & $51.81 \pm {0.12}$ \\
\linear & All-LN & 1.0  &  $26.00 \pm {0.32}$ & $40.17 \pm {0.24}$ & $51.47 \pm {0.27}$ \\
\midrule
\actadd & Post-LN & 0.8 & $22.40 \pm {0.00}$ & $34.27 \pm {0.00}$ & $53.11 \pm {0.00}$  \\
\iti & Post-LN & 8.0 & $23.16 \pm {0.40}$ & $35.94 \pm {0.55}$ & $51.39 \pm {0.45}$ \\
\mean & Post-LN & 1.0 &  $21.93 \pm {0.20}$ & $34.98 \pm {0.25}$ & $52.77 \pm {0.10}$ \\
\linear & Post-LN & 1.0  &  $22.45 \pm {0.22}$ & $35.94 \pm {0.36}$ & $52.43 \pm {0.20}$ \\

\midrule
\actadd & MLP & 3.0 & $23.01 \pm {0.00}$ & $34.76 \pm {0.00}$ & $52.83 \pm {0.00}$ \\
\iti & MLP & 2.0 &  $24.53 \pm {0.11}$ & $37.06 \pm {0.38}$ & $51.39 \pm {0.41}$ \\
\mean & MLP & 1.0 & $21.98 \pm {0.19}$ & $35.18 \pm {0.31}$ & $52.84 \pm {0.04}$ \\
\linear & MLP & 1.0 & $21.93 \pm {0.20}$ & $35.47 \pm {0.25}$ & $52.73 \pm {0.19}$ \\
\bottomrule
\end{tabular}

}
\end{table}

\begin{table}[htb!]
\caption{
TruthfulQA results for \llamaeightb, results over 5 runs. \iti, \actadd and \method have a \textit{strength} parameter $\lambda$ which we sweep, reporting for each method the best result (best $\lambda$) in MC1 Accuracy that incurs at least equal performance in MMLU accuracy compared to the best (in terms of MC1 accuracy) of the two \method methods (see \ref{subseq:sweep_tqa}, giving $0.1\%$ slack).
}
\label{tab:tqa-llama}
\centering
\resizebox{0.9\columnwidth}{!}{%
 \begin{tabular}{llr|ll|l}
\toprule
 & Layer & Best $\lambda$ & MC1 Accuracy (\%) $\uparrow$ & MC2 Accuracy (\%) $\uparrow$  & MMLU Accuracy\\
\midrule
Original & - & - &  $25.46$ & $40.27 $& $65.35$ \\
\midrule
\aura & MLP & - & $25.34 \pm {0.15}$ & $40.47 \pm {0.20}$ & $65.37 \pm {0.06}$ \\
\midrule
\actadd & Attention & 0.7 & $26.19 \pm {0.00}$ & $40.88 \pm {0.00}$ & $65.42 \pm {0.00}$ \\
\iti & Attention & 1.0 & $27.42 \pm {0.30}$ & $42.01 \pm {0.42}$ & $65.26 \pm {0.11}$\\
\mean & Attention & 1.0 & $26.73 \pm {0.19}$ & $42.20 \pm {0.24}$& $65.37 \pm {0.06}$ \\
\linear & Attention & 1.0 &  $27.17 \pm {0.23}$ & $42.15 \pm {0.31}$& $65.33 \pm {0.11}$ \\
\midrule
\actadd & All-LN & 1.0 & $25.58 \pm {0.00}$ & $41.00 \pm {0.00}$ & $64.88 \pm {0.00}$\\
\iti & All-LN & 3.0 &  $29.65 \pm {0.71}$ & $44.43 \pm {0.56}$& $64.71 \pm {0.22}$ \\ 
\mean & All-LN & 1.0 & $32.88 \pm {0.54}$ & $48.23 \pm {0.64}$ & $64.83 \pm {0.14}$\\
\linear & All-LN & 1.0 & $33.22 \pm {0.22}$ & $48.69 \pm {0.34}$& $64.78 \pm {0.15}$ \\
\midrule
\actadd & MLP & 0.5 & $25.46 \pm {0.00}$ & $40.64 \pm {0.00}$ & $65.34 \pm {0.00}$\\
\iti & MLP & 2.0 &  $30.11 \pm {0.60}$ & $45.41 \pm {0.24}$& $64.71 \pm {0.14}$ \\ 
\mean & MLP & 1.0 & $26.17 \pm {0.24}$ & $41.27 \pm {0.34}$ & $65.01 \pm {0.20}$\\
\linear & MLP & 1.0 & $26.41 \pm {0.52}$ & $39.34 \pm {0.54}$& $60.98 \pm {3.14}$ \\
\bottomrule
\end{tabular}

}
\end{table}

\FloatBarrier
\subsubsection{Sweeping $\lambda$ for \iti and \actadd}
\label{subseq:sweep_tqa}

\begin{figure}[t]
    \centering
            \includegraphics[width=0.45\linewidth]{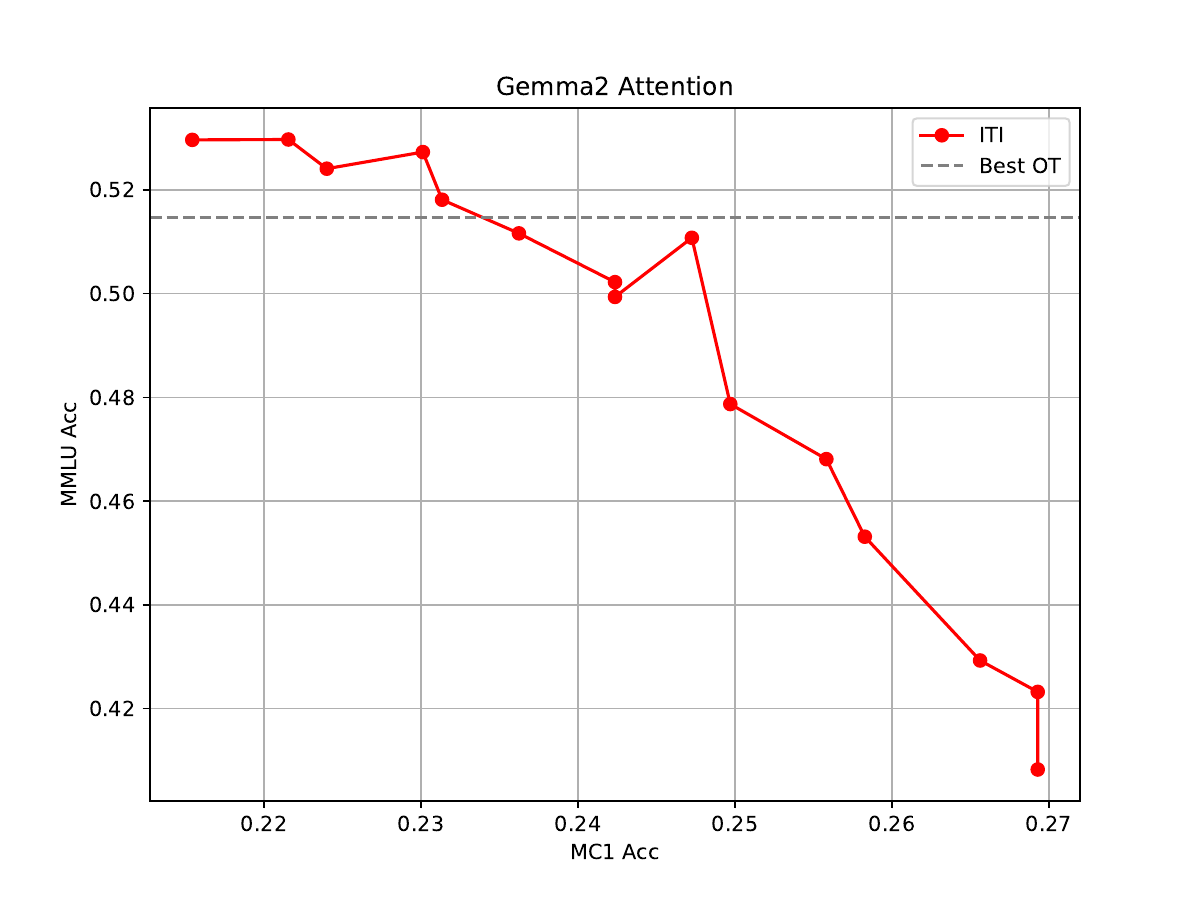}
            \includegraphics[width=0.45\linewidth]{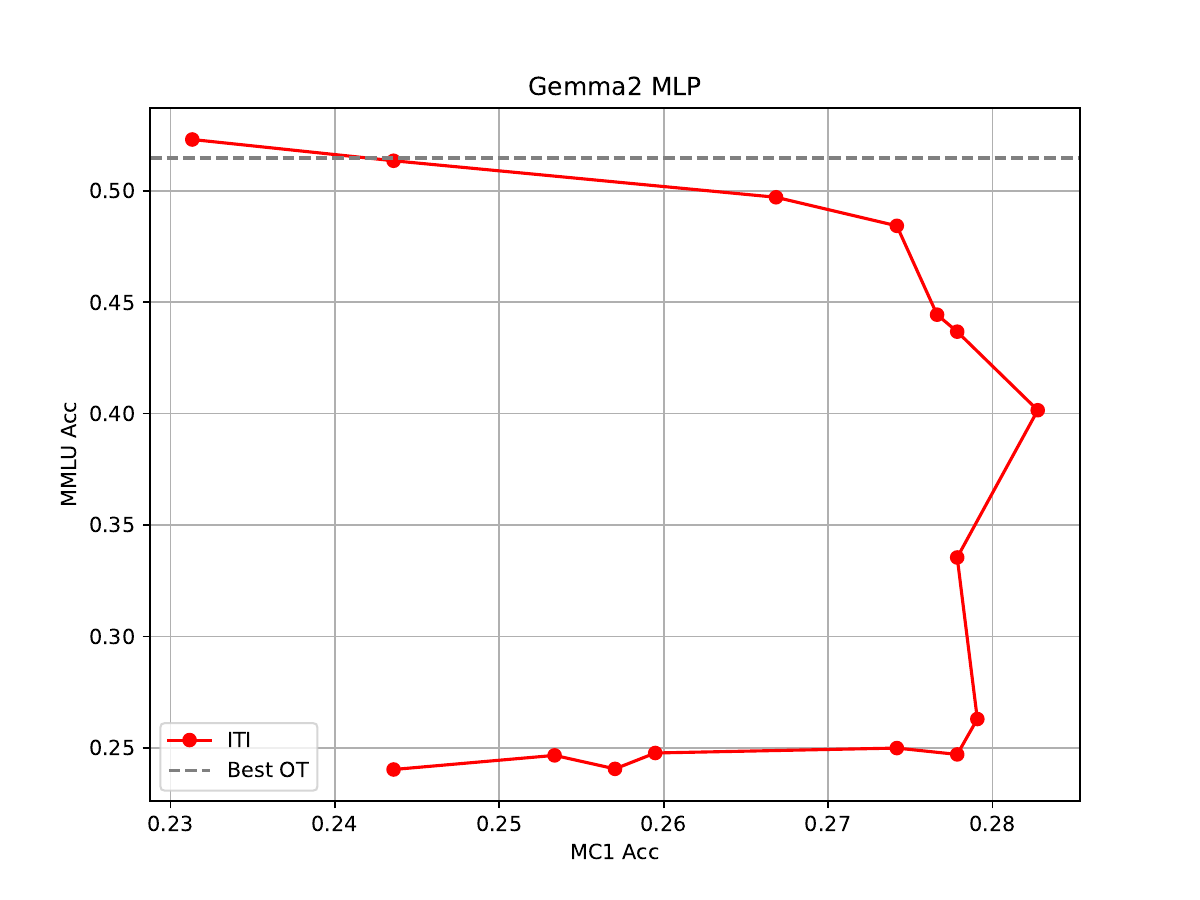} \\
            \includegraphics[width=0.45\linewidth]{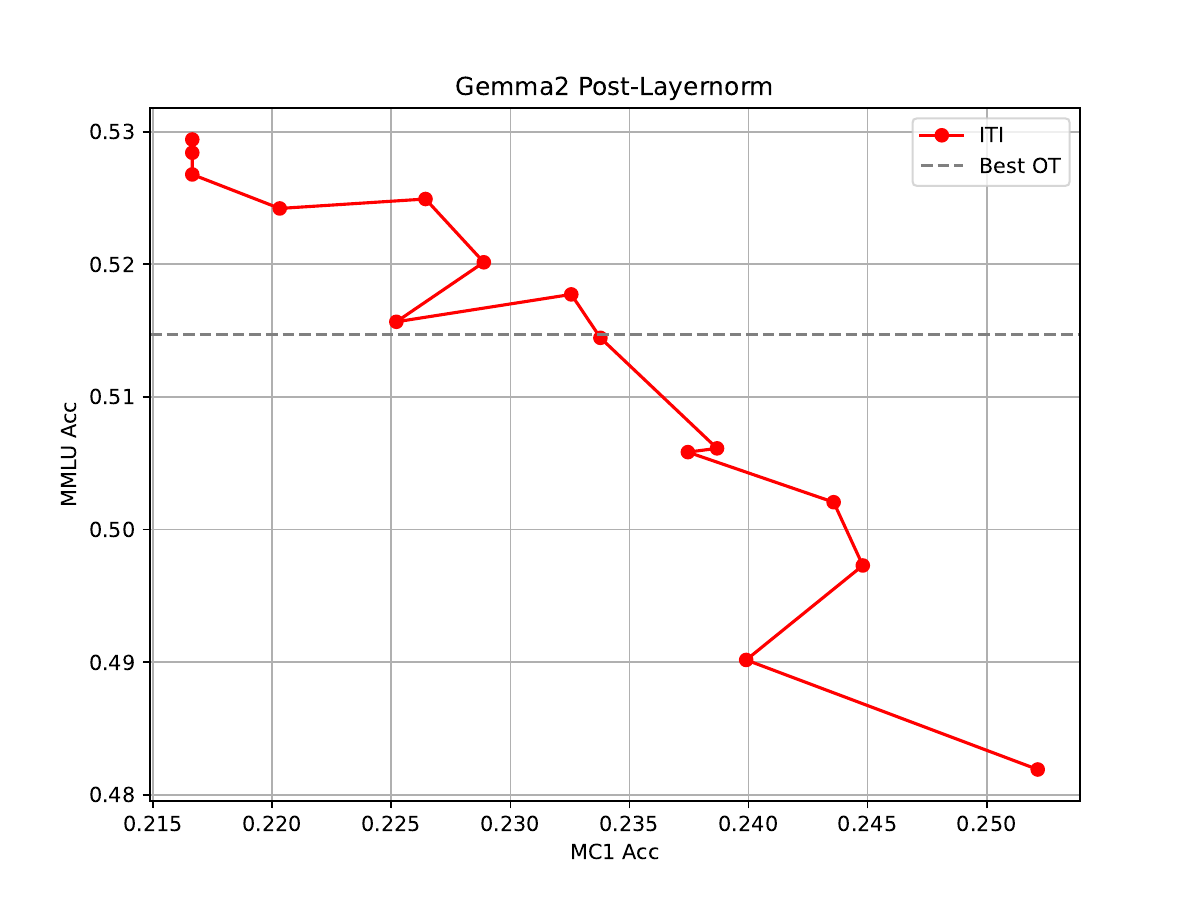}
            \includegraphics[width=0.45\linewidth]{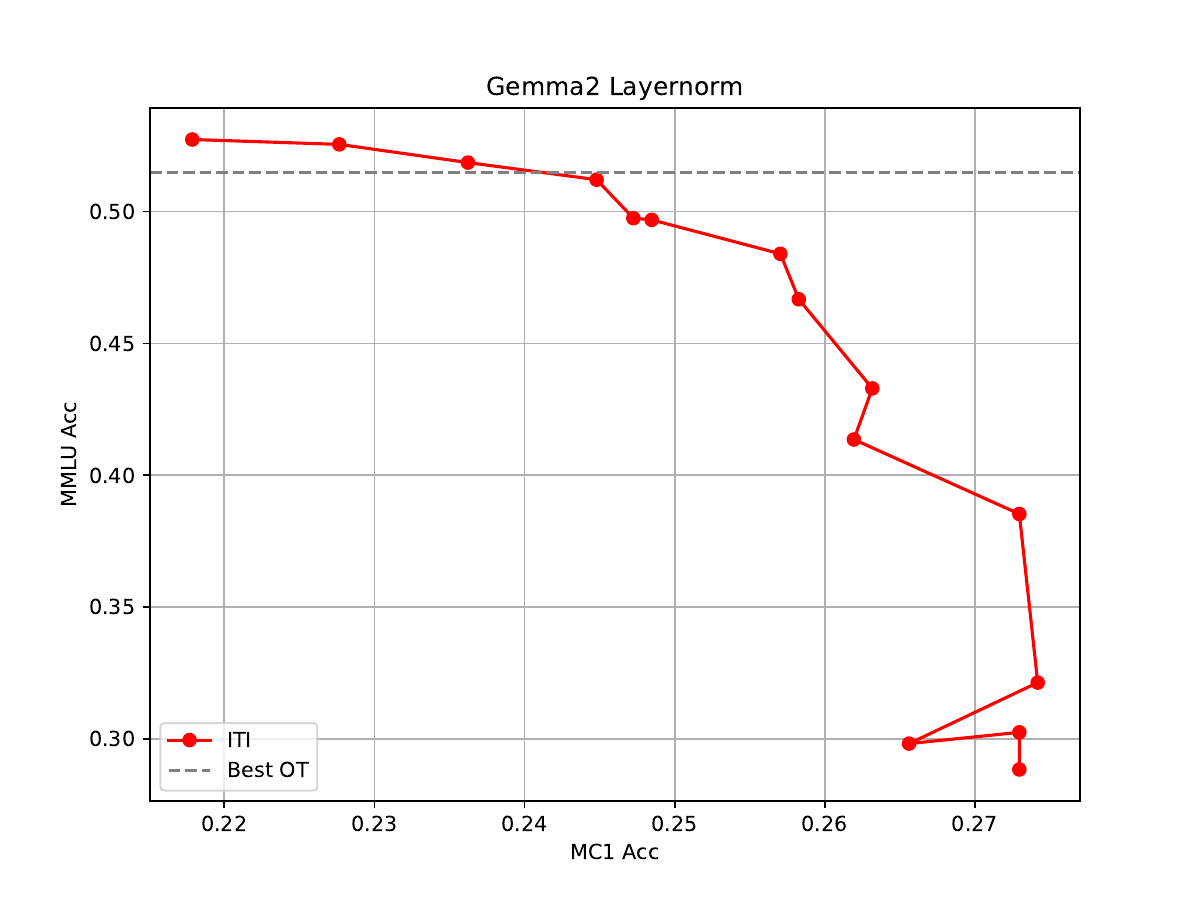}
    \caption{Sweeping $\lambda$ for inducing truthfulness with \iti on \gemmatwob. Left endpoint of line is $\lambda=1.0$, right endpoint of line is $\lambda=15.0$ (each point increasing $\lambda$ by $1.0$). Note this is for $1$ seed only.}
    \label{fig:gemma2_iti_sweep}
\end{figure}

\begin{figure}[t]
    \centering
            \includegraphics[width=0.45\linewidth]{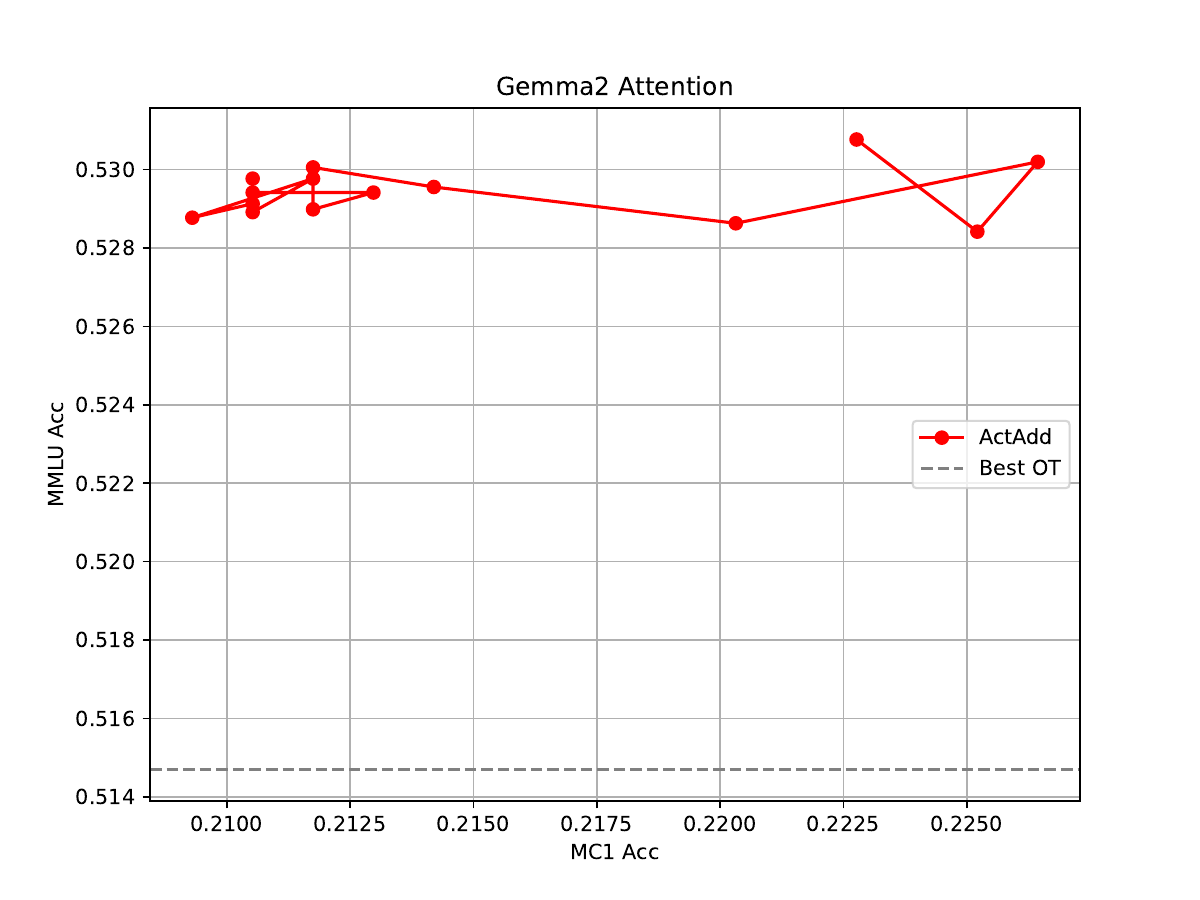}
            \includegraphics[width=0.45\linewidth]{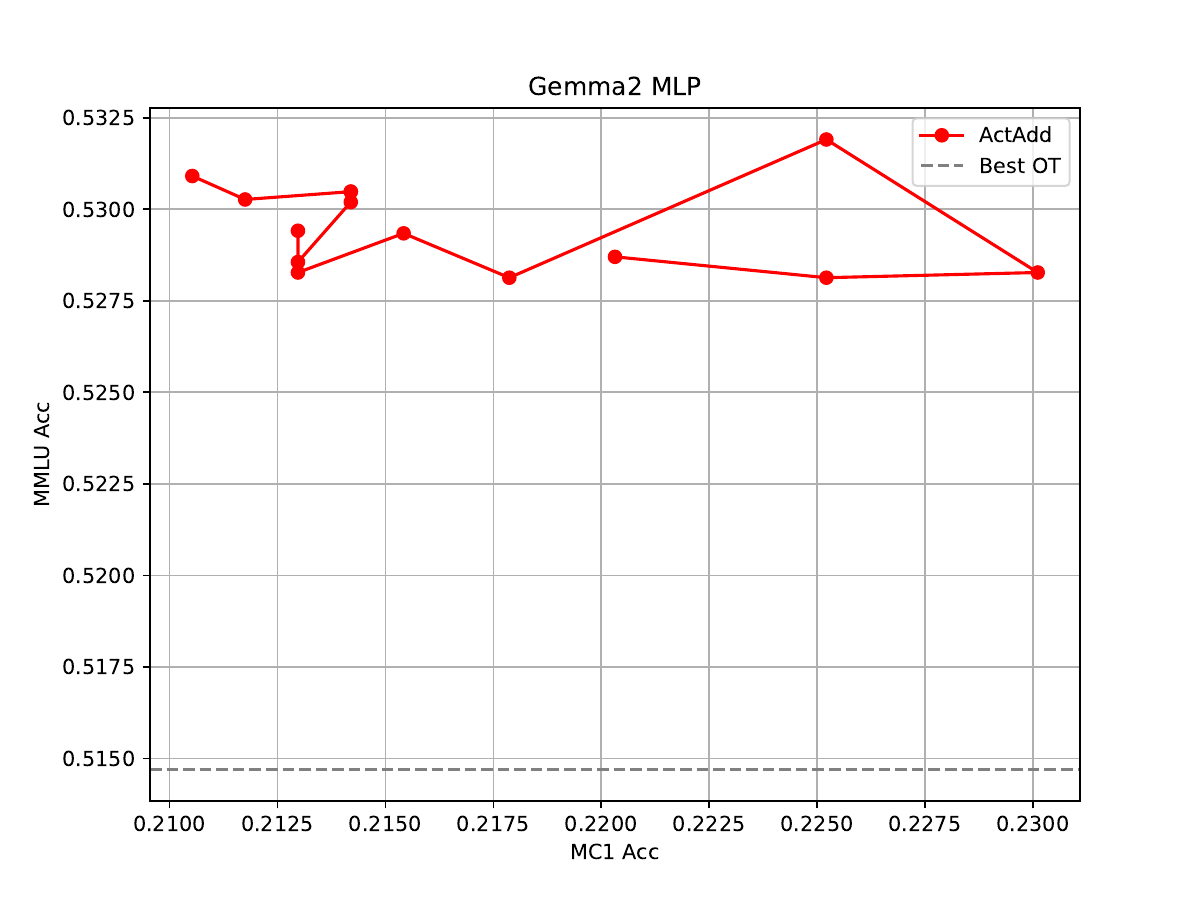} \\
            \includegraphics[width=0.45\linewidth]{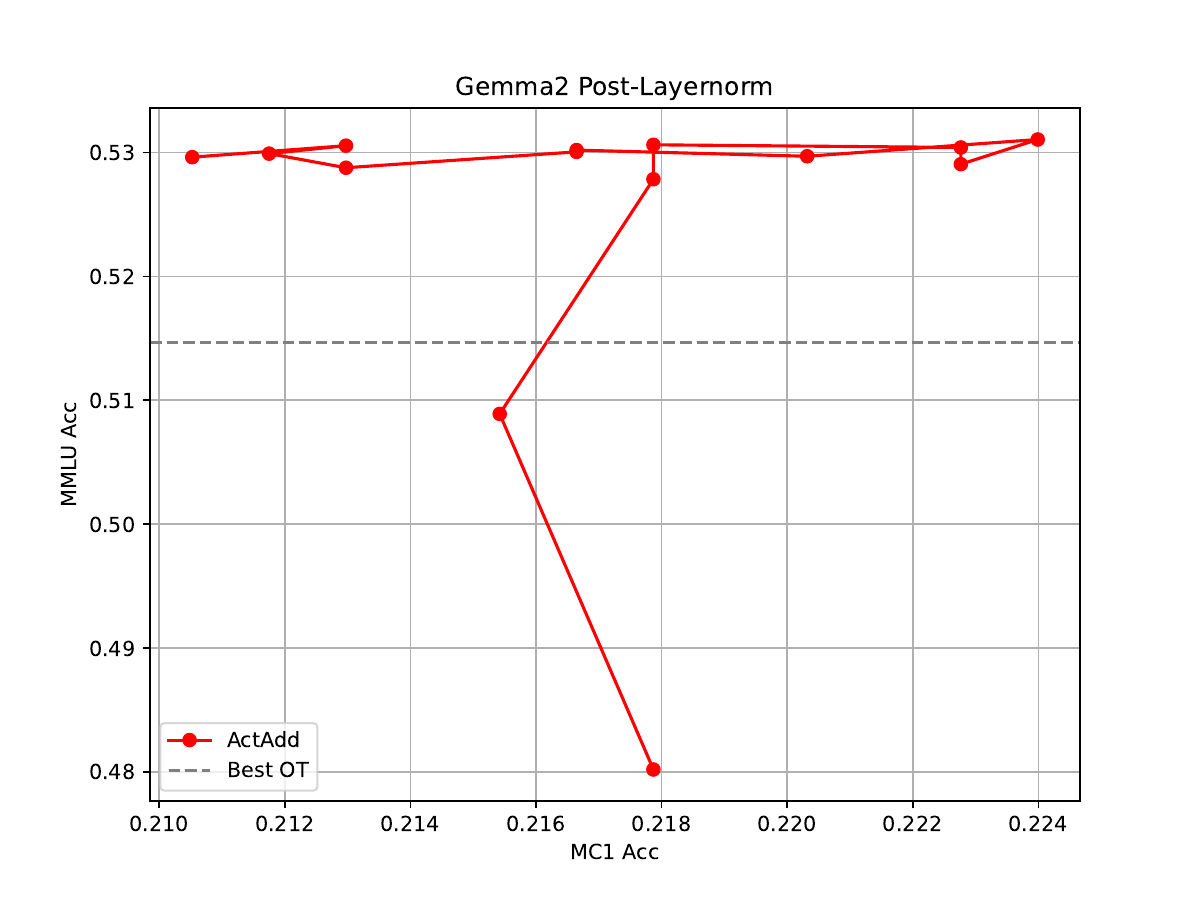}
            \includegraphics[width=0.45\linewidth]{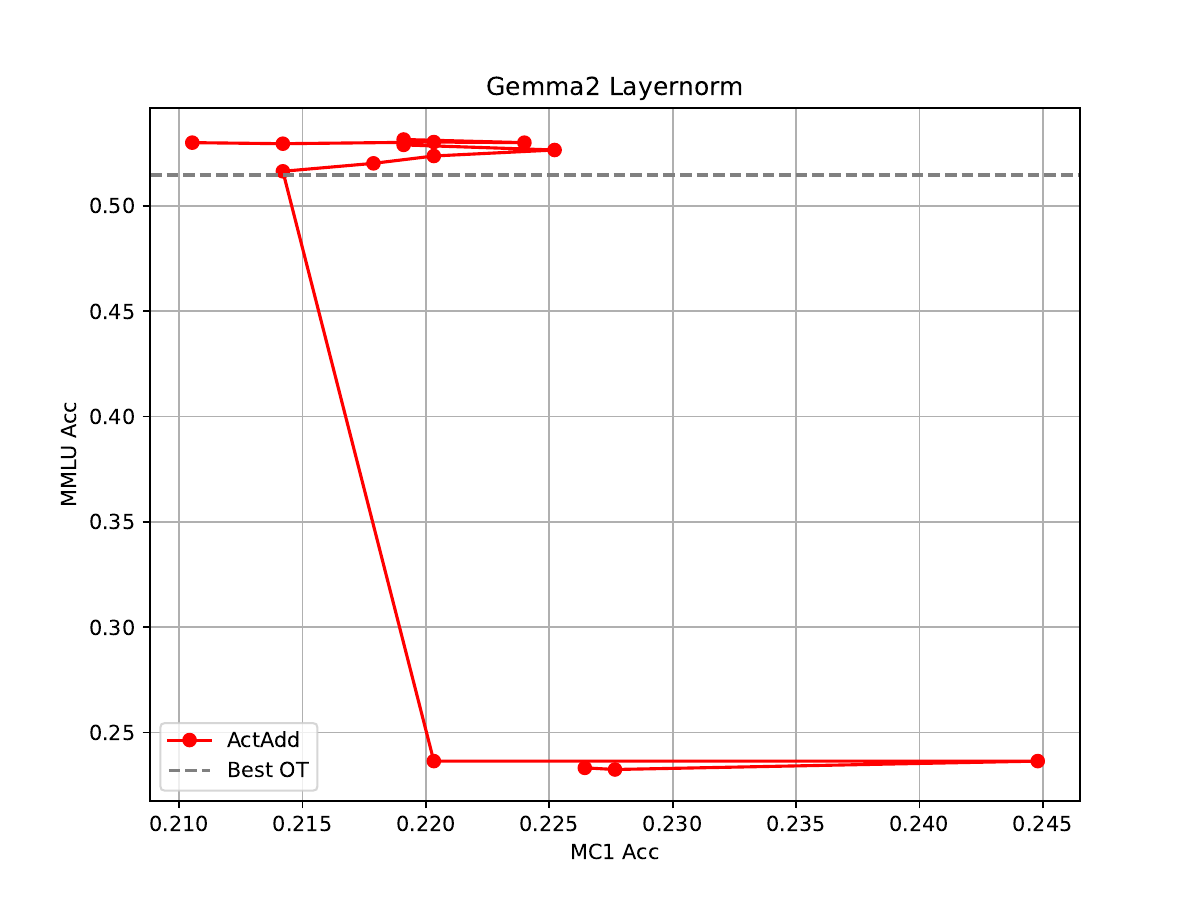}
    \caption{Sweeping $\lambda$ for inducing truthfulness with \actadd on \gemmatwob. Left endpoint of line is $\lambda=0.1$, right endpoint of line is $\lambda=5.0$ ($\lambda \in [0.1, 0.2, 0.3, 0.4, 0.5, 0.6, 0.7, 0.8, 0.9, 1.0, 2.0, 3.0, 4.0, 5.0]$). Note this is for $1$ seed only.}
    \label{fig:gemma2_actadd_sweep}
\end{figure}

\begin{figure}[t]
    \centering
            \includegraphics[width=0.45\linewidth]{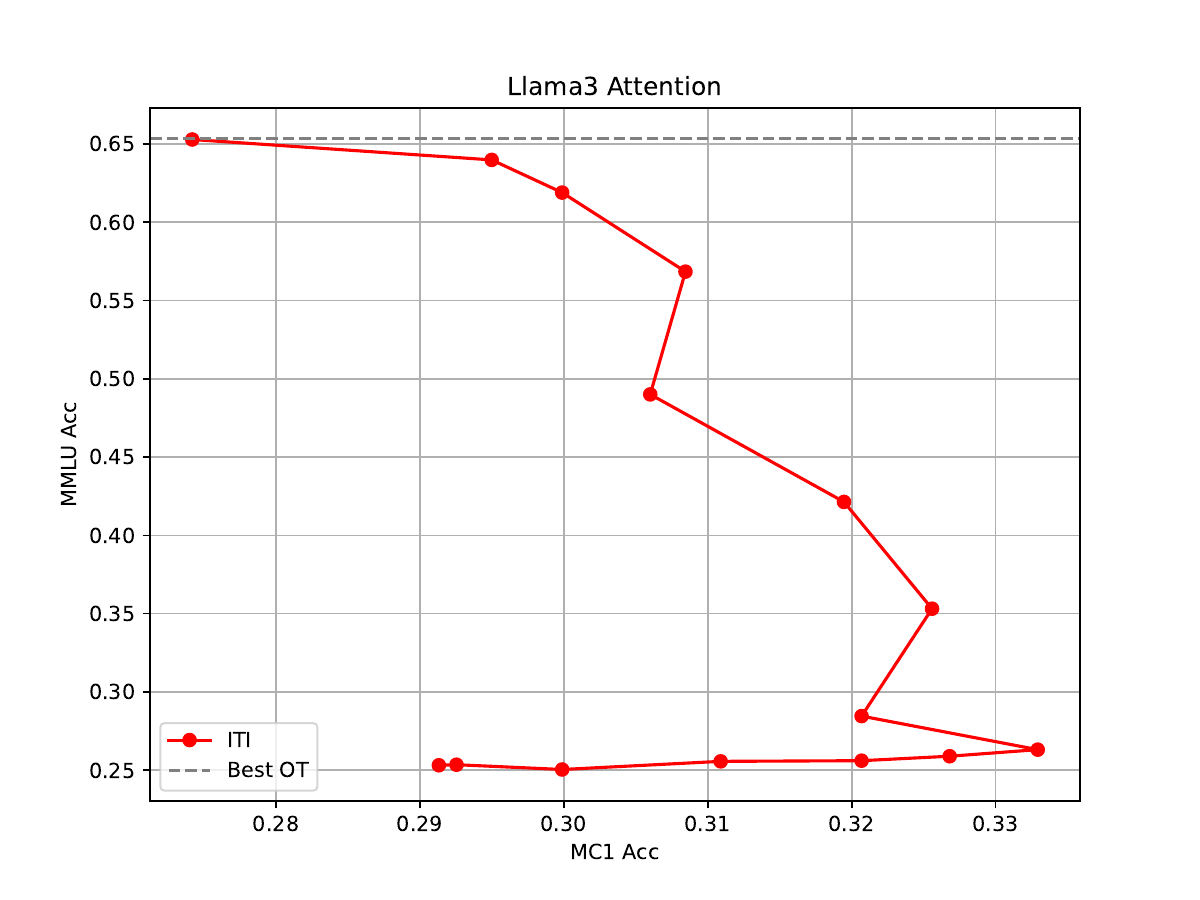}
            \includegraphics[width=0.45\linewidth]{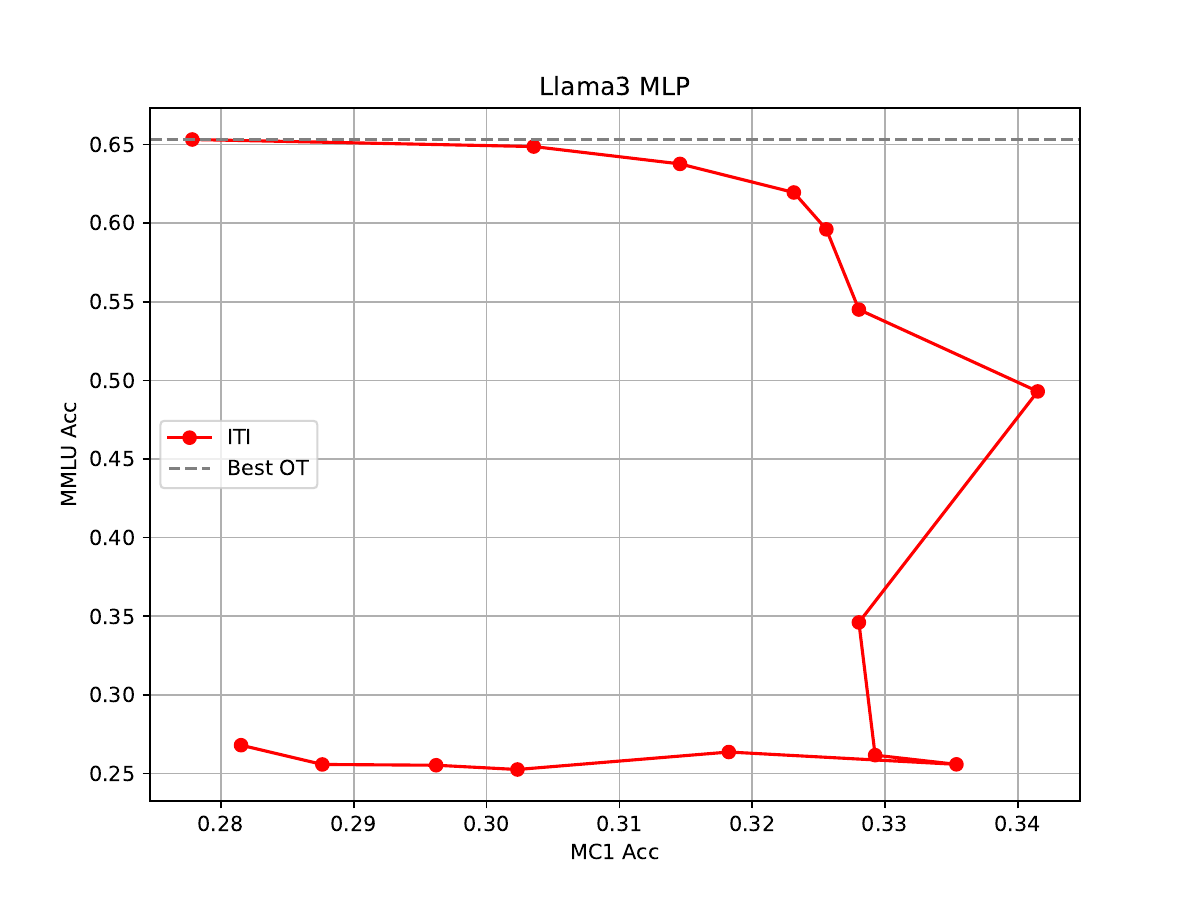} \\
            \includegraphics[width=0.45\linewidth]{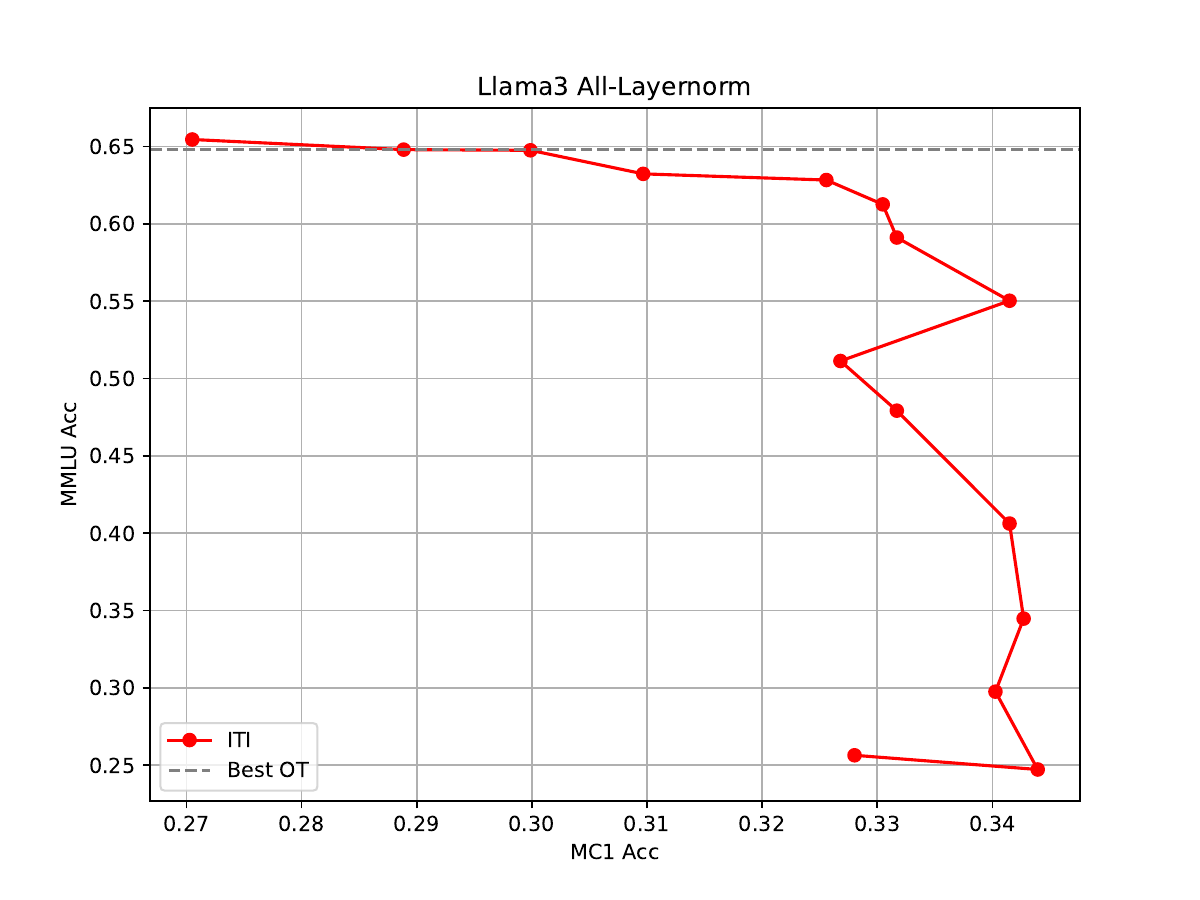}
    \caption{Sweeping $\lambda$ for inducing truthfulness with \iti on \llamaeightb. Left endpoint of line is $\lambda=1.0$, right endpoint of line is $\lambda=15.0$ (each point increasing $\lambda$ by $1.0$). Note this is for $1$ seed only.}
    \label{fig:llama3_iti_sweep}
\end{figure}

\begin{figure}[t]
    \centering
            \includegraphics[width=0.45\linewidth]{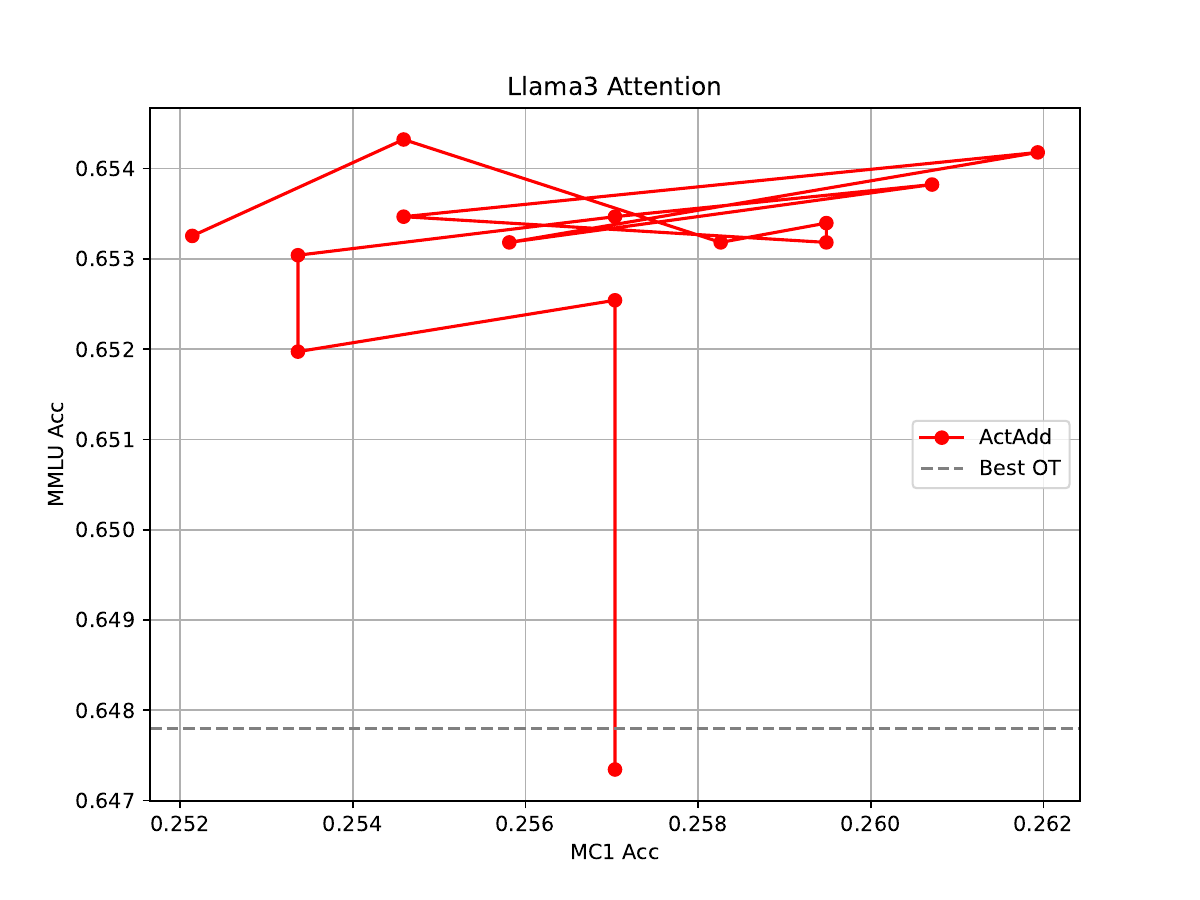}
            \includegraphics[width=0.45\linewidth]{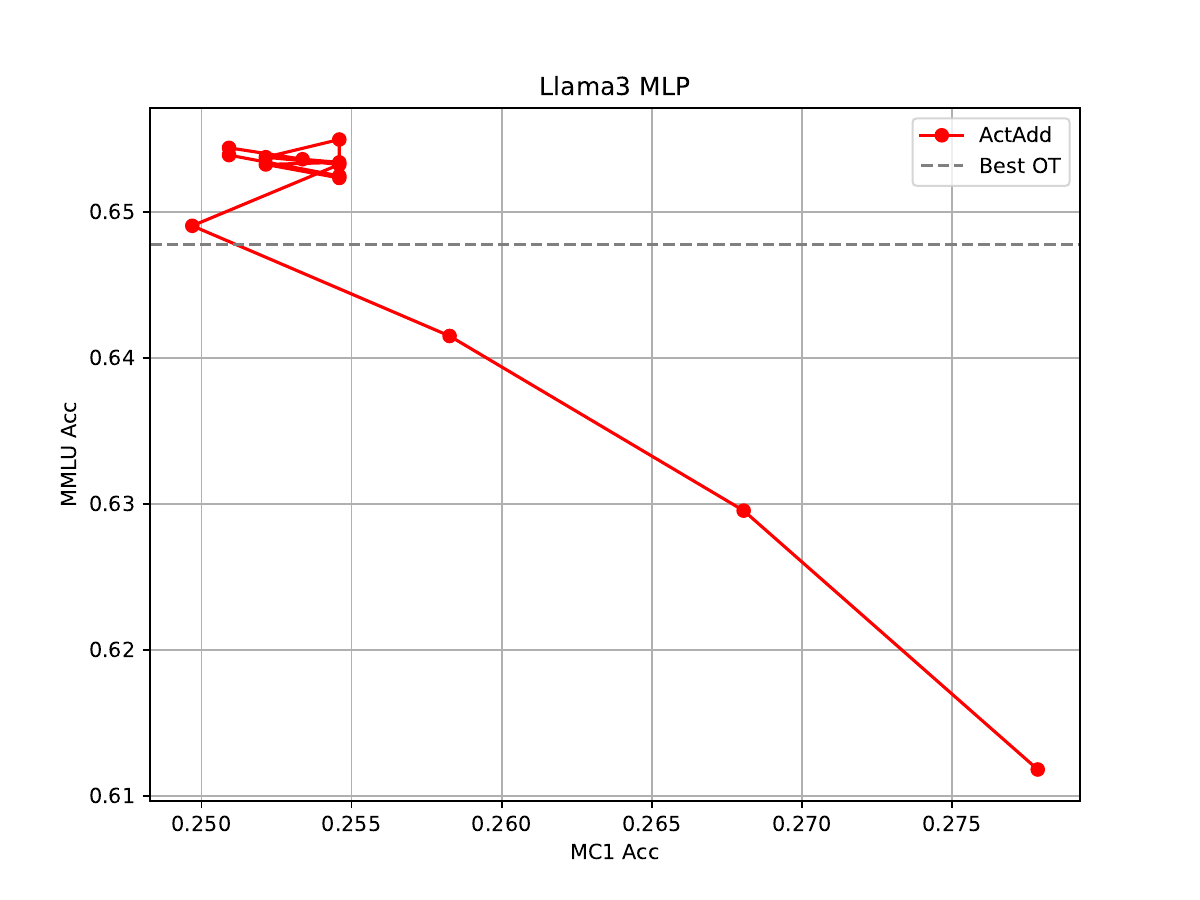} \\
            \includegraphics[width=0.45\linewidth]{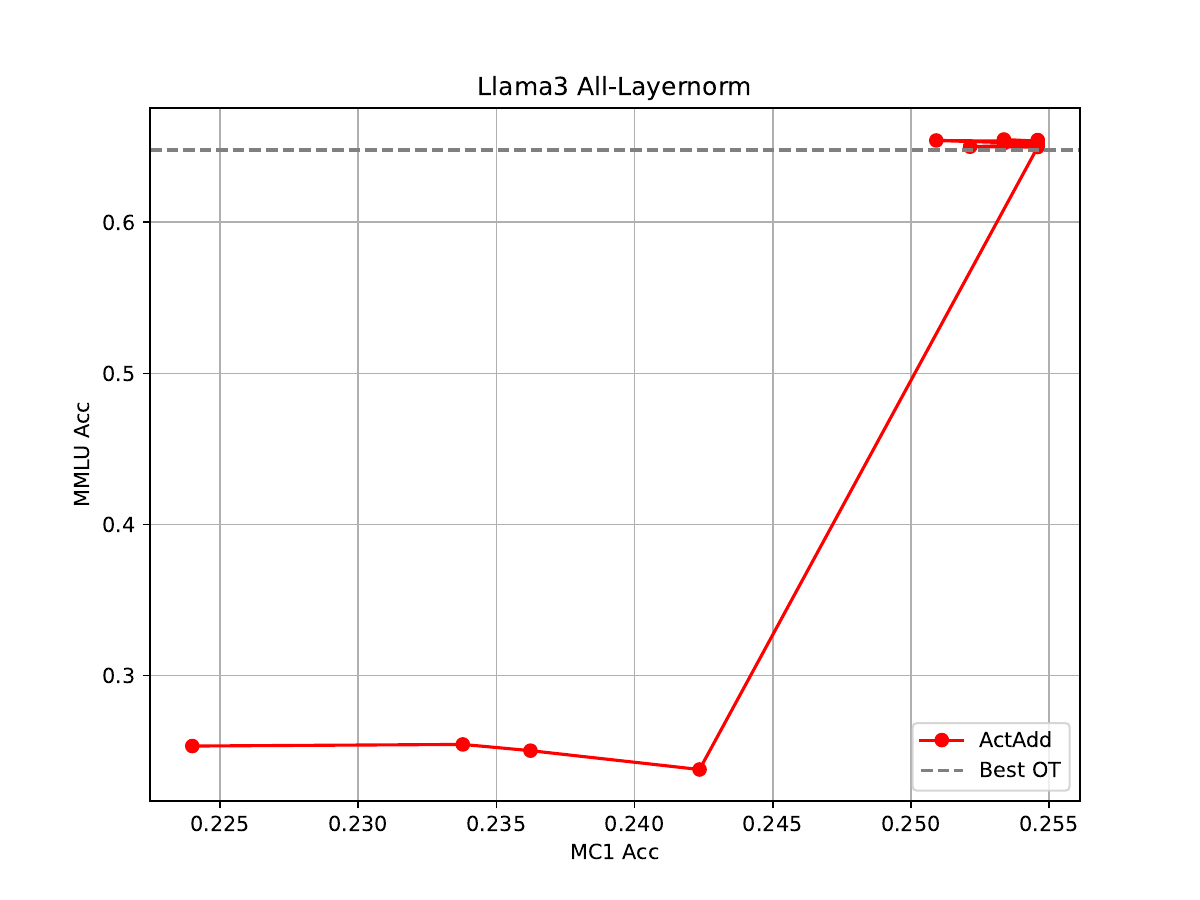}
    \caption{Sweeping $\lambda$ for inducing truthfulness with \actadd on \llamaeightb. Left endpoint of line is $\lambda=0.1$, right endpoint of line is $\lambda=5.0$ ($\lambda \in [0.1, 0.2, 0.3, 0.4, 0.5, 0.6, 0.7, 0.8, 0.9, 1.0, 2.0, 3.0, 4.0, 5.0]$). Note this is for $1$ seed only.}
    \label{fig:llama3_actadd_sweep}
\end{figure}

In Figures \ref{fig:gemma2_iti_sweep} - \ref{fig:llama3_actadd_sweep}, we show the results of sweeping the value of $\lambda$ for \iti and \actadd for both \gemmatwob and \llamaeightb. For each model, we also indicate the MMLU accuracy of the best \method method for that model with a horizontal grey dashed line, as this is our point of reference for choosing $\lambda$ for \iti and \actadd: we choose the value of $\lambda$ that achieves the best MC1 accuracy, while achieving at least equal MMLU accuracy to this grey dotted line (up to a slack of $0.1 \%$).

For \iti, where we see a clear relationship between MMLU and MC1 accuracy as $\lambda$ varies, we sweep $\lambda \in [1.0, 2.0, 3.0, 4.0, 5.0, 6.0, 7.0, 8.0, 9.0, 10.0, 11.0, 12.0, 13.0, 14.0, 15.0]$.
For \actadd, where the relationship can be more erratic, we also sweep values $<1.0$. Here, we sweep  $\lambda \in [0.1, 0.2, 0.3, 0.4, 0.5, 0.6, 0.7, 0.8, 0.9, 1.0, 2.0, 3.0, 4.0, 5.0]$. 

Overall we see that $\lambda$ can have a strong impact on performance for \iti, but in a different way for each layer and model. In particular, it can decrease MMLU performance to catastrophic levels (more than halving performance on \gemmatwob for MLP layers and on \llamaeightb for both attention and MLP layers), making it necessary to sweep $\lambda$ to find its value that provides a reliable control method using \iti for the problem at hand. Similar things can be found about \actadd (e.g. when interventing upon on all Layernorm layers on \gemmatwob, Figure \ref{fig:gemma2_actadd_sweep}).

\clearpage
\FloatBarrier
\section{Experimental Details and Extended Results for T2I generation}
\label{app:t2i}
\Cref{app:guidance} illustrates the effect of the guidance parameter in SDXL. \Cref{app:negative-prompts} illustrates the problem of concept negation when using negative prompts.  \Cref{app:style-control} contains additional qualitative examples of style control on SDXL and FLUX. \Cref{app:concept-negation} contains additional qualitative examples for concept negation in SDXL and FLUX. \Cref{app:style-prompts,app:concept-prompts} contain the list of tags used as prompt modifiers to generate the target/source distribution of activations for each style/concept respectively. \Cref{app:flux-details} contains details on FLUX's architecture conditioning.
\FloatBarrier
\subsection{Guidance Parameter in Existing Diffusion Models}
\label{app:guidance}

\begin{figure}[htb!]
    \begin{tikzpicture}
			\tikzstyle{lambdablock} = [rectangle,font=\footnotesize,rounded corners=1pt,fill=white,text opacity=1,fill opacity=0.5,inner sep=0.5pt,scale=0.7];
			
			\node[anchor=north west, inner sep=0pt] (img1)  at (1,0)
			{\includegraphics[width=0.95\textwidth]{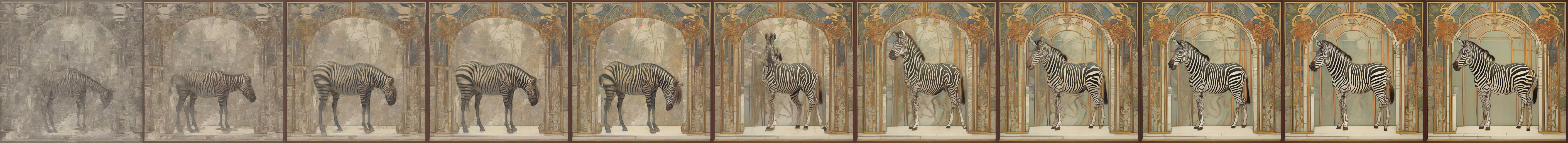}};
                                          
            \foreach \v/\i in {1.0/0, 1.5/1, 2.0/2, 2.5/3, 3.0/4, 3.5/5, 4.0/6, 4.5/7, 5.0/8, 5.5/9, 6.0/10}
			{
				\node[lambdablock] at (0.086\textwidth*\i + 35, -1.1)  {$\v$};
			}    
            
		\end{tikzpicture}
    \caption{SDXL with \textit{art nouveau} tags  appended to the prompt as described in  \Cref{app:style-control} and  guidance strength linearly increasing from 1 to 6. Note how for low guidance (left most images) the semantic content is almost completely lost.}
    \label{fig:guidance_scale}
\end{figure}

We show in \Cref{fig:guidance_scale} the effect of changing the guidance scale parameter in SDXL. While large values lead to effective conditioning, lower values destroy content. This makes guidance non intuitive and harder to use by users.

\FloatBarrier
\subsection{Negative prompting}
\label{app:negative-prompts}

\begin{figure}[h!]
    \centering
    \begin{subfigure}[t]{0.30\linewidth}
    \includegraphics[width=\linewidth]{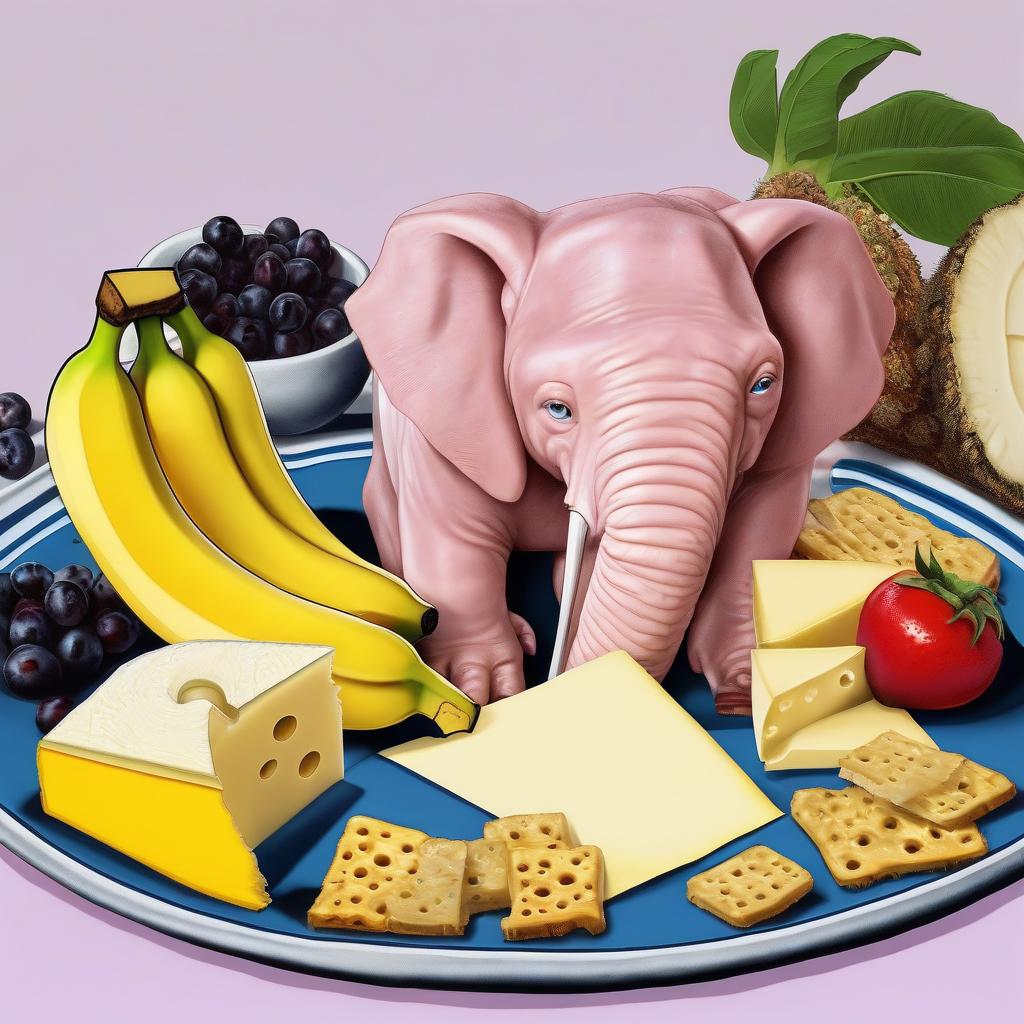}
    \end{subfigure}
    \begin{subfigure}[t]{0.30\linewidth}
    \includegraphics[width=\linewidth]{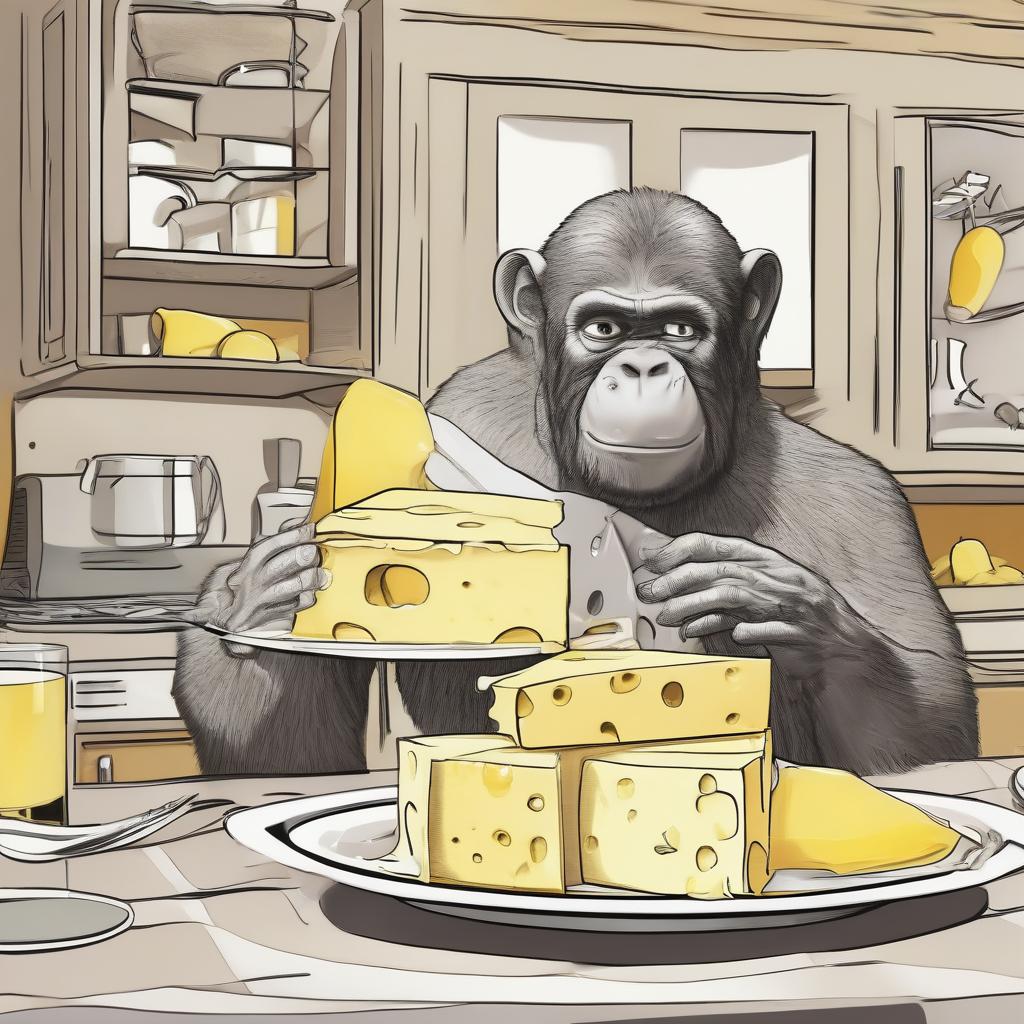}
    \end{subfigure}
    \begin{subfigure}[t]{0.30\linewidth}
    \includegraphics[width=\linewidth]{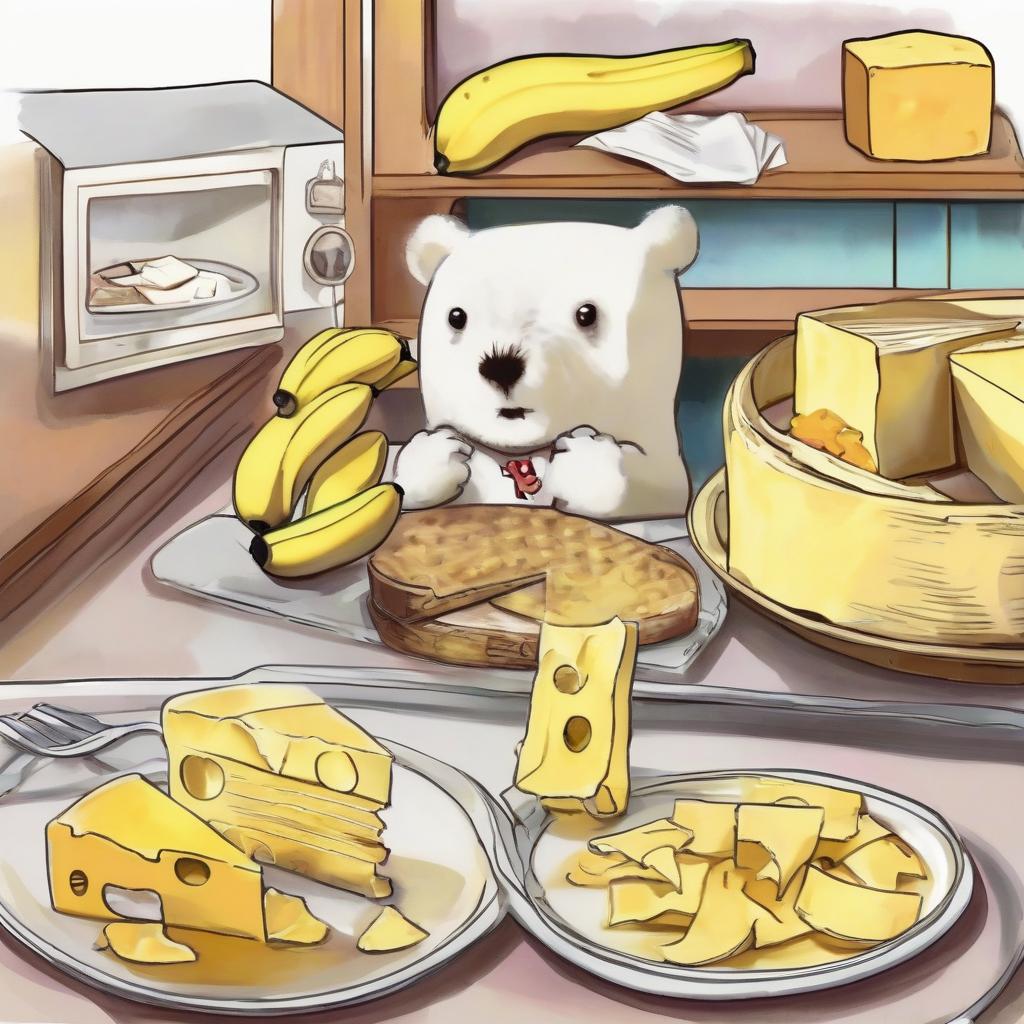}
    \end{subfigure}
    \caption{\textbf{SDXL with Negative Prompt.} Prompt: ``There is a banana and two pieces of cheese on a plate. A \texttt{\{pink elephant, gorilla, white bear\}} cannot be seen anywhere.''. Negative prompt: ``A \texttt{\{pink elephant, gorilla, white bear\}}''.}
    \label{fig:negative-prompt-sdxl}
\end{figure}

\begin{figure}[h!]
    \centering
    \begin{subfigure}[t]{0.3\linewidth}
    \includegraphics[width=\linewidth]{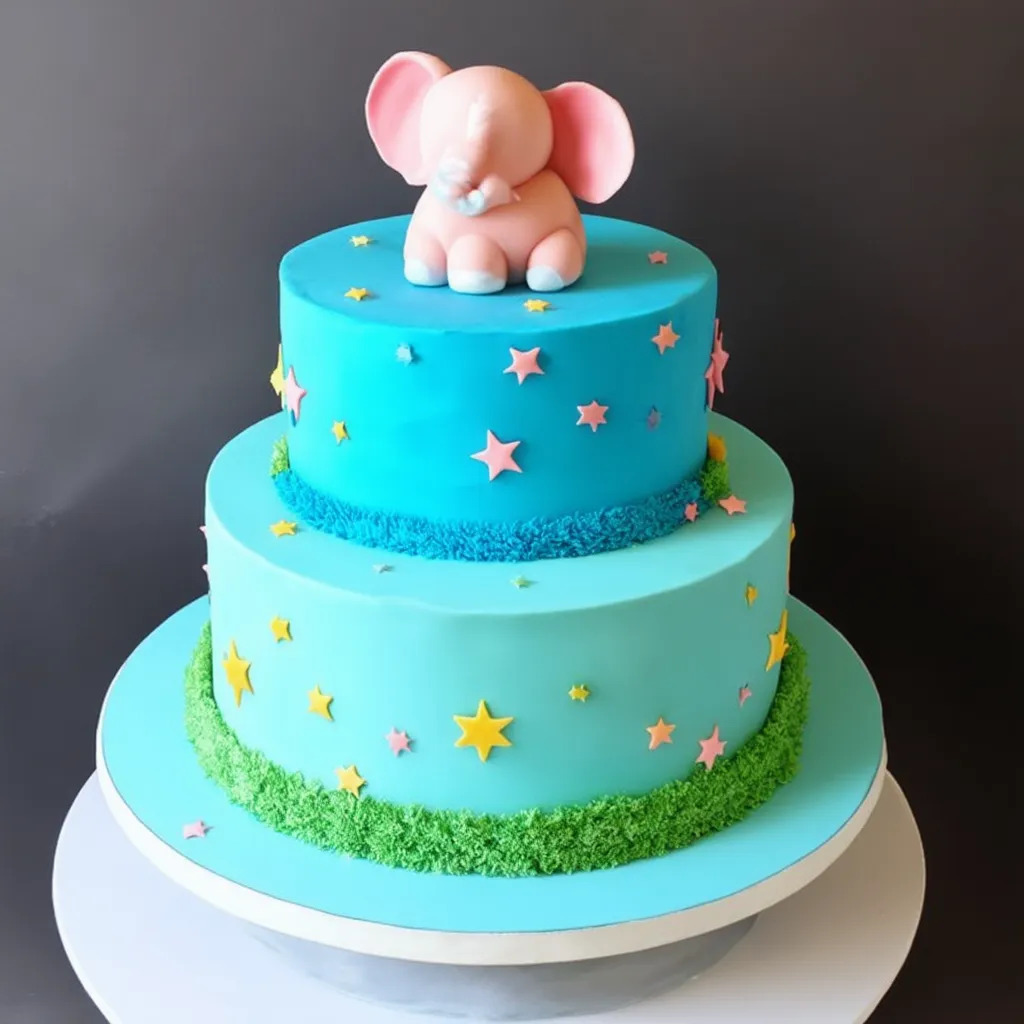}
    \end{subfigure}
    \begin{subfigure}[t]{0.3\linewidth}
    \includegraphics[width=\linewidth]{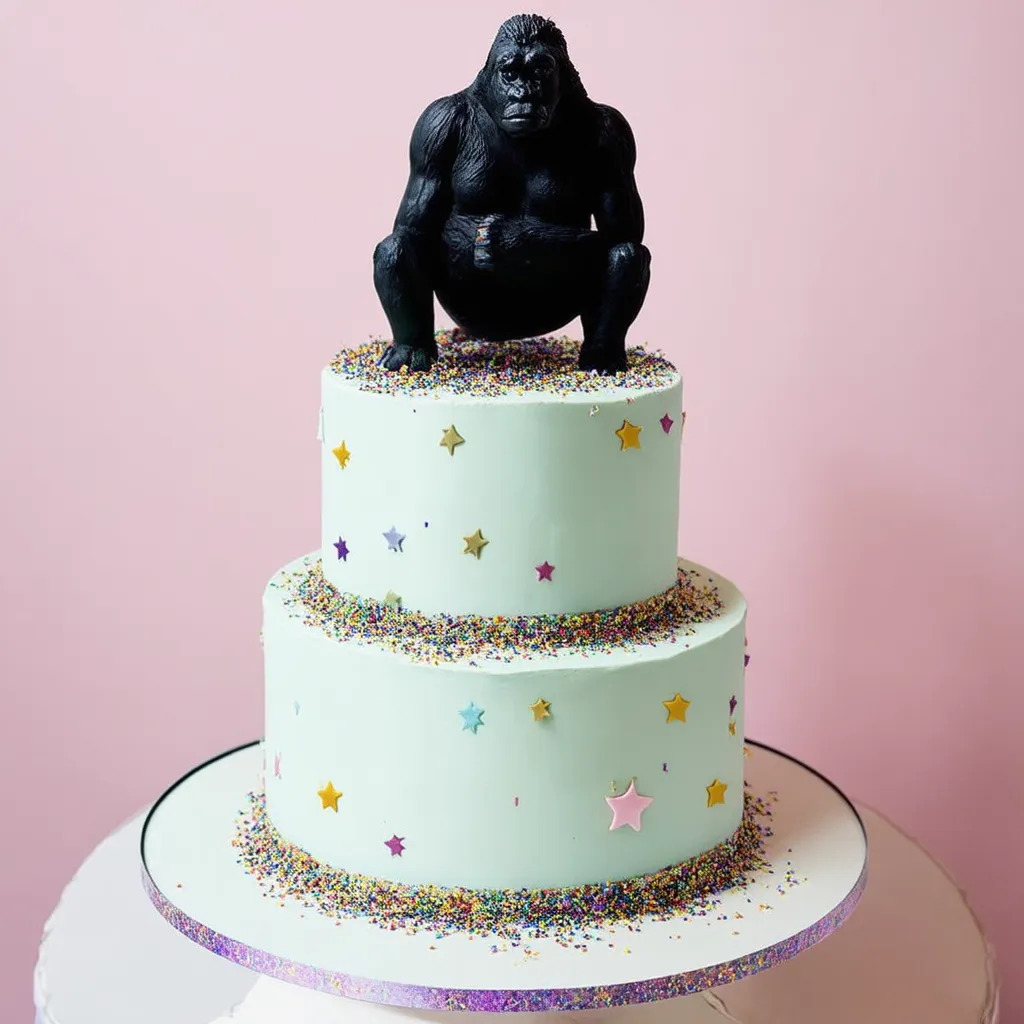}
    \end{subfigure}
    \begin{subfigure}[t]{0.3\linewidth}
    \includegraphics[width=\linewidth]{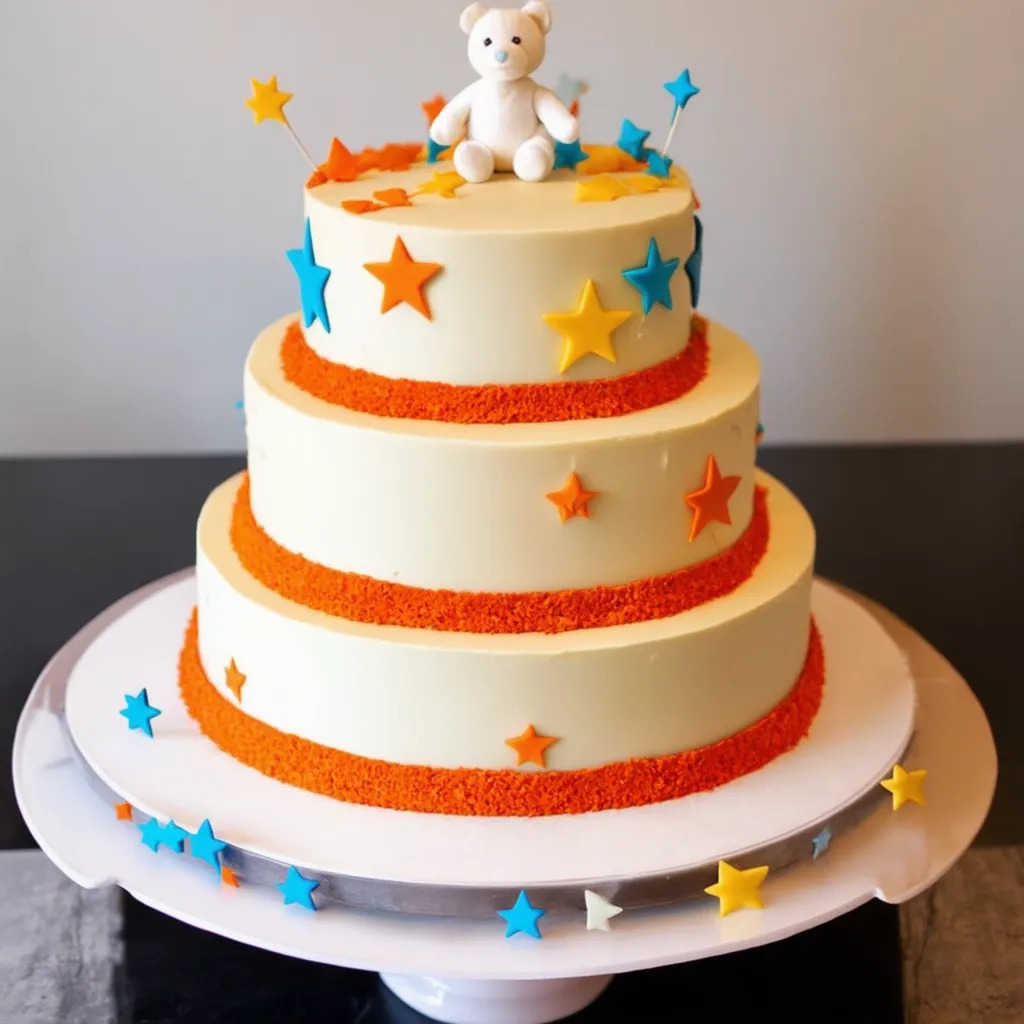}
    \end{subfigure}
    \caption{\textbf{Stable Diffusion 3 with Negative Prompt.} Prompt: ``2 tier cake with multicolored stars attached to it. A \texttt{\{pink elephant, gorilla, white bear\}} cannot be seen anywhere.'' Negative prompt: ``A \texttt{\{pink elephant, gorilla, white bear\}}.''.}
    \label{fig:negative-prompt-SD3}
\end{figure}

Stable diffusion models allow using negative prompts to avoid unwanted elements in the generated images~\citep{rombach2022high,podellsdxl}. Here, we show that this method is ineffective at removing \textit{pink elephant}, \textit{white bear}, and \textit{gorilla}. \Cref{fig:negative-prompt-sdxl,fig:negative-prompt-SD3} contain some failure cases of SDXL and Stable Diffusion 3~\citep{esser2024scaling} at removing unwanted concepts. \Cref{fig:negation-I} and \Cref{fig:negation-II} show results intervening SDXL with \method, showing its effectiveness at removing these concepts with the same prompts. In \Cref{fig:negation-III} we show some failure cases at concept negation.


\FloatBarrier
\subsection{Style Control}
\label{app:style-control}
\begin{figure}[htb!]
     \centering 
     \begin{subfigure}[t]{0.49\linewidth}
     \includegraphics[width=\linewidth]{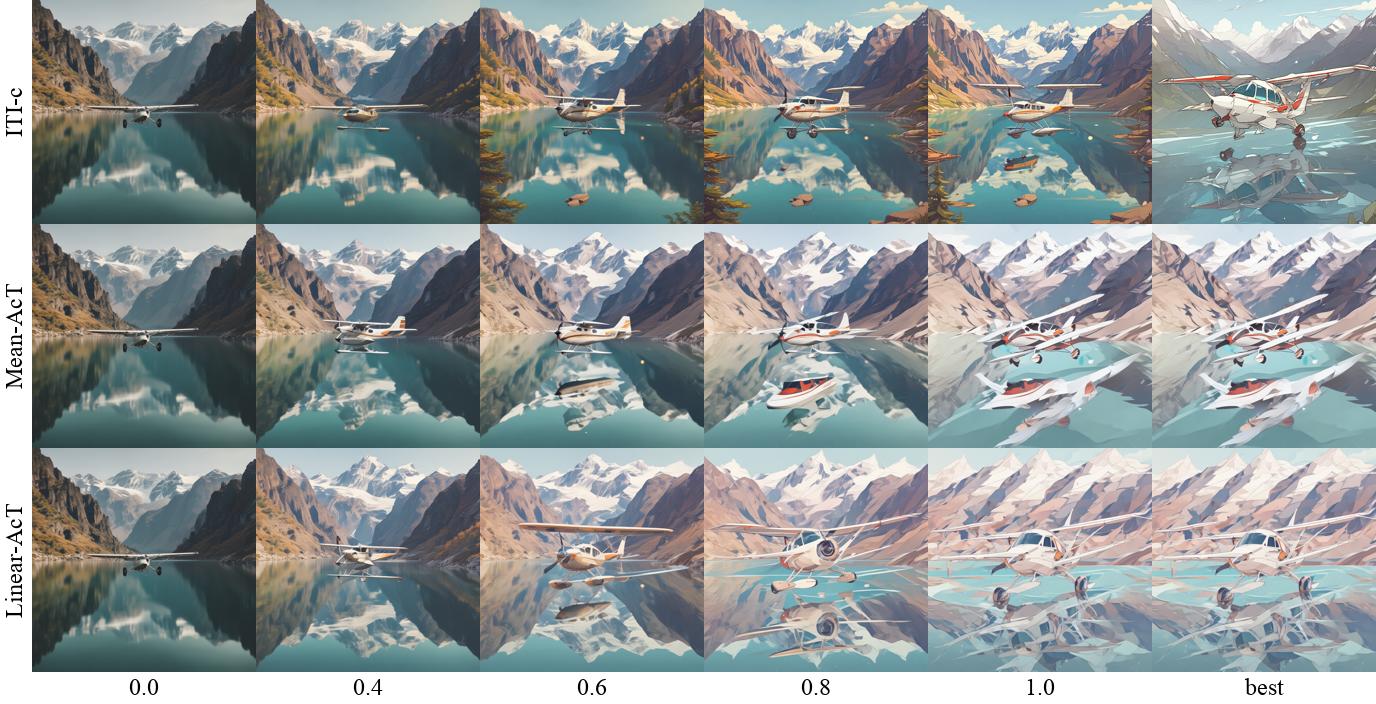}
     \caption{Anime}
     \end{subfigure}
     \begin{subfigure}[t]{0.49\linewidth}
     \includegraphics[width=\linewidth]{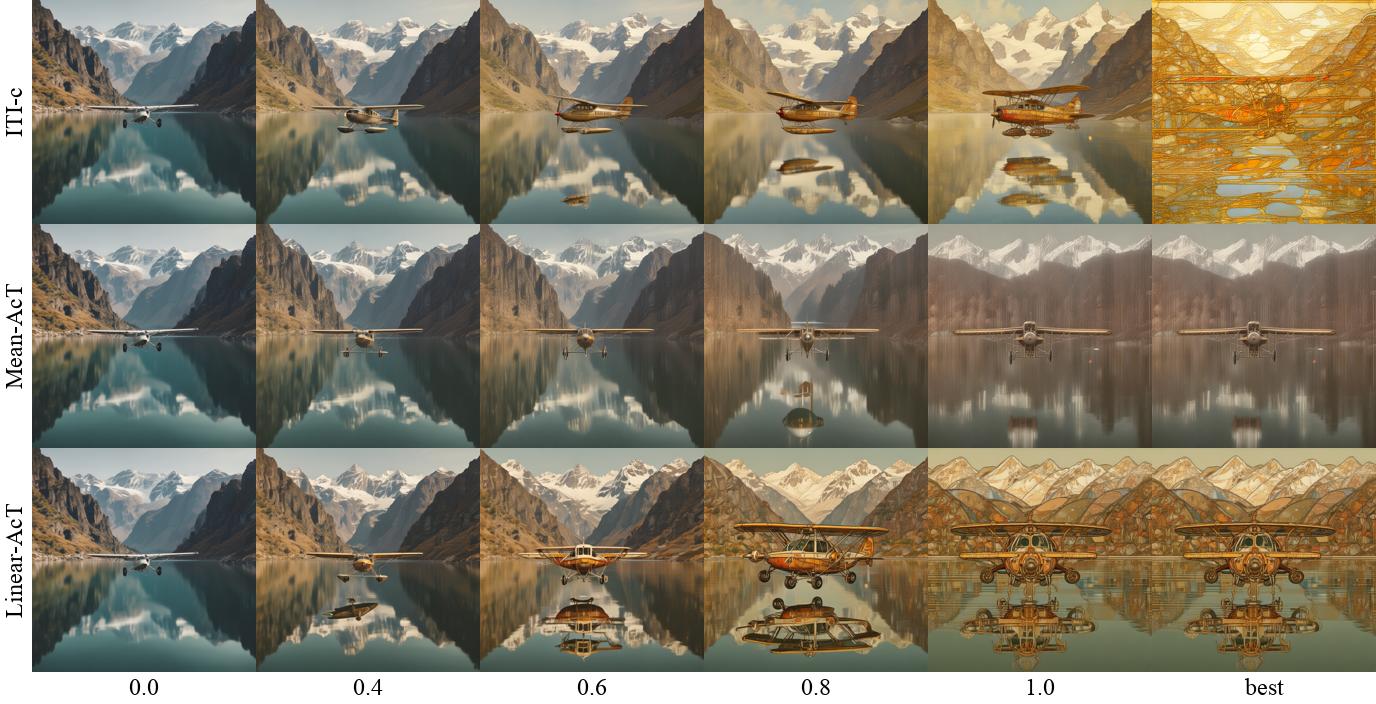}
     \caption{Art Nouveau}
     \end{subfigure}
     \begin{subfigure}[t]{0.49\linewidth}
     \includegraphics[width=\linewidth]{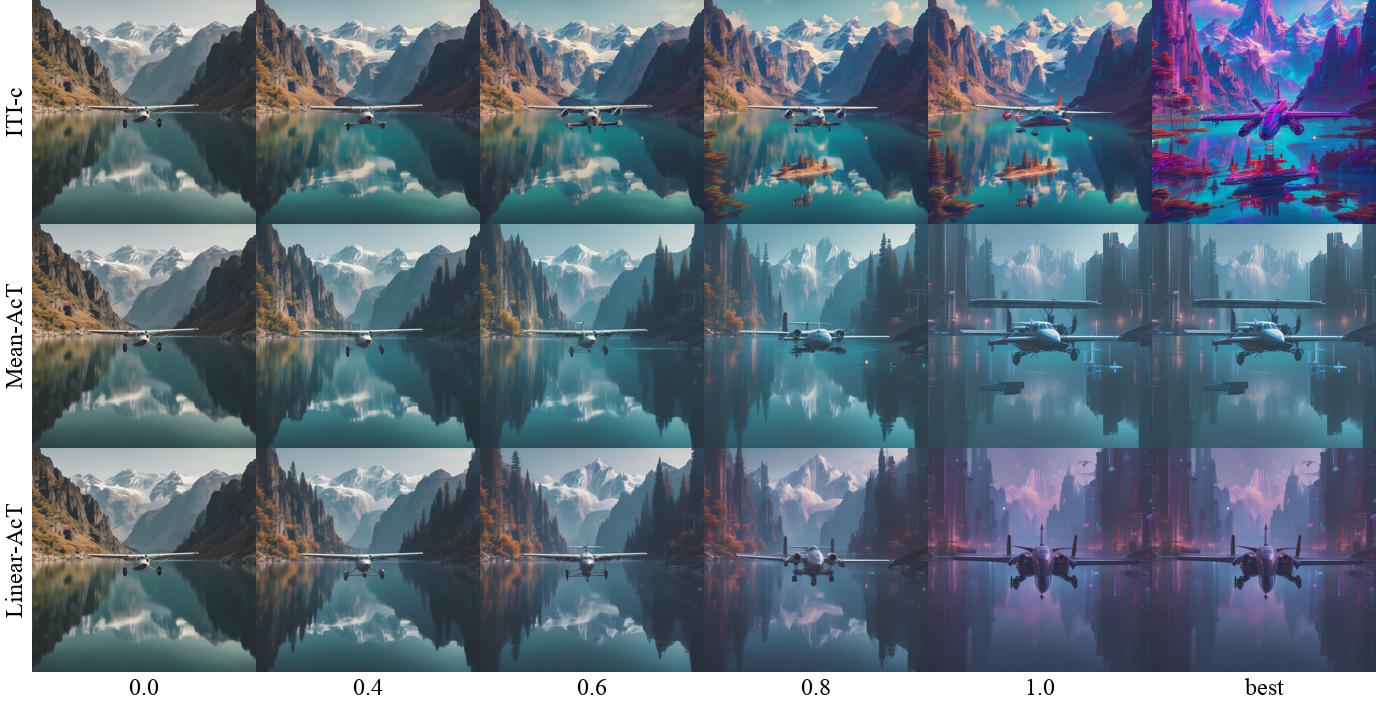}
     \caption{Cyberpunk}
     \end{subfigure}
     \begin{subfigure}[t]{0.49\linewidth}
     \includegraphics[width=\linewidth]{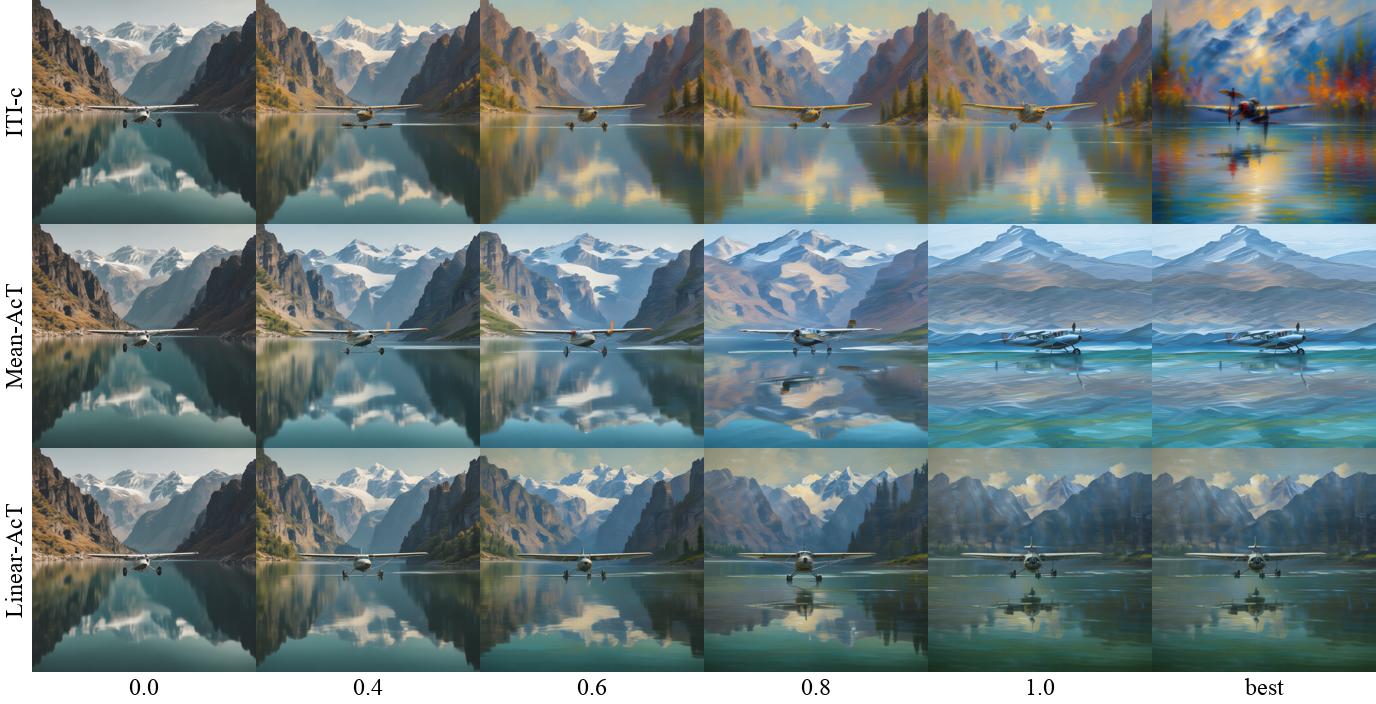}
     \caption{Impressionism}
     \end{subfigure}
     \begin{subfigure}[t]{0.49\linewidth}
     \includegraphics[width=\linewidth]{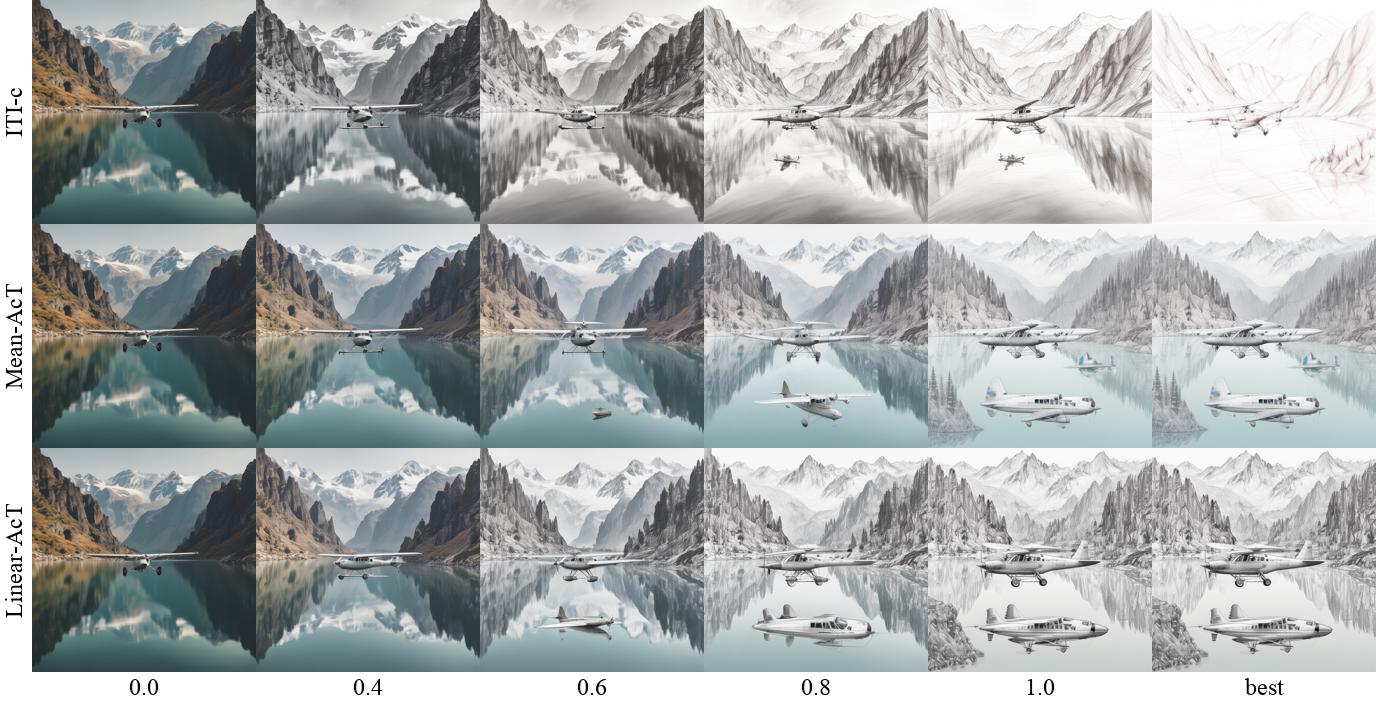}
     \caption{Sketch.}
     \end{subfigure}
     \begin{subfigure}[t]{0.49\linewidth}
     \includegraphics[width=\linewidth]{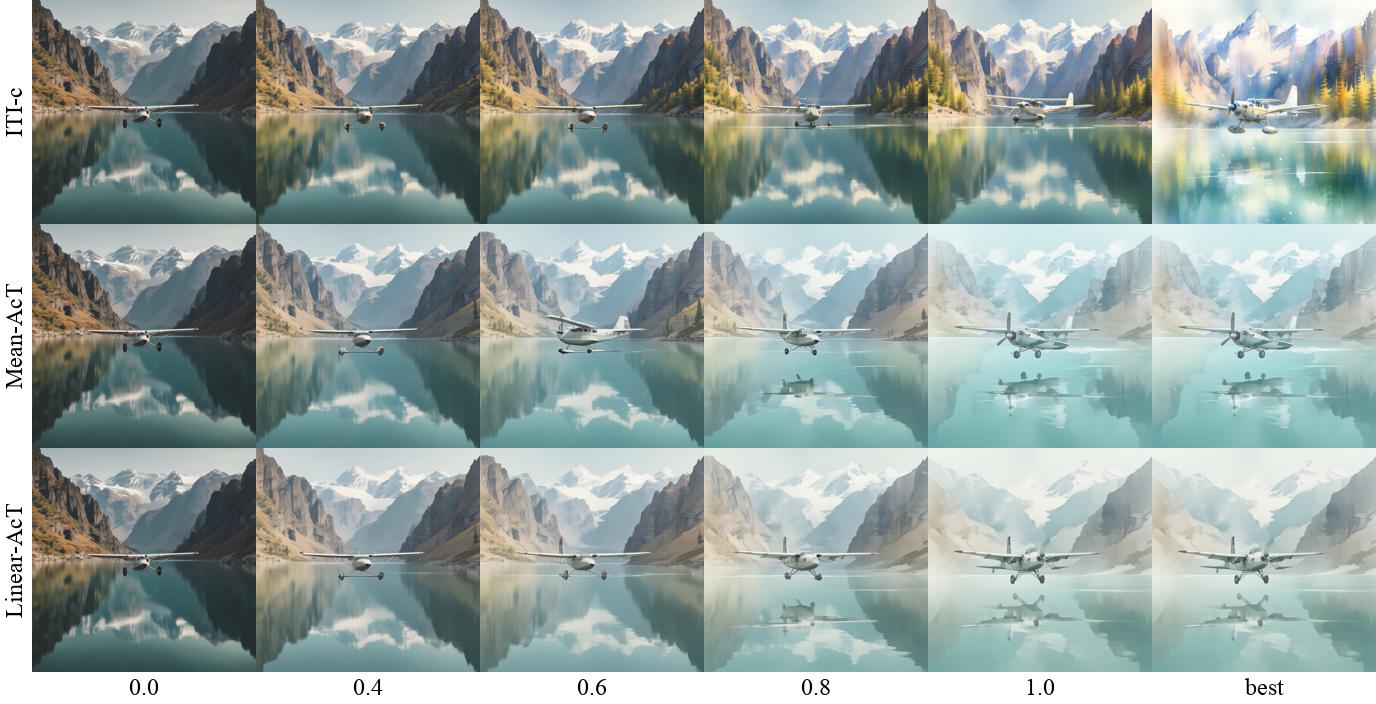}
     \caption{Watercolor}
     \end{subfigure}
\caption{\textbf{SDXL - A plane floating on top of a lake surrounded by mountains.} From left to right conditioning strength $\lambda$ increases from 0 to 1. Rightmost column corresponds to the best strength found in \Cref{fig:clip_score} ($\lambda=1$ for \method and $\lambda=2$ for \iti). \linear succeeds at inducing different styles. \mean fails at inducing \textit{art nouveau}. \iti introduces noise for \textit{art nouveau} and \textit{cyberpunk}.}
\label{fig:style-I}
\end{figure}

\begin{figure}[t]
     \centering 
     \begin{subfigure}[t]{0.49\linewidth}
     \includegraphics[width=\linewidth]{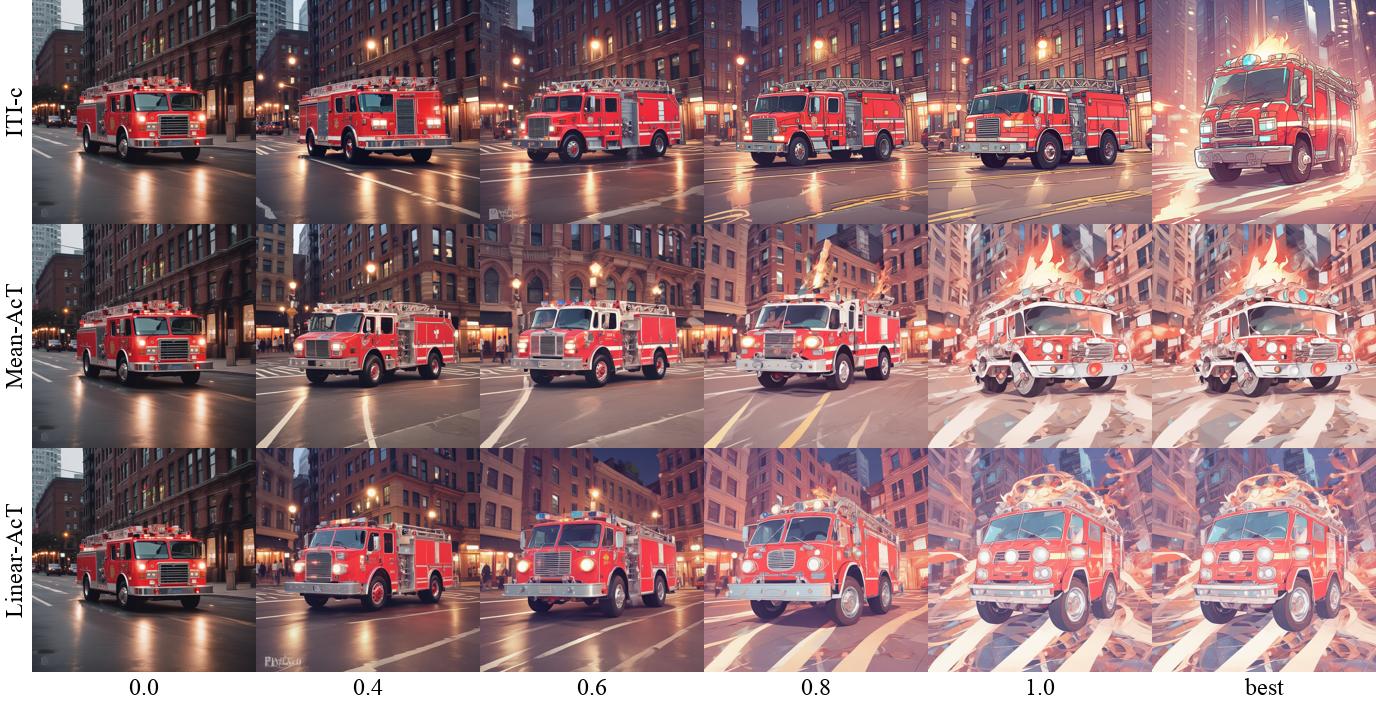}
     \caption{Anime}
     \end{subfigure}
     \begin{subfigure}[t]{0.49\linewidth}
     \includegraphics[width=\linewidth]{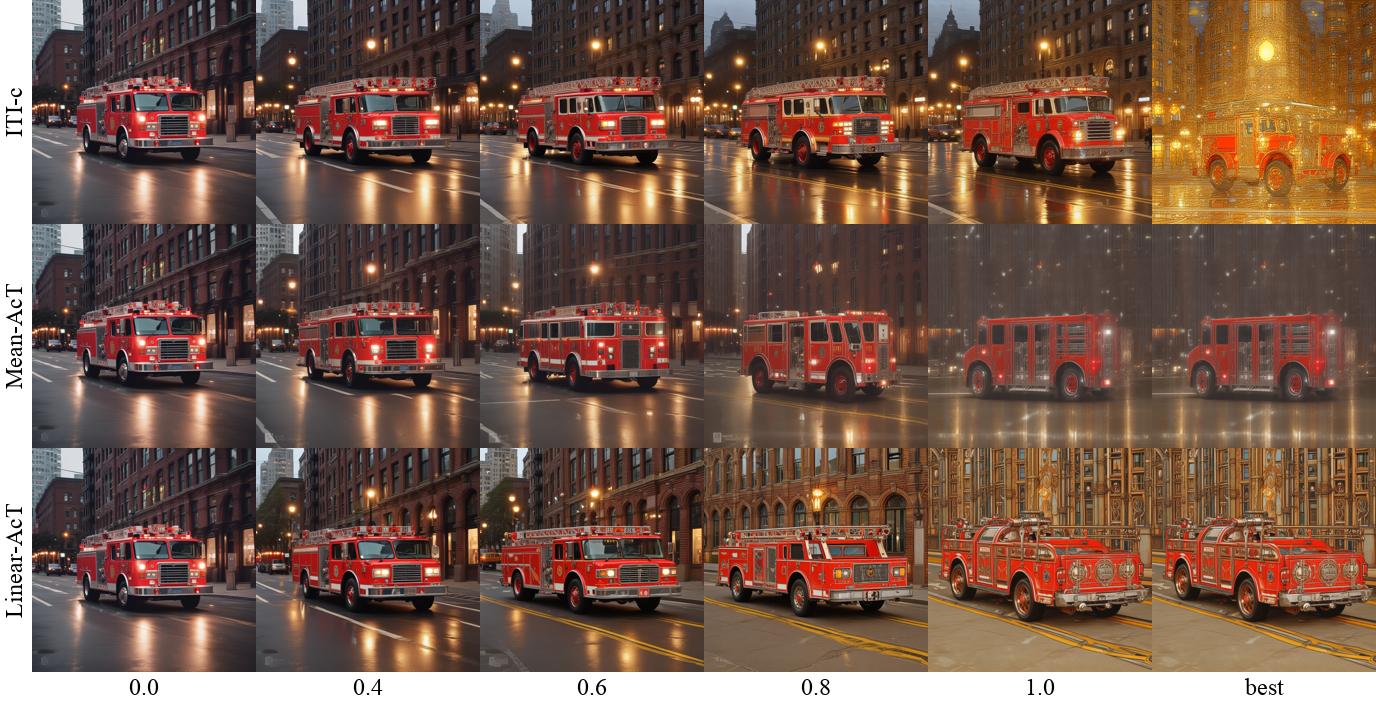}
     \caption{Art Nouveau}
     \end{subfigure}
     \begin{subfigure}[t]{0.49\linewidth}
     \includegraphics[width=\linewidth]{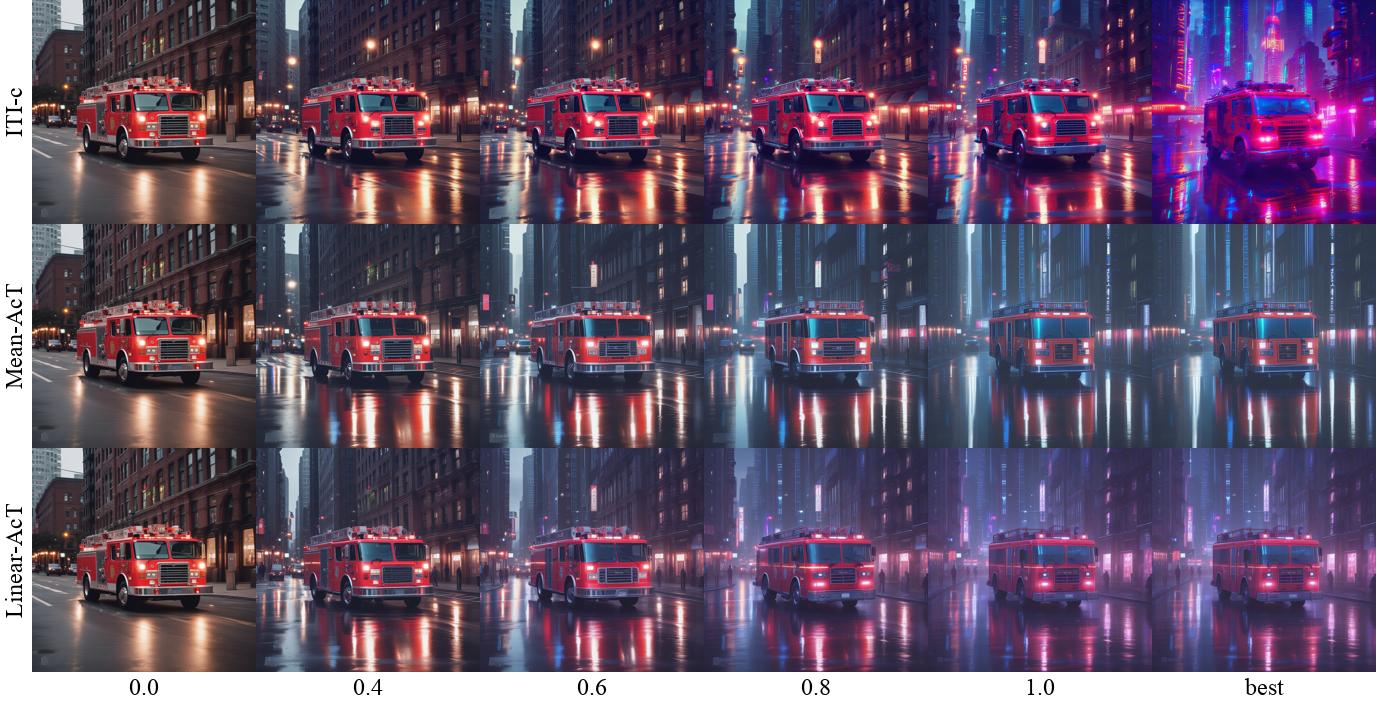}
     \caption{Cyberpunk}
     \end{subfigure}
     \begin{subfigure}[t]{0.49\linewidth}
     \includegraphics[width=\linewidth]{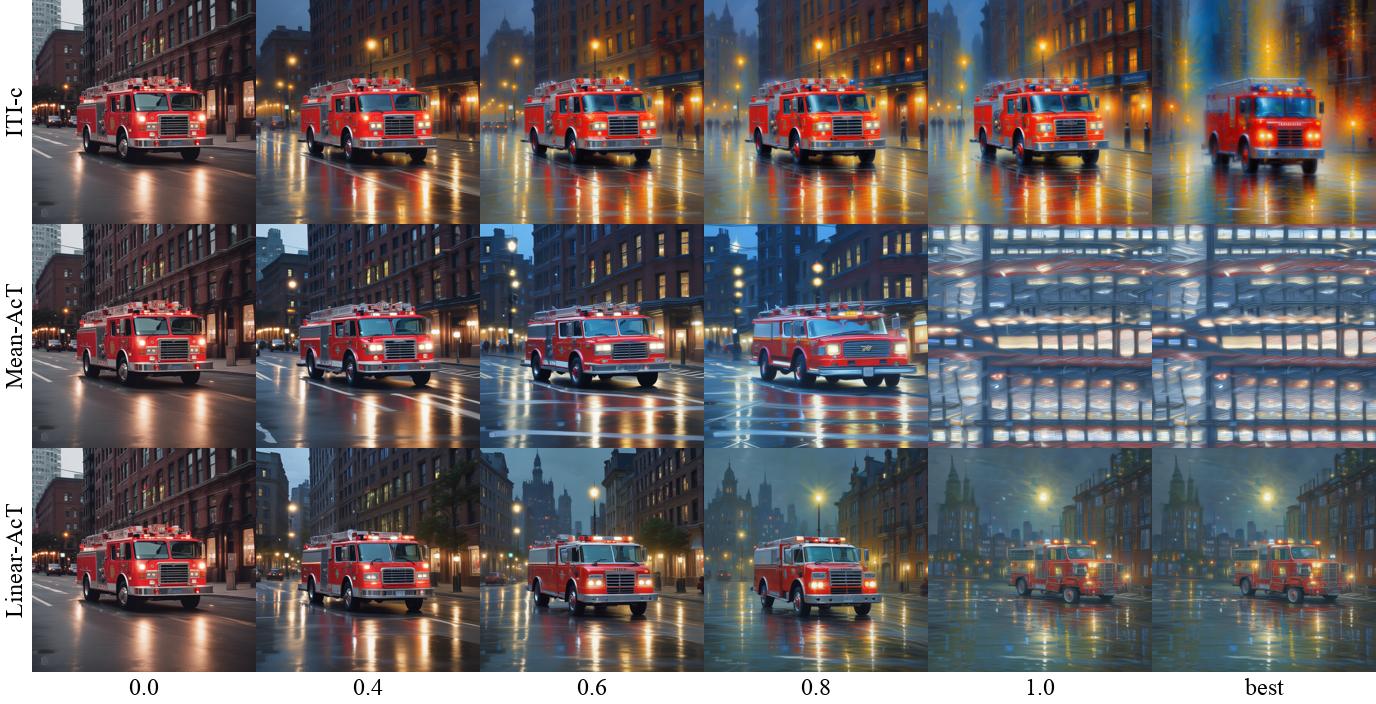}
     \caption{Impressionism}
     \end{subfigure}
     \begin{subfigure}[t]{0.49\linewidth}
     \includegraphics[width=\linewidth]{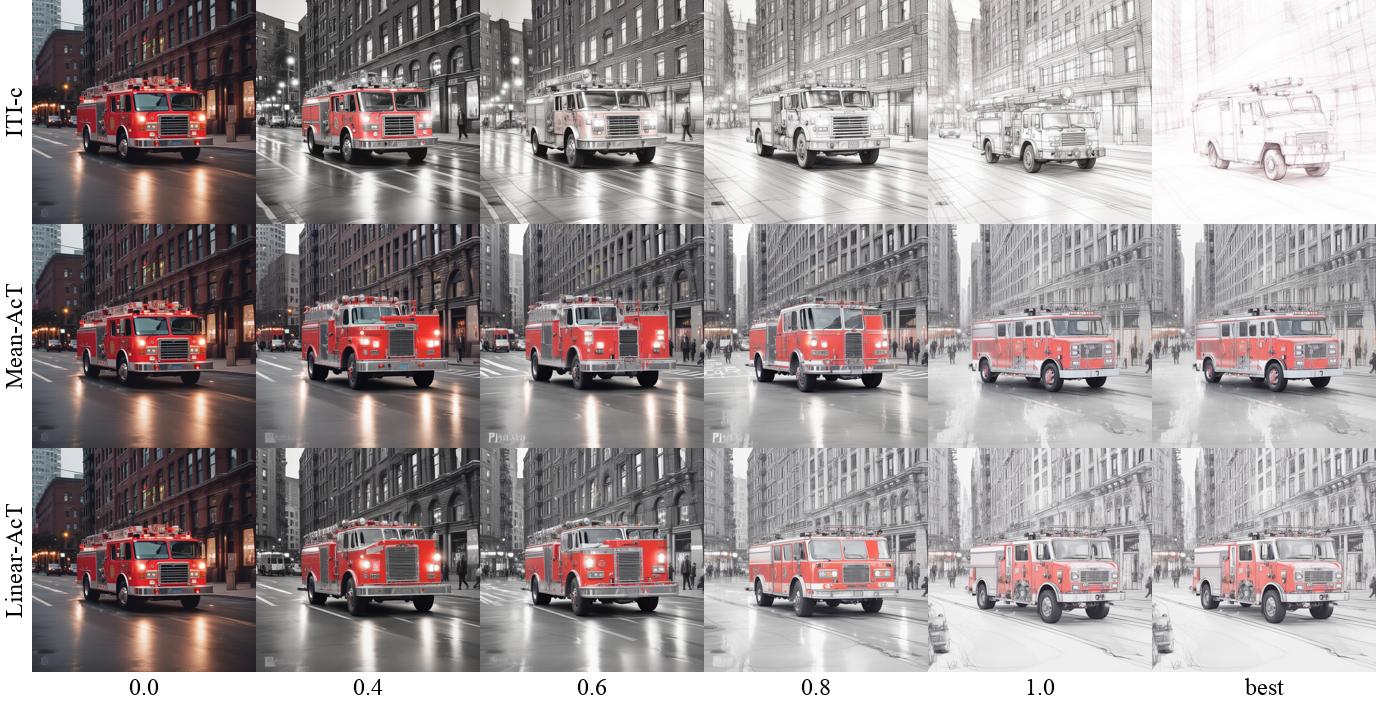}
     \caption{Sketch.}
     \end{subfigure}
     \begin{subfigure}[t]{0.49\linewidth}
     \includegraphics[width=\linewidth]{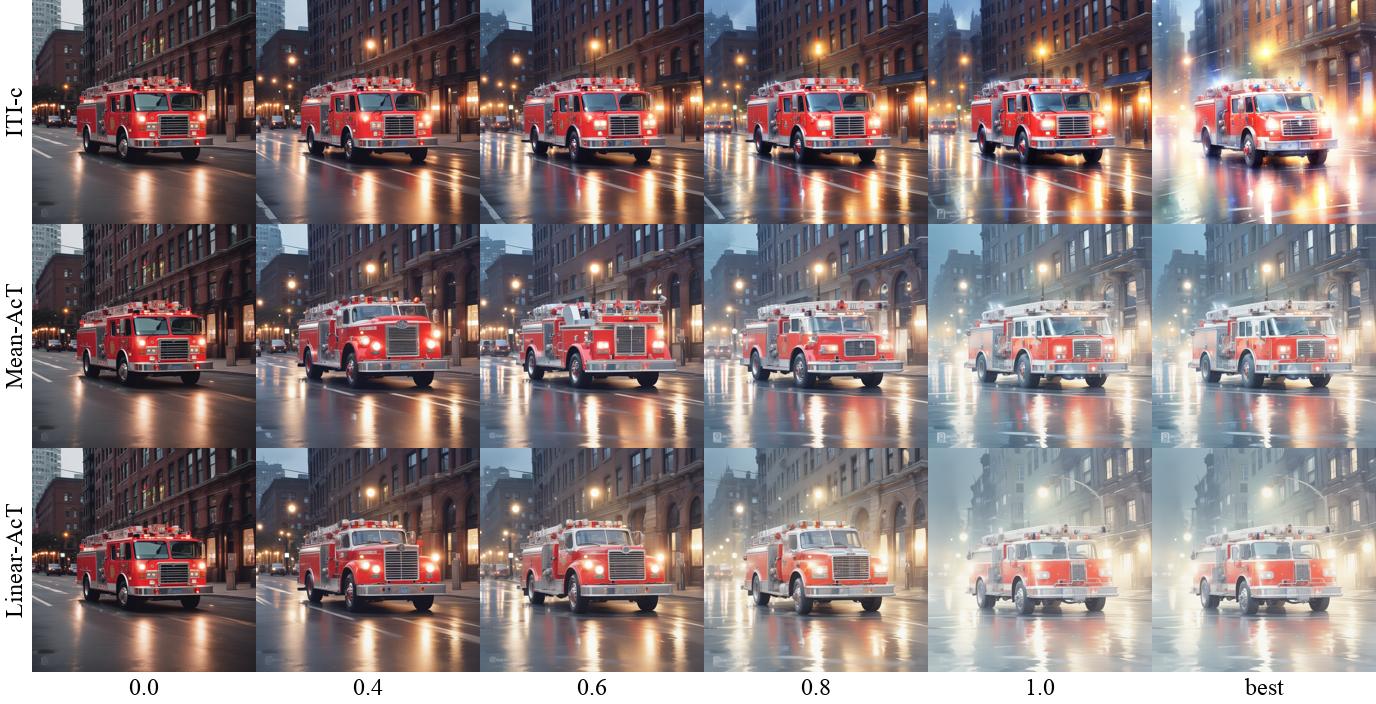}
     \caption{Watercolor}
     \end{subfigure}
\caption{\textbf{SDXL - A firetruck with lights on is on a city street.} Rightmost column corresponds to the best strength found in \Cref{fig:clip_score} ($\lambda=1$ for \method and $\lambda=2$ for \iti). \mean fails at inducing \textit{impressionism} and \textit{art nouveau}. \iti  achieves the strongest conditioning and generates a noisy image for \textit{art nouveau}.}
\label{fig:style-II}
\end{figure}

\begin{figure}[t]
     \centering 
     \begin{subfigure}[t]{0.49\linewidth}
     \includegraphics[width=\linewidth]{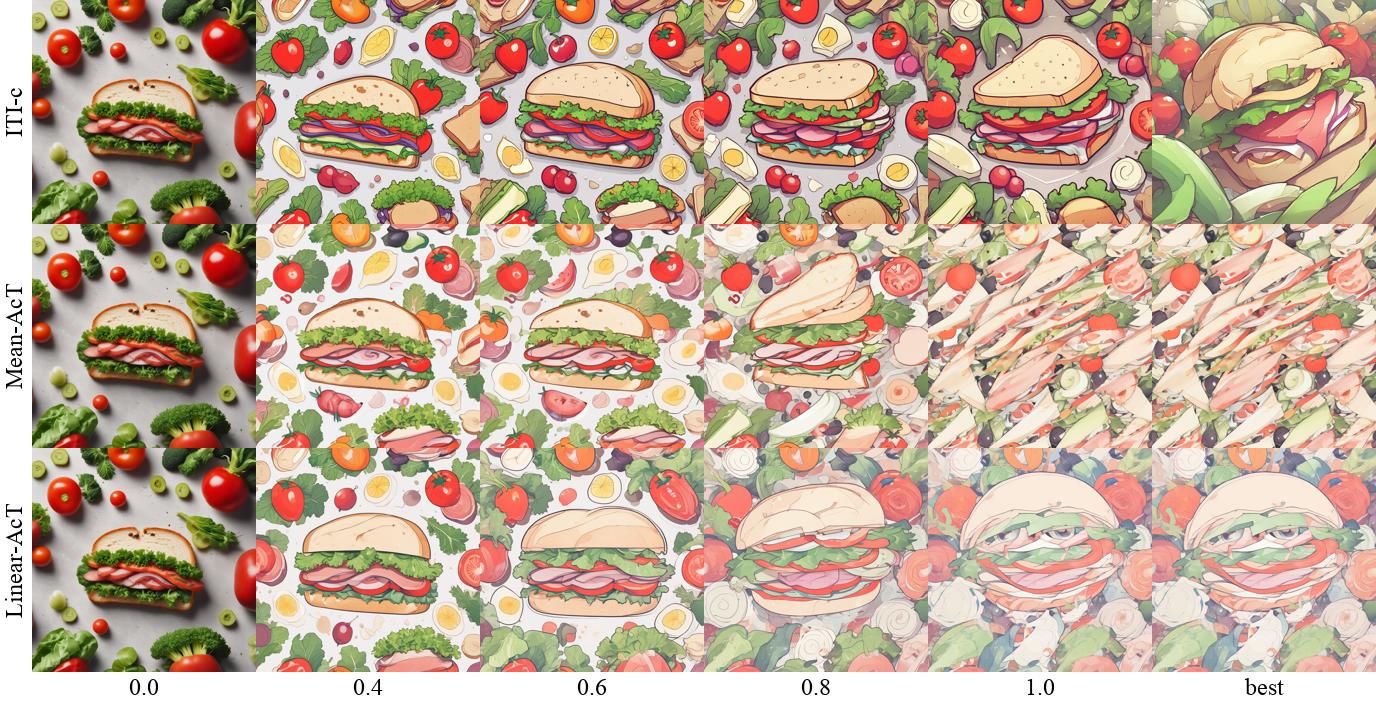}
     \caption{Anime}
     \end{subfigure}
     \begin{subfigure}[t]{0.49\linewidth}
     \includegraphics[width=\linewidth]{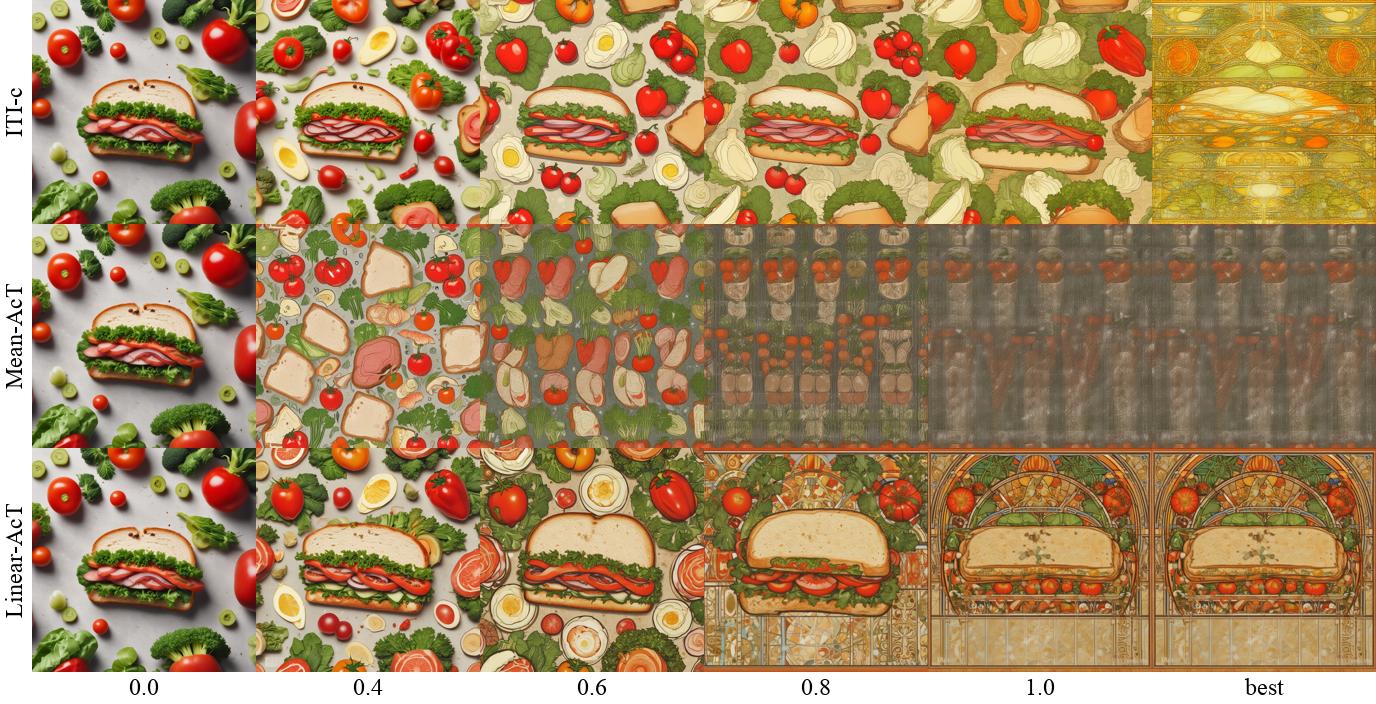}
     \caption{Art Nouveau}
     \end{subfigure}
     \begin{subfigure}[t]{0.49\linewidth}
     \includegraphics[width=\linewidth]{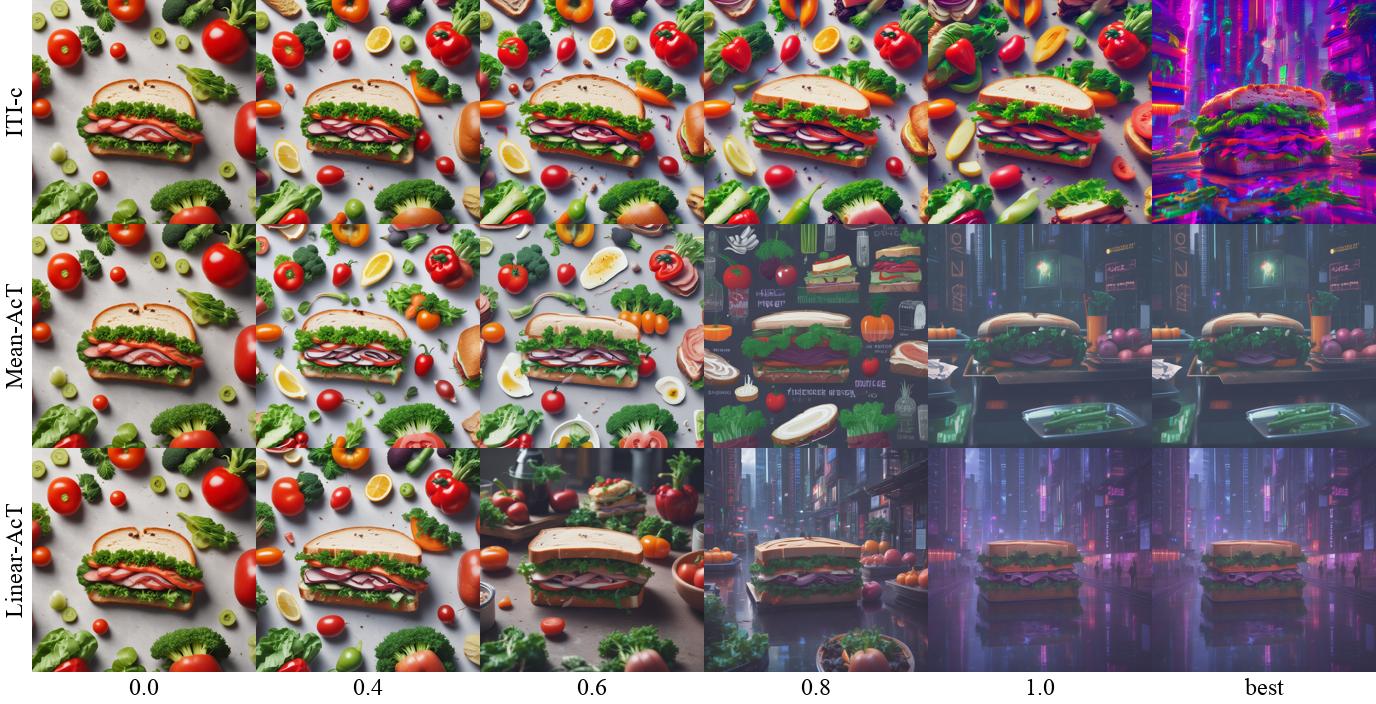}
     \caption{Cyberpunk}
     \end{subfigure}
     \begin{subfigure}[t]{0.49\linewidth}
     \includegraphics[width=\linewidth]{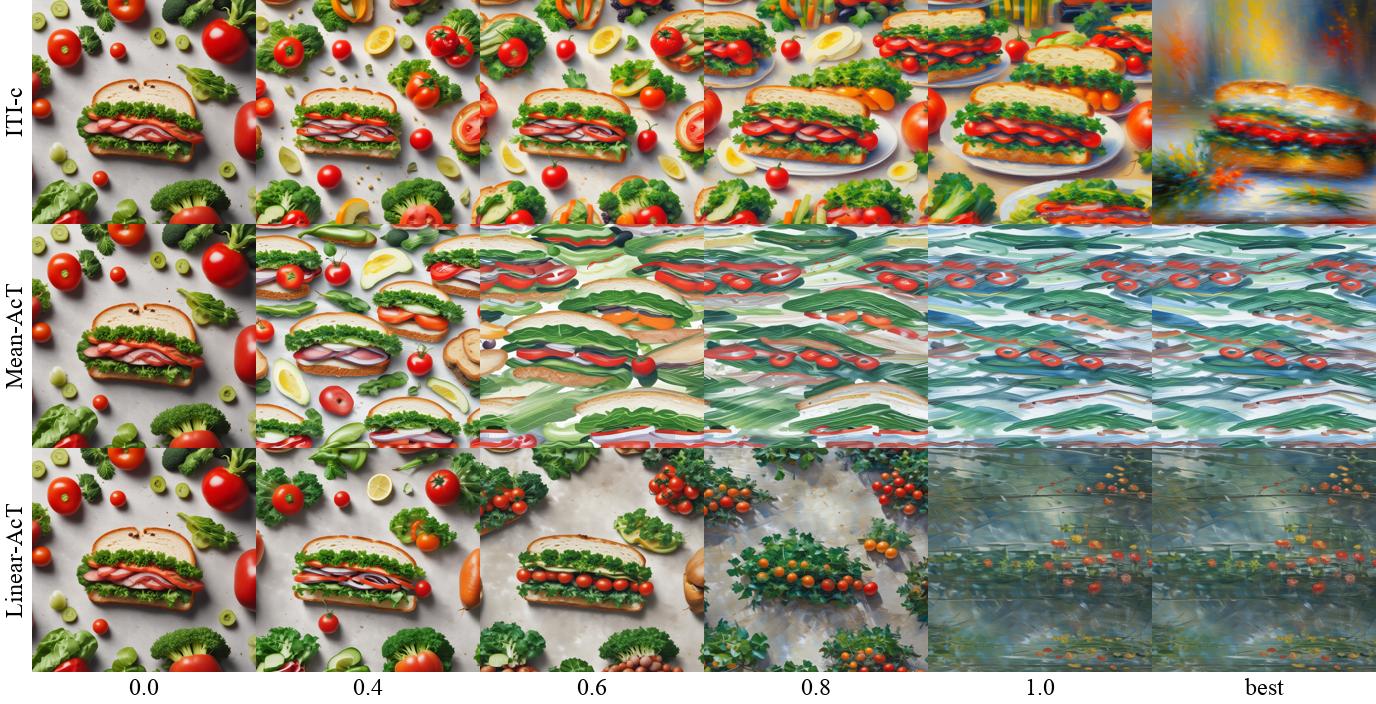}
     \caption{Impressionism}
     \end{subfigure}
     \begin{subfigure}[t]{0.49\linewidth}
     \includegraphics[width=\linewidth]{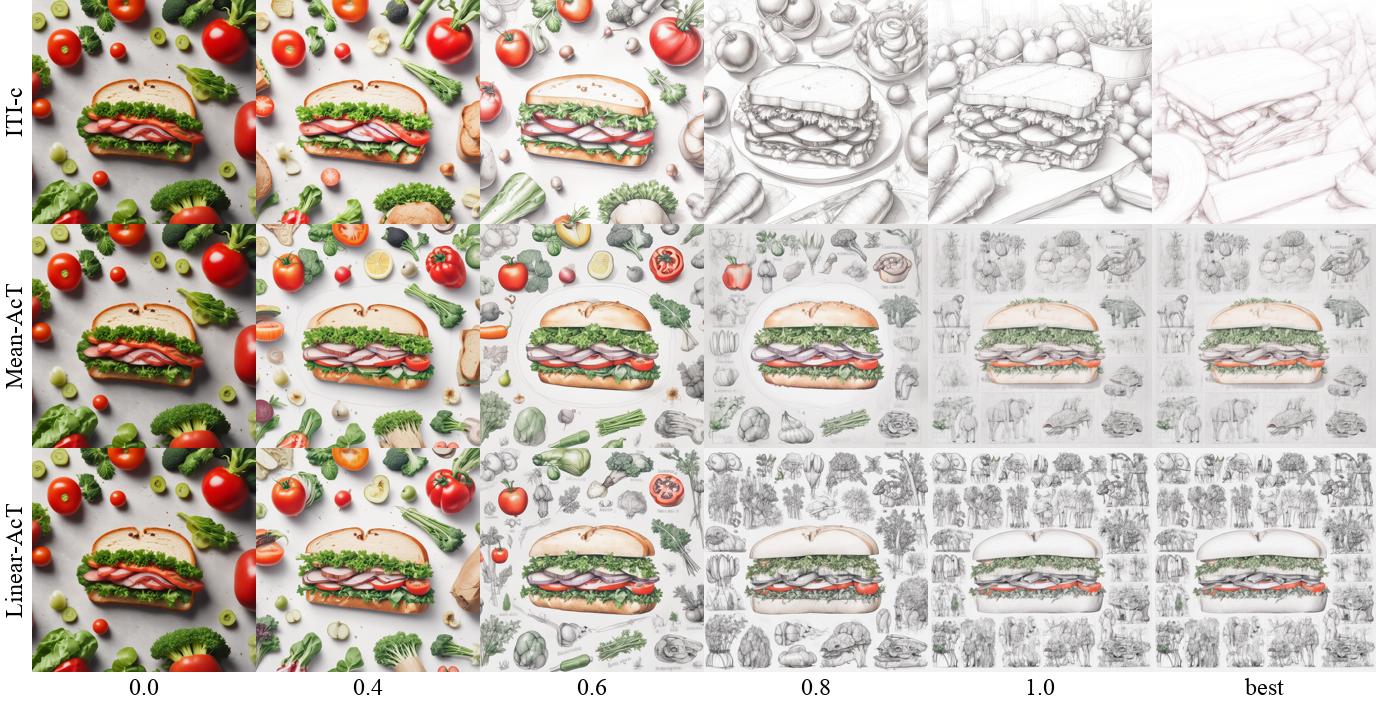}
     \caption{Sketch.}
     \end{subfigure}
     \begin{subfigure}[t]{0.49\linewidth}
     \includegraphics[width=\linewidth]{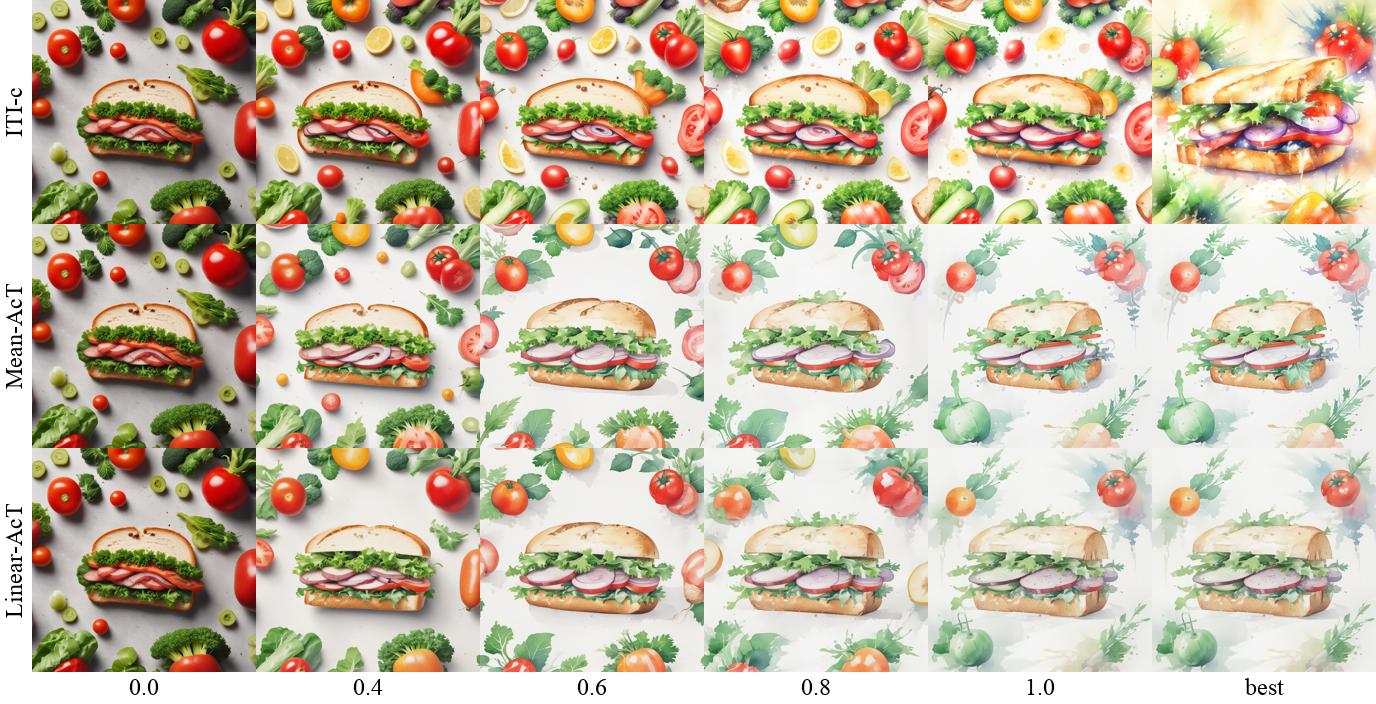}
     \caption{Watercolor}
     \end{subfigure}
\caption{\textbf{SDXL - A sandwich is placed next to some vegetables.} Rightmost column corresponds to the best strength found in \Cref{fig:clip_score} ($\lambda=1$ for \method and $\lambda=2$ for \iti). \iti  fails at inducing style progressively (e.g. (c) \textit{cyberpunk}).}
\label{fig:style-III}
\end{figure}

\Cref{fig:style-I,fig:style-II,fig:style-III} complement the results shown in \Cref{subsec:style-control}.


\begin{figure}[t]
     \centering 
     \begin{subfigure}[t]{0.49\linewidth}
     \includegraphics[width=\linewidth]{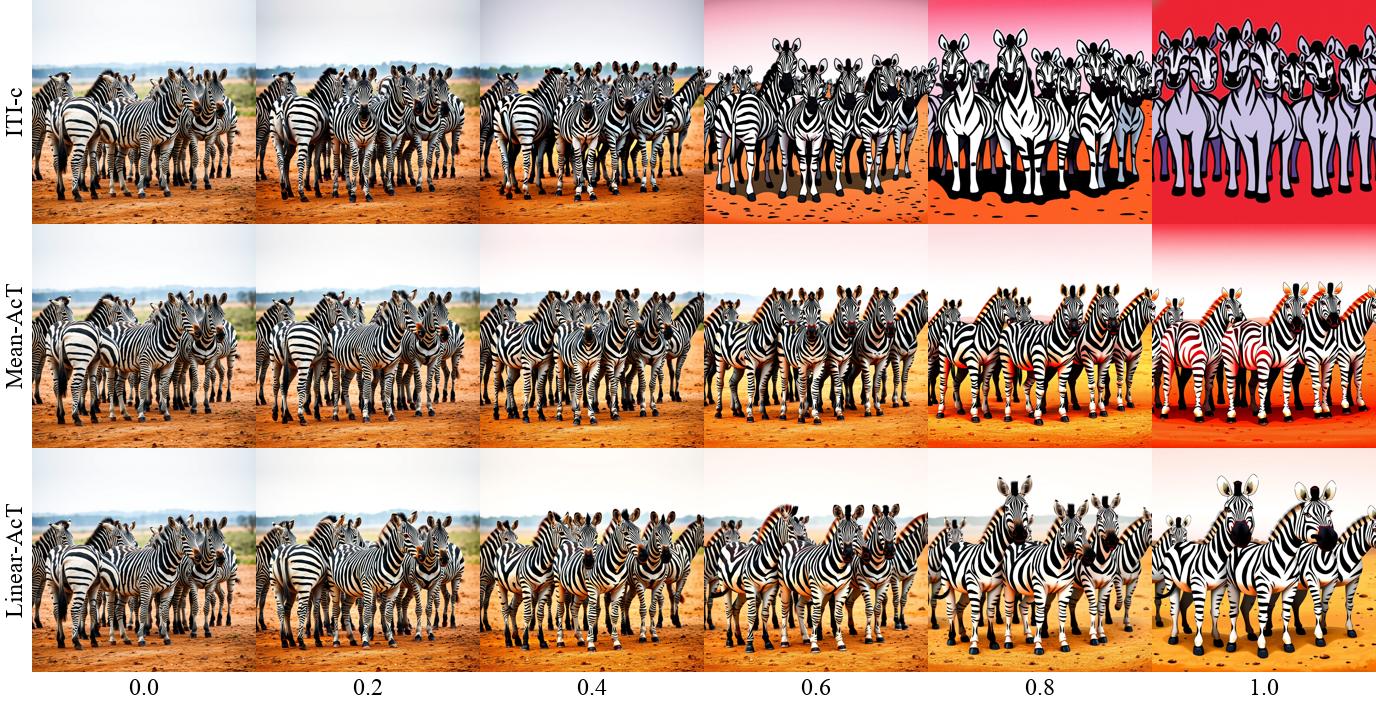}
     \caption{Anime}
     \end{subfigure}
     \begin{subfigure}[t]{0.49\linewidth}
     \includegraphics[width=\linewidth]{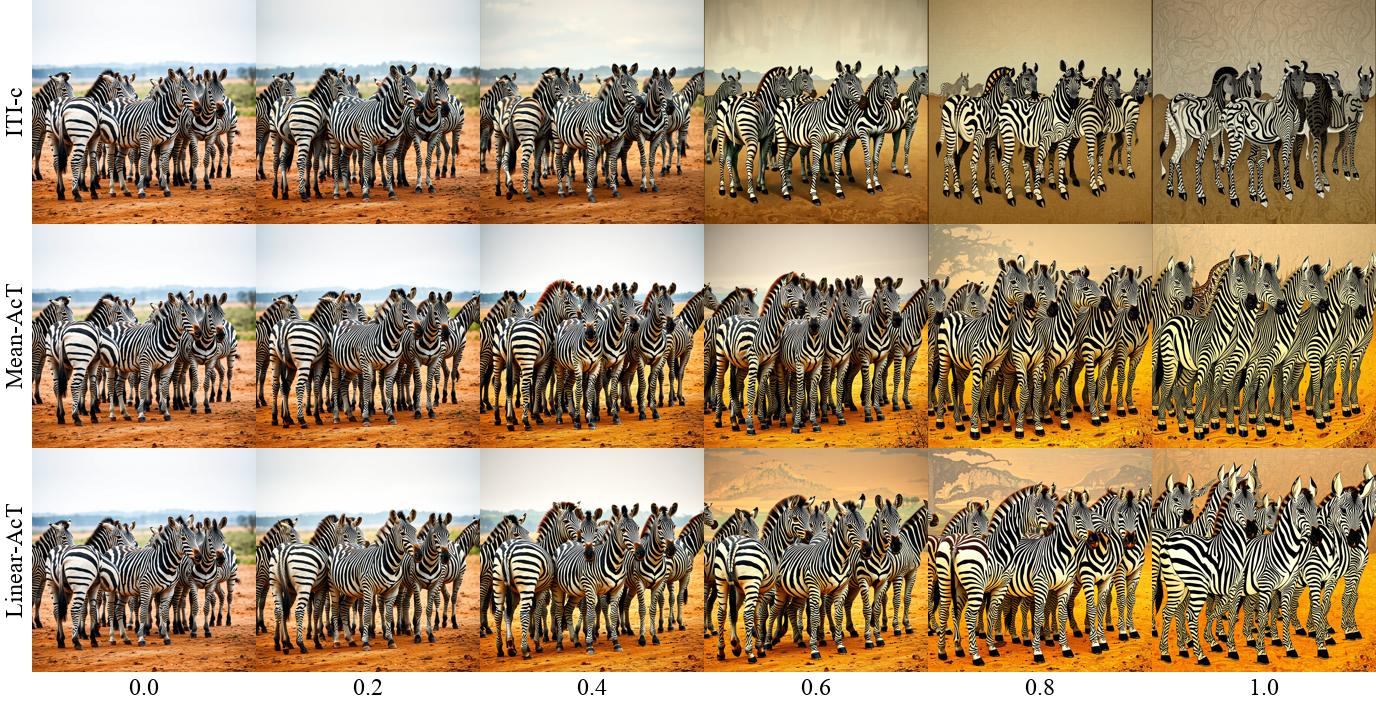}
     \caption{Art Nouveau}
     \end{subfigure}
     \begin{subfigure}[t]{0.49\linewidth}
     \includegraphics[width=\linewidth]{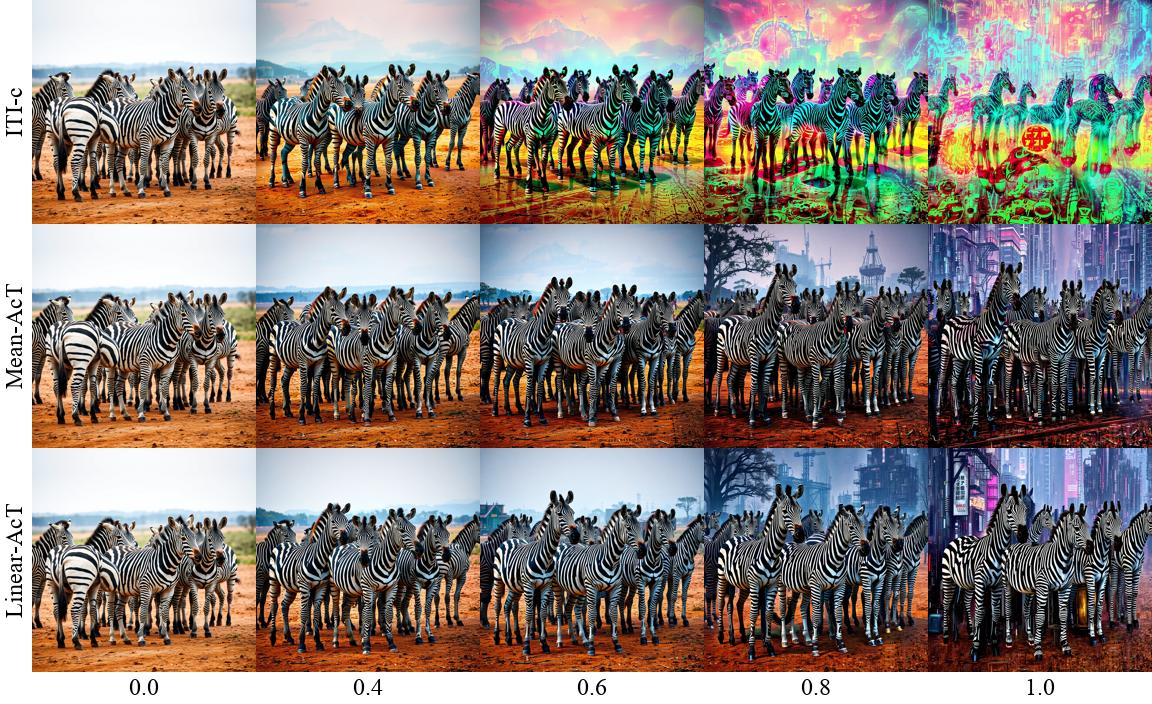}
     \caption{Cyberpunk}
     \end{subfigure}
     \begin{subfigure}[t]{0.49\linewidth}
     \includegraphics[width=\linewidth]{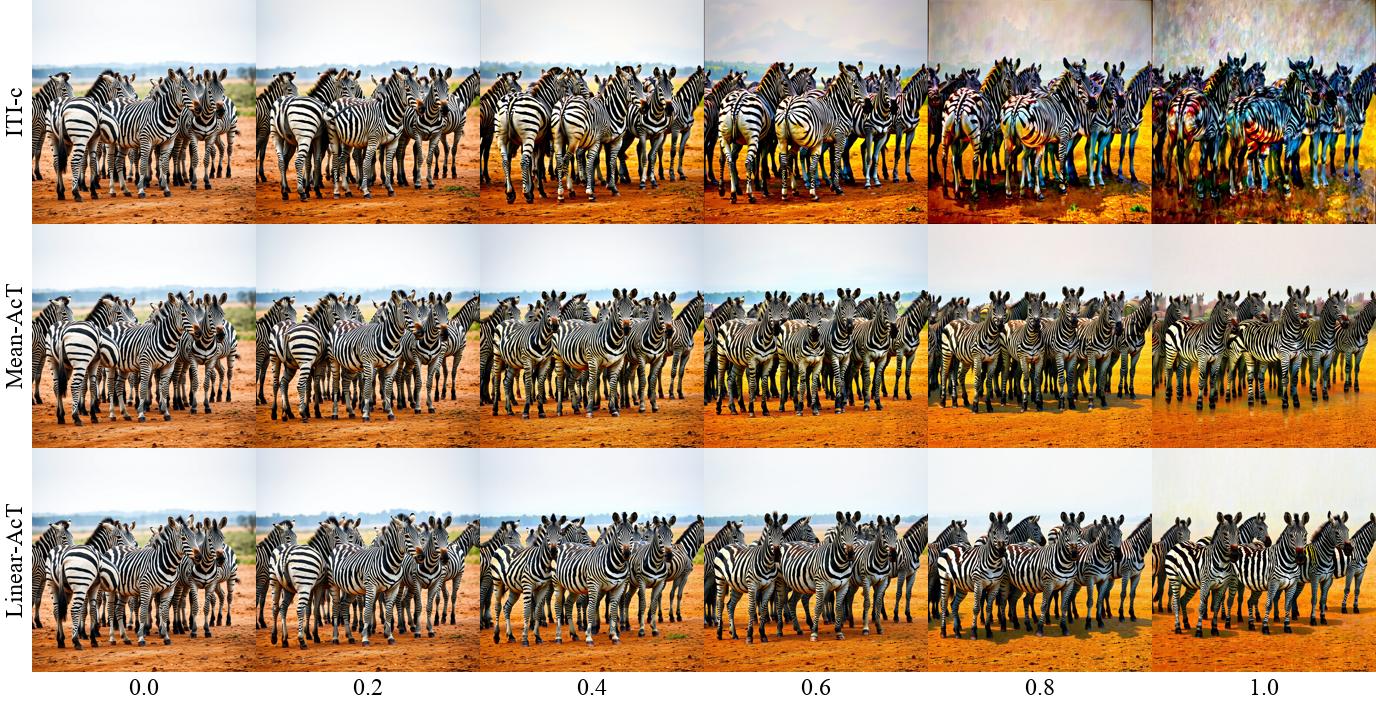}
     \caption{Impressionism}
     \end{subfigure}
     \begin{subfigure}[t]{0.49\linewidth}
     \includegraphics[width=\linewidth]{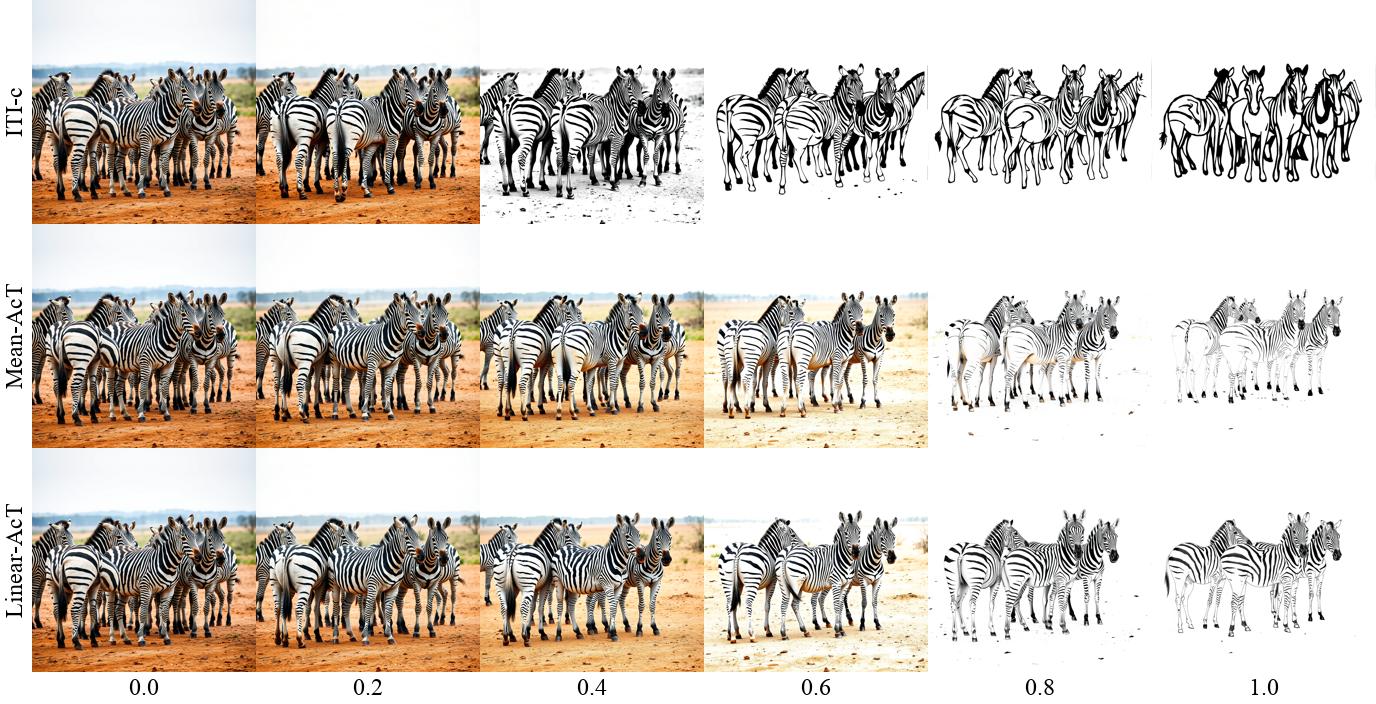}
     \caption{Sketch.}
     \end{subfigure}
     \begin{subfigure}[t]{0.49\linewidth}
     \includegraphics[width=\linewidth]{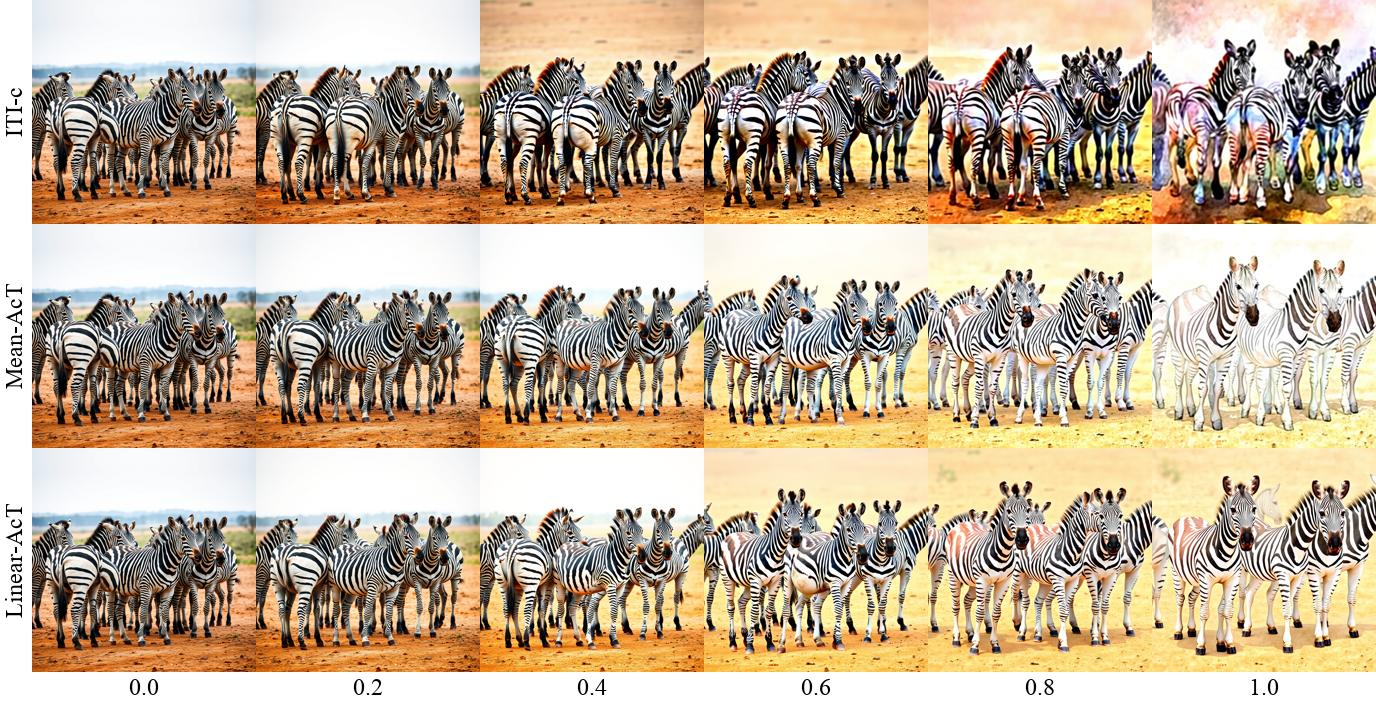}
     \caption{Watercolor}
     \end{subfigure}
\caption{\textbf{FLUX - A group of zebra standing next to each other on a dirt field.} Rightmost column corresponds to the best strength found in \Cref{fig:clip_score} ($\lambda=1$ for all methods). \linear is successful at inducing all styles. \iti fails at inducing \textit{cyberpunk} and \textit{anime}.}
\label{fig:flux-style-I}
\end{figure}
\begin{figure}[t]
     \centering 
     \begin{subfigure}[t]{0.49\linewidth}
     \includegraphics[width=\linewidth]{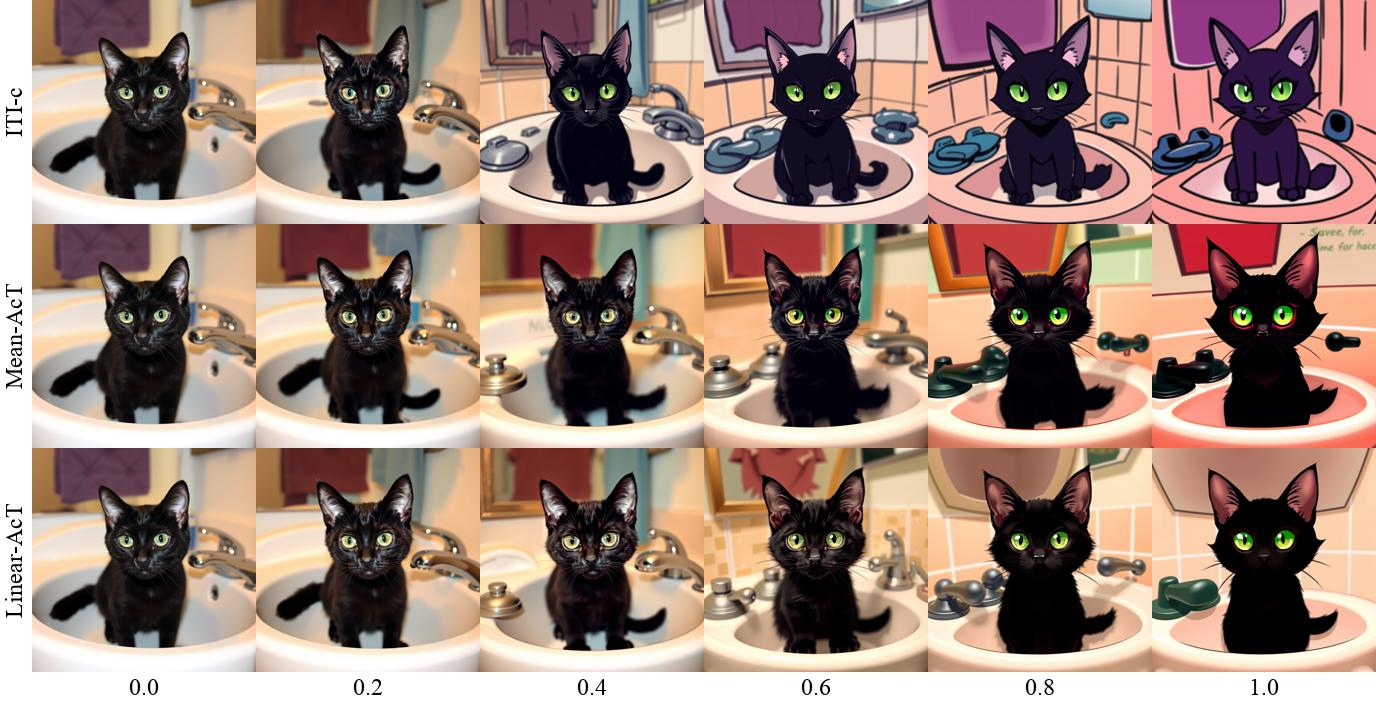}
     \caption{Anime}
     \end{subfigure}
     \begin{subfigure}[t]{0.49\linewidth}
     \includegraphics[width=\linewidth]{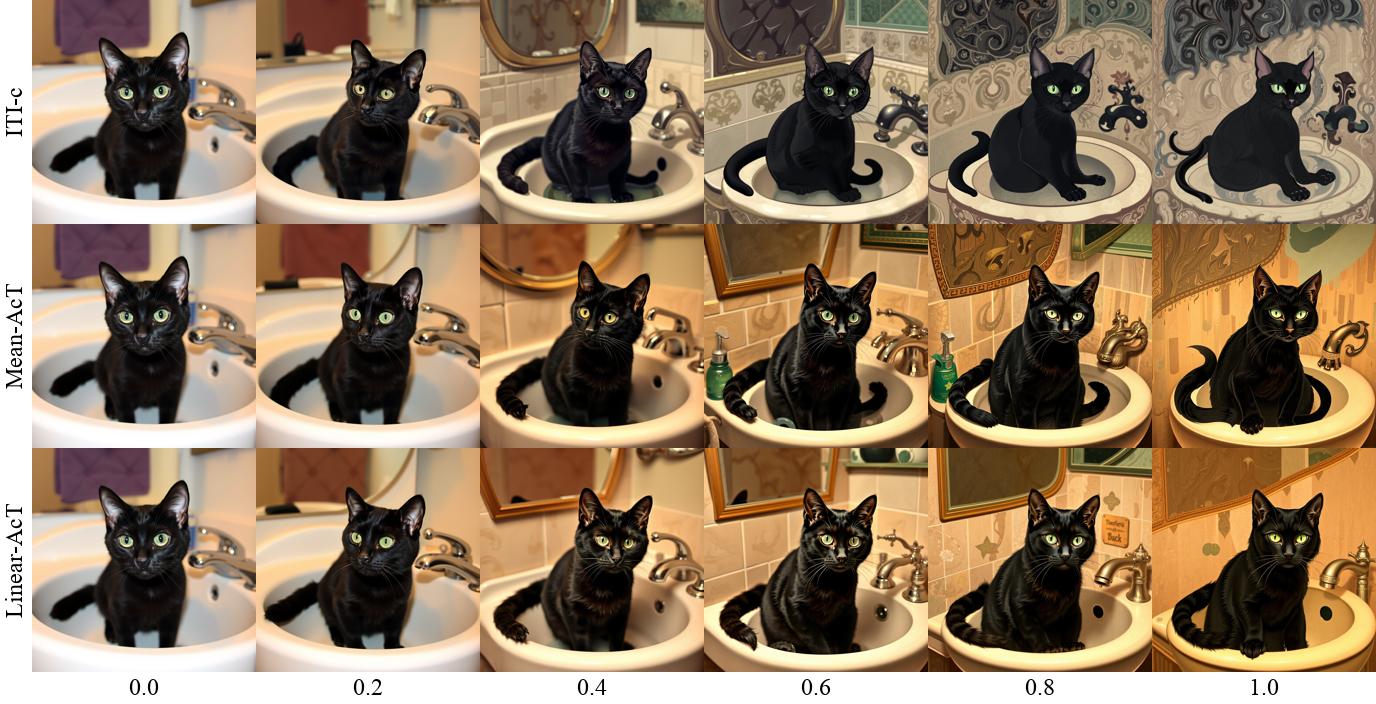}
     \caption{Art Nouveau}
     \end{subfigure}
     \begin{subfigure}[t]{0.49\linewidth}
     \includegraphics[width=\linewidth]{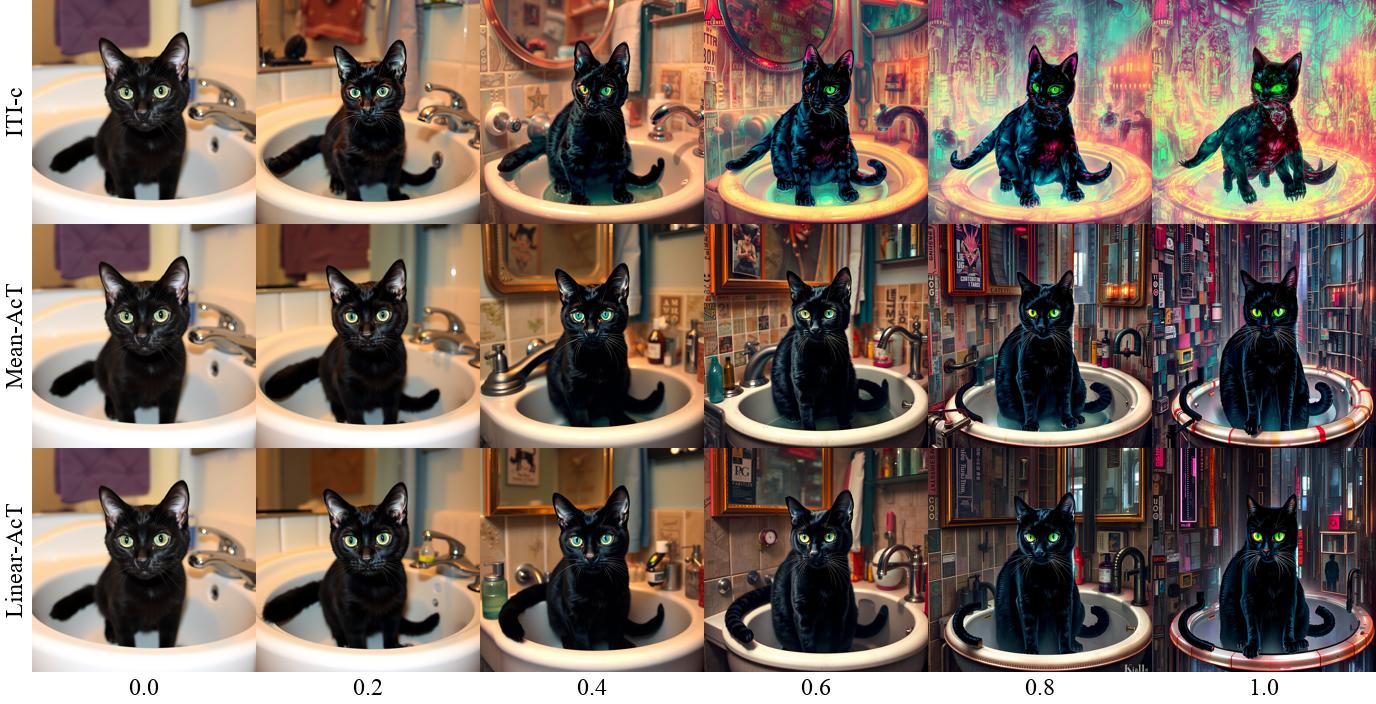}
     \caption{Cyberpunk}
     \end{subfigure}
     \begin{subfigure}[t]{0.49\linewidth}
     \includegraphics[width=\linewidth]{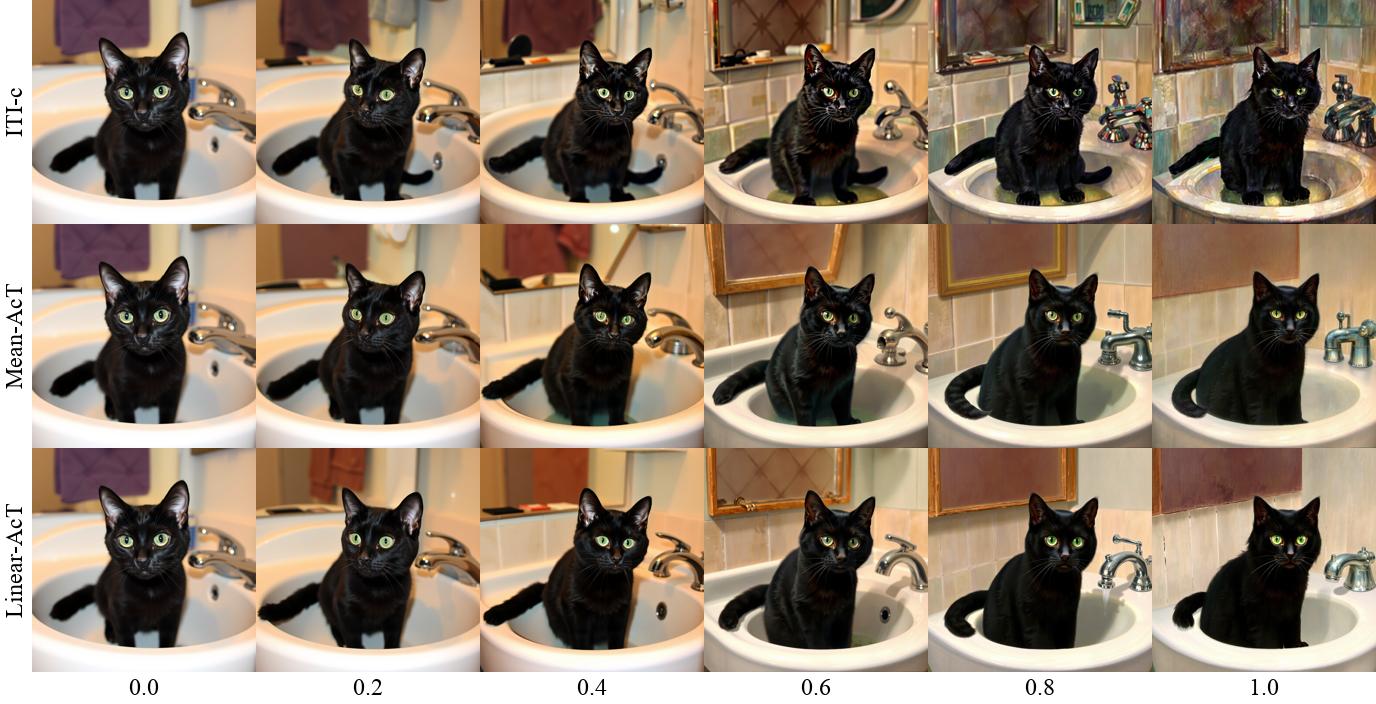}
     \caption{Impressionism}
     \end{subfigure}
     \begin{subfigure}[t]{0.49\linewidth}
     \includegraphics[width=\linewidth]{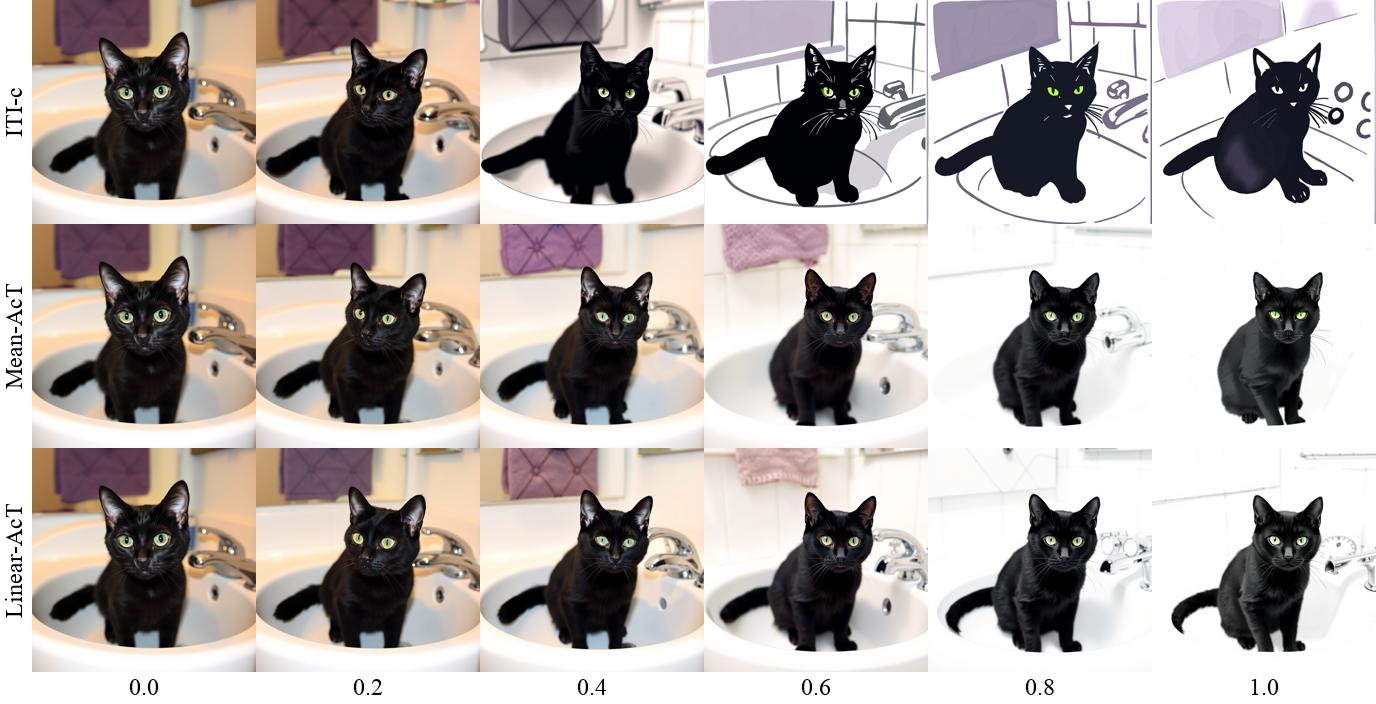}
     \caption{Sketch.}
     \end{subfigure}
     \begin{subfigure}[t]{0.49\linewidth}
     \includegraphics[width=\linewidth]{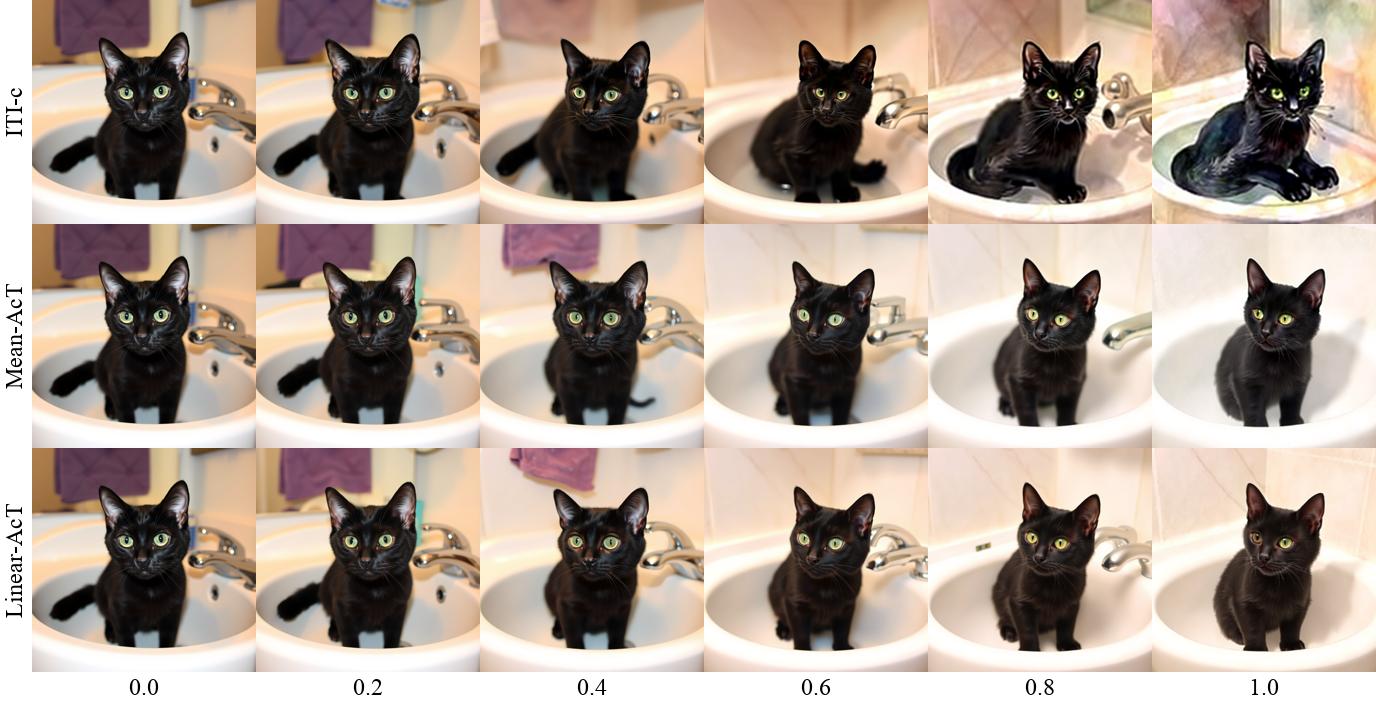}
     \caption{Watercolor}
     \end{subfigure}
\caption{\textbf{FLUX - Black cat with green eyes sitting in a bathroom sink.} Rightmost column corresponds to the best strength found in \Cref{fig:clip_score} ($\lambda=1$ for all methods). \method's conditioning is weak for \textit{sketch} and \textit{watercolor}. \iti fails at inducing \textit{cyberpunk}.}
\label{fig:flux-style-II}
\end{figure}
\begin{figure}[t]
     \centering 
     \begin{subfigure}[t]{0.49\linewidth}
     \includegraphics[width=\linewidth]{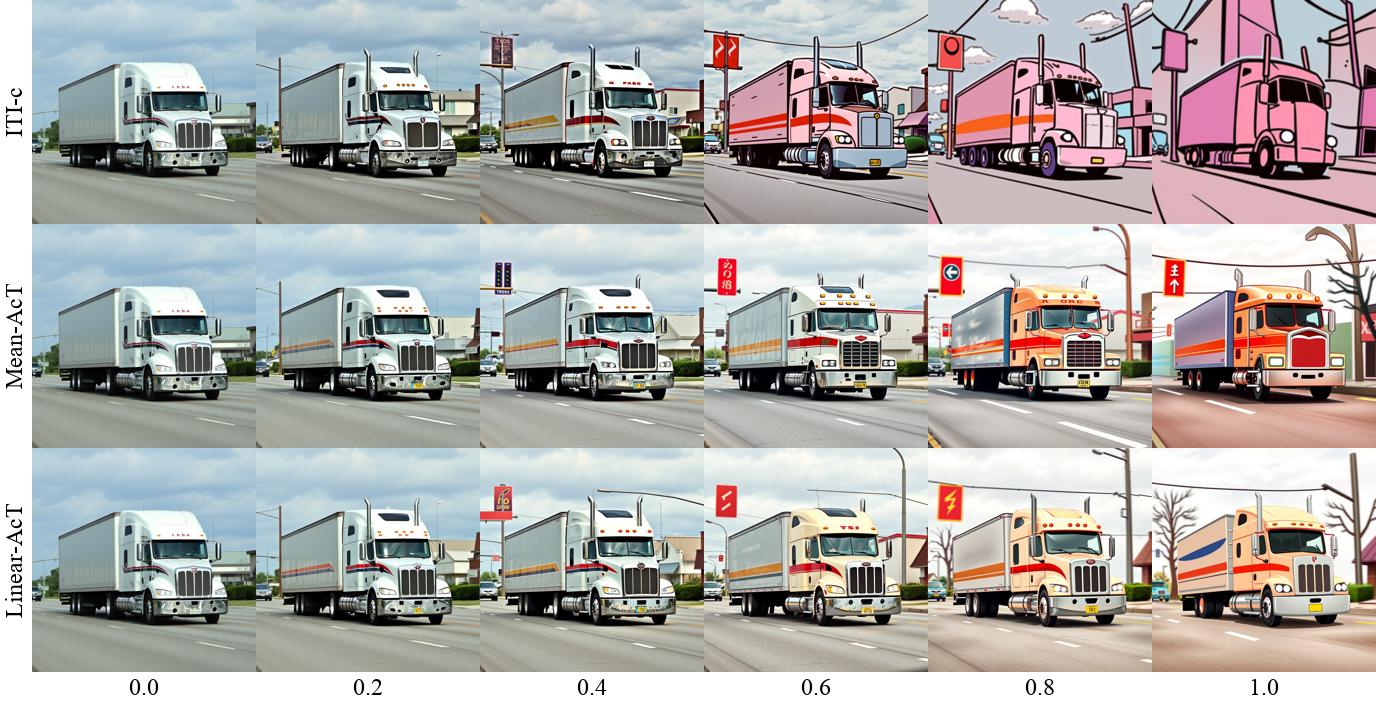}
     \caption{Anime}
     \end{subfigure}
     \begin{subfigure}[t]{0.49\linewidth}
     \includegraphics[width=\linewidth]{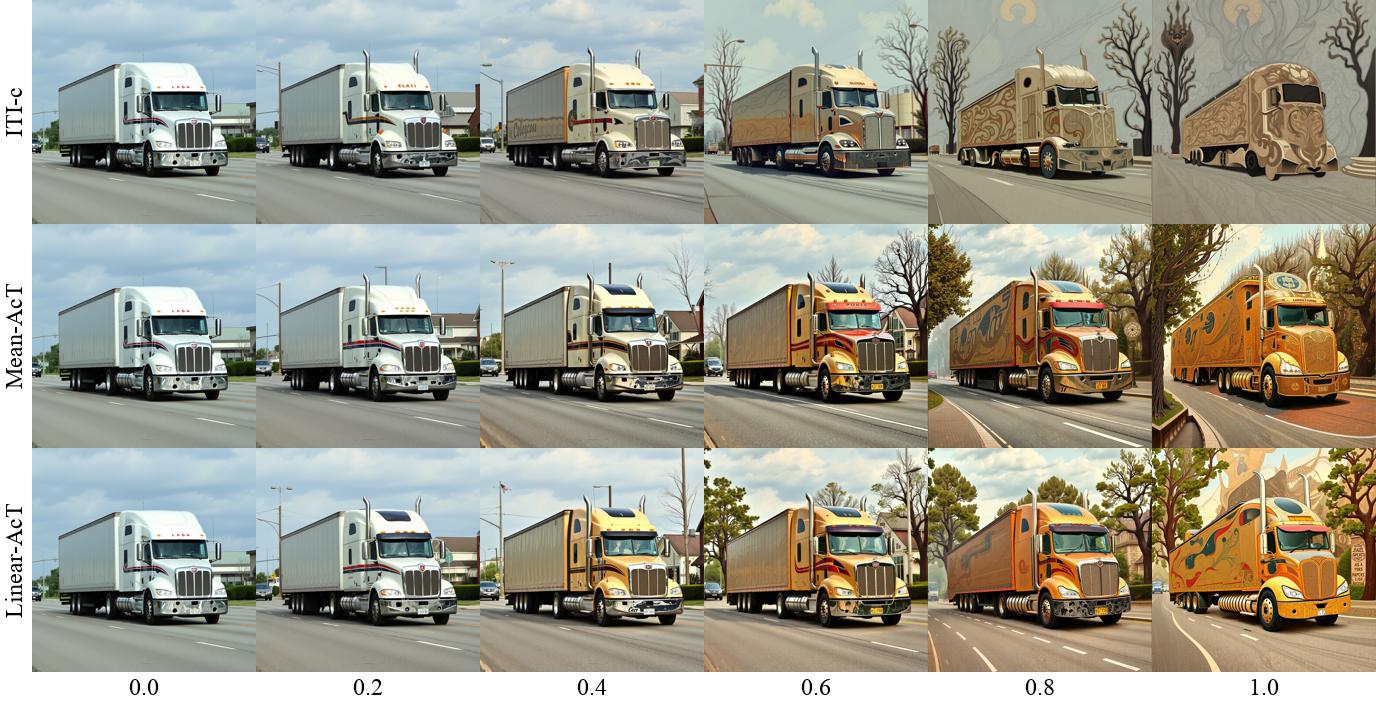}
     \caption{Art Nouveau}
     \end{subfigure}
     \begin{subfigure}[t]{0.49\linewidth}
     \includegraphics[width=\linewidth]{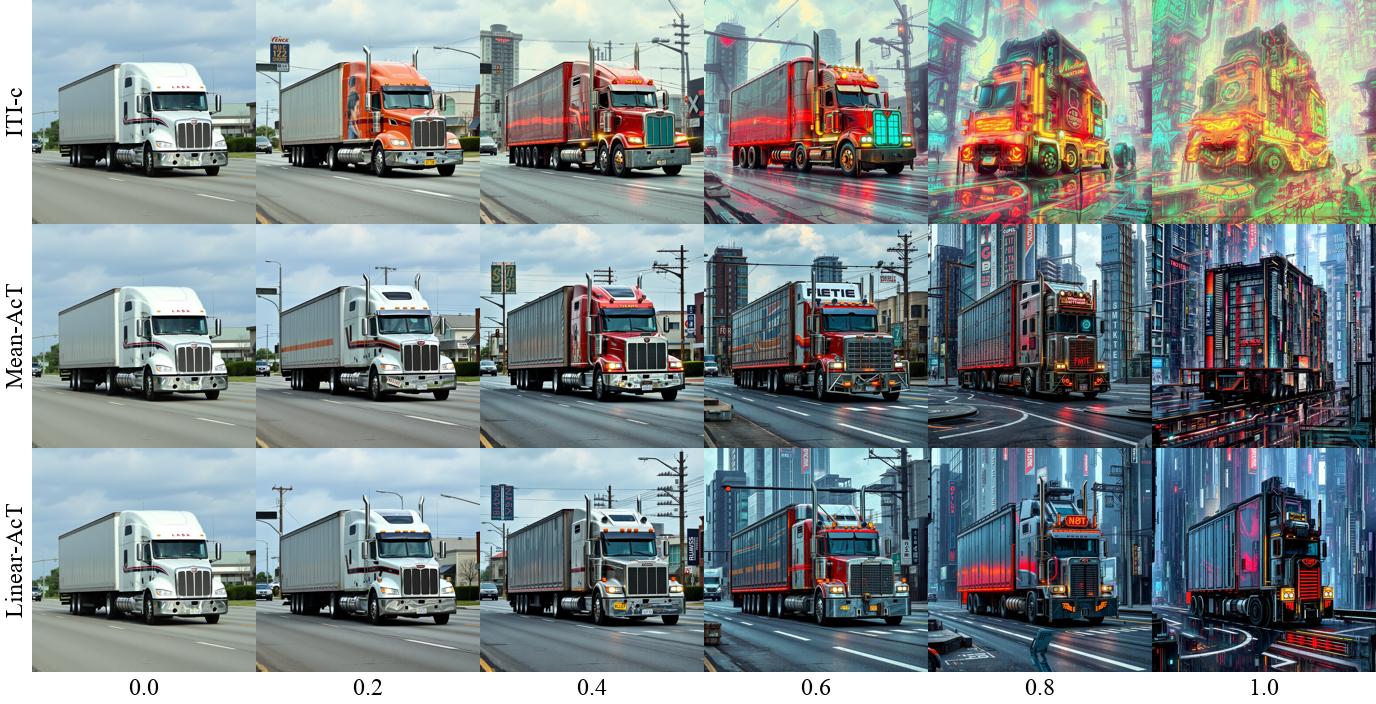}
     \caption{Cyberpunk}
     \end{subfigure}
     \begin{subfigure}[t]{0.49\linewidth}
     \includegraphics[width=\linewidth]{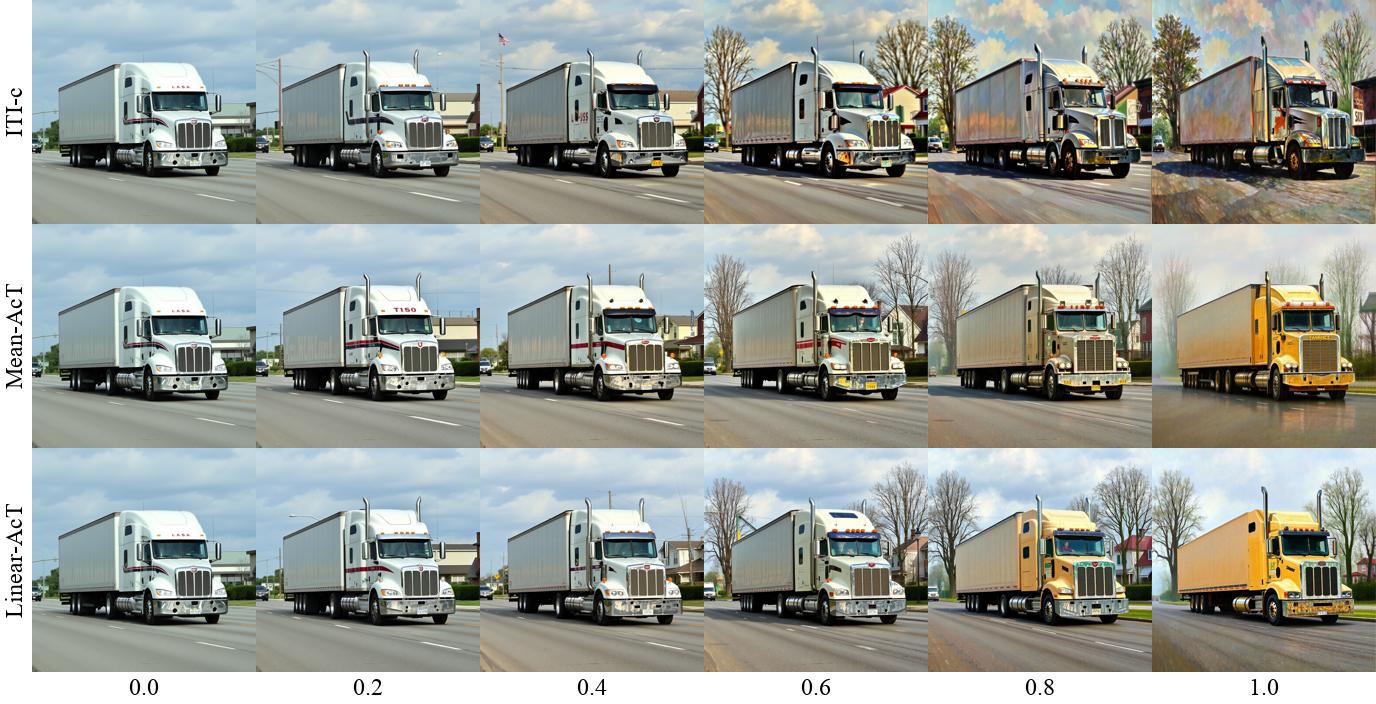}
     \caption{Impressionism}
     \end{subfigure}
     \begin{subfigure}[t]{0.49\linewidth}
     \includegraphics[width=\linewidth]{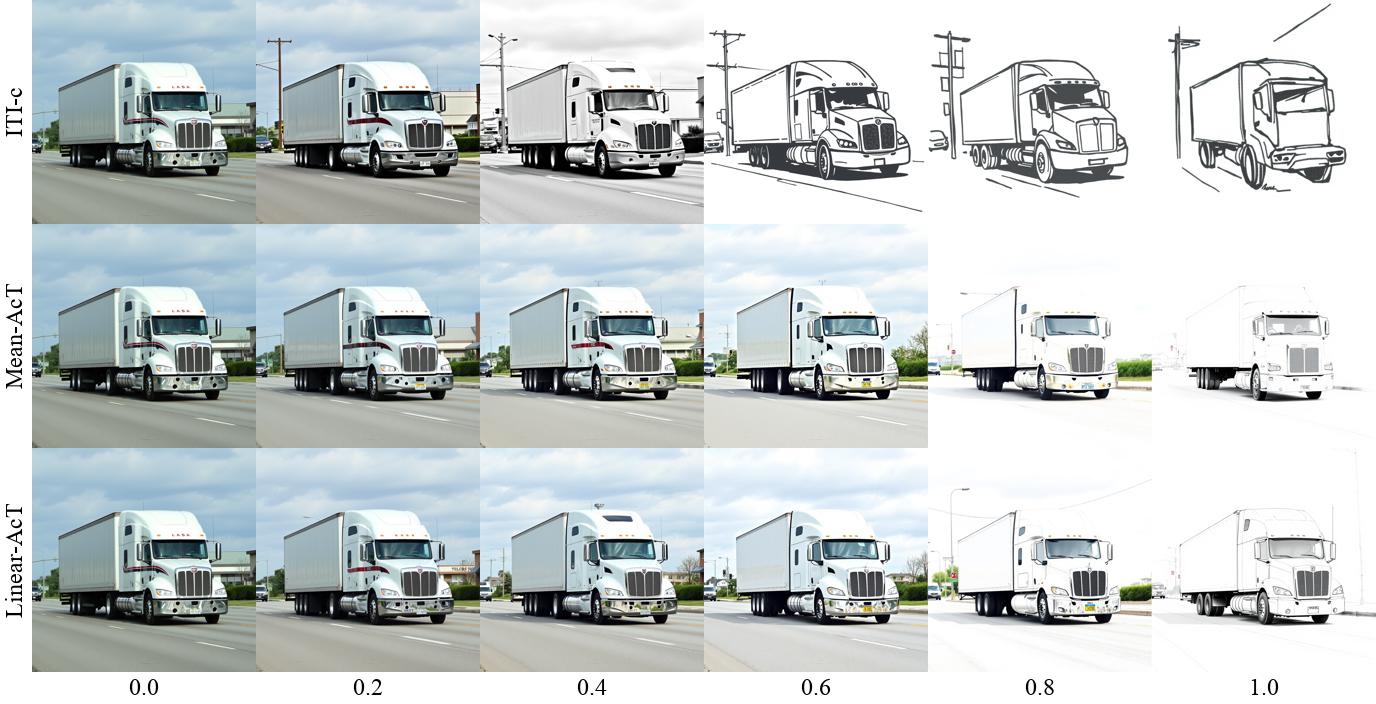}
     \caption{Sketch.}
     \end{subfigure}
     \begin{subfigure}[t]{0.49\linewidth}
     \includegraphics[width=\linewidth]{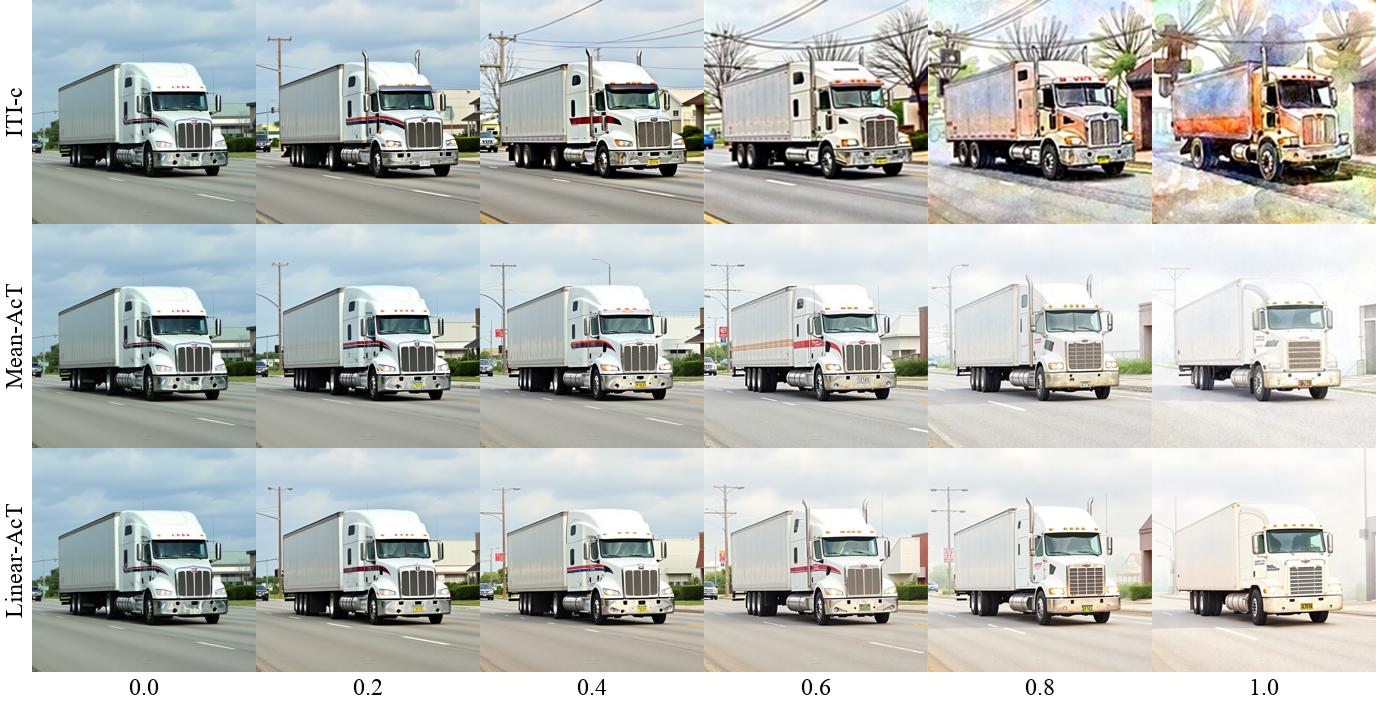}
     \caption{Watercolor}
     \end{subfigure}
\caption{\textbf{FLUX - A semi truck is driving down a street.} Rightmost column corresponds to the best strength found in \Cref{fig:clip_score} ($\lambda=1$ for all methods). \method is able to preserve the semantics for all styles and we observe only mild conditioning for \textit{impressionism} and \textit{watercolor}. \iti fails at inducing \textit{anime} and \textit{cyberpunk}.}
\label{fig:flux-style-III}
\end{figure}

\FloatBarrier
\subsection{Concept negation}
\label{app:concept-negation}
\begin{figure}[h!]
     \centering 
     \begin{subfigure}[t]{0.8\linewidth}
     \includegraphics[width=\linewidth]{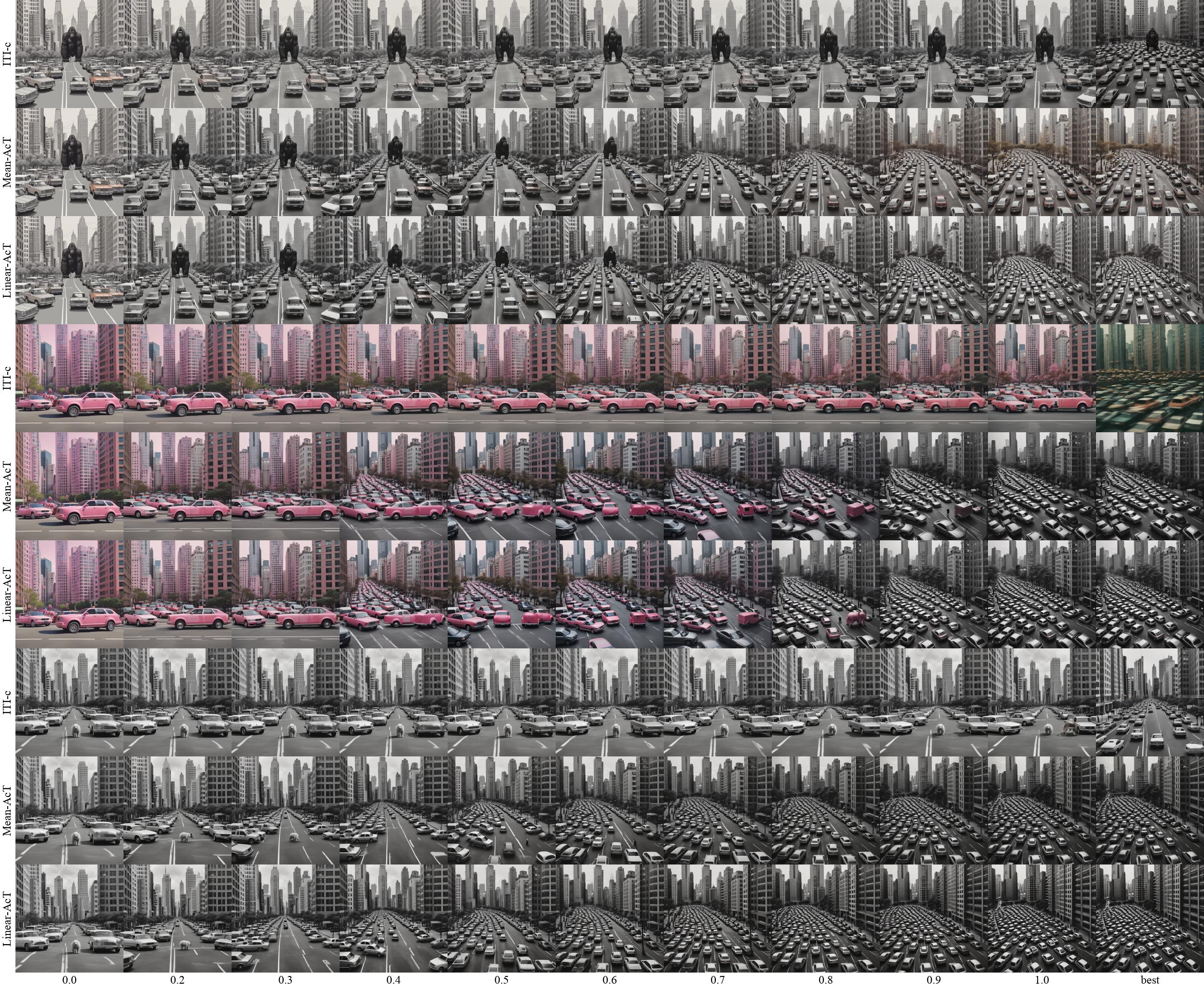}
     \caption{Many cars parked on a city street with tall buildings in the background.}
     \end{subfigure}
     \begin{subfigure}[t]{0.8\linewidth}
     \includegraphics[width=\linewidth]{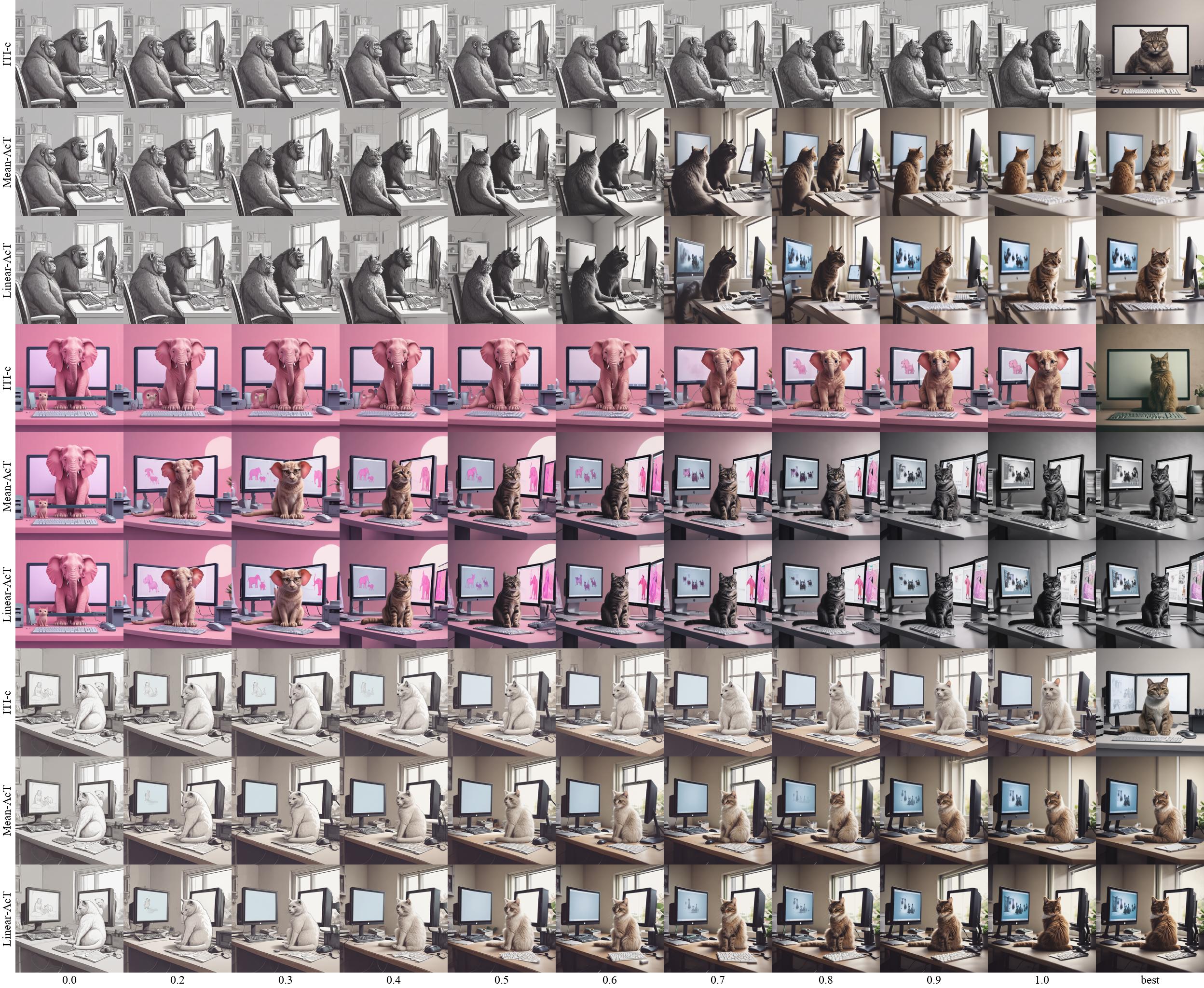}
     \caption{A cat sitting in front of a large computer monitor.}
     \end{subfigure}
\caption{\textbf{SDXL - Concept negation examples I.} Rightmost column corresponds to the best strength found in \Cref{fig:clip_score} ($\lambda=1$ for \method and $\lambda=4$ for \iti). Every 3 rows represent a different concept in \texttt{\{gorilla, pink elephant, white bear\}} which was negated at the input of the image generator. \mean and \linear succeed at removing the unwanted concept. \iti fails for \textit{gorilla} and produces a blurry image for \textit{pink elephant}.}
\label{fig:negation-I}
\end{figure}
\begin{figure}[t]
     \centering 
     \begin{subfigure}[t]{0.8\linewidth}
     \includegraphics[width=\linewidth]{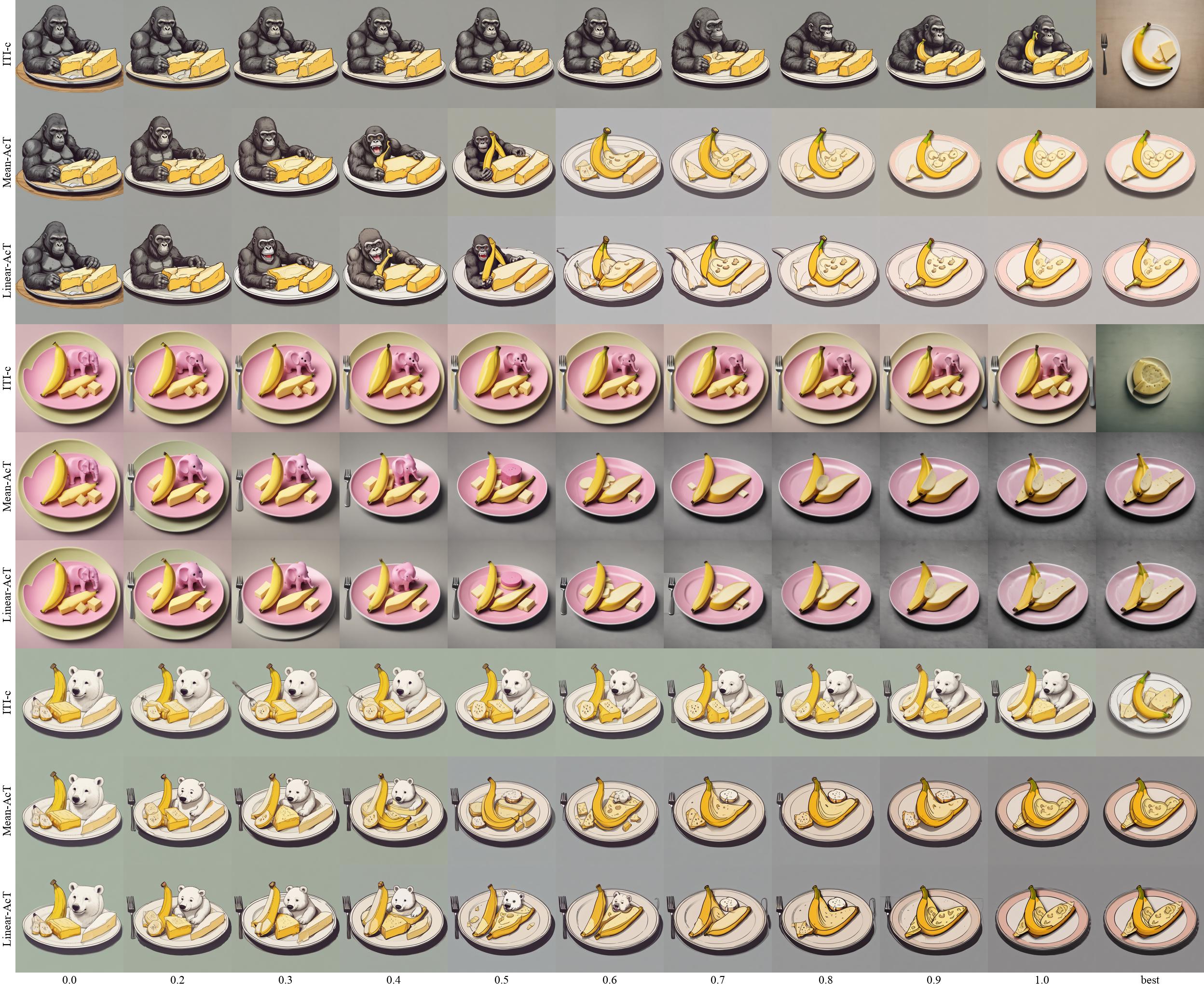}
     \caption{There is a banana and two pieces of cheese on a plate.}
     \end{subfigure}
     \begin{subfigure}[t]{0.8\linewidth}
     \includegraphics[width=\linewidth]{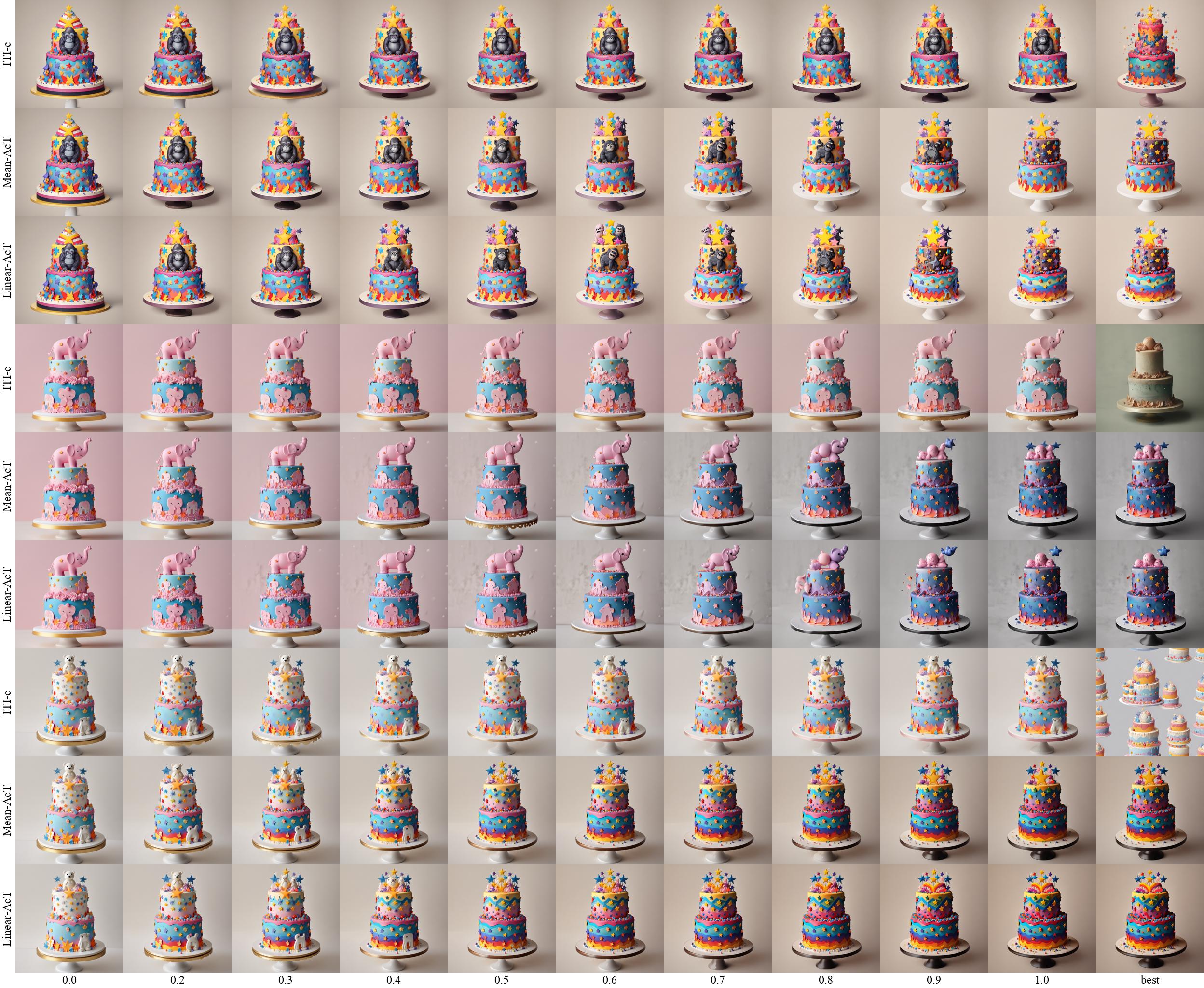}
     \caption{2 tier cake with multicolored stars attached to it.}
     \end{subfigure}
\caption{\textbf{SDXL - Concept negation examples II.} Rightmost column corresponds to the best strength found in \Cref{fig:clip_score} ($\lambda=1$ for \method and $\lambda=4$ for \iti). Every 3 rows represent a different concept in \texttt{\{gorilla, pink elephant, white bear\}} which was negated at the input of the image generator. \linear and \mean succeed at removing the negated concepts while \iti tends to modify the semantics of the image.}
\label{fig:negation-II}
\end{figure}
\begin{figure}[t]
     \centering 
     \begin{subfigure}[t]{0.8\linewidth}
     \includegraphics[width=\linewidth]{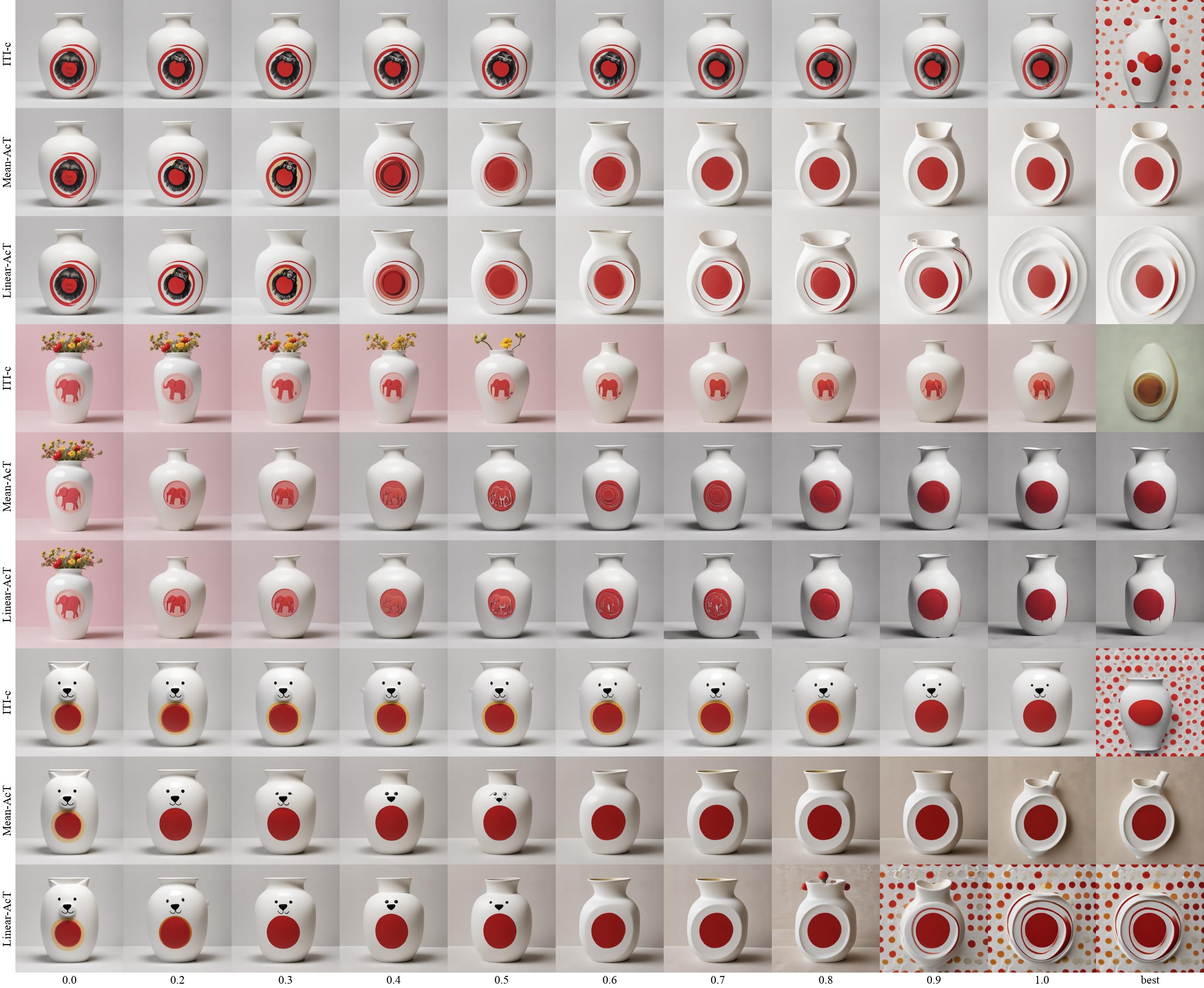}
     \subcaption{Closeup of a white and yellow vase with a red circle at the bottom.}
     \end{subfigure}
     \begin{subfigure}[t]{0.8\linewidth}
     \includegraphics[width=\linewidth]{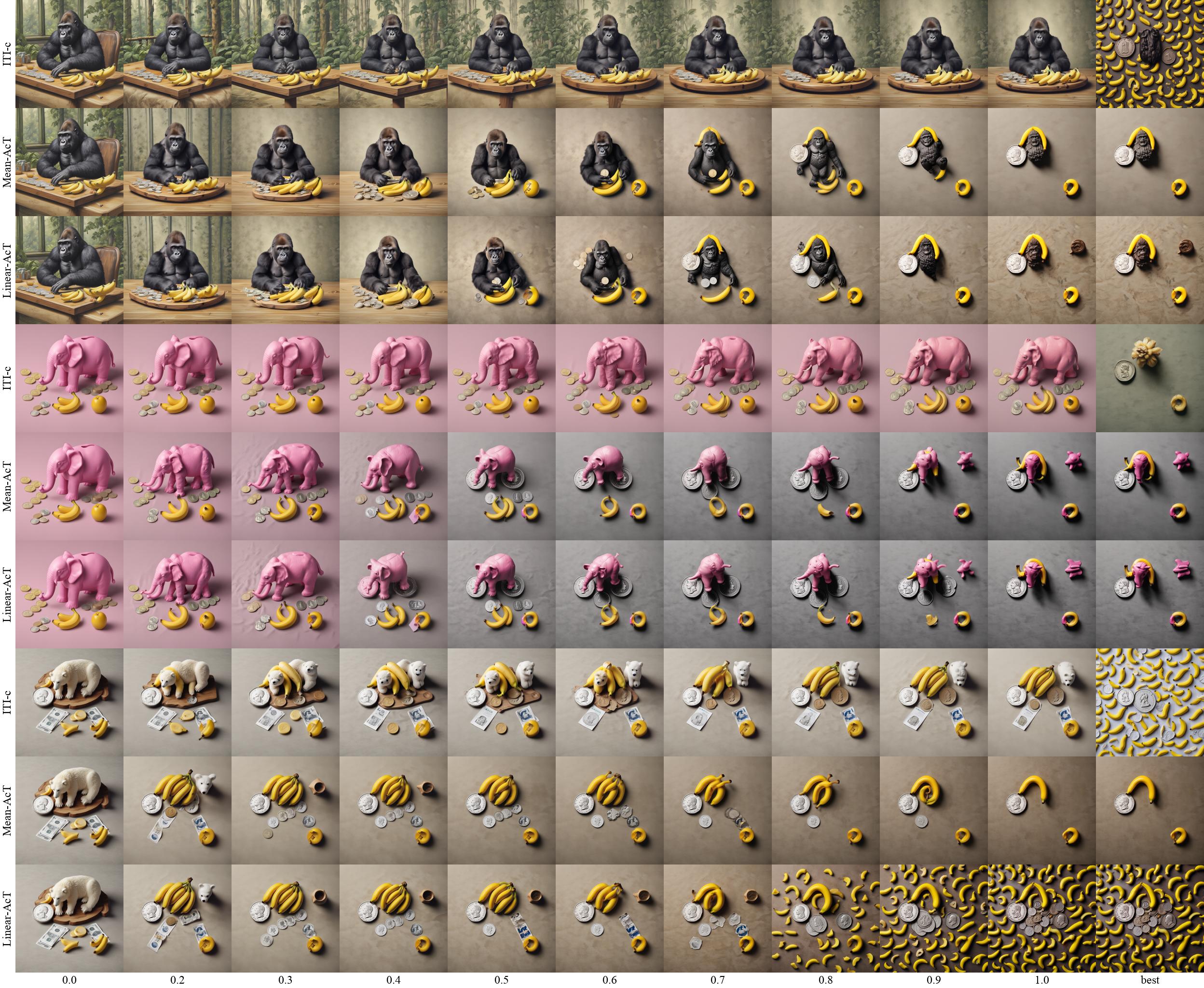}
    \subcaption{A table topped with bananas next to a coin.}
     \end{subfigure}
\caption{\textbf{SDXL - Concept negation examples III (failures)} Rightmost column corresponds to the best strength found in \Cref{fig:clip_score} ($\lambda=1$ for \method and $\lambda=4$ for \iti). Every 3 rows represent a different concept in \texttt{\{gorilla, pink elephant, white bear\}} which was negated at the input of the image generator. While \mean and \linear are successful at removing the concept, there is sometimes a change in semantics of the image for the maximum strength. \iti at best strength ($\lambda=4$) changes semantics for all concepts. }
\label{fig:negation-III}
\end{figure}
\begin{figure}[t]
     \centering 
     \begin{subfigure}[t]{0.8\linewidth}
     \includegraphics[width=\linewidth]{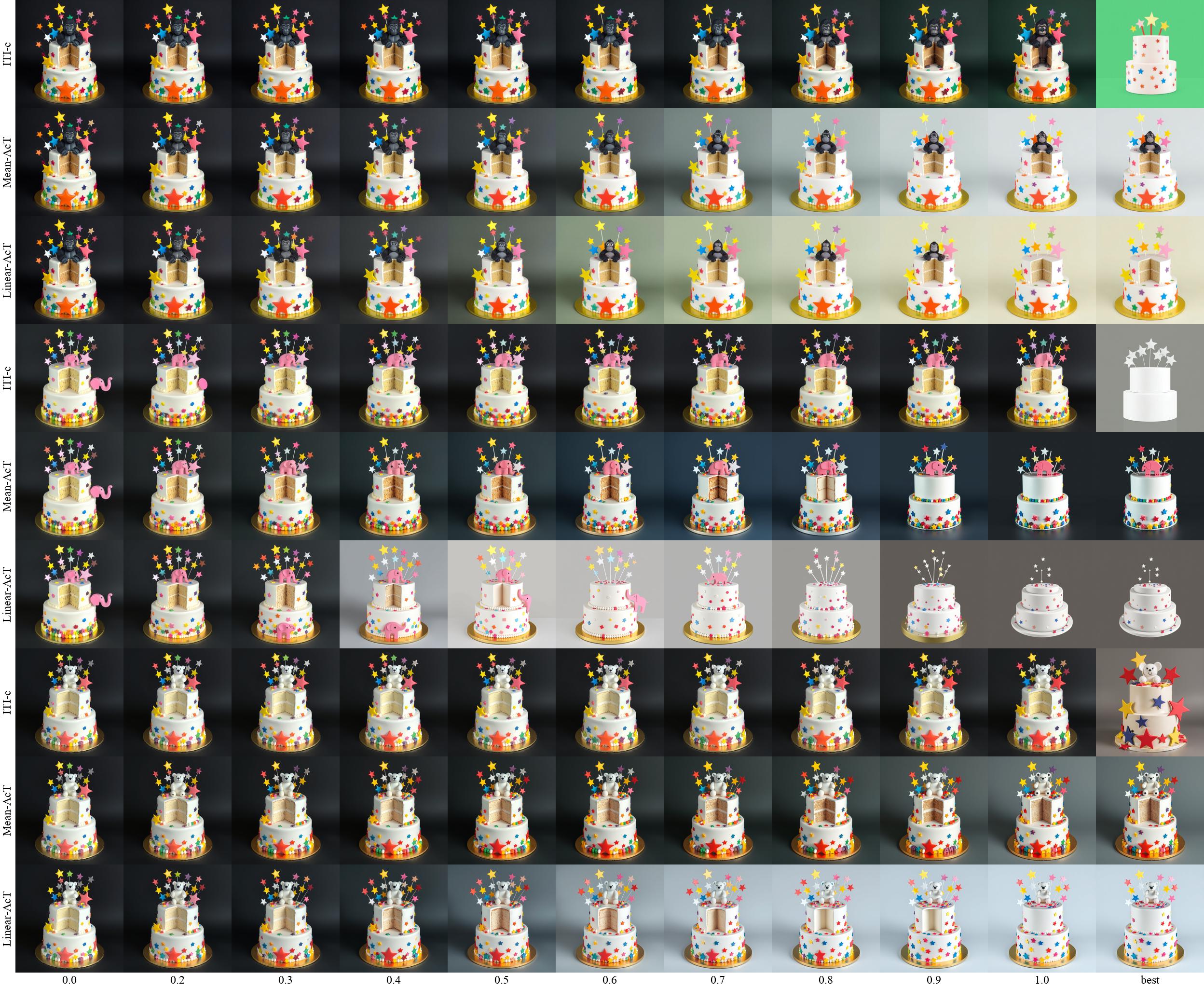}
     \subcaption{2 tier cake with multicolored stars attached to it.}
     \end{subfigure}
     \begin{subfigure}[t]{0.8\linewidth}
     \includegraphics[width=\linewidth]{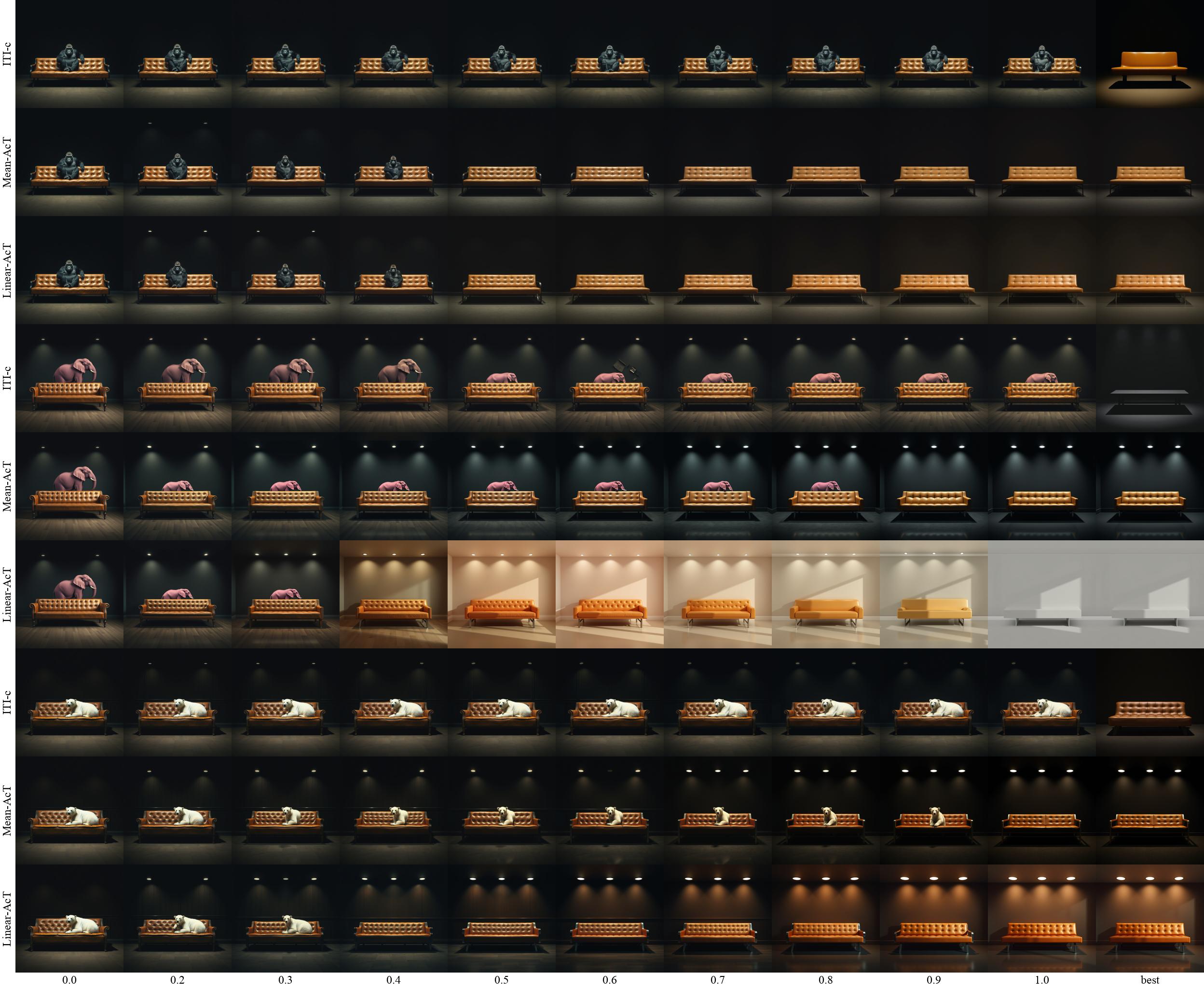}
    \subcaption{A table topped with bananas next to a coin.}
     \end{subfigure}
\caption{\textbf{FLUX - Concept negation examples I.} Rightmost column corresponds to the best strength found in \Cref{fig:clip_score} ($\lambda=1$ for \method and $\lambda=5$ for \iti). Every 3 rows represent a different concept in \texttt{\{gorilla, pink elephant, white bear\}} which was negated at the input of the image generator. \linear removes the negated concepts except for \textit{white bear} in (a). \iti is effective at ``best'' ($\lambda=5$). At high strengths, \linear and \iti also affect other image semantics.  }
\label{fig:flux-negation-I}
\end{figure}
\begin{figure}[t]
     \centering 
     \begin{subfigure}[t]{0.8\linewidth}
     \includegraphics[width=\linewidth]{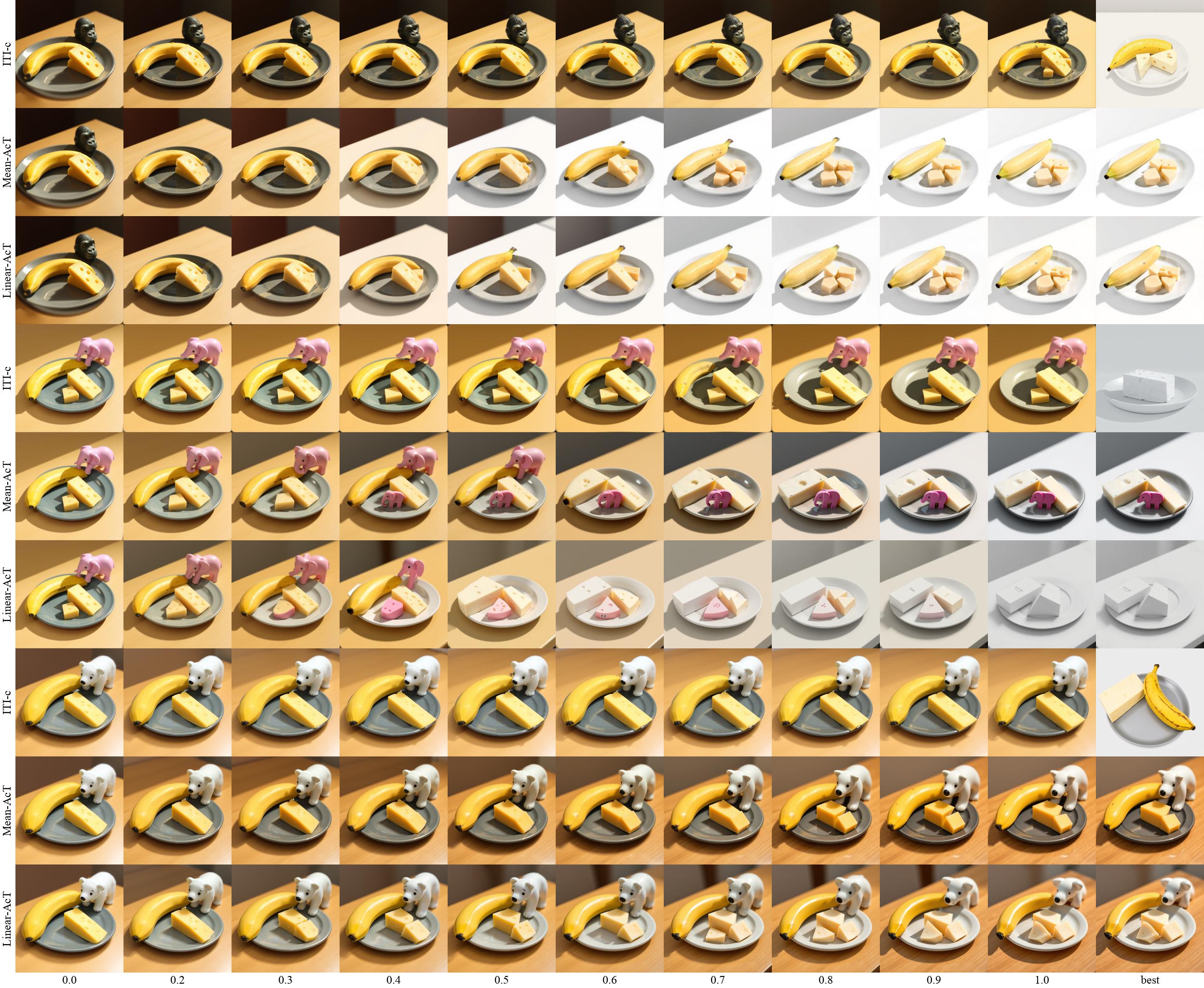}
     \subcaption{There is a banana and two pieces of cheese on a plate.}
     \end{subfigure}
     \begin{subfigure}[t]{0.8\linewidth}
     \includegraphics[width=\linewidth]{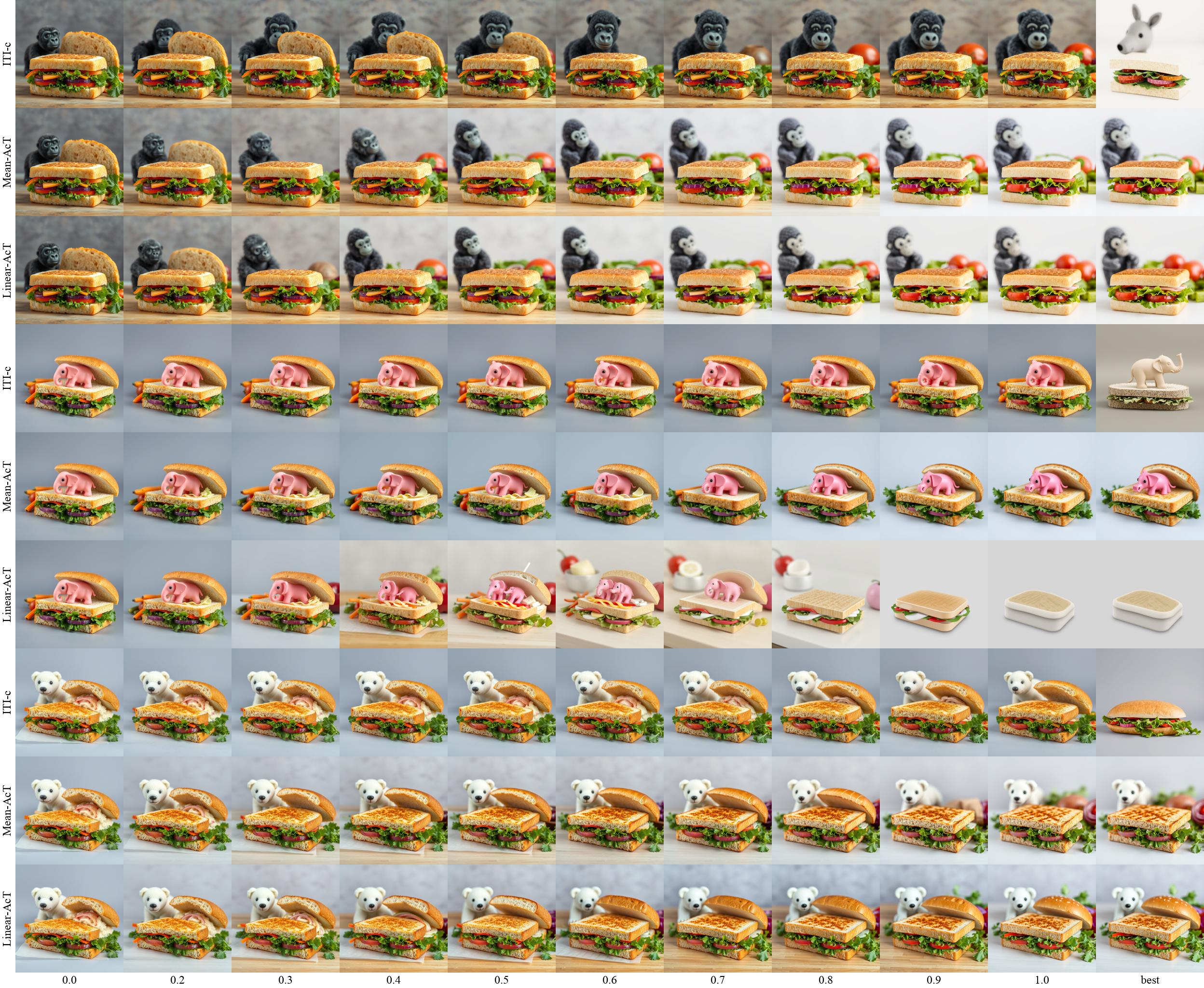}
    \subcaption{A sandwich is placed next to some vegetables. }
     \end{subfigure}
\caption{\textbf{FLUX - Concept negation examples II (Failures)} Rightmost column corresponds to the best strength found in \Cref{fig:clip_score} ($\lambda=1$ for \method and $\lambda=5$ for \iti). Every 3 rows represent a different concept in \texttt{\{gorilla, pink elephant, white bear\}} which was negated at the input of the image generator. \method does not remove \textit{white bear}, and fails to remove \textit{gorilla} in (b). For high $\lambda$, \linear modifies the semantics of the image. \iti removes the unwanted concept for $\lambda=5$.  }
\label{fig:flux-negation-II}
\end{figure}
\FloatBarrier
\subsection{Detailed Results}
\begin{figure}[h!]
    \centering
    \begin{subfigure}[t]{0.29\linewidth}
        \includegraphics[width=\linewidth]{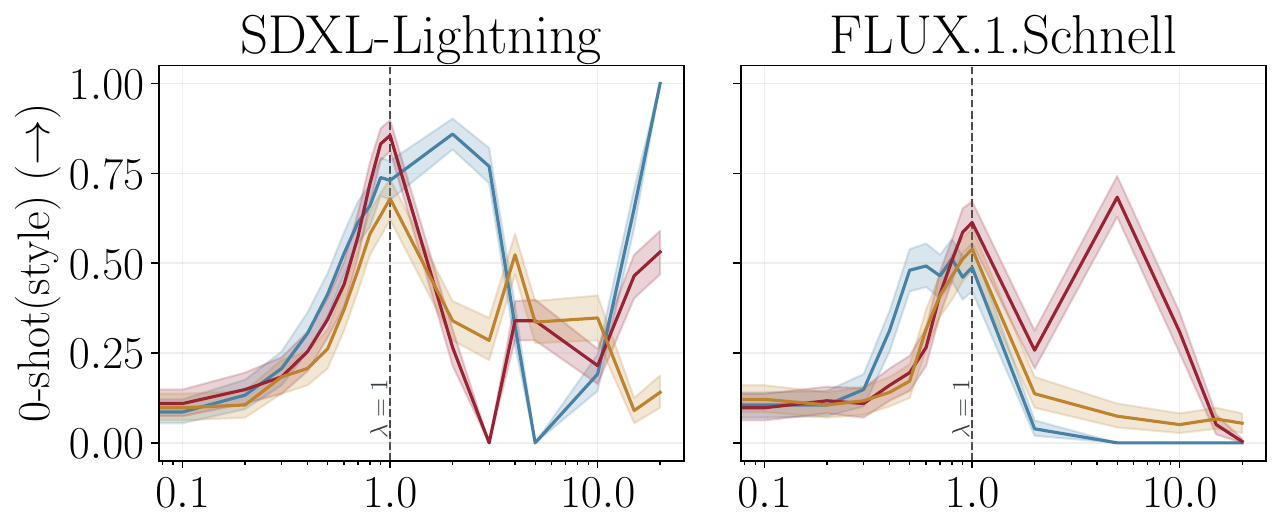}
        \includegraphics[width=\linewidth]{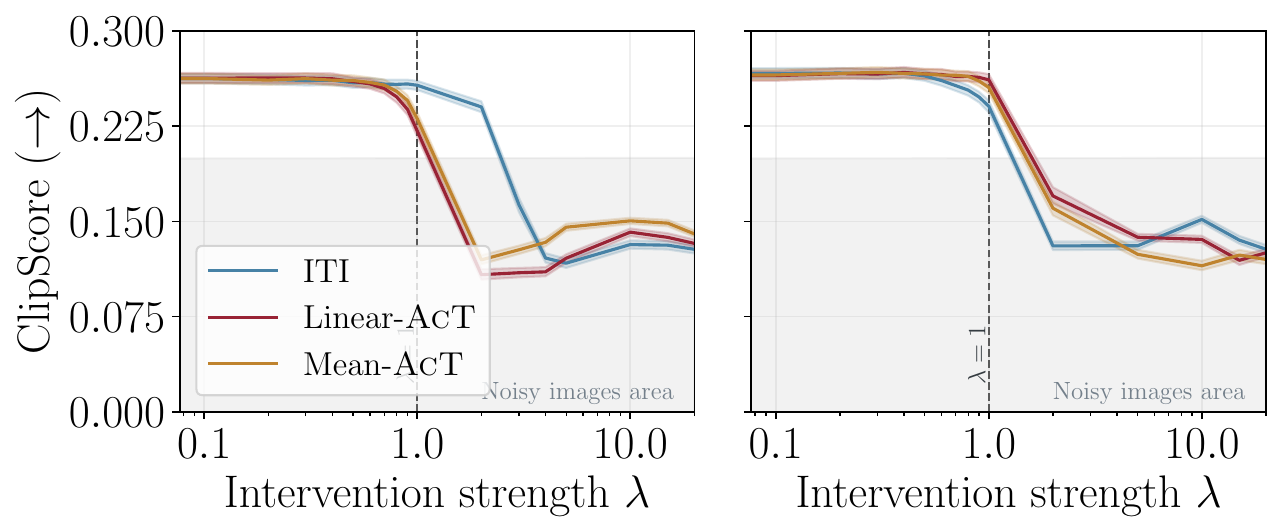}
        \caption{Anime}
    \end{subfigure}
    \begin{subfigure}[t]{0.29\linewidth}
        \includegraphics[width=\linewidth]{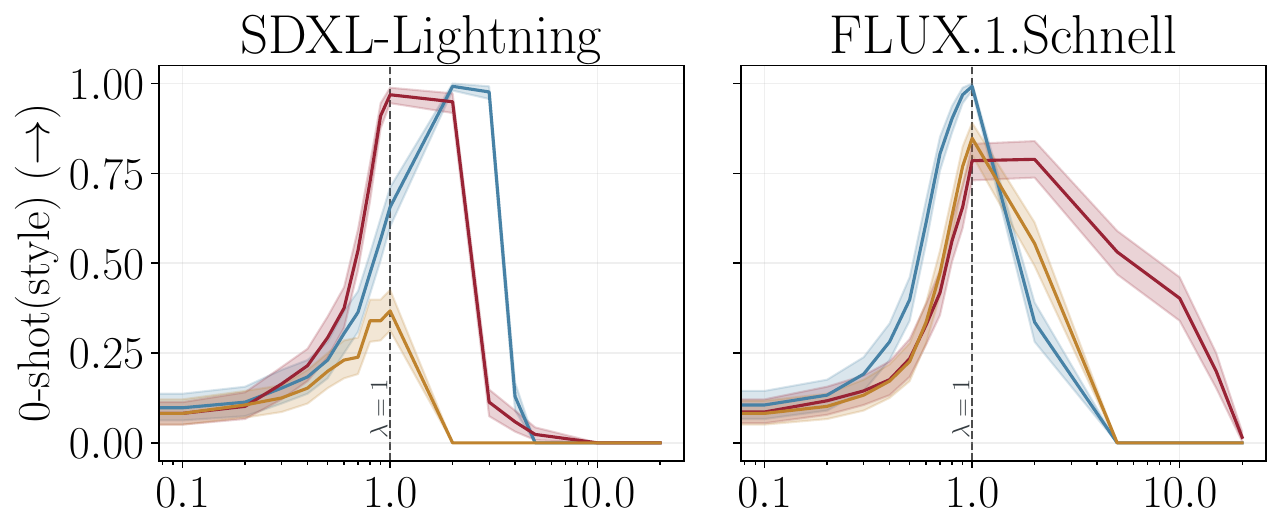}
        \includegraphics[width=\linewidth]{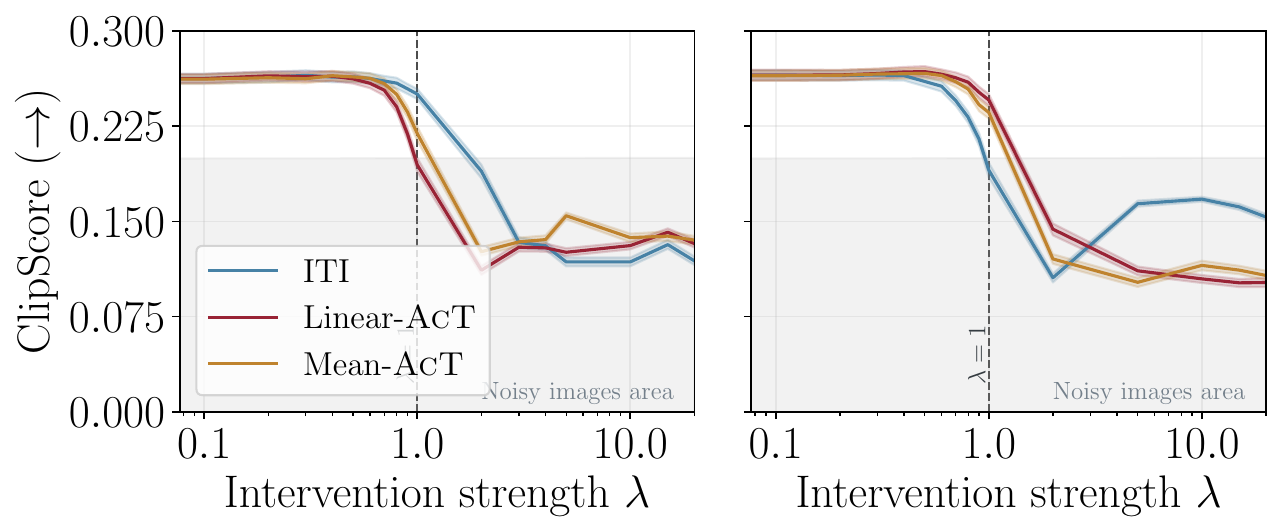}
        \caption{Art Nouveau}
    \end{subfigure}
    \begin{subfigure}[t]{0.29\linewidth}
        \includegraphics[width=\linewidth]{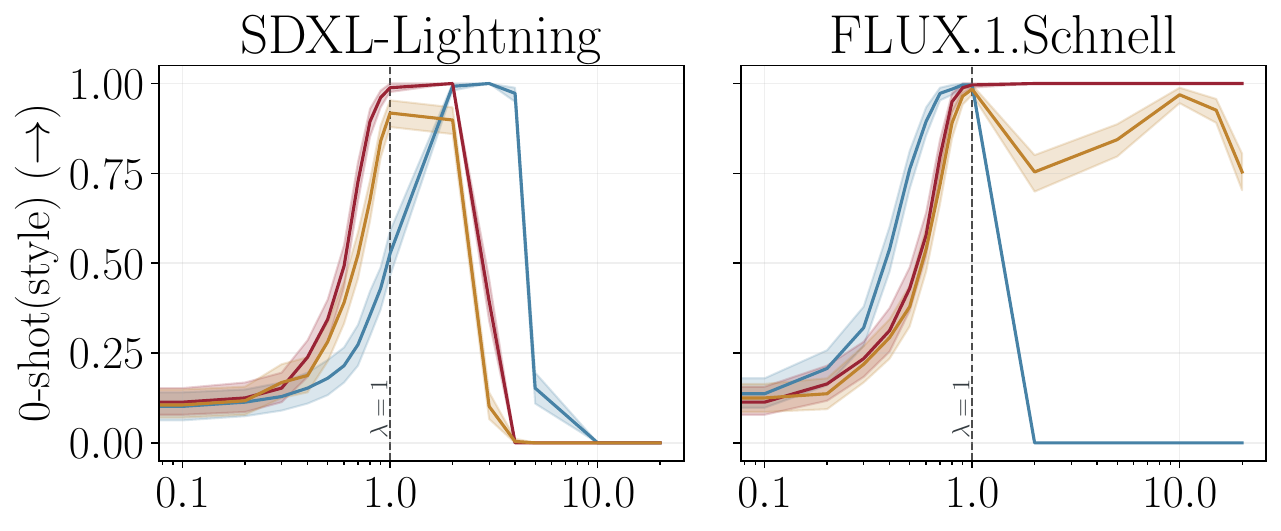}
        \includegraphics[width=\linewidth]{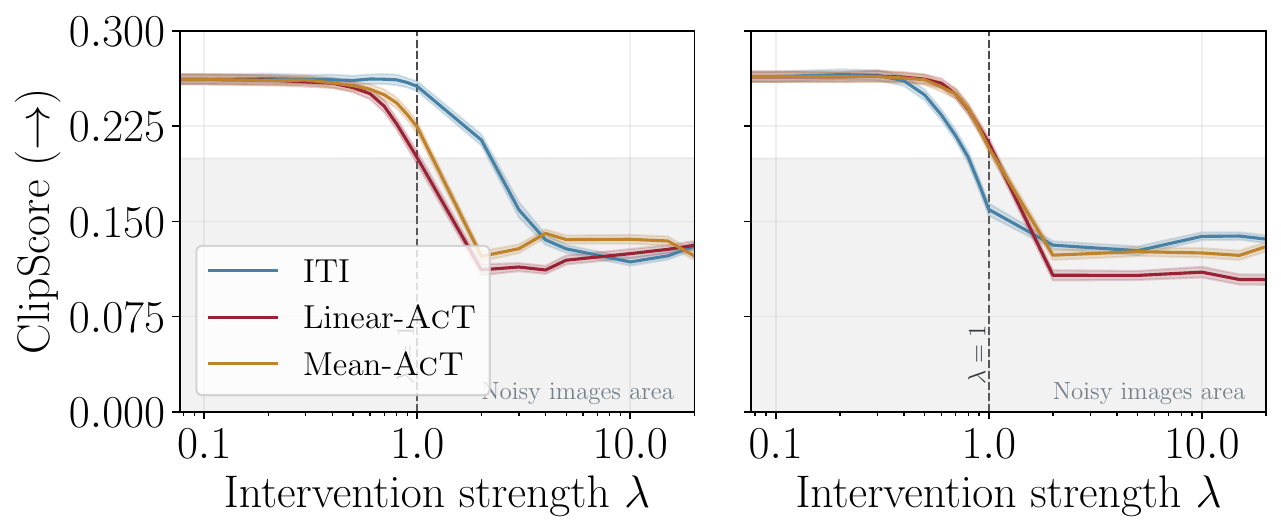}
        \caption{Cyberpunk}
    \end{subfigure}
    \begin{subfigure}[t]{0.29\linewidth}
        \includegraphics[width=\linewidth]{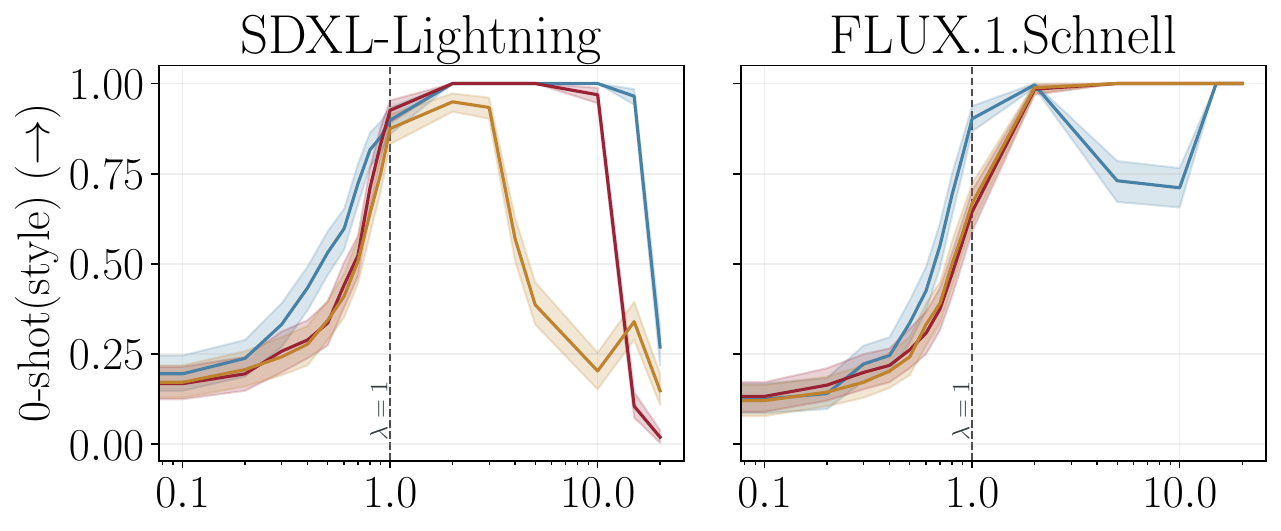}
        \includegraphics[width=\linewidth]{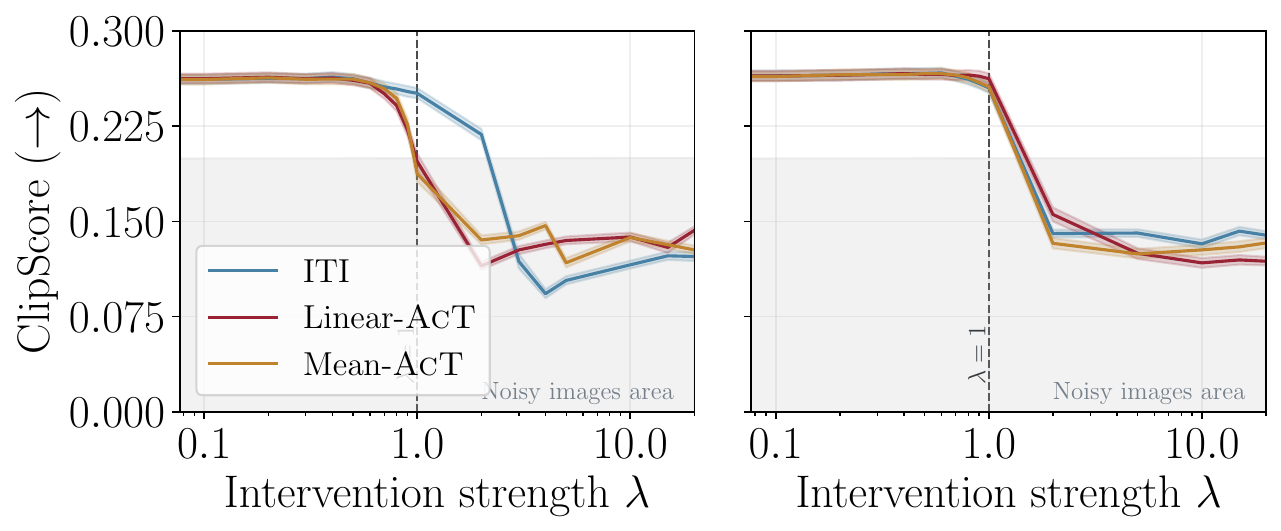}
        \caption{Impressionism}
    \end{subfigure}
    \begin{subfigure}[t]{0.29\linewidth}
        \includegraphics[width=\linewidth]{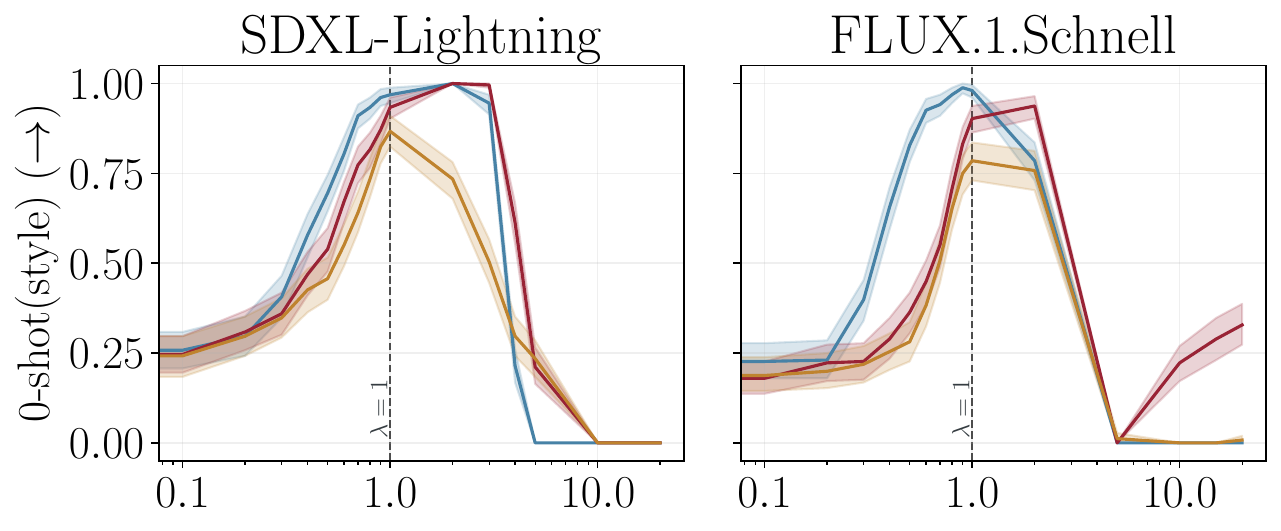}
        \includegraphics[width=\linewidth]{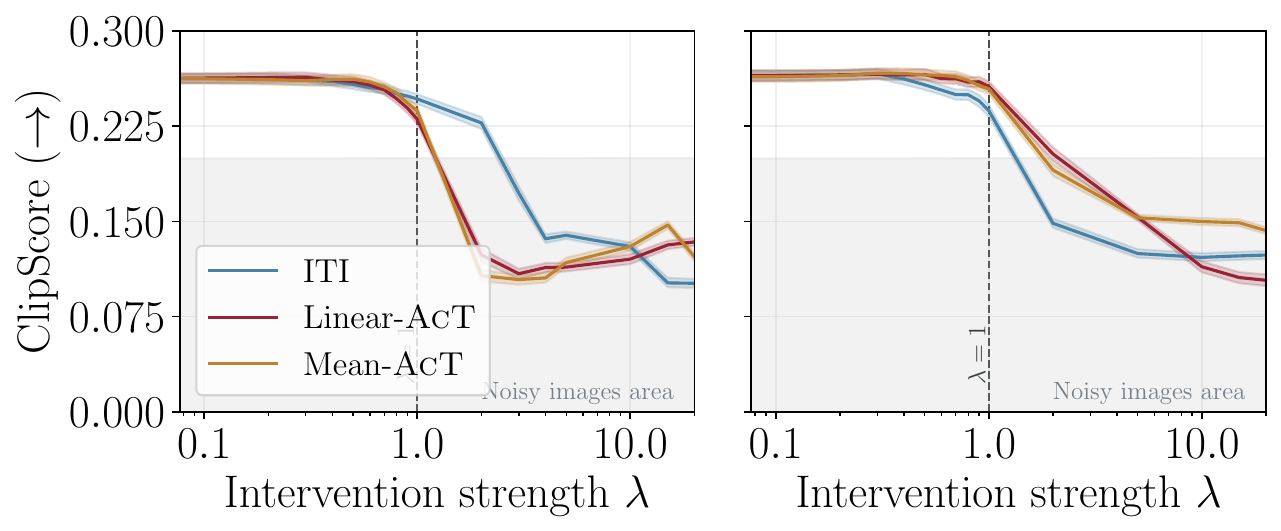}
        \caption{Sketch}
    \end{subfigure} 
    \begin{subfigure}[t]{0.29\linewidth}
        \includegraphics[width=\linewidth]{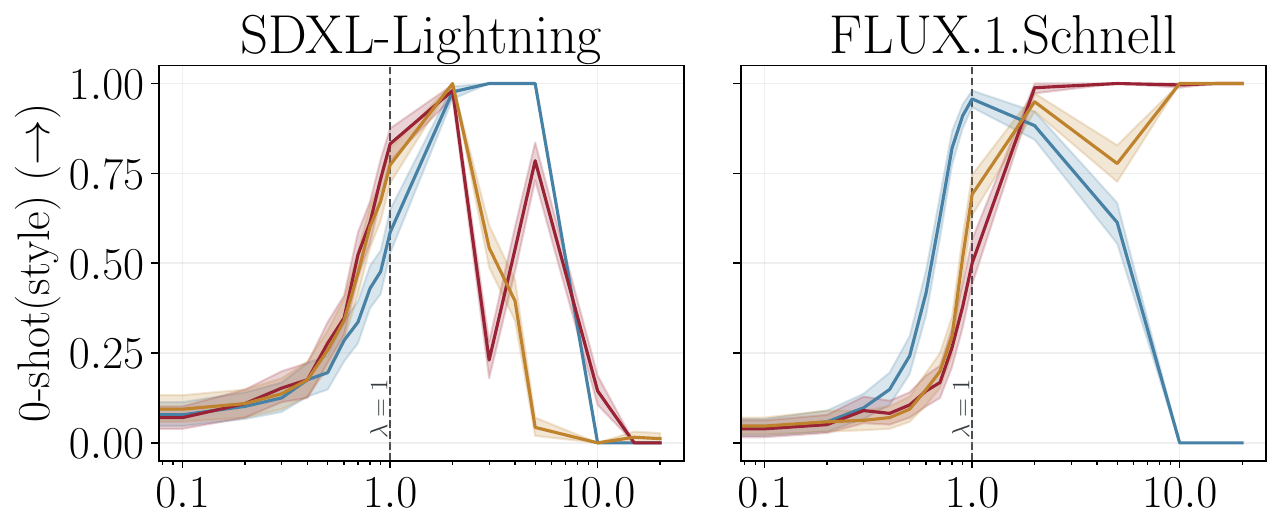}
        \includegraphics[width=\linewidth]{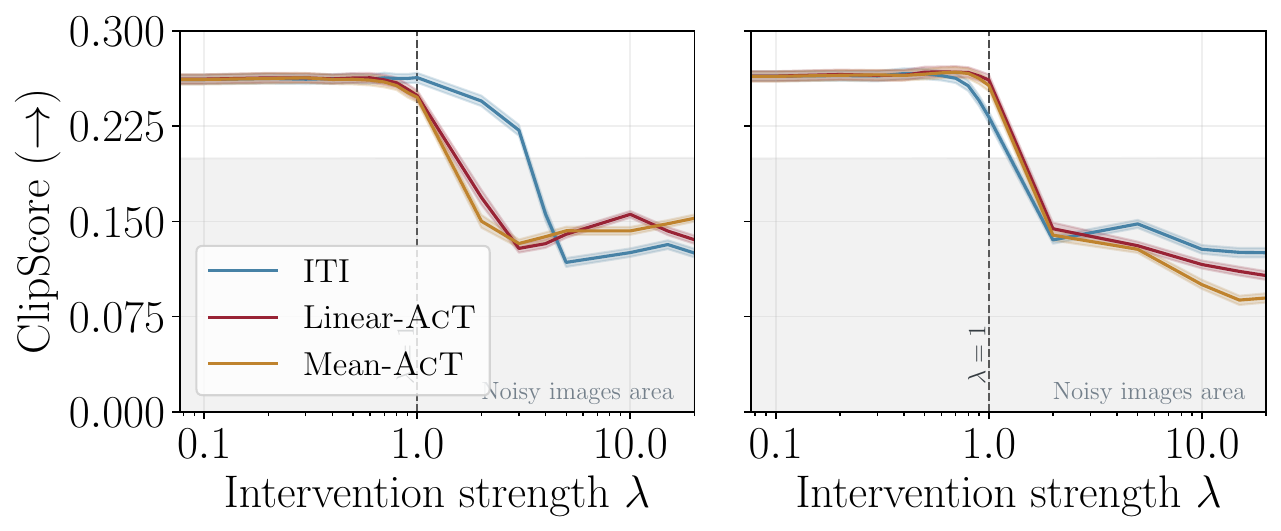}
        \caption{Watercolor}
    \end{subfigure}
    \caption{\textbf{Style induction.} For each style (a-f) and model (left-right), we show the 0-shot classification score for the style being present in the generated images (top) and the ClipScore to track how much generated images deviate from the unconditional prompt (bottom). The gray area indicates images that have lost their semantic content.}
    \label{fig:details-style}
\end{figure}
\begin{figure}[h!]
    \centering
    \begin{subfigure}[t]{0.29\linewidth}
        \includegraphics[width=\linewidth]{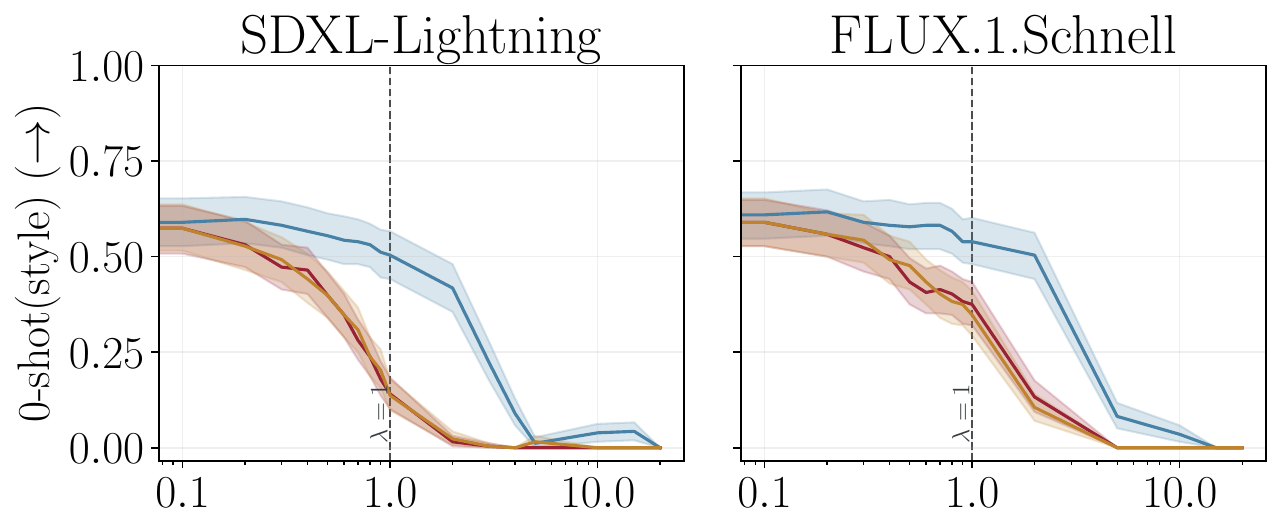}
        \includegraphics[width=\linewidth]{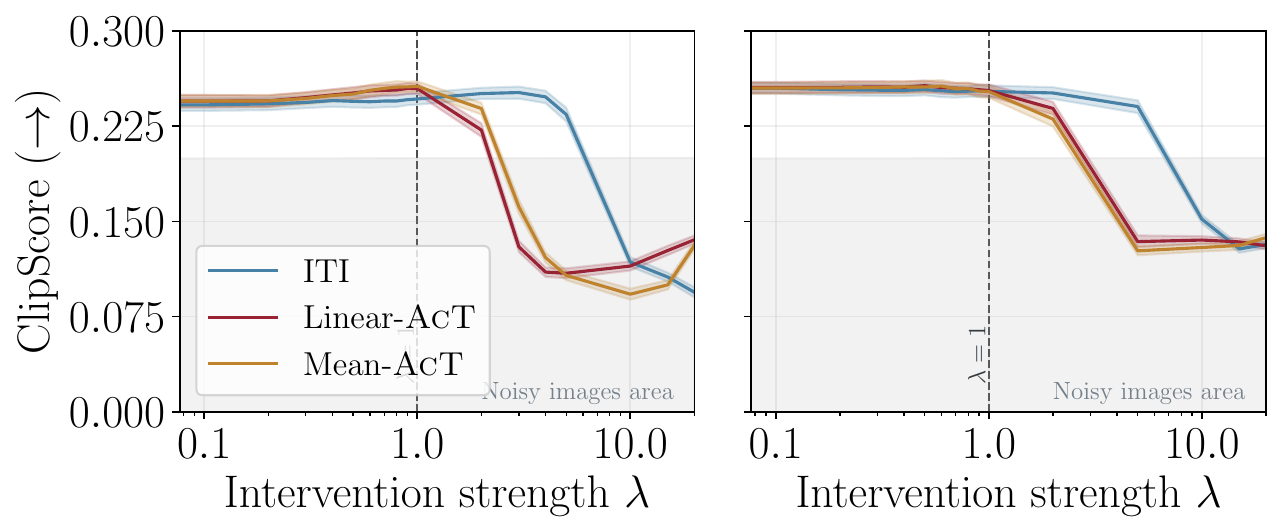}
        \caption{Gorilla}
    \end{subfigure}
    \begin{subfigure}[t]{0.29\linewidth}
        \includegraphics[width=\linewidth]{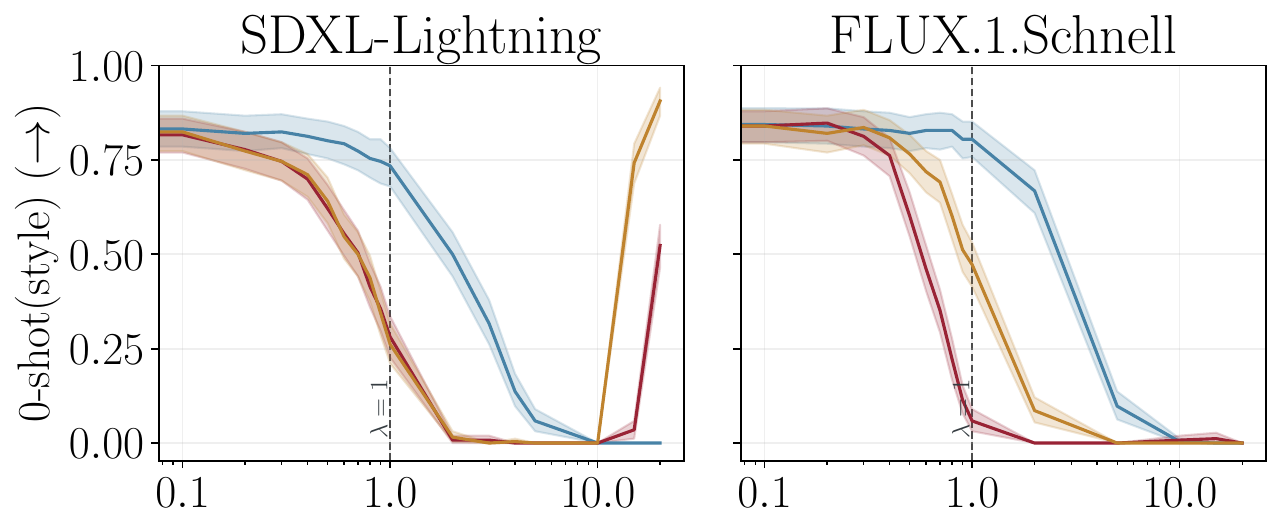}
        \includegraphics[width=\linewidth]{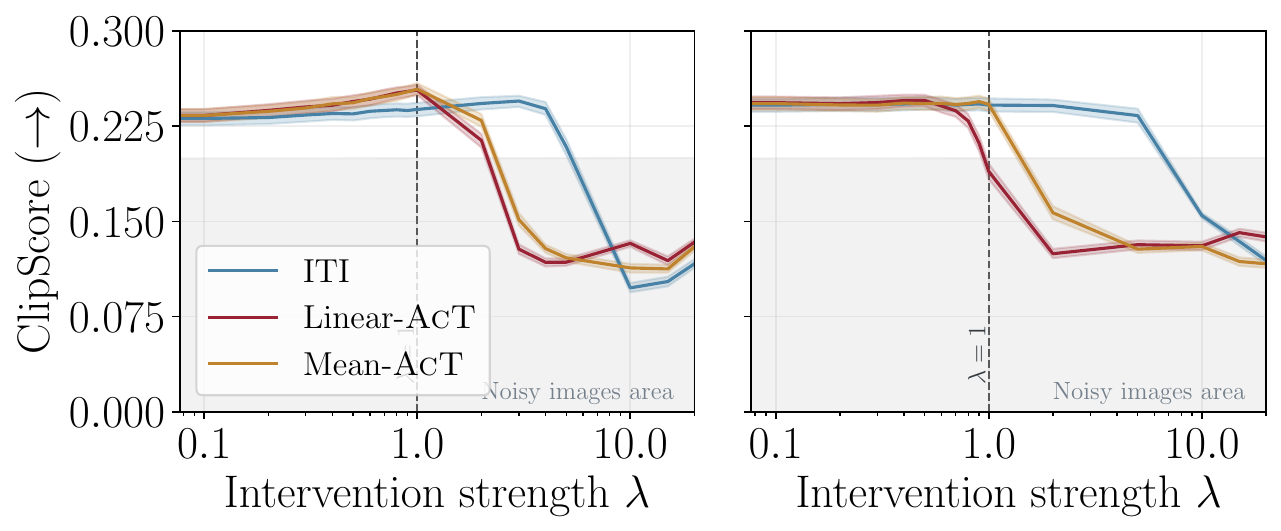}
        \caption{Pink elephant}
    \end{subfigure}
    \begin{subfigure}[t]{0.29\linewidth}
        \includegraphics[width=\linewidth]{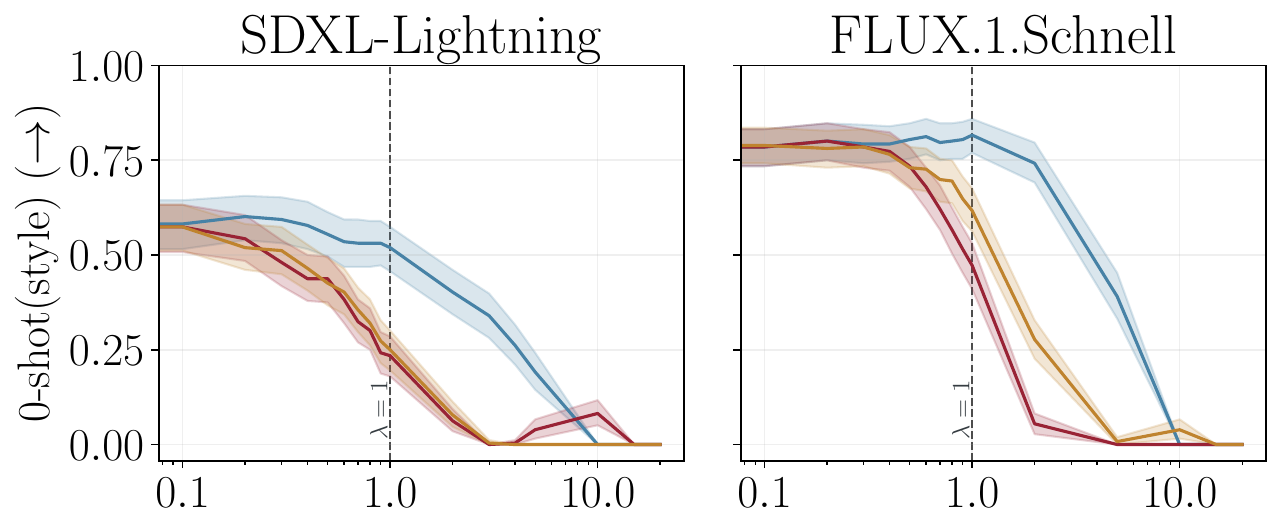}
        \includegraphics[width=\linewidth]{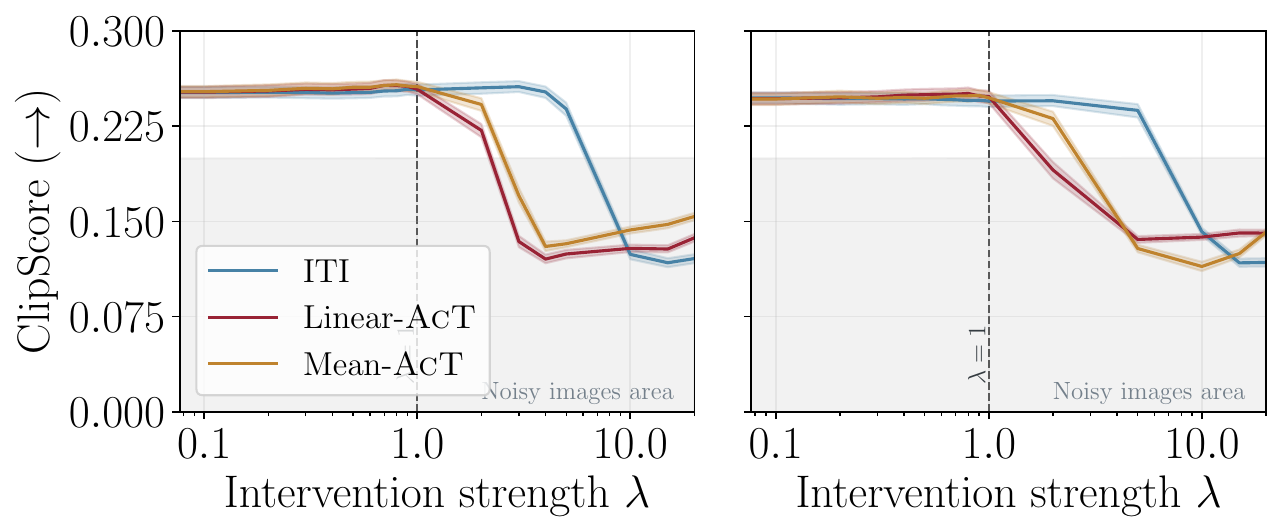}
        \caption{White bear}
    \end{subfigure}
    \caption{\textbf{Concept negation.} For each concept (a-c) and model (left-right), we show the 0-shot classification score for the concept being present in the generated images (top) and the ClipScore (bottom) to track how much generated images deviate from the unconditional prompt. The gray area indicates images that have lost their semantic content.}
    \label{fig:details-style}
\end{figure}
\FloatBarrier
\clearpage
\subsection{Style prompts}
\label{app:style-prompts}
\begin{table}[h!]
\caption{List of tags generated with Llama-8B-instruct (right) to induce different styles (left).}
\label{app:style-tags}
\begin{tabular}{rp{12cm}}
\toprule
Anime & \texttt{anime style, large expressive eyes, stylized hair, bold outlines, simplified colors, dynamic perspective, exaggerated features, angular shapes, chibis, manga inspired, emotive facial expressions, action sequences, speed lines, cell shading, graphic backgrounds, vibrant palettes}\\
Art nouveau & \texttt{Art Nouveau, Alphonse Mucha, Gustav Klimt, flowing lines, organic shapes, floral motifs, geometric patterns, ornamental designs, Jugendstil, Secessionism, symbolism, female figures, gold leaf, intricate details, turn of the century art, early 20th century}\\
Impressionism & \texttt{impressionism, Claude Monet, brush strokes, light, color, outdoor scenes, water lilies, haystacks, Rouen Cathedral, reflections, nature, atmospheric, vibrant colors, visible textures, 19th century art, French impressionism}\\
Cyberpunk & \texttt{cyberpunk, neon lights, urban jungles, high-tech architecture, augmented reality, AI technology, biopunk, futuristic cities, post-apocalyptic scenes, digital hacking, megacorporations, androids, dystopian societies, cybernetic enhancements, chromed details, glowing neon signs, rain-soaked streets}\\
Photorealism & \texttt{photorealism, hyperrealism, optical precision, photographic quality, fine detail, lifelike textures, realistic lighting, accurate perspective, human figures, still life, cityscapes, landscapes, skin tones, reflections and shadows, everyday objects, documentary style art, contemporary realism}\\
Sketch & \texttt{sketches, pencil drawing, charcoal sketches, ink illustrations, gestural lines, quick studies, figure drawing, perspective sketching, urban sketching, landscape sketches, still life drawings, sketchbook art, doodles, minimalist lines, expressive mark-making, observational drawing}\\
Watercolor & \texttt{watercolor style, transparent media, wet-on-wet application, dry brush strokes, soft blending, delicate touches, gentle shading, luminous hues, atmospheric lighting, ethereal quality, subtle textures, color gradients, painterly aesthetics, fluid paint behavior, watercolor paper texture}\\
\bottomrule
\end{tabular}
\end{table}
\FloatBarrier
\clearpage
\subsection{Concept prompts}
\label{app:concept-prompts}
\begin{table}[h!]
\caption{List of tags generated with Llama-8B-instruct (right) to induce different concepts (upper left) or to prompt models not to generate them (lower left).}
\label{app:concept-tags}
\begin{tabular}{rp{12cm}}
\toprule
Pink elephant & \texttt{a pink elephant. containing a pink elephant. with a pink elephant in plain view. and a pink elephant. it displays a pink elephant. featuring a pink elephant. in addition to a pink elephant. and also a pink elephant. and a pink elephant as well. the pink elephant can be clearly seen.}\\
Gorilla & \texttt{a gorilla. containing a gorilla. with a gorilla in plain view. and a gorilla. it displays a gorilla. featuring a gorilla. in addition to a gorilla. and also a gorilla. and a gorilla as well. the gorilla can be clearly seen.}\\
White bear & \texttt{a white bear. containing a white bear. with a white bear in plain view. and a white bear. it displays a white bear. featuring a white bear. in addition to a white bear. and also a white bear. and a white bear as well. the white bear can be clearly seen.}\\
\midrule
No pink elephant & \texttt{without a pink elephant. not containing a pink elephant. without a pink elephant in plain view. and a pink elephant that cannot be seen. it does not display a pink elephant. not featuring a pink elephant. lacking a pink elephant. and not a pink elephant. and a pink elephant is missing. the pink elephant cannot be seen.}\\
No gorilla & \texttt{without a gorilla. not containing a gorilla. without a gorilla in plain view. and a gorilla that cannot be seen. it does not display a gorilla. not featuring a gorilla. lacking a gorilla. and not a gorilla. and a gorilla is missing. the gorilla cannot be seen.}\\
No white bear & \texttt{without a white bear. not containing a white bear. without a white bear in plain view. and a white bear that cannot be seen. it does not display a white bear. not featuring a white bear. lacking a white bear. and not a white bear. and a white bear is missing. the white bear cannot be seen.}\\
\bottomrule
\end{tabular}
\end{table}
\FloatBarrier
\subsection{Details on FLUX conditioning}
\label{app:flux-details}
FLUX's diffusion architecture~\footnote{https://blackforestlabs.ai/announcing-black-forest-labs/} is based on the transformer architecture~\citep{vaswani2017attention}. Concretely, it is composed of $N$ consecutive multi-modal fusion transformer residual blocks followed by $M$ uni-modal transformer residual blocks. We found that the most effective strategy for strong conditioning is to intervene upon the output of all blocks. However, we found that conditioning blocks closest to the output tends to deteriorate the generated images. Thus, we condition all the $N$ multi-modal blocks and the first $15$ uni-modal blocks. 

\end{document}